\documentclass[preprint]{elsarticle}
\usepackage{geometry}

\makeatletter                                           
\def\ps@pprintTitle{
    \let\@oddhead\@empty
    \let\@evenhead\@empty
}
\makeatother

\usepackage{amssymb}

\usepackage{float}     
\usepackage{subfigure}
\usepackage{amsmath}   
\usepackage{enumitem}  
\usepackage{siunitx}   
\usepackage{tabularx}  
\usepackage{xcolor}
\usepackage{tabularx,booktabs}
\usepackage{graphicx}
\usepackage{tabularx}
\usepackage{multirow}
\usepackage{colortbl}
\newcolumntype{L}{>{\raggedright\arraybackslash}X}

\begin{document}

\begin{frontmatter}




\title{Density Uncertainty Quantification with NeRF-Ensembles: Impact of Input Data. Towards Robust 3D Reconstruction and Artifact Removal}
\title{Density Uncertainty Quantification with NeRF-Ensembles: Impact of Input Data}
\title{Density Uncertainty Quantification with NeRF-Ensembles: \newline Impact of Data and Scene Constraints}

\author{Miriam Jäger\corref{cor1}\fnref{label1}}
\ead{miriam.jaeger@kit.edu}
\author{Steven Landgraf\fnref{label1}}
\author{Boris Jutzi\fnref{label1}}
\address[label1]{Institute of Photogrammetry and Remote Sensing (IPF), Karlsruhe Institute of Technology (KIT), Karlsruhe, Germany}
\cortext[cor1]{Corresponding author.}

\begin{abstract}
In the fields of computer graphics, computer vision and photogrammetry, Neural Radiance Fields (NeRFs) are a major topic driving current research and development. However, the quality of NeRF-generated 3D scene reconstructions and subsequent surface reconstructions, heavily relies on the network output, particularly the density. Regarding this critical aspect, we propose to utilize NeRF-Ensembles that provide a density uncertainty estimate alongside the mean density. We demonstrate that data constraints such as low-quality images and poses lead to a degradation of the training process, increased density uncertainty and decreased predicted density. Even with high-quality input data, the density uncertainty varies based on scene constraints such as acquisition constellations, occlusions and material properties. NeRF-Ensembles not only provide a tool for quantifying the uncertainty but exhibit two promising advantages: Enhanced robustness and artifact removal. Through the utilization of NeRF-Ensembles instead of single NeRFs, small outliers are removed, yielding a smoother output with improved completeness of structures. Furthermore, applying percentile-based thresholds on density uncertainty outliers proves to be effective for the removal of large (foggy) artifacts in post-processing. We conduct our methodology on 3 different datasets: (i) synthetic benchmark dataset, (ii) real benchmark dataset, (iii) real data under realistic recording conditions and sensors.
\end{abstract}


\end{frontmatter}


\section{Introduction}\label{sec:introduction}

The groundbreaking research on Neural Radiance Fields (NeRFs) \cite{mildenhall_et_al_2020} initiated a new era in computer graphics. NeRFs take a sparse set of images and corresponding camera poses to predict position-dependent density and view-dependent color values, creating new views through the so-called novel-view synthesis. Furthermore, NeRFs provide opportunities in computer vision and photogrammetry by enabling 3D surface and point cloud reconstructions. However, these density-dependent reconstructions often yield noisy and incomplete surfaces \citep{neuralangelo,NeuS}. Unfortunately, the resulting geometry of 3D scene reconstruction depends on the network output, especially the predicted density. Since the network optimization is based on the error between training images and rendered images, the density is not controlled with respect to a ground truth information and only learned implicitly. Additionally, each neural network generally leads to a slightly different prediction, which invariably leads to different results.\newline

For this reason, the application of Deep Ensemble to approximate the predictive uncertainty is of interest. In the context of NeRFs, our investigations focus on the position-dependent density in 3D space to quantify the density uncertainty. For this purpose, we apply NeRF-Ensembles. Such NeRF-Ensembles consist of randomly initialized, independent members of individual NeRFs, providing a quantification of the density uncertainty of the network output, as well as a mean density.\newline


In this paper, we aim to identify crucial factors for achieving optimal 3D scene reconstructions with NeRFs, building the foundation for subsequent 3D surface and point cloud reconstructions. In this context, we analyze the impact of data constraints, such as the quality of input images and camera poses, and explore the impact of scene constraints, including acquisition constellations, occlusions, and material properties. We assess quality in terms of training performance and density uncertainty of NeRFs. 
In addition to evaluating data and scene constraints, with a specific focus on the density uncertainty, we explore additional applications and potentials of NeRF-Ensembles. This includes analyzing the performance of NeRF-Ensembles with mean density compared to single NeRFs. Furthermore, we investigate the usage of NeRF-Ensembles for the density uncertainty-dependent removal of (foggy) artifacts in post-processing, also referring to fog and ghostly, or floater artifacts \cite{cleannerf, vipnerf, nerfbusters, artifacts_postprocessing}.

\begin{figure}[H] 
\begin{center}
		\includegraphics[width=0.9\columnwidth]{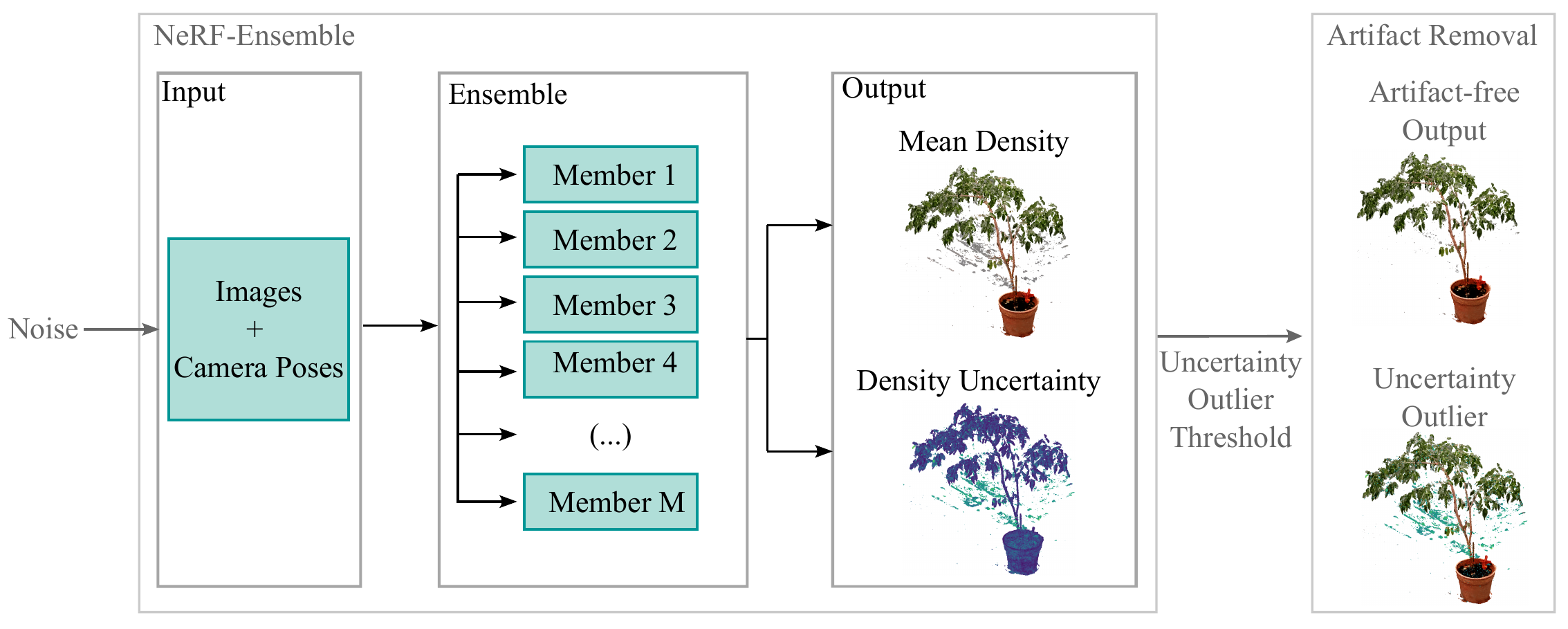}
	\vspace{-3mm}
	\caption{Flowchart of the methodology: Input data are the images and corresponding camera poses. A NeRF-Ensemble with a total number of M members is trained. The NeRF-Ensemble provides an average network output with mean density values over all NeRFs in the ensemble and a corresponding density uncertainty quantification of the prediction in 3D space. Optionally (shown in gray) we add noise to the input data and investigate the effect of data constraints on the results. In addition, a subsequent removing of high uncertainty points, e.g. with percentiles, offers the opportunity to remove (foggy) artifacts.}
\label{fig:flowchart}
\end{center}
\end{figure}

\vspace{-5mm}
Our main contributions can be summarized as follows:
\begin{itemize}[itemsep=0.5pt]
\item \textbf{Data constraints}. Low-quality images and poses have an impact on the training process, as indicated by the Peak Signal-to-Noise Ratio (PSNR) and the density uncertainty of the NeRF output: 
Image noise $\sigma_{\text{Im}}$ results in lower PSNR, but not necessary higher density uncertainty. While pose noise in translation $\sigma_{\text{t}}$, rotation $\sigma_{\text{R}}$ and their combination $\sigma_{\text{t,R}}$ results in both lower PSNR and higher density uncertainty. Overall, the density decreases significantly under data constraints, accompanied by the occurrence of (foggy) artifacts and a noisy scene reconstruction.
\item \textbf{Scene constraints}. Even with high-quality input data the density uncertainty in the scene varies based on acquisition constellations, occlusions and material properties. Areas of the scene that are poorly covered or obscured exhibit high density uncertainty, resulting in the presence of (foggy) artifacts.
\end{itemize}

In addition to the density uncertainty quantification, NeRF-Ensembles provides two further advantages:
\begin{itemize}[itemsep=0.5pt]
\item \textbf{Robustness}. The usage of the mean density from NeRF-Ensembles eliminates small outliers of the network output and yields a higher completeness of fine structures and low-textured surfaces, along with a kind of smoothing.
\item \textbf{Artifact Removal}. Large (foggy) artifacts due to low-quality poses, unsuitable acquisition constellations and occlusions can be removed in post-processing by density uncertainty-dependent outlier removal.
\end{itemize}

\newpage

\section{Related Work}\label{sec:related_work}
In this section, we summarize related work to our study. Thereby, we give an overview on basic and related research and development on NeRFs, Uncertainty Estimation in Deep Learning and Uncertainty Estimation in Neural Radiance Fields.

\paragraph{\textbf{Neural Radiance Fields}}
The foundation for the Neural Radiance Fields (NeRFs) was established by Scene Representation Networks \cite{SceneRepresenationNetworks}. Their underlying principle is modeling the scene as a function of 3D coordinates within it. It was followed by the groundbreaking research work of Neural Radiance Fields (NeRFs) \cite{mildenhall_et_al_2020}. These enable estimation of color values and densities for each 3D coordinate through 6D camera poses and associated 2D images by training a multi-layer perceptron (MLP).

The vanilla NeRF was followed by thousands of publications driving research and development in various domains.
Camera poses are an essential part of NeRFs. Therefore, some works deal with the efficient training in the case of incorrect or insufficient camera poses and their refinement. BaRF \cite{Barf} relies on bundle block adjustment to train NeRFs with insufficient camera poses. CBARF \cite{CBARF} cascades multiple bundle adjustment modules at different levels for stepwise pose refinement. GaRF \cite{Garf} introduces gaussian-activated radiance field to avoid the drop into local minima during training. Also, embedding pose refinement during training through image data is realized based on residues between pixels of the rendered and training images \cite{instantngp_refinement}. In contrast to the latter mentioned methods, the following do not require proper initial camera poses. GNeRF \cite{GNerf} trains a coarse NeRF with random camera poses for approximate poses, which are then refined. NoPe-NeRF \cite{NoPe-NeRF} incorporates undistorted monocular depth priors to constraint the positions of relative poses between images. \newline
Since the vanilla NeRF is subject to a training duration of hours to days, some works such as AdaNeRF \cite{adanerf}, FastNeRF \cite{fastnerf} and Instant NGP \cite{mueller_et_al_2022} focus on faster training and rendering. Plenoxels, a non-neural radiance field \cite{plenoxels} also manages this by representing the scene as a sparse 3D grid. A similar approach is taken by the latest work, 3D Gaussian Splatting \cite{gaussiansplatting}. Starting from a sparse point cloud as the initial Gaussian and learning the scene by splitting, growing and expanding Gaussians. However, since we want to investigate the quality of common radiance fields, we use the Instant NGP NeRF implementation in our study, which uses a combination of small MLPs and spatial hash table encoding for real-time training and rendering. \newline
Several works are also dedicated to enable artifact-free scene reconstruction, as artifacts often occur in the 3D scene. This is realized for example by visibility analysis that cause artifacts \cite{nerfbusters, vipnerf}. Whereby \cite{nerfbusters} refines the estimated density and color values in cubes to remove floater artifacts. Another approach performs post-processing removal based on consistency metrics in local neighborhoods to identify floater artifacts \cite{artifacts_postprocessing}. Clean-NeRF \cite{cleannerf} identifies view-dependent appearances e.g. from sparse input views during the training. Current work removes artifacts in the rendered images based on uncertainties from Bayesian Laplace approximation in null spaces due to the ray geometry \cite{bayesrays}. \newline

\paragraph{\textbf{Uncertainty Estimation in Deep Learning}}
Because of the large number of model parameters and non-linearities in deep neural networks, the exact posterior probability distribution of a network's output prediction is intractable \cite{blundell2015BayesByBackprop, loquercio2020UncertaintyFrameworkDronet}. Therefore, approximative uncertainty quantification approaches like Bayesian Neural Networks (BNN) \cite{mackay1992PracticalBayesian}, as well as Monte Carlo Dropout (MCD) \cite{gal2016DropoutBayesian}, and Deep Ensembles (DE) \cite{lakshminarayanan2017SimpleScalable} have been introduced.\newline
BNNs provide a mathematical approach where a deterministic neural network is transformed into a stochastic one. This can be achieved by placing probability distributions over the activations or the weights \cite{jospin2020HandsOnBaysianNetworks}. Sampling from these distributions at test time creates an ensemble of models which is used to sample from the posterior distribution of the predictions \cite{blundell2015BayesByBackprop}.
Gal et al. \cite{gal2016DropoutBayesian} propose MCD as an approximation of a stochastic process to overcome the high computational cost of BNNs. MCD introduces dropout, the well-established regularization method \cite{srivastava2014dropout}, at test time to sample from the posterior distribution of the predictions. Just like softmax probabilities, MCD uncertainties are not calibrated \cite{guo2017CalibrationModerna,gal2016DropoutBayesian}. This drawback is overcome by DEs, where an ensemble of trained models produces a sample of predictions at test time \cite{lakshminarayanan2017SimpleScalable}. By a random weight initialization or different data augmentations across the ensemble members, a diverse set of models can be trained \cite{fort2020DeepEnsembles}. Due to the introduced randomness across the ensemble members, DEs are well-calibrated \cite{lakshminarayanan2017SimpleScalable} and outperform other uncertainty quantification methods like softmax probability, BNNs, or MCD \cite{ovadia2019DatasetShift,wursthorn2022}. To overcome the substantial computational cost of DEs, the concept of knowledge distillation \cite{hinton2015DistillingKnowledgea} has been successfully proposed by multiple prior works \cite{simpson2022LearningStructured,holder2021EfficientUncertainty,landgraf2023dudes}.

\paragraph{\textbf{Uncertainty Estimation in Neural Radiance Fields}}
Uncertainty estimation techniques can also be applied to NeRFs. Stochastic Neural Radiance Fields (S-NeRF) \cite{S_nerf} learns a probability distribution over all the possible radiance fields modeling the scene. While Conditional-Flow NeRF (CF-NeRF) \cite{CF_nerf} learns a distribution over all possible complex radiance fields modelling the scene by coupling Latent Variable Modelling and Conditional Normalizing Flows without prior assumptions. In order to capture potential transient objects in the modeled scene, uncertainty on a pixel-level was addressed within the network by NeRF in the Wild (NeRF-W) \cite{W_nerf}. To ensure robustness in case of limited input data, ActiveNeRF \cite{active_nerf} are using the uncertainty estimation within the training process. The identification of uncertainty caused by a lack of input data is addressed through research on the termination of probabilities along individual rays \cite{ensemble_densityaware}. \newline

In contrast to other work, we use the NeRF-Ensemble technique for the density uncertainty quantification, while focusing on the impact of data and scene constraints on the uncertainty. Since the density is a position-dependent parameter, we investigate the scene implicitly described by the continuous field by discretizing it as a dense 3D grid. In addition, we use NeRF-Ensembles along with density uncertainty quantification for enhanced robustness of the 3D reconstruction and artifact-removal in post-processing.

\section{Methodology}\label{sec:method} 

In this section, we describe our methodology according to Figure \ref{fig:flowchart}, where an ensemble of several NeRFs are trained on the same input data to build a NeRF-Ensemble. The NeRF-Ensemble provides an average network output with mean density over all NeRFs in the ensemble and a corresponding quantification of the density uncertainty of the prediction in 3D space. Section \ref{sec:uncertainty} outlines the process of the density uncertainty estimation and mean density calculation using NeRF-Ensembles. The input data is given by the images and corresponding camera poses.
Furthermore we add noise to the input data (Section \ref{sec:input_data}), image noise and pose noise, and investigate the impact of data constraints on the NeRF training process and density uncertainty.


\subsection{Density Uncertainty}\label{sec:uncertainty}
NeRFs implicitly learn the 3D scene by optimizing the weights and bias of the network. Since the learned field represents a continuous implicit function, we discretize the implicit field by a 3D grid. This is realized by an equidistant fine sampling of positions within a 3D cube of the scene.
This allows us to achieve a comprehensive approximated coverage of the entire continuous 3D scene, which would not be possible with several single ray-based samplings, for instance.\newline 

The estimation of the density uncertainty is based on the output of a NeRF-Ensemble. Each NeRF represents a member of an ensemble of NeRFs trained on the same input data. Thereby the varying number of NeRFs are randomly initialized and independently trained. The randomly initialization of the weights \cite{weights} of the NeRFs is achieved in the Instant NGP \cite{mueller_et_al_2022} implementation. On this basis, the density uncertainty estimation is approximated by the density output of all members in the ensemble. For the position-wise uncertainty estimation, the same position for each NeRF member is regarded.\newline

For each 3D position $\text{X}=\text{x,y,z}$, the density uncertainty \text{$\text{U}_{\delta}$} and mean density $\overline{\delta}$ results from Equations \ref{equ:2} and \ref{equ:2_2}:
\begin{equation}\label{equ:2}
  \begin{aligned}
\text{$\text{U}_{\delta}$}(\text{X}) &= \sqrt{\frac{1}{\text{M}-1} \sum_{\text{m = 1}}^{\text{M}} (\delta_{\text{m}}(\text{X}) - \overline{\delta}(\text{X}))^2},
  \end{aligned}
\end{equation}
with
\begin{equation}\label{equ:2_2}
\begin{aligned}
 \overline{\delta}(\text{X}) &= \frac{1}{\text{M}} \sum_{\text{m = 1}}^{\text{M}} \delta_{\text{m}}(\text{X}),
   \end{aligned}
\end{equation}

where M is the total number of members per ensemble and $\overline{\delta}(\text{X})$ is the arithmetic mean value of the density values $\delta(\text{X})$ per position $\text{X}$.\newline

A specification of the mean uncertainty in the 2D image domain is derived from the uncertainty of all pixels \cite{holder2021EfficientUncertainty,landgraf2023dudes}. Since the 3D scene reconstruction is performed in 3D space when it comes to NeRFs, we calculate the mean density uncertainty \text{$\text{mU}_{\delta}$} based on all positions in a 3D density grid which approximated the scene, respectively.
Note that the density is a unit-less parameter that represents information about the volume density in 3D to render images with the so-called volume rendering \cite{mildenhall_et_al_2020}. Accordingly, the density uncertainty is also unit-less.

\subsection{Constraints}\label{sec:input_data}
The output of the network differs within the identical neural network for different training runs with the same input data. Additionally, suggesting that the quality of the input data, due to data constraints, such as image and pose noise, is directly correlated to the density uncertainty of the network output. Furthermore, we assume that increased density uncertainty results from other factors like scene constraints such as acquisition constellation, occlusions, material properties.

\paragraph{\textbf{Data Constraints: Image Noise, Pose Noise}}
 For the purpose of analyzing the impact of data constraints on the training process as well as density uncertainty, different types of input data configurations are investigated. Thereby synthetic noise is applied to both the images and the camera poses. The following modifications are considered: 

\begin{itemize}[itemsep=1pt]
\item \textbf{Image Noise \textbf{$\sigma_{\text{Im}}$}}: Random additive Gaussian noise of standard deviations in RGB color space of \textbf{$\sigma_{\text{Im}}$} is added to the images.
\item \textbf{Pose Noise Translation \textbf{$\sigma_{\text{t}}$}}: Random additive Gaussian noise of standard deviations of \textbf{$\sigma_{\text{t}}$} is added to the translation component t of the transformation matrix.
\item \textbf{Pose Noise Rotation \textbf{$\sigma_{\text{R}}$}}:  Random additive Gaussian noise of standard deviations of \textbf{$\sigma_{\text{R}}$} is added on the Euler angles of the rotation component R of the transformation matrix.
\item \textbf{Pose Noise Translation and Rotation \textbf{$\sigma_{\text{t,R}}$}}: Combines the rotation \textbf{$\sigma_{\text{t}}$} and translation noise \textbf{$\sigma_{\text{R}}$} as mentioned above, resulting in \textbf{$\sigma_{\text{t,R}}$}.
\end{itemize}

\paragraph{\textbf{Scene Constraints: Acquisition constellation, occlusions, material properties}}
In addition to data constraints due to low-quality input data, images and camera poses, we aim to investigate other possible sources of an increased density uncertainty. Under real recording conditions, data is not always recorded with a proper acquisition constellation and coverage of the scene. In addition, the objects in the scene are not always well textured and can reflect complexly in the images depending on the recording situation and material properties. For this reason, we analyze the impact of the following scene constraints on the density uncertainty: Acquisition constellations, occlusions, material properties.

\section{Experimental Setup}\label{sec:experiments_setup}
In this section, we introduce the setup for our experiments. We present the datasets which include different scene constraints in Section \ref{sec:dataset}, the training metric in Section \ref{sec:training_metric}, the implementation details in Section \ref{sec:implementation} and the configurations on data constraints in Section \ref{sec:investigations}.

\subsection{Datasets} \label{sec:dataset}
Our experiments are based on three different datasets, which differ in their characteristics and thereby include different levels and types of scene constraints such as acquisition constellation, occlusions and material properties.

\paragraph{Synthetic data} Since synthetic data have highly accurate camera poses and images and cover a wide range of object properties, they are particularly suitable for comprehensive analyses involving their modification. Therefore, we use the NeRF synthetic benchmark dataset \cite{mildenhall_et_al_2020}. The dataset consists of synthetic blender images with corresponding ideal camera poses, whereas we use 100 images each. The acquisition constellation provides images in an upper hemisphere or complete sphere that covers a majority of the scene, with a few gaps.

\paragraph{Real data} We focus on two different real datasets, to apply our analysis to real conditions.\newline
On the one hand, we use the DTU benchmark dataset \cite{dtu}, which includes scenes featuring real objects, including images and corresponding camera poses. We focus on six scenes, each containing either 49 or 64 RGB images. 
The acquisition constellation provides images in an upper half hemisphere that covers the scene from one side only. \newline
On the other hand, we use real HoloLens data \cite{jaeger2023hololens} captured with the Microsoft HoloLens generation 2 to investigate the effects of realistic recording conditions and sensors. It consists of 64 images from the 1920×1080 photo/video RGB camera of the HoloLens and corresponding internal camera poses of the device, as well as refined poses by Instant NGPs \cite{mueller_et_al_2022} pose refinement \cite{instantngp_refinement}. The acquisition constellation provides images in an upper hemisphere that covers the scene from all sides.

\subsection{Training metric}\label{sec:training_metric}
The evaluation of the impact of the input data on the training performance is done using the Peak Signal-to-Noise Ratio (PSNR) in \si{\dB}. It is a common metric to evaluate the quality of the synthetically rendered images compared to the training images, e.g. for NeRFs \cite{mildenhall_et_al_2020} and gives an indication of the training accuracy. It can be defined as follows: 

\begin{equation}\label{equ:3}
\text{{PSNR}} = -10 \cdot \log_{10}\left(\frac{{\text{{max\_value}}^2}}{{\text{{MSE}}}}\right),
\end{equation}

whereas the maximum value $\text{{max\_value}}$ is 255 for RGB color images and MSE der Mean Squared Error between the pixel values of two images.

\subsection{Implementation Details}\label{sec:implementation}
For all investigations, the Instant NGP \cite{mueller_et_al_2022} NeRF implementation was taken into account, since it enables real-time training and rendering. Regarding the network architecture, the basic NeRF architecture with ReLu activations and hash encoding is selected, while the training incorporates 50 000 training steps on a NVIDIA RTX3090 GPU.
Unless otherwise noted, we use 10 member per NeRF-Ensemble for all of our experiments, following prior work \cite{landgraf2023dudes,lakshminarayanan2017SimpleScalable,fort2020DeepEnsembles}. \newline

The uncertainty estimation in our application refers to the direct density network output of the NeRF, which is not yet affected by any further activation function that takes place for neural rendering. An exponential density activation before the rendering in the Instant NGP \cite{mueller_et_al_2022} implementation would cause a non-linear scaling of the data. This increases uncertainties in high density scale ranges between the members of the NeRF-Ensemble non-linearly and would make them less comparable to the density values in low ranges. 
Furthermore, for part of the analyses we only consider the network output with a density greater than 15. High density values lead to large alpha channels \cite{mildenhall_et_al_2020} in NeRFs. 
Therefore, points with high density are visible in the rendered image through neural view synthesis, while low density areas hardly have an effect on the scene reconstruction as well as the rendered images. This makes them less relevant in case of noise. 

\subsection{Data Constraints: Image Noise, Pose Noise} \label{sec:investigations}
Several configurations are investigated in order to study the impact of the data constraints (see Section \ref{sec:input_data}) on the NeRF output. For this purpose, the images and corresponding original ground truth camera poses of each dataset are used directly and modified by adding random gaussian noise on images and poses.
For the NeRF synthetic and DTU dataset, in addition to the original input data without noise, the configurations of Table \ref{tab:data_constraints} are analyzed. For the HoloLens data, we apply both the HoloLens' internal camera poses and the poses refined under Instant NGPs pose refinement \cite{instantngp_refinement} with the same corresponding images. 

\begin{table}[H]
\caption{Data Constraints. Configurations of the random additive Gaussian Image and Pose Noise on the NeRF synthetic and DTU dataset. For the NeRF synthetic dataset, the cameras on a (hemi-) sphere with a non-metric circumference of circa 25.33. We add Gaussian noise of a percentage of the circumference for $\sigma_{\text{t}}$.}
\label{tab:data_constraints}
\begin{tabularx}{0.6\columnwidth}{@{\hspace{1.3cm}} L l l @{}}
\toprule
\multicolumn{1}{@{}>{\hsize=3.5cm}X}{\textbf{}} & NeRF synthetic & DTU \\
\midrule
\text{Image Noise} & 5\% & 15 \\
$\sigma_{\text{Im}}$ & 10\% & \\
 & 15\% & \\
 & 20\% & \\ \arrayrulecolor{black!30}\midrule
\text{Pose Noise} & 0.01\% & 0.1\,mm \\
$\sigma_{\text{t}}$ & 0.02\% & \\
 & 0.03\% & \\
 & 0.04\% & \\ \arrayrulecolor{black!30}\midrule
\text{Pose Noise} & 0.1° & 0.01° \\
$\sigma_{\text{R}}$ & 0.2° & \\
 & 0.3° & \\
 & 0.4° & \\ \arrayrulecolor{black!30}\midrule
\text{Pose Noise} & 0.01\%, 0.1° & 0.1\,mm, 0.01° \\
$\sigma_{\text{t,R}}$ & 0.02\%, 0.2° & \\
 & 0.03\%, 0.3° & \\
 & 0.04\%, 0.4° & \\ \arrayrulecolor{black}\bottomrule
\end{tabularx}
\end{table}


\section{Experimental Results} \label{sec:experimental_results}
The evaluation is categorized into two main points: Firstly, \textit{Data Constraints: Image Noise, Pose Noise} in Section \ref{sec:data_constraints_results}. The qualitative variation of the training process and density uncertainty based on different qualities of input data. Secondly, \textit{Scene Constraints: Acquisition constellation, occlusions, material properties} in Section \ref{sec:scene_constraints_results}. The density uncertainty of noticeable anomalies from scene constraints in the results. \newline
In addition, this Section provides an outlook on the potential of NeRF-Ensembles in Section \ref{sec:results_potential} regarding robustness and removal of (foggy) artifacts.

\subsection{Data Constraints: Image Noise, Pose Noise} \label{sec:data_constraints_results}
This Section presents the impact of data constraints. It includes a quantitative evaluation of the training process (Section \ref{sec:results_training}), as well as a quantitative and qualitative evaluation of the density uncertainty (Section \ref{sec:results_density_uncertainty}) due to data constraints.

\subsubsection{\textbf{Training}}\label{sec:results_training}
Concerning the NeRF synthetic dataset, the quantitative results indicate a consistent trend, as shown in the Table \ref{tab:PSNR}.
There is a rapid drop in PSNR for the lowest additive noise per noise type \text{$\sigma_{\text{Im}}$}, \text{$\sigma_{\text{t}}$}, \text{$\sigma_{\text{R}}$} and \text{$\sigma_{\text{t,R}}$}. Image noise \text{$\sigma_{\text{Im}}$=5} leads to a decrease in PSNR of 1 bis 3 \si{\dB}.
After an initial rapid drop of the PSNR when adding the lowest additive noise, the PSNR per scene continues to decrease for additional noise. Adding noise with a steady magnitude also leads to a steady decrease in PSNR almost continuously in steps of 1 to 4 \si{\dB} depending on noise type and scene. In addition, no further reduction in PSNR is observed for the combination of \text{$\sigma_{\text{t}}$} and \text{$\sigma_{\text{R}}$} to \text{$\sigma_{\text{t,R}}$}.
A similar trend is noticed for the DTU dataset, as shown in Table \ref{tab:PSNR_dtu}. The occurrence of noise leads to a lower PSNR value. Again, the PSNR value for all scenes decreases due to additive Gaussian noise of all types \text{$\sigma_{\text{Im}}$}, \text{$\sigma_{\text{t}}$}, \text{$\sigma_{\text{R}}$} between 1 and 3 \si{\dB}.
The combination of noise in translation and rotation \text{$\sigma_{\text{t,R}}$} again does not lead to a corresponding additive decrease in the PSNR, just as with the NeRF synthetic dataset. 
The real HoloLens data provides a PSNR of 25 \si{\dB} with the internal, unrefined poses. Again, the PSNR is lower than for input data where pose refinement takes place with a PSNR of 27.5 \si{\dB}.

\begin{table}[H]
\caption{\textbf{Mean PSNR}$\uparrow$ in \si{\dB} on the NeRF synthetic dataset per NeRF-Ensemble.}
\label{tab:PSNR}
\begin{tabularx}{\textwidth}{@{} l *{1}{>{\hsize=1.5\hsize}X} *{9}{X} @{}}
\toprule
 & & chair & drums & ficus & hotdog & lego & materials & mic & ship \\
\midrule
Original		 & 		& \cellcolor{green!25}37.07 & \cellcolor{green!25}30.68 & \cellcolor{green!25}36.14 & \cellcolor{green!25}38.98 & \cellcolor{green!25}38.44 & \cellcolor{green!25}33.71 & \cellcolor{green!25}38.55 & \cellcolor{green!25}32.85  \\	\arrayrulecolor{black!30}\midrule 
Image Noise & 5 			& \cellcolor{red!5}35.63 & \cellcolor{red!5}30.13 & \cellcolor{red!5}32.35 & \cellcolor{red!5}35.29 & \cellcolor{red!5}35.21 & \cellcolor{red!5}32.34 & \cellcolor{red!5}37.13 & \cellcolor{red!5}30.67 \\	
$\sigma_{\text{Im}}$ 		    & 10 		& \cellcolor{red!10}33.28 & \cellcolor{red!10}29.17 & \cellcolor{red!10}31.65 & \cellcolor{red!10}31.36 & \cellcolor{red!10}31.60 & \cellcolor{red!10}30.58 & \cellcolor{red!10}34.91 & \cellcolor{red!10}28.12 \\

		    & 15 		& \cellcolor{red!15}30.94 & \cellcolor{red!15}28.06 & \cellcolor{red!15}30.45 & \cellcolor{red!15}28.68 & \cellcolor{red!15}29.19 & \cellcolor{red!15}28.83 & \cellcolor{red!15}32.78 & \cellcolor{red!15}26.22 \\ 	
		    & 20 		& \cellcolor{red!20}29.24 & \cellcolor{red!20}27.09 & \cellcolor{red!20}29.55 & \cellcolor{red!20}26.84 & \cellcolor{red!20}27.44 & \cellcolor{red!20}27.32 & \cellcolor{red!20}31.26 & \cellcolor{red!20}24.78 \\\arrayrulecolor{black!30}\midrule	
Pose Noise  & 0.01\%		& \cellcolor{red!5}31.57 & \cellcolor{red!5}28.67 & \cellcolor{red!5}32.98 & \cellcolor{red!5}34.61 & \cellcolor{red!5}31.10 & \cellcolor{red!5}31.62 & \cellcolor{red!5}33.29 & \cellcolor{red!5}29.70 \\	
$\sigma_{\text{t}}$& 0.02\% 	& \cellcolor{red!10}29.51 & \cellcolor{red!10}26.93 & \cellcolor{red!10}30.85 & \cellcolor{red!10}32.46 & \cellcolor{red!10}28.61 & \cellcolor{red!10}29.74 & \cellcolor{red!10}31.63 & \cellcolor{red!10}28.17 \\ 	
			& 0.03\% 	& \cellcolor{red!15}28.60 & \cellcolor{red!15}26.27 & \cellcolor{red!15}29.87 & \cellcolor{red!15}31.03 & \cellcolor{red!15}27.44 & \cellcolor{red!15}28.65 & \cellcolor{red!15}30.34 & \cellcolor{red!15}27.47 \\
			& 0.04\% 	& \cellcolor{red!20}27.91 & \cellcolor{red!20}25.76 & \cellcolor{red!20}29.10 & \cellcolor{red!20}30.36 & \cellcolor{red!20}26.44 & \cellcolor{red!20}27.88 & \cellcolor{red!20}29.69 & \cellcolor{red!20}26.76 \\\arrayrulecolor{black!30}\midrule
Pose Noise & 0.1° 		& \cellcolor{red!5}28.69 & \cellcolor{red!5}26.76 & \cellcolor{red!5}29.96 & \cellcolor{red!5}31.54 & \cellcolor{red!5}27.91 & \cellcolor{red!5}29.14 & \cellcolor{red!5}30.60 & \cellcolor{red!5}27.57 \\
$\sigma_{\text{R}}$ & 0.2° 		& \cellcolor{red!10}27.12 & \cellcolor{red!10}25.31 & \cellcolor{red!10}28.47 & \cellcolor{red!10}29.62 & \cellcolor{red!10}25.74 & \cellcolor{red!10}27.24 & \cellcolor{red!10}28.55 & \cellcolor{red!10}26.16 \\	
		   & 0.3° 		& \cellcolor{red!15}26.09 & \cellcolor{red!15}24.56 & \cellcolor{red!15}27.62 & \cellcolor{red!15}27.88 & \cellcolor{red!15}25.10 & \cellcolor{red!15}26.41 & \cellcolor{red!15}28.33 & \cellcolor{red!15}25.60 \\ 	
		   & 0.4° 		& \cellcolor{red!20}25.64 & \cellcolor{red!20}23.78 & \cellcolor{red!20}27.34 & \cellcolor{red!20}27.50 & \cellcolor{red!20}24.40 & \cellcolor{red!20}25.53 & \cellcolor{red!20}27.09 & \cellcolor{red!20}25.19 \\\arrayrulecolor{black!30}\midrule
Pose Noise & 0.01\%,\,0.1° 	& \cellcolor{red!5}28.71 & \cellcolor{red!5}26.47 & \cellcolor{red!5}29.87 & \cellcolor{red!5}31.12 & \cellcolor{red!5}27.48 & \cellcolor{red!5}28.73 & \cellcolor{red!5}30.27 & \cellcolor{red!5}27.47 \\
$\sigma_{\text{t,R}}$ &0.02\%,\,0.2° 	& \cellcolor{red!10}27.01 & \cellcolor{red!10}24.90 & \cellcolor{red!10}28.29 & \cellcolor{red!10}29.37 & \cellcolor{red!10}25.80 & \cellcolor{red!10}26.85 & \cellcolor{red!10}28.67 & \cellcolor{red!10}26.23 \\ 	
 	&0.03\%,\,0.3° 	& \cellcolor{red!15}26.07 & \cellcolor{red!15}24.11 & \cellcolor{red!15}27.47 & \cellcolor{red!15}27.91 & \cellcolor{red!15}24.99 & \cellcolor{red!15}26.11 & \cellcolor{red!15}26.89	   & \cellcolor{red!15}25.46 \\
			&0.04\%,\,0.4° 	& \cellcolor{red!20}25.59 & \cellcolor{red!20}23.87 & \cellcolor{red!20}27.03 & \cellcolor{red!20}27.16 & \cellcolor{red!20}24.46 & \cellcolor{red!20}25.15 & \cellcolor{red!20}25.94 & \cellcolor{red!20}25.10 \\
\arrayrulecolor{black}\bottomrule
\end{tabularx}
\end{table}

\begin{table}[H]
\caption{\textbf{Mean PSNR}$\uparrow$ in \si{\dB} on the DTU dataset per NeRF-Ensemble.}
\label{tab:PSNR_dtu}
\begin{tabularx}{\textwidth}{@{} l *{1}{>{\hsize=1.5\hsize}X} *{7}{X} @{}}
\toprule
 & & scan24 & scan37 & scan40 & scan55 & scan63 & scan114 \\\midrule
Original			 & 				& \cellcolor{green!25}30.27& \cellcolor{green!25}28.31& \cellcolor{green!25}28.22& \cellcolor{green!25}29.17& \cellcolor{green!25}36.41&\cellcolor{green!25}32.31  \\	\arrayrulecolor{black!30}\midrule	 
Image Noise & 15  			& 24.54 & 24.34 & 24.11&  23.79& 26.39& 24.49\\
$\sigma_{\text{Im}}$ & & & & & &\\\arrayrulecolor{black!30}\midrule
Pose Noise  & 0.01\,mm			& 26.83&27.08 & 26.25& 26.16& 35.33&29.72\\	
$\sigma_{\text{t}}$ &	& & & & & &\\ 	\arrayrulecolor{black!30}\midrule
Pose Noise & 0.1° 				& 27.01& 27.03& 26.42& 25.86&35.33 &29.89 \\
$\sigma_{\text{R}}$ & & & & & & & \\\arrayrulecolor{black!30}\midrule
Pose Noise & 0.01\,mm,\,0.1° & 27.05& 26.97& 26.26& 25.36& 35.60 & 29.88\\
$\sigma_{\text{t,R}}$& & & & & & &\\ 	
\arrayrulecolor{black}\bottomrule
\end{tabularx}
\end{table}

\subsubsection{\textbf{Density Uncertainty}}\label{sec:results_density_uncertainty}
\paragraph{Quantitative}

\begin{figure}[H]
	\centering
\subfigure[]{\label{fig:spiderchart_synthetic}
	\includegraphics[width=0.35\linewidth]{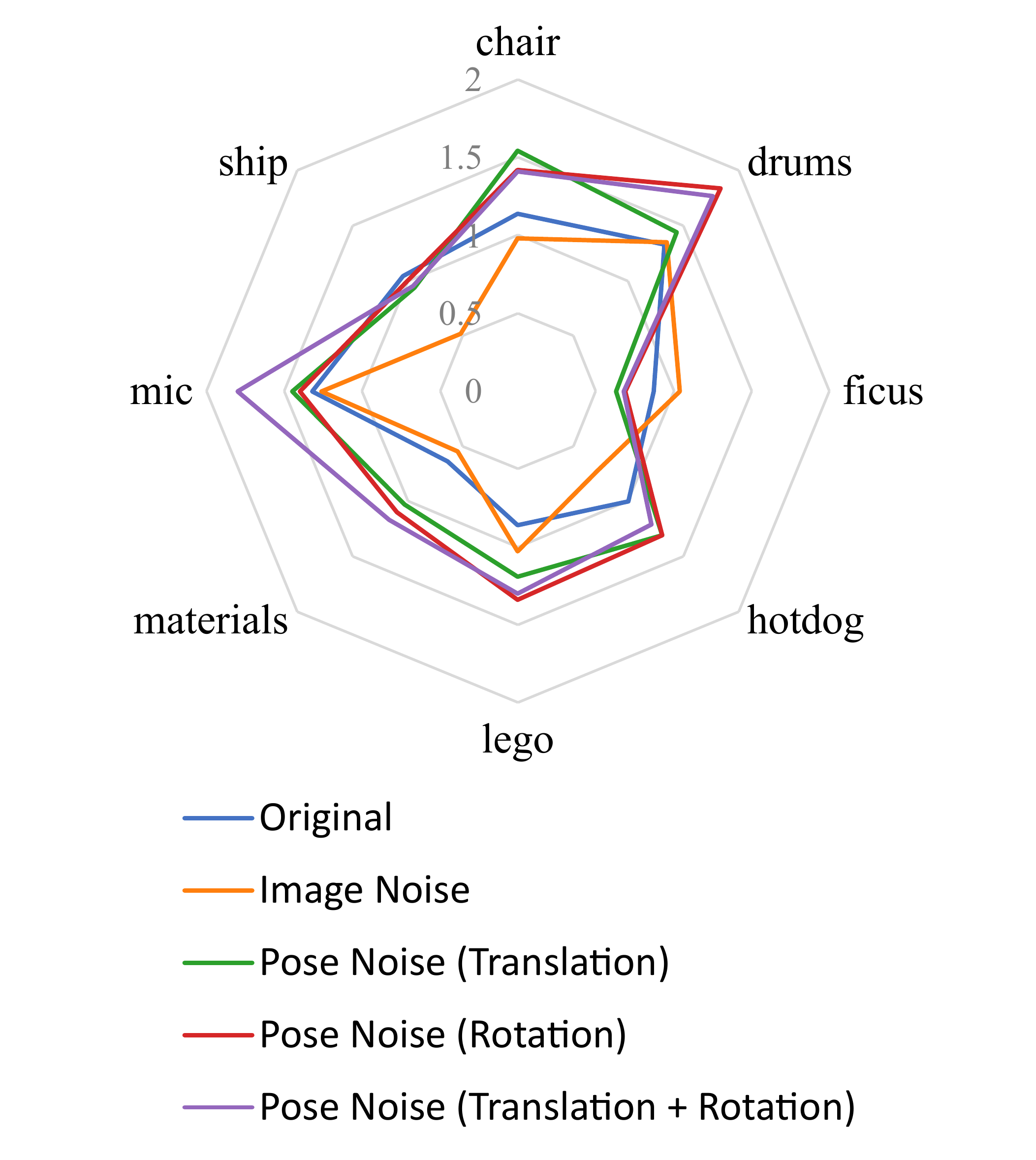}} 
\subfigure[]{\label{fig:spider__dtu}  
     \includegraphics[width=0.35\linewidth]{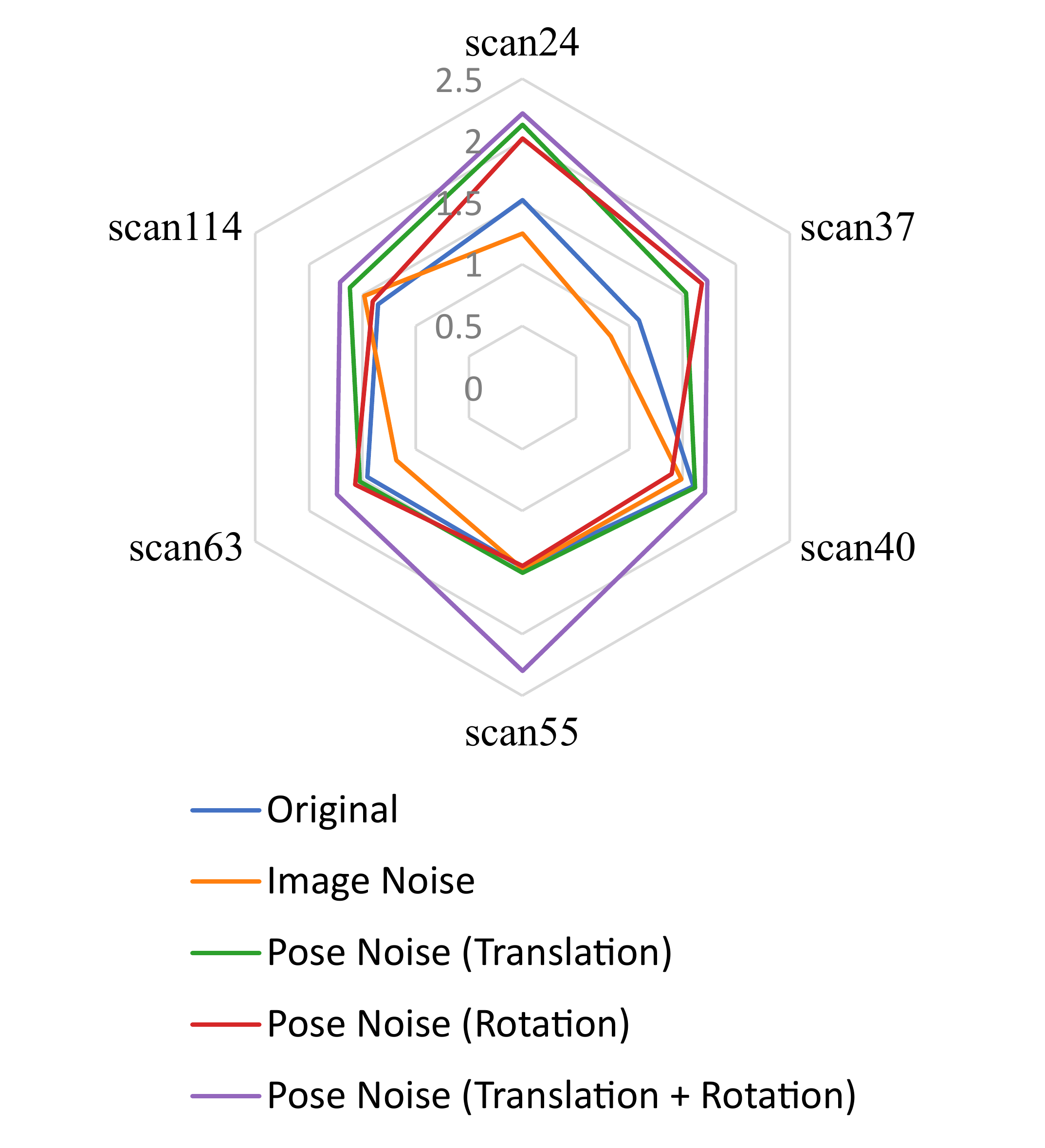}}
     \vspace{-3mm}
	\caption{Spider chart showing the density uncertainty for NeRF synthetic (left) and DTU dataset (right). Displayed are the mean density uncertainty \text{$\text{mU}_{\delta}$} in the 3D grid for different types of input data: original, image noise $\sigma_{\text{Im}}$, pose noise (translation) $\sigma_{\text{t}}$, pose noise (rotation) $\sigma_{\text{R}}$ and pose noise (translation + rotation) $\sigma_{\text{R,t}}$.}
\label{fig:spider}
\end{figure}

The quantitative analysis of the training process using the PSNR has shown a decrease in the quality of the rendered images with increasing data constraints from additive noise on input data. The mean density uncertainty \text{$\text{mU}_{\delta}$} in the 3D grid results show a similar behavior to the PSNR results for different input data configurations.
For the synthetic NeRF dataset as well as DTU dataset in Tables \ref{tab:Uncertainty_neu} and \ref{tab:Uncertainty_dtu}, an increase in density uncertainty with increasing noise is noticeable for almost all scenes. It is striking that although noise on the images leads to a lower PSNR, the density uncertainty does not necessarily increase rapidly. Another noticeable aspect in addition to the increase of the density uncertainty from data constraints is the significant decrease of the mean density \text{m$\overline{\delta}$} in the 3D grid for each scene. \newline
To illustrate the impact of noise on the scenes, Figure \ref{fig:spider} shows the spider charts for the highest noise in each scene. This highlights the overall trend that almost all scenes in the synthetic dataset for non-noisy input data range from 0 to 1. With increasing noise, the uncertainty increases from 1 to 2. For the DTU dataset, results are more balanced across the scenes. A clear trend is evident between the density uncertainty of noisy and original data without noise. \newline

\begin{figure}[h]
	\centering
\subfigure[]{\label{fig:histogramm_chair_b}
	\includegraphics[width=0.475\linewidth]{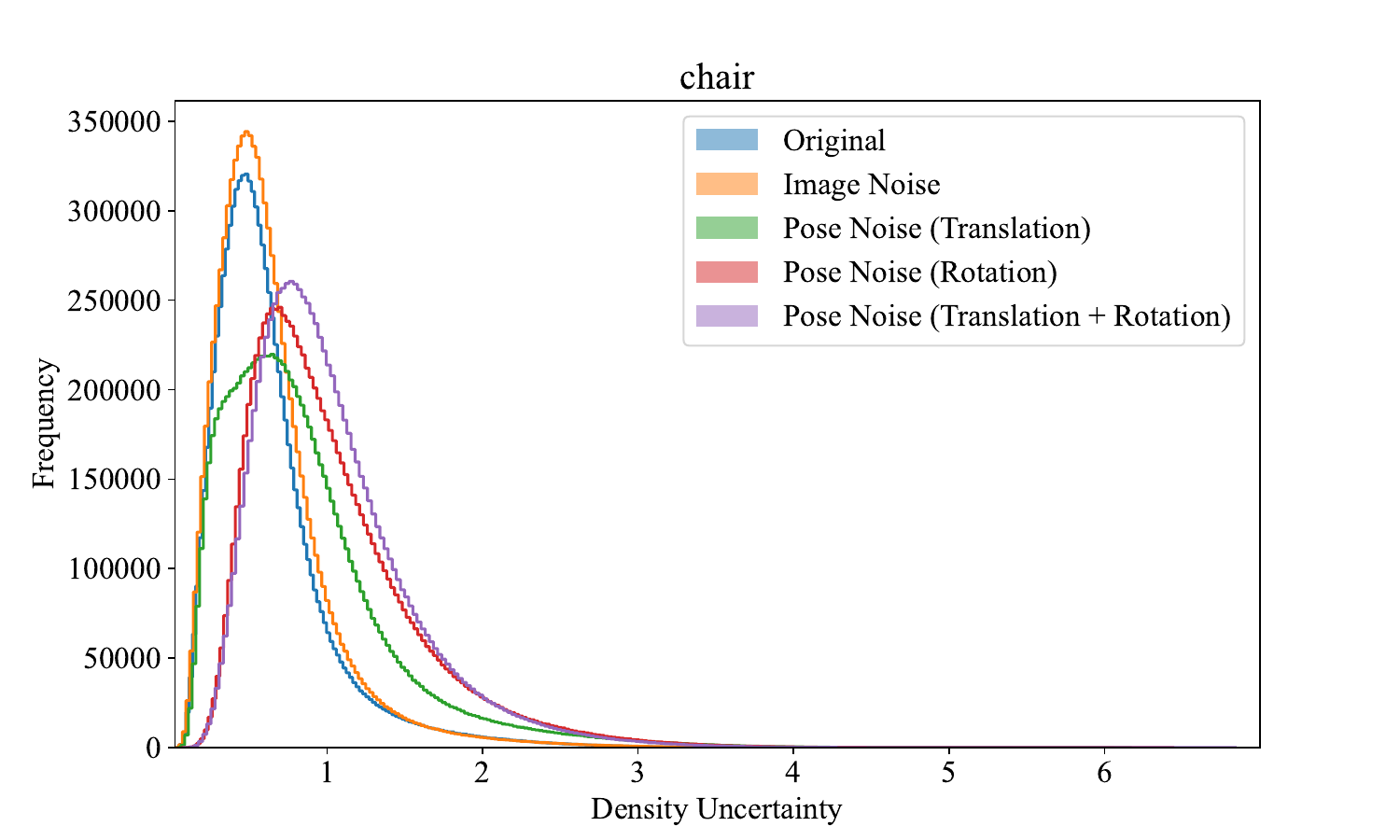}} 
\hspace{-1cm}
\subfigure[]{\label{fig:histogramm_lego_b}  
     \includegraphics[width=0.475\linewidth]{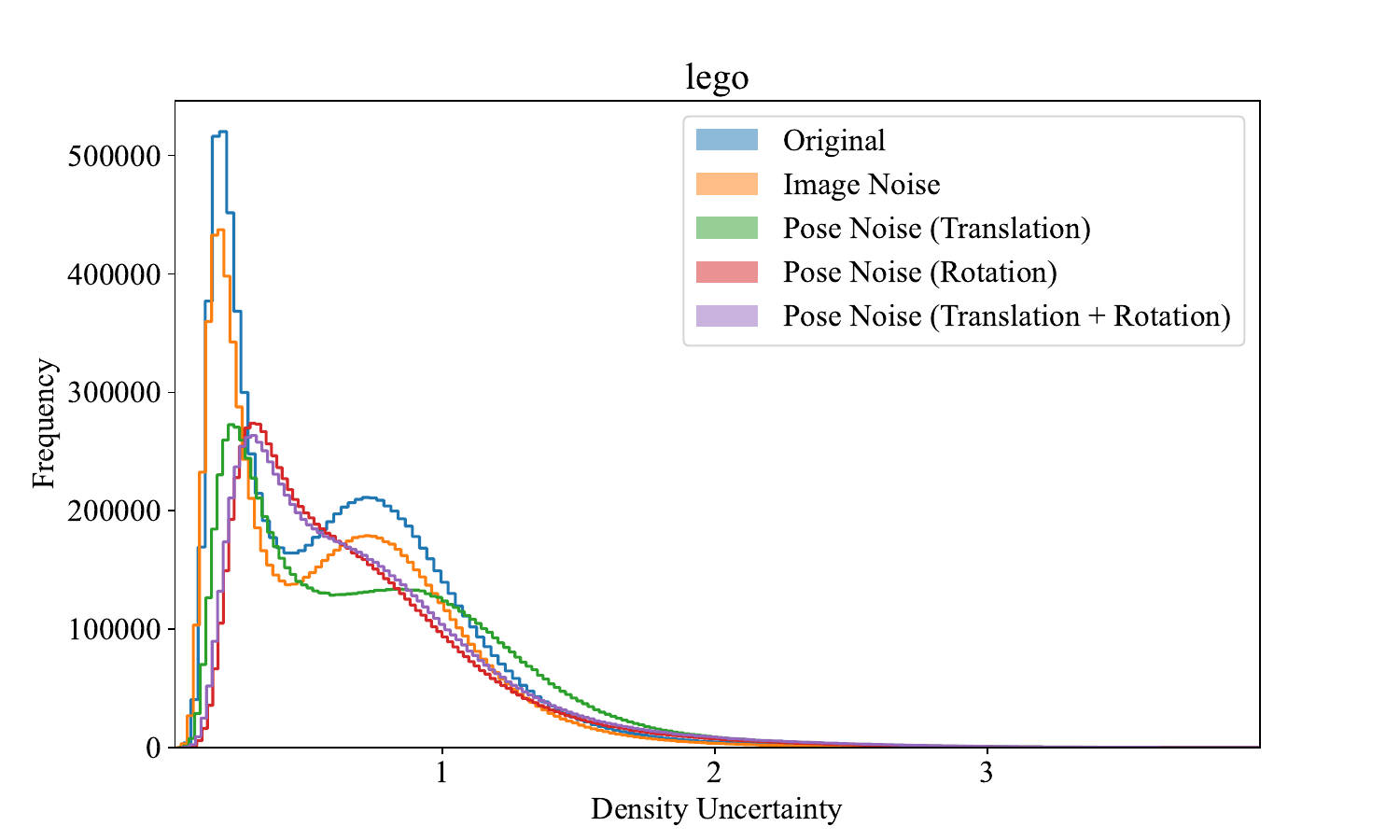}}
\vspace{-0.3cm}
	\caption{Histograms displaying the frequency of the density uncertainty \text{$\text{U}_{\delta}$} of all points in the 3D grid for the NeRF synthetic dataset. Shown are the scenes under different input data types: Original, image noise $\sigma_{\text{Im}}$, pose noise $\sigma_{\text{t}}$ (translation), pose noise $\sigma_{\text{R}}$ (rotation) and pose noise $\sigma_{\text{t,R}}$ (translation + rotation).}
\label{fig:histograms_synthetic}
\end{figure}

\begin{figure}[h]
	\centering

\subfigure[]{\label{fig:histogramm_40_b}  
     \includegraphics[width=0.475\linewidth]{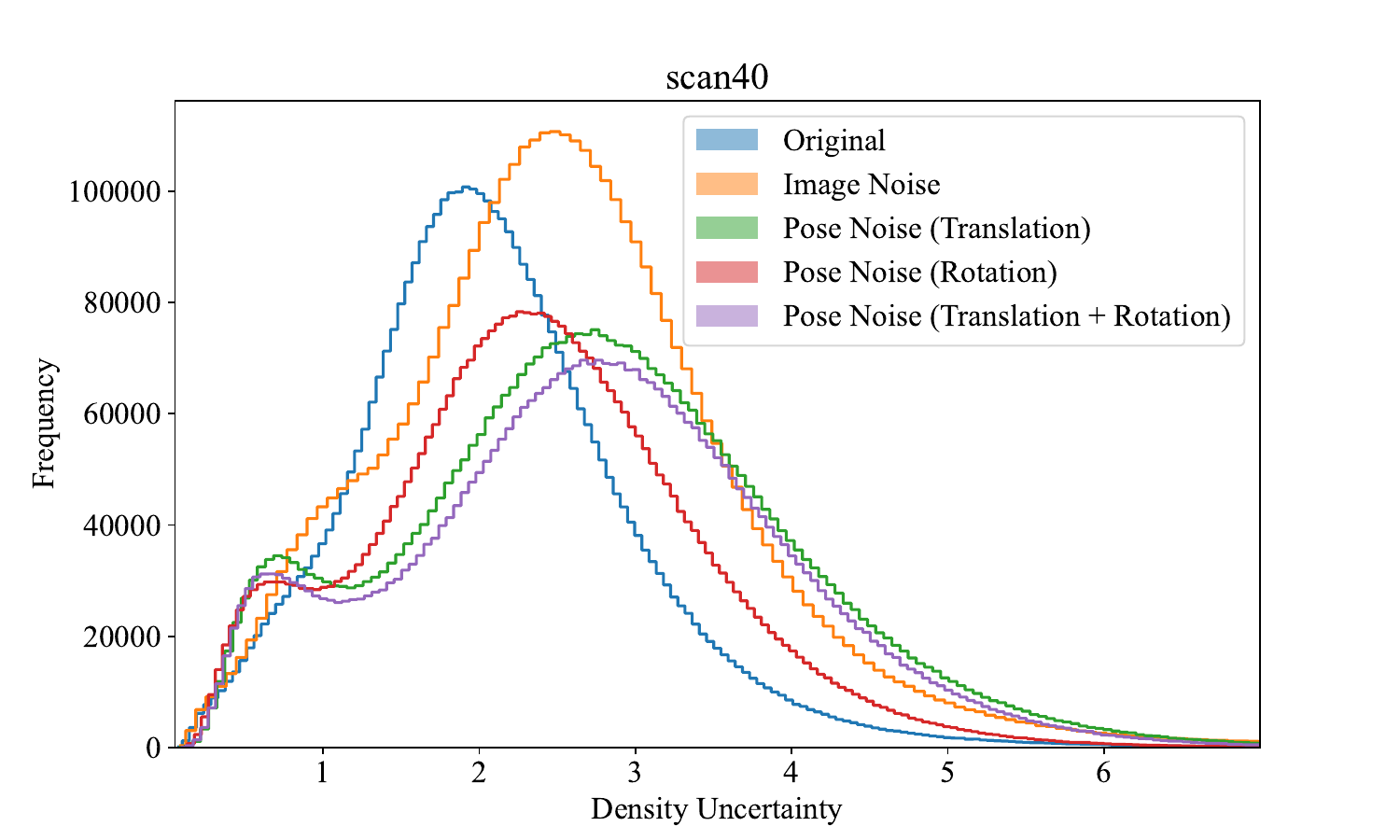}}
\hspace{-1cm}
\subfigure[]{\label{fig:histogramm_114_b}  
     \includegraphics[width=0.475\linewidth]{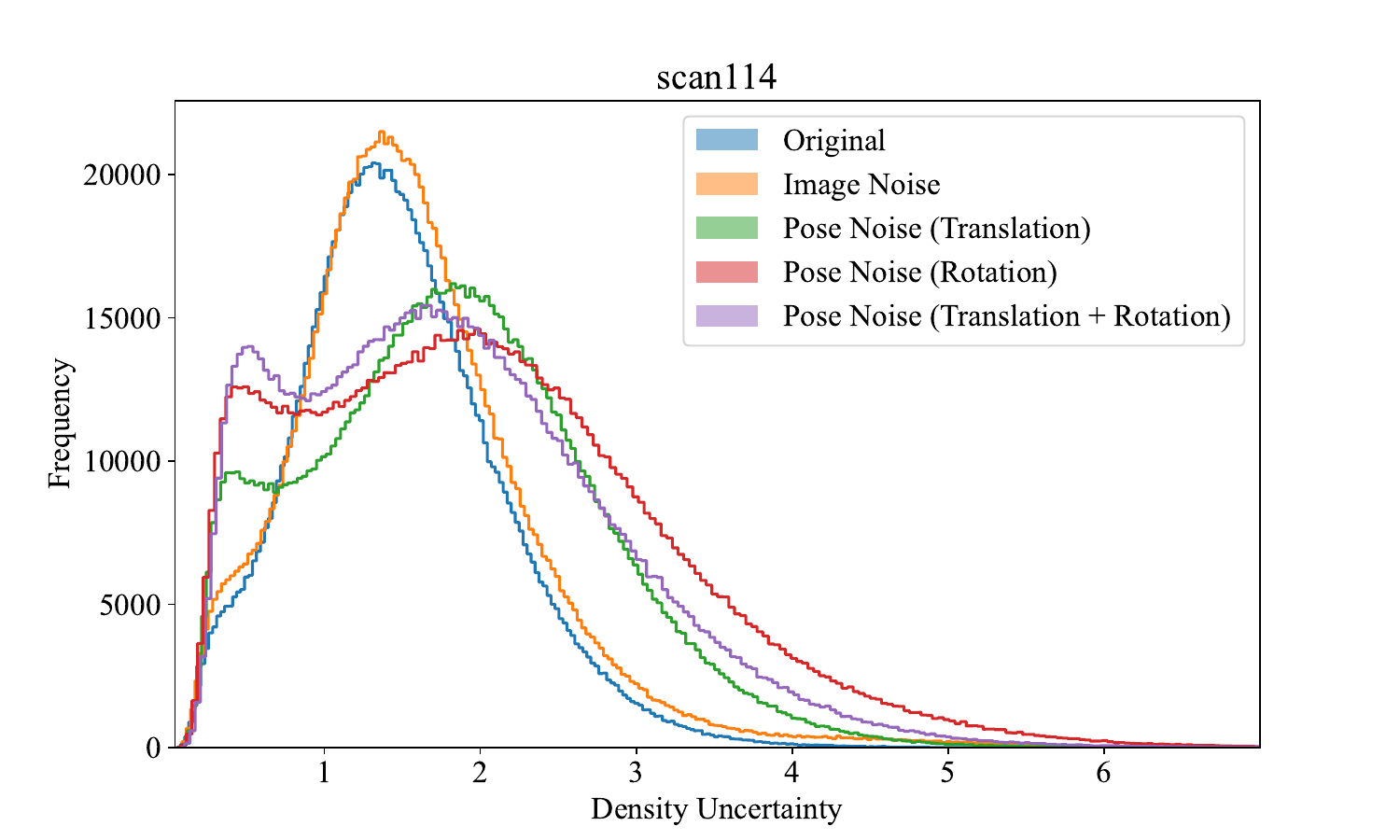}}
     \vspace{-0.3cm}
	\caption{Histograms displaying the frequency of the density uncertainty \text{$\text{U}_{\delta}$} of all points in the 3D grid for the DTU dataset. Shown are the scenes under different input data types: Original, image noise $\sigma_{\text{Im}}$, pose noise $\sigma_{\text{t}}$ (translation), pose noise $\sigma_{\text{R}}$ (rotation) and pose noise $\sigma_{\text{t,R}}$ (translation + rotation).}
\label{fig:histograms_dtu}
\end{figure}

The additional visualization of histograms for the entire discretized 3D space in the form of points in the 3D grid shows the distribution of density uncertainty by data constraints.
The histograms in Figures \ref{fig:histograms_synthetic} and \ref{fig:histograms_dtu} illustrate the overall frequency distribution of the density uncertainties in the grid. For the scenes of the NeRF synthetic dataset, it shows that the uncertainty of the density values of the grid are increasingly low for non-noisy data. The values then shift towards higher uncertainty with noise. In addition, the combination of pose noise in the translation and rotation components leads to a further shift towards higher uncertainties. 
The scenes of the DTU dataset show similar results. Here, peaks with low uncertainties occur more frequently in non-noisy data.

\vspace{-0.25cm}

\begin{table}[H]
\caption{Mean \textbf{density uncertainty} \text{$\text{mU}_{\delta}$} and mean \textbf{mean density} \text{m$\overline{\delta}$} in brackets per NeRF-Ensemble in the discretized 3D grid for the NeRF synthetic dataset.}
\label{tab:Uncertainty_neu}
\begin{tabularx}{\textwidth}{@{} l *{1}{>{\hsize=1.5\hsize}X} *{9}{X} @{}}
\toprule
 & & chair & drums & ficus & hotdog & lego & materials & mic & ship  \\
\midrule

Original			 &		& 1.14 (1355.76) & 1.33 (960.98) & 0.87 (273.40) & 1.00 (6464.55) & 0.86 (1473.02) & 0.64 (277.65) & 1.32 (375.84)   & 1.04 (326.37)\\	\arrayrulecolor{black!30}\midrule 	 
Image Noise  & 5 			& 0.64 (1253.29) & 1.38 (719.84) & 0.93 (232.71) & 0.78 (1508.90) & 0.84 (1251.83) & 0.62 (202.08) & 1.23 (360.59) & 0.71 (260.58) \\
$\sigma_{\text{Im}}$    & 10 		& 1.13 (1183.46) & 1.37 (630.44) & 0.91 (206.77) & 0.70 (935.41)  & 0.90 (1088.10) & 0.61 (161.90) & 1.21 (364.08)   & 0.61 (219.51)\\
		     & 15 		& 1.13 (1076.47) & 1.35 (559.19) & 0.87 (206.90) & 0.72 (547.41)  & 0.96 (944.62)  & 0.54 (127.83) & 1.29 (318.88)   & 0.60 (156.06)\\
		    & 20 		& 0.98 (694.58)  & 1.35 (521.85) & 1.04 (274.78) & 0.72 (342.16)  & 1.03 (779.80)  & 0.55 (111.82) & 1.26 (289.47)   & 0.52 (88.69) \\  \arrayrulecolor{black!30}\midrule 
Pose Noise  	  & 0.01\%	& 1.57 (756.92)  & 1.44 (800.25) & 0.86 (228.00) & 1.07 (1116.22) & 1.10 (721.74)  & 0.75 (239.66) & 1.32 (323.07)   & 0.92 (248.41)\\
$\sigma_{\text{t}}$ & 0.02\% 	& 1.60 (472.92)  & 1.58 (620.29) & 0.81 (196.26) & 1.08 (581.12)  & 1.13 (354.94)  & 0.86 (198.60) & 1.31 (314.27)   & 0.90 (185.94) \\
			  & 0.03\% 	& 1.60 (369.09)  & 1.64 (473.94) & 0.68 (140.31) & 1.22 (348.64)  & 1.12 (259.16)  & 0.97 (195.23) & 1.43 (278.88)   & 0.88 (180.62) \\
			  & 0.04\% 	& 1.54 (294.23)  & 1.83 (395.66) & 0.63 (116.55) & 1.31 (321.44)  & 1.19 (226.63)  & 1.03 (180.30) & 1.45 (256.00)   & 0.94 (166.25) \\  \arrayrulecolor{black!30}\midrule
Pose Noise & 0.1° 		& 1.58 (391.66)  & 1.62 (513.99) & 0.67 (146.19) & 1.26 (439.23)  & 1.09 (276.89)  & 0.92 (197.37) & 1.38 (317.08) & 0.90 (177.62)\\
$\sigma_{\text{R}}$ & 0.2° 		& 1.06 (215.00)  & 1.66 (365.51) & 0.63 (100.87) & 1.28 (258.87)  & 1.23 (173.44)  & 1.05 (164.31) & 1.41 (234.44) & 0.94 (148.93) \\
		   & 0.3° 		& 1.41 (149.26)  & 1.95 (269.80) & 0.64 (86.97)  & 1.21 (170.35)  & 1.26 (144.43)  & 1.03 (132.70) & 1.40 (165.83) & 0.93 (113.11) \\
		   & 0.4° 		& 1.42 (132.26)  & 1.84 (223.46) & 0.69 (78.20)  & 1.31 (151.77)  & 1.34 (110.57)  & 1.10 (121.82) & 1.40 (142.94) &  1.00 (120.88) \\  \arrayrulecolor{black!30}\midrule  
Pose Noise 	& 0.01\%,\,0.1° 	& 1.51 (363.79)  & 1.60 (524.51) & 0.67 (152.87) & 1.31 (367.99)  & 1.14 (262.80)  & 0.92 (193.13) & 1.37 (268.45) & 0.86 (175.37)\\
$\sigma_{\text{t,R}}$ & 0.02\%,\,0.2° 	& 1.50 (228.43)  & 1.93 (336.29) & 0.63 (96.10)  & 1.24 (220.46)  & 1.20 (167.69)  & 1.09 (143.01) & 1.13 (212.43) & 0.89 (140.22) \\
            & 0.03\%,\,0.3° 	& 1.46 (163.45)  & 1.99 (249.97) & 0.62 (85.97)  & 1.30 (161.31)  & 1.30 (136.88)  & 1.10 (129.60) & 0.97 (136.218) & 0.93 (132.94) \\
			& 0.04\%,\,0.4°	& 1.41 (126.09)  & 1.77 (225.10) & 0.68 (77.69)  & 1.21 (132.76)  & 1.30 (117.13)  & 1.17 (128.76) & 1.87 (130.108) & 0.95 (107.65) \\
\bottomrule
\end{tabularx}
\end{table}

\begin{table}[H]
 \vspace{-0.3cm}
\caption{Mean \textbf{density uncertainty} \text{$\text{mU}_{\delta}$} per NeRF-Ensemble in the discretized 3D grid for the DTU dataset.}
\label{tab:Uncertainty_dtu}
\begin{tabularx}{\textwidth}{@{} l *{1}{>{\hsize=1.5\hsize}X} *{7}{X} @{}}
\toprule
 & & scan24 & scan37 & scan40 & scan55 & scan63 & scan114 \\
\midrule
Original 			 			  & & 1.52 & 1.09 & 1.60 & 1.46 & 1.45 & 1.35\\ \arrayrulecolor{black!30}\midrule
Image Noise & 15  				& 1.25 & 0.83 & 1.49 & 1.47 & 1.18 & 1.48 \\
$\sigma_{\text{Im}}$          & & & & & &\\ \arrayrulecolor{black!30}\midrule
Pose Noise  & 0.01\,mm			& 2.13 & 1.53 & 1.62 & 1.50 & 1.52 & 1.62 \\	
 $\sigma_{\text{t}}$ 					& & & & & &\\ 	\arrayrulecolor{black!30}\midrule
Pose Noise & 0.1° 				& 2.02 & 1.68 & 1.40 & 1.45 & 1.57 & 1.40 \\
$\sigma_{\text{R}}$					  & & & & & & & \\ \arrayrulecolor{black!30}\midrule
Pose Noise & 0.01\,mm,\,0.1° 	  	& 2.22 & 1.73 & 1.71 & 2.30 & 1.74 & 1.71 \\
$\sigma_{\text{t,R}}$				    & & & & & &\\
\arrayrulecolor{black}\bottomrule
\end{tabularx}
\end{table}

 \vspace{-0.6cm}


\paragraph{Qualitative}

\begin{figure}[H]
	\centering
\hspace{-0.25cm}
\raisebox{\dimexpr 0cm-\height}{NeRF}
\hspace{1cm}
\raisebox{\dimexpr 0cm-\height}{NeRF-Ensemble}
\hspace{0.8cm}
\raisebox{\dimexpr 0cm-\height}{Original}
\hspace{1.6cm}
\raisebox{\dimexpr 0cm-\height}{$\sigma_{\text{Im}}$}
\hspace{2.1cm}
\raisebox{\dimexpr 0cm-\height}{$\sigma_{\text{t}}$}
\hspace{2.1cm}
\raisebox{\dimexpr 0cm-\height}{$\sigma_{\text{R}}$}\\
	\vspace{3mm}
\rotatebox{90}{$\,\,\,\,\,\,\,\,\,\,\,\,\,\,\,\,$chair}
\subfigure{\label{fig:chair_member_}
	\includegraphics[width=0.13\linewidth]{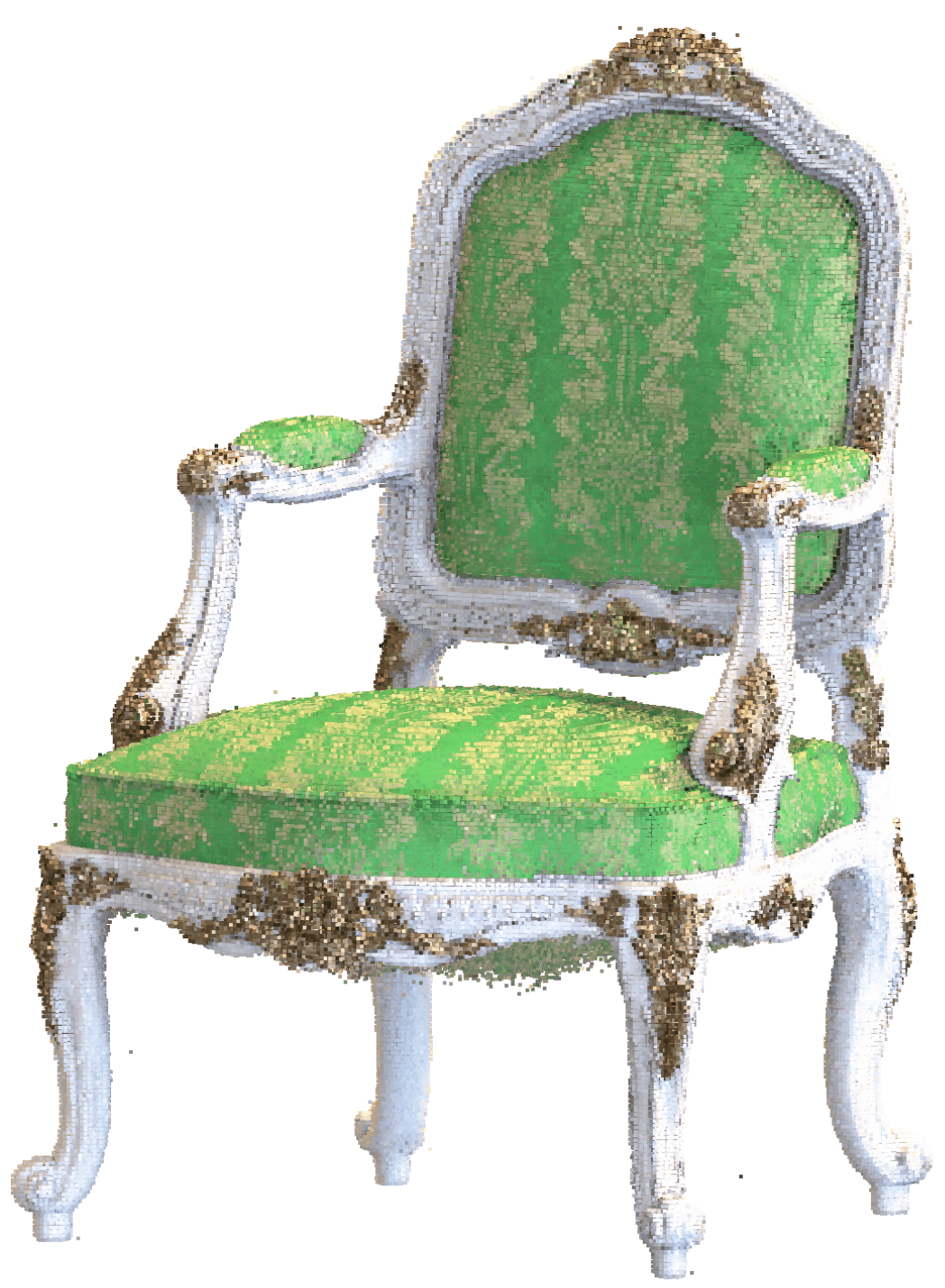}}
	\hspace{0.3cm}
\subfigure{\label{fig:chair_ensemble_}
     \includegraphics[width=0.13\linewidth]{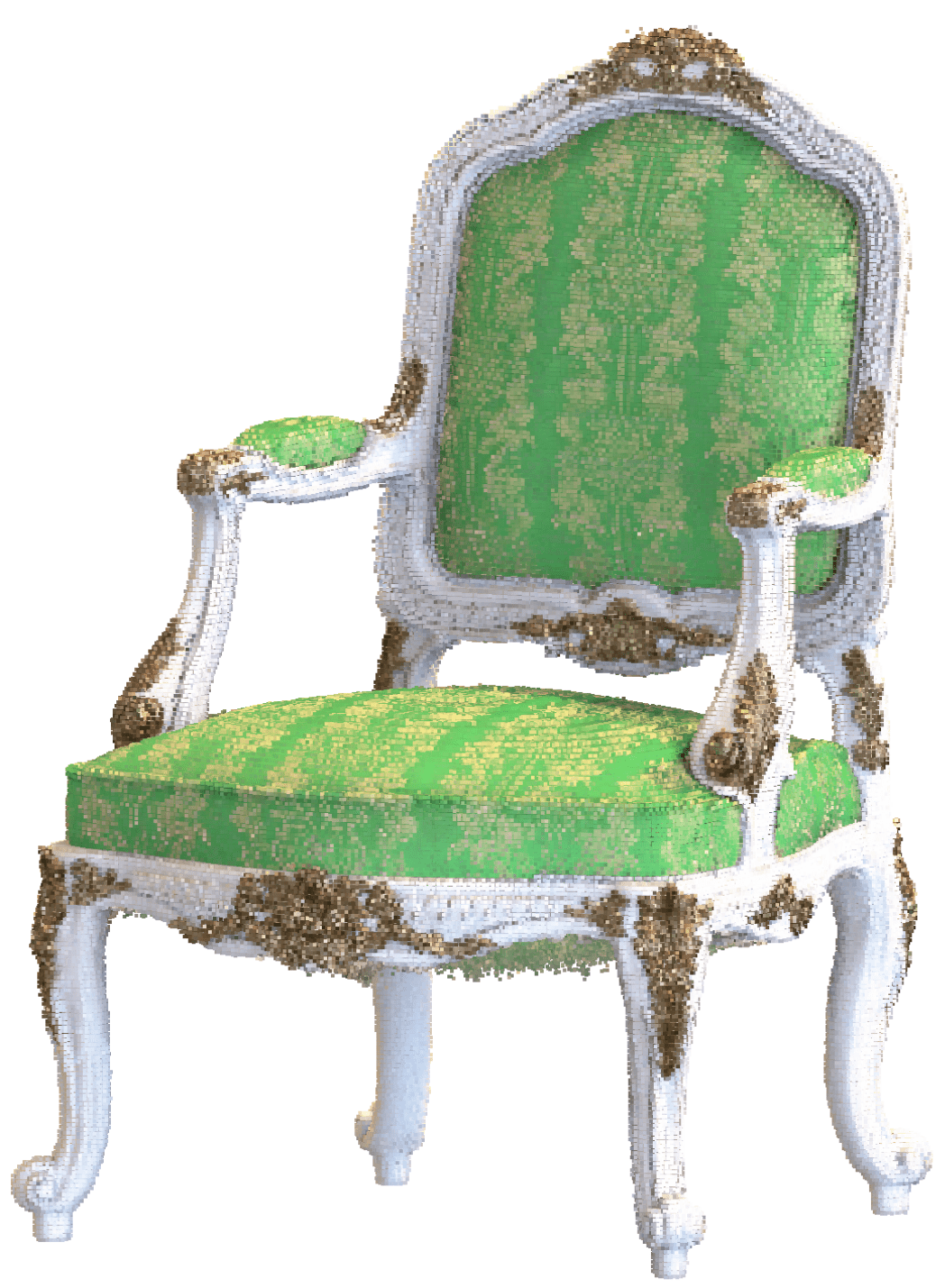}} 
     \hspace{0.3cm}
\subfigure{\label{fig:chair_unc_}  
     \includegraphics[width=0.13\linewidth]{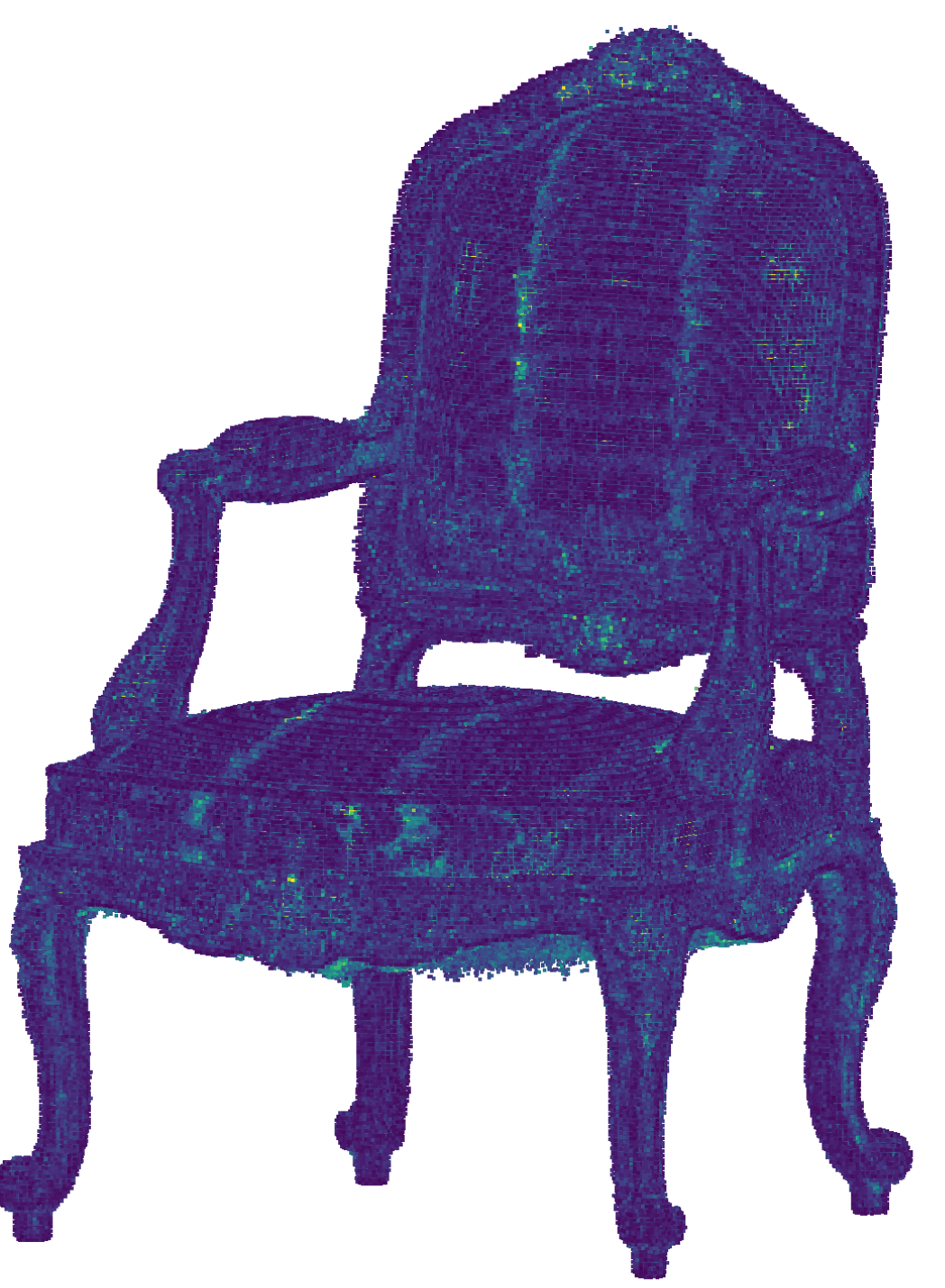}}
     \hspace{0.3cm}
\subfigure{\label{fig:chair_noise_image_unc_}
	\includegraphics[width=0.13\linewidth]{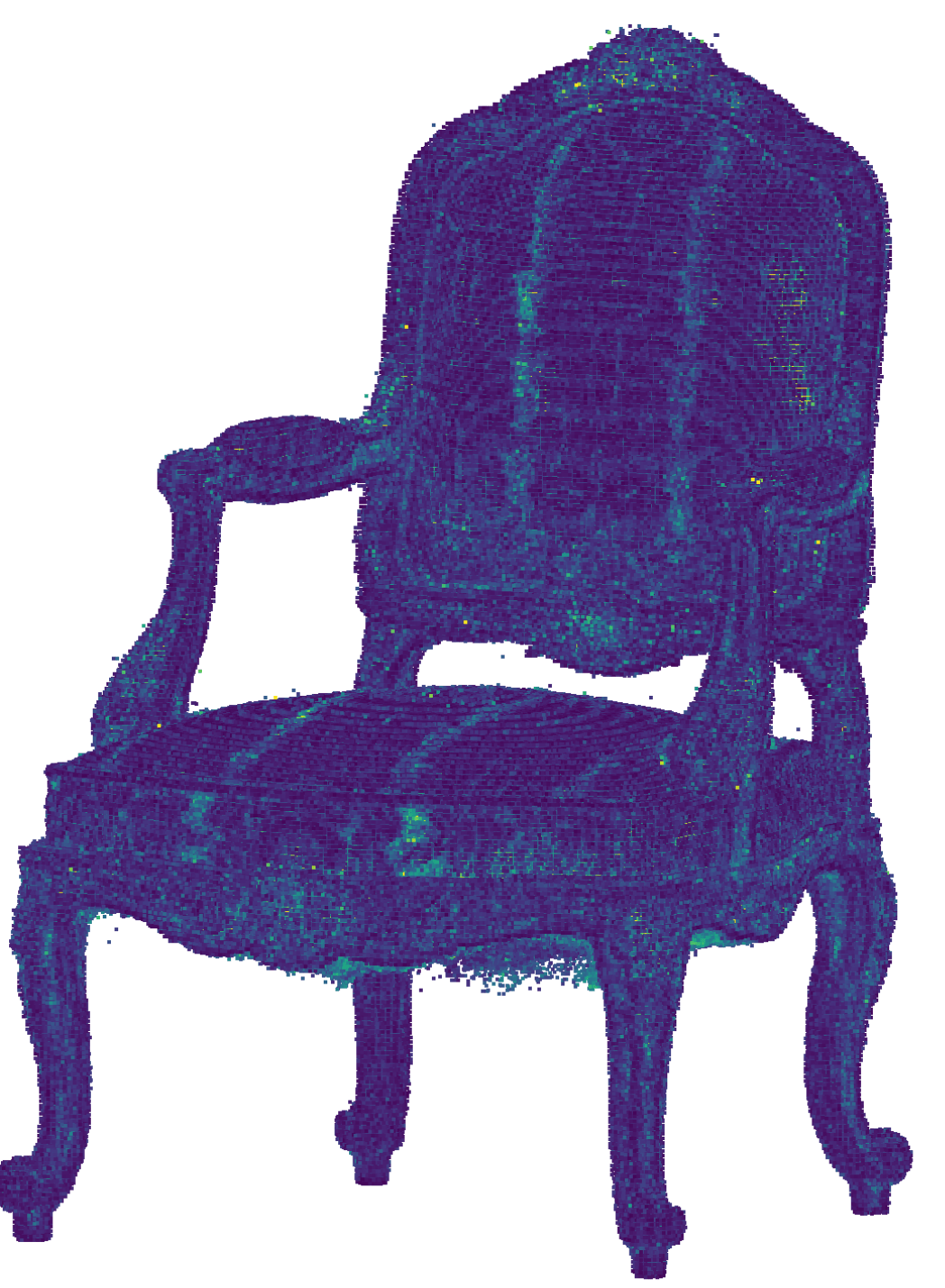}}
	\hspace{0.3cm}
\subfigure{\label{fig:chair_noise_trans_unc_}
	\includegraphics[width=0.13\linewidth]{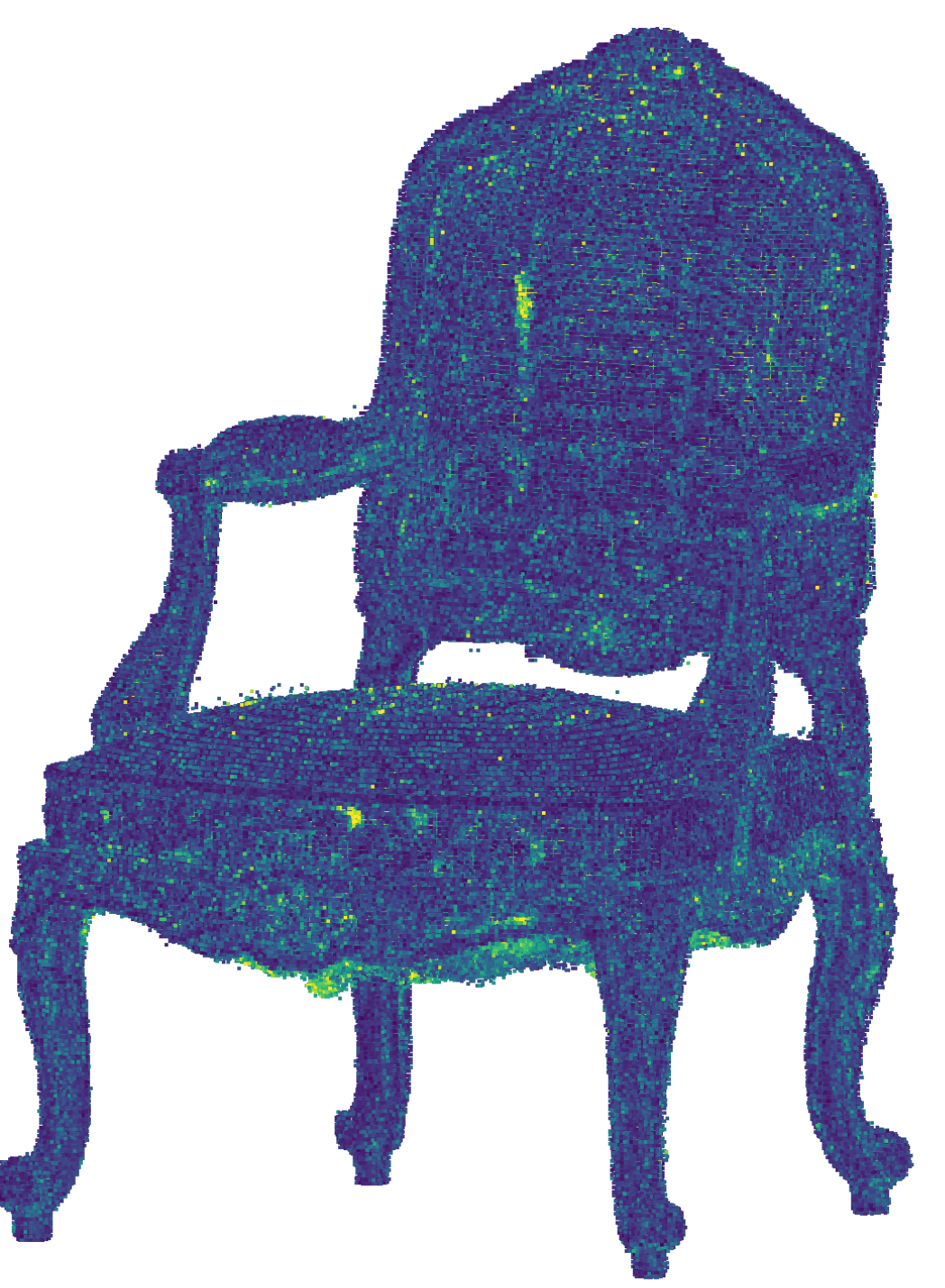}}
	\hspace{0.3cm}
\subfigure{\label{fig:chair_noise_rot_unc_}
	\includegraphics[width=0.13\linewidth]{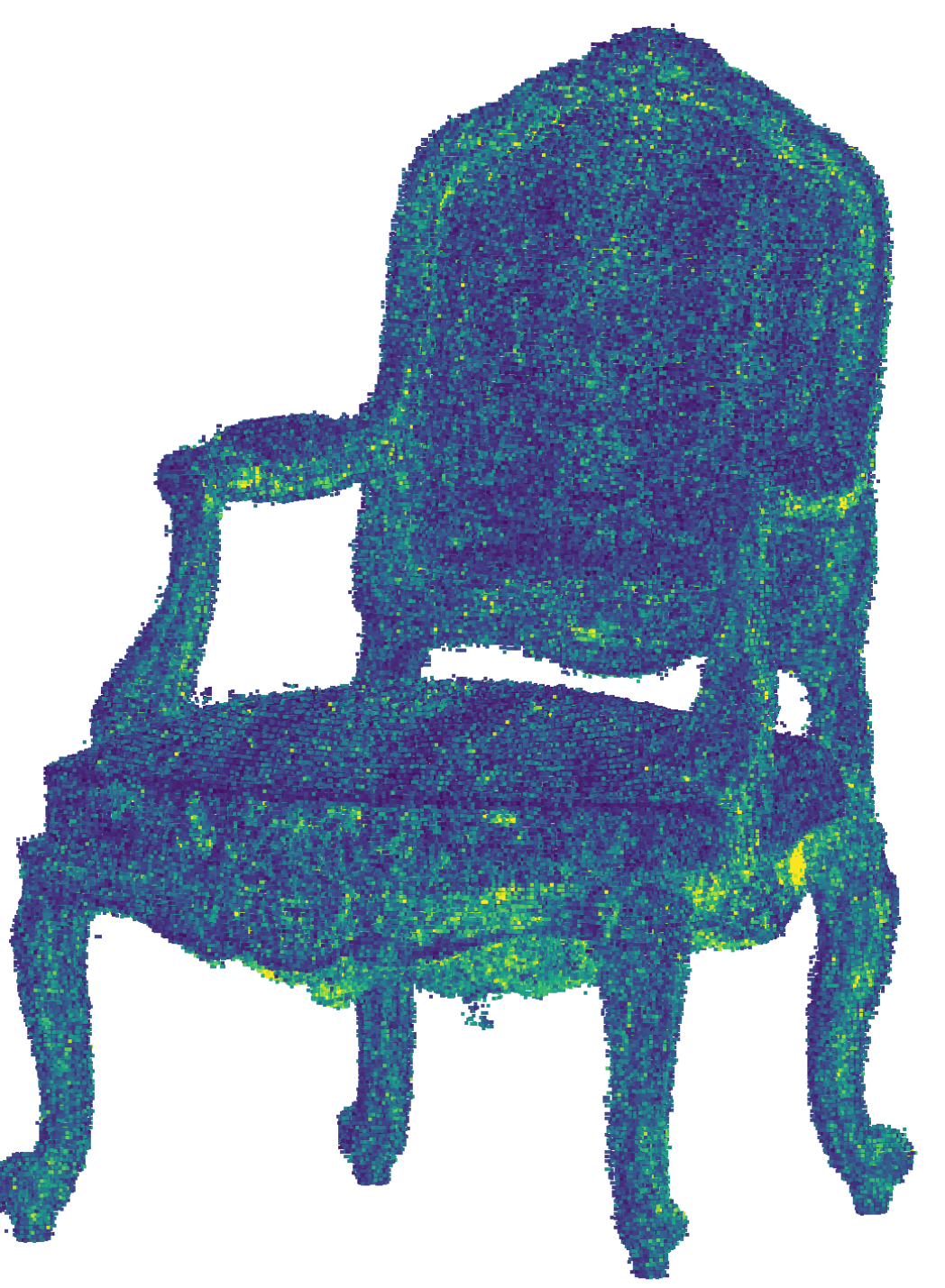}}\\
\rotatebox{90}{$\,\,\,\,\,\,\,\,\,\,\,\,\,\,\,\,$ficus}
\subfigure{\label{fig:ficus_member_}
	\includegraphics[width=0.11\linewidth]{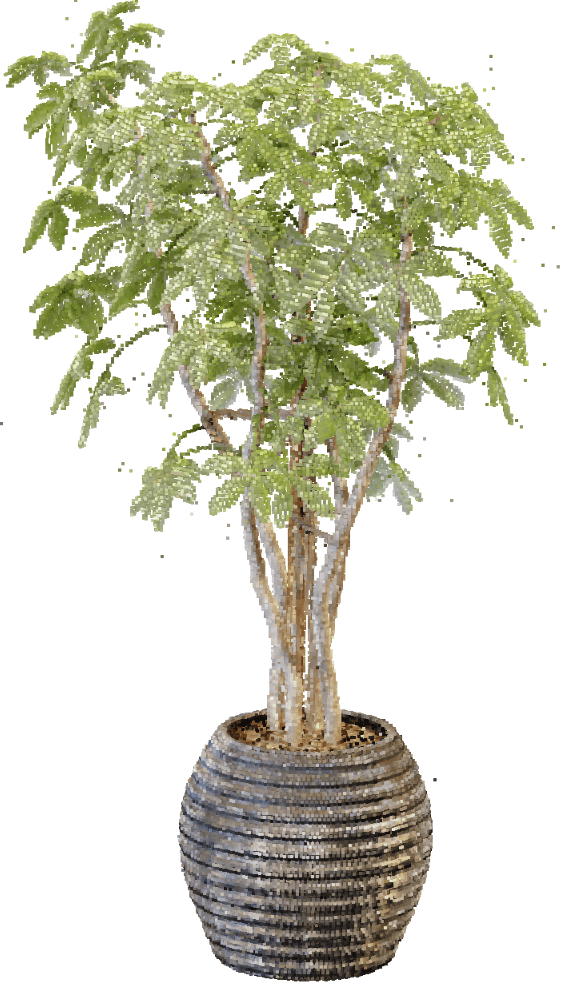}}
     \hspace{7.1mm}
\subfigure{\label{fig:ficus_ensemble_}  
     \includegraphics[width=0.11\linewidth]{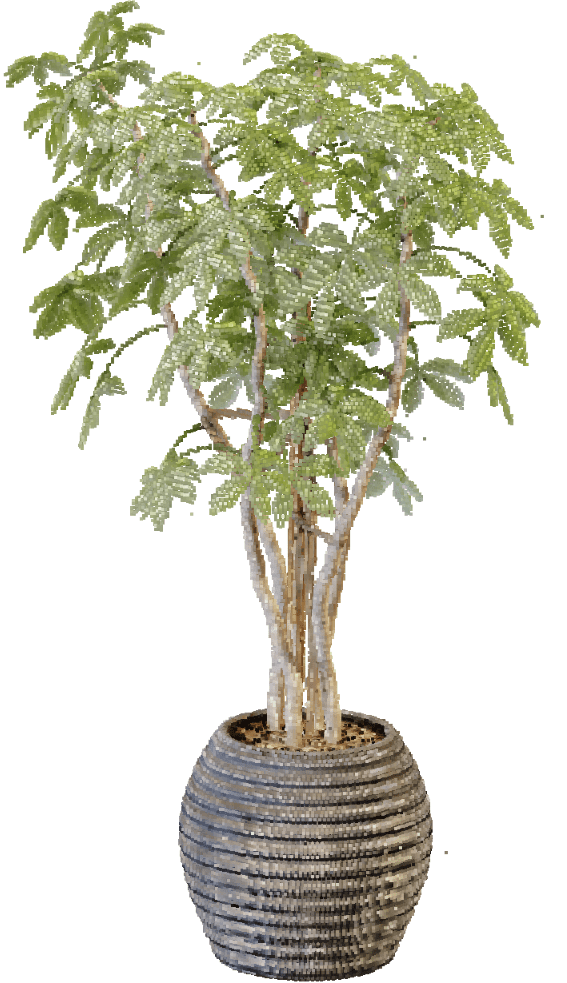}} 
    \hspace{7.1mm}
\subfigure{\label{fig:ficus_ensemble_unc_}  
     \includegraphics[width=0.11\linewidth]{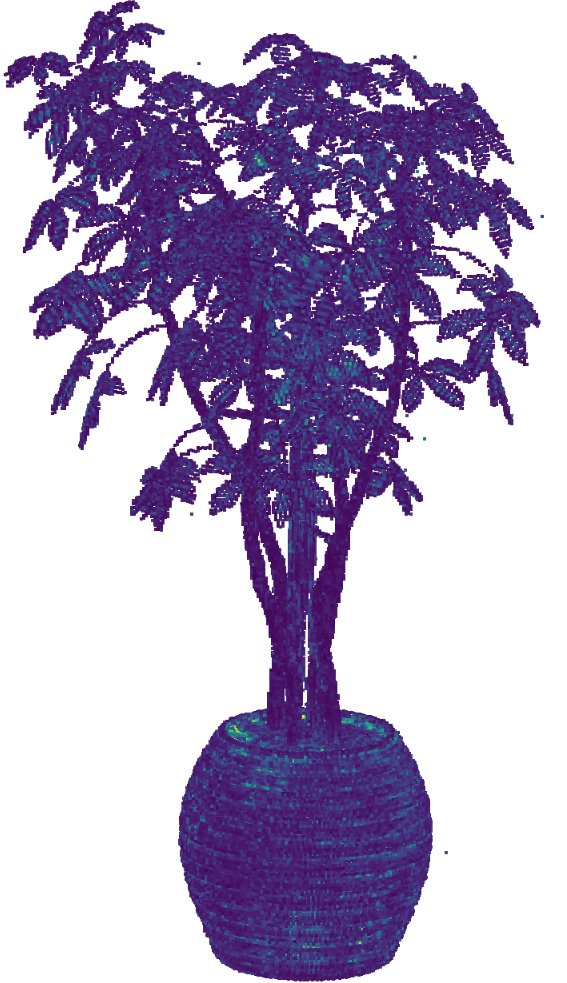}}
      \hspace{7.1mm}
\subfigure{\label{fig:ficus_ensemble_image_unc_}
	\includegraphics[width=0.11\linewidth]{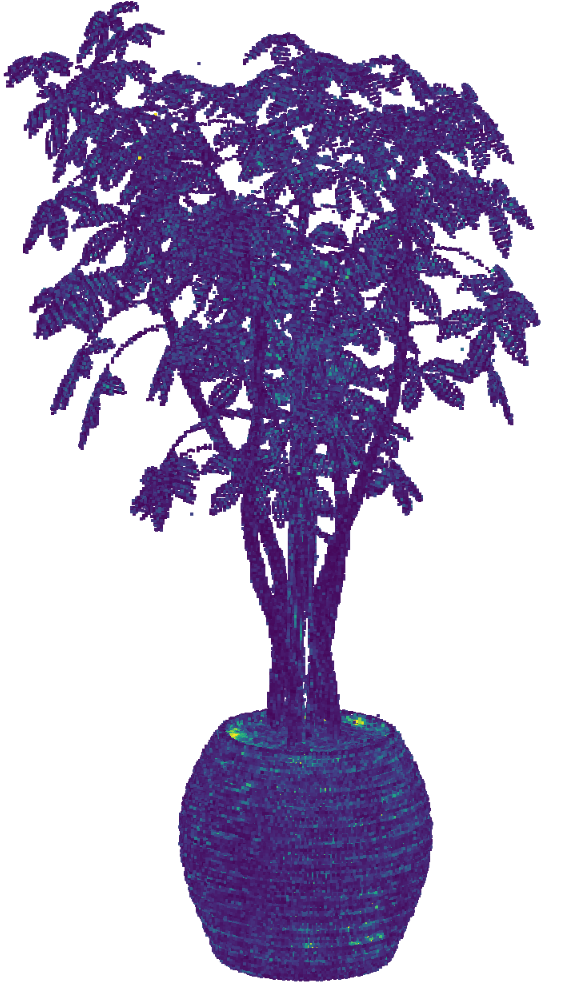}}
  \hspace{7.1mm}
\subfigure{\label{fig:ficus_ensemble_trans_unc_}
	\includegraphics[width=0.11\linewidth]{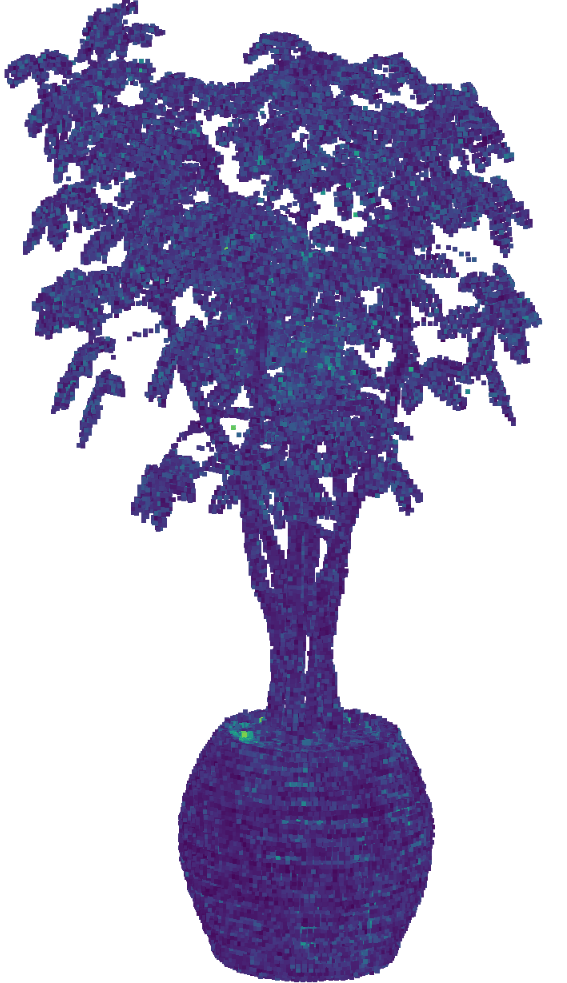}}
  \hspace{7.1mm}
\subfigure{\label{fig:ficus_ensemble_rot_unc_}
	\includegraphics[width=0.11\linewidth]{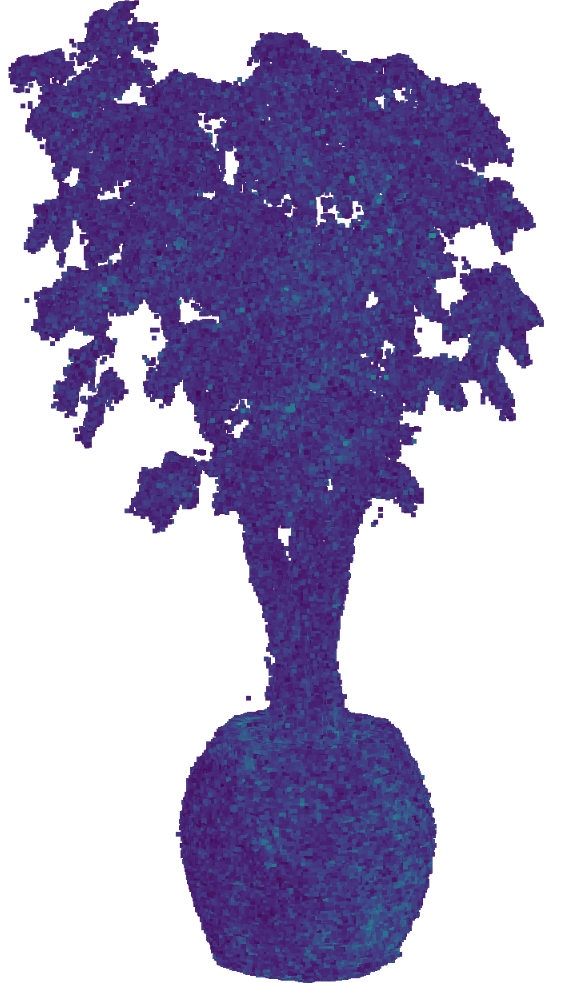}}\\
\rotatebox{90}{$\,\,\,\,\,\,\,\,\,\,\,\,\,\,\,\,$drums}
\subfigure{\label{fig:drums_member_}
	\includegraphics[width=0.15\linewidth]{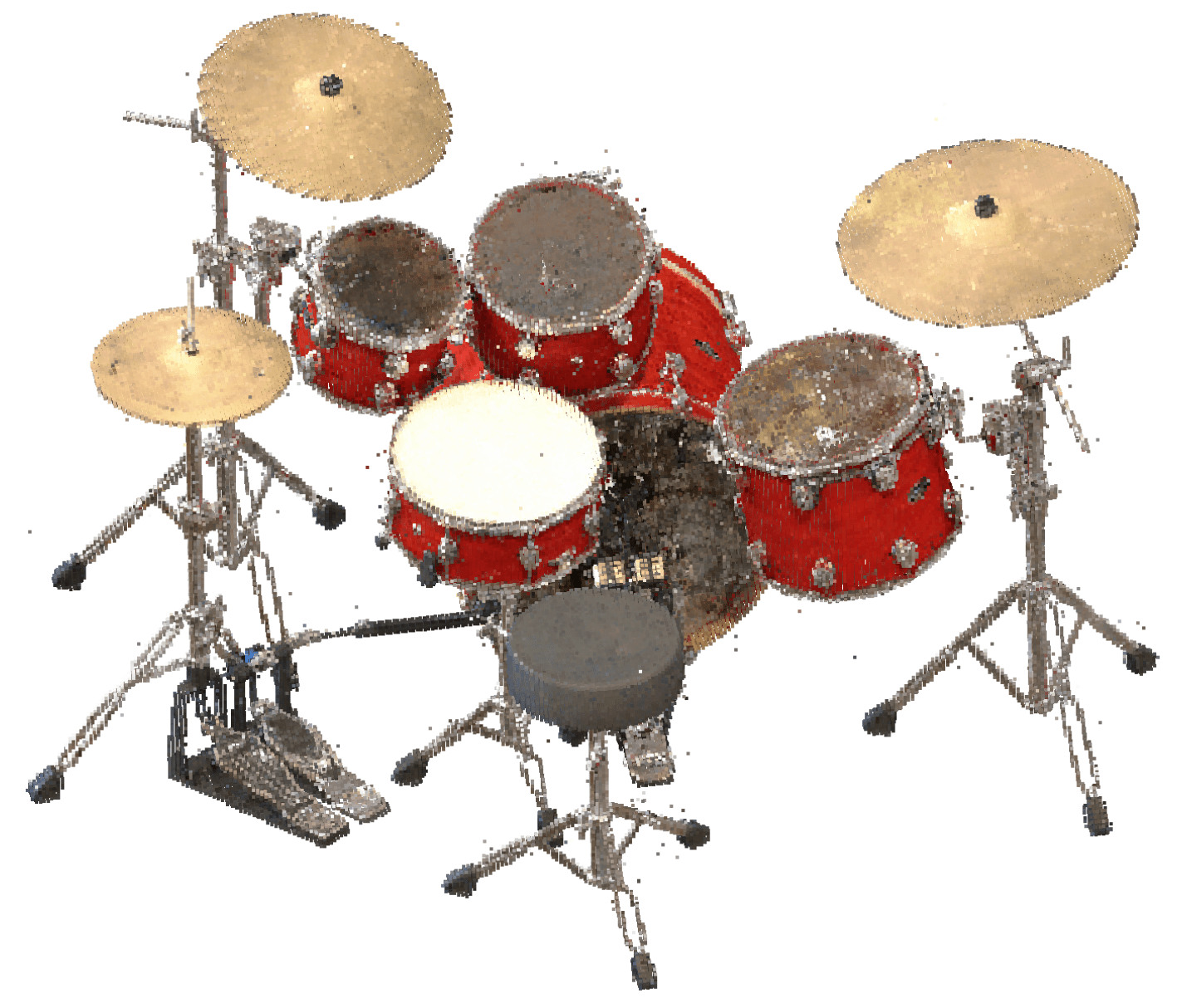}}
\subfigure{\label{fig:drums_ensemble_}  
     \includegraphics[width=0.15\linewidth]{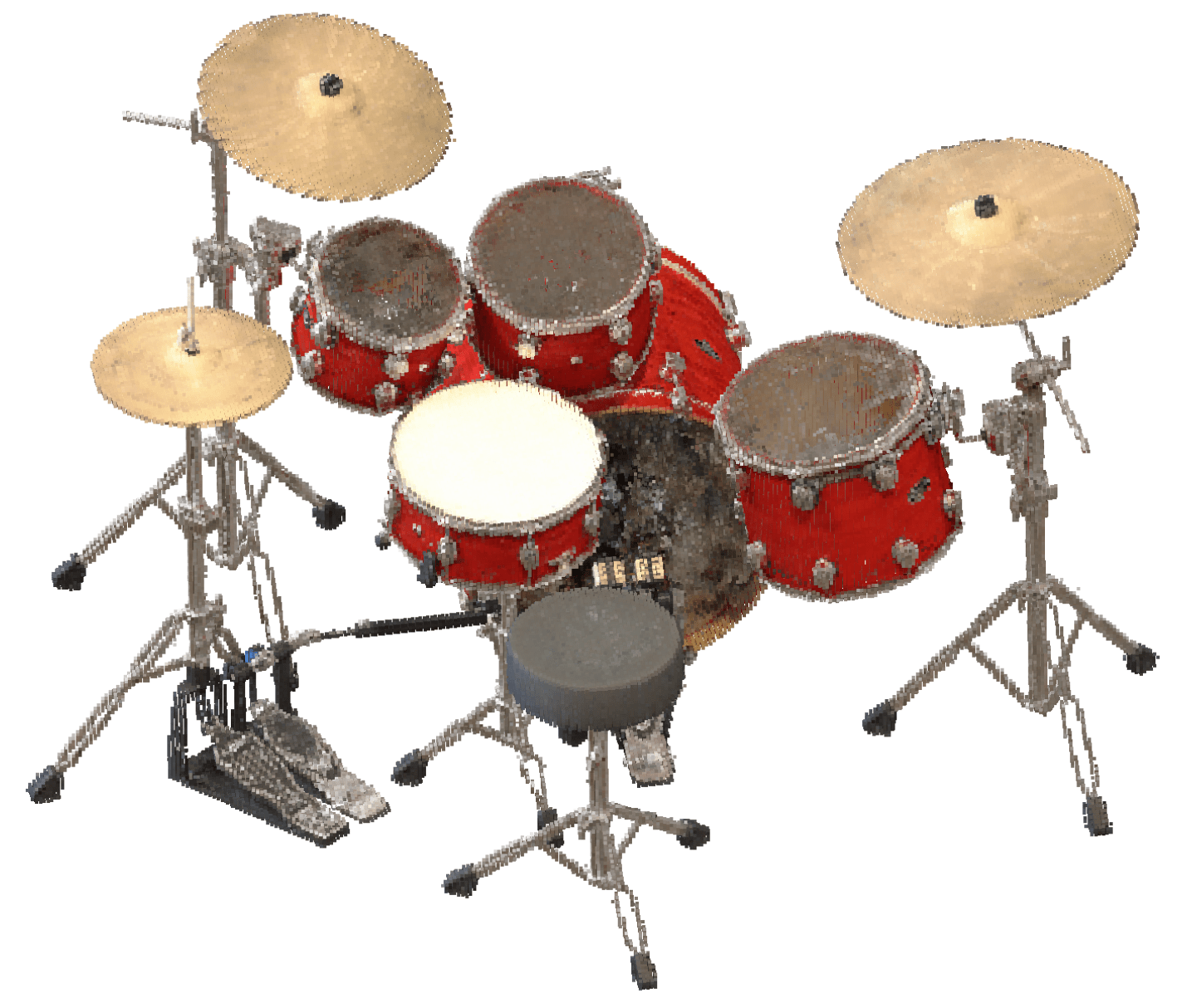}} 
\subfigure{\label{fig:drums_unc_}  
     \includegraphics[width=0.15\linewidth]{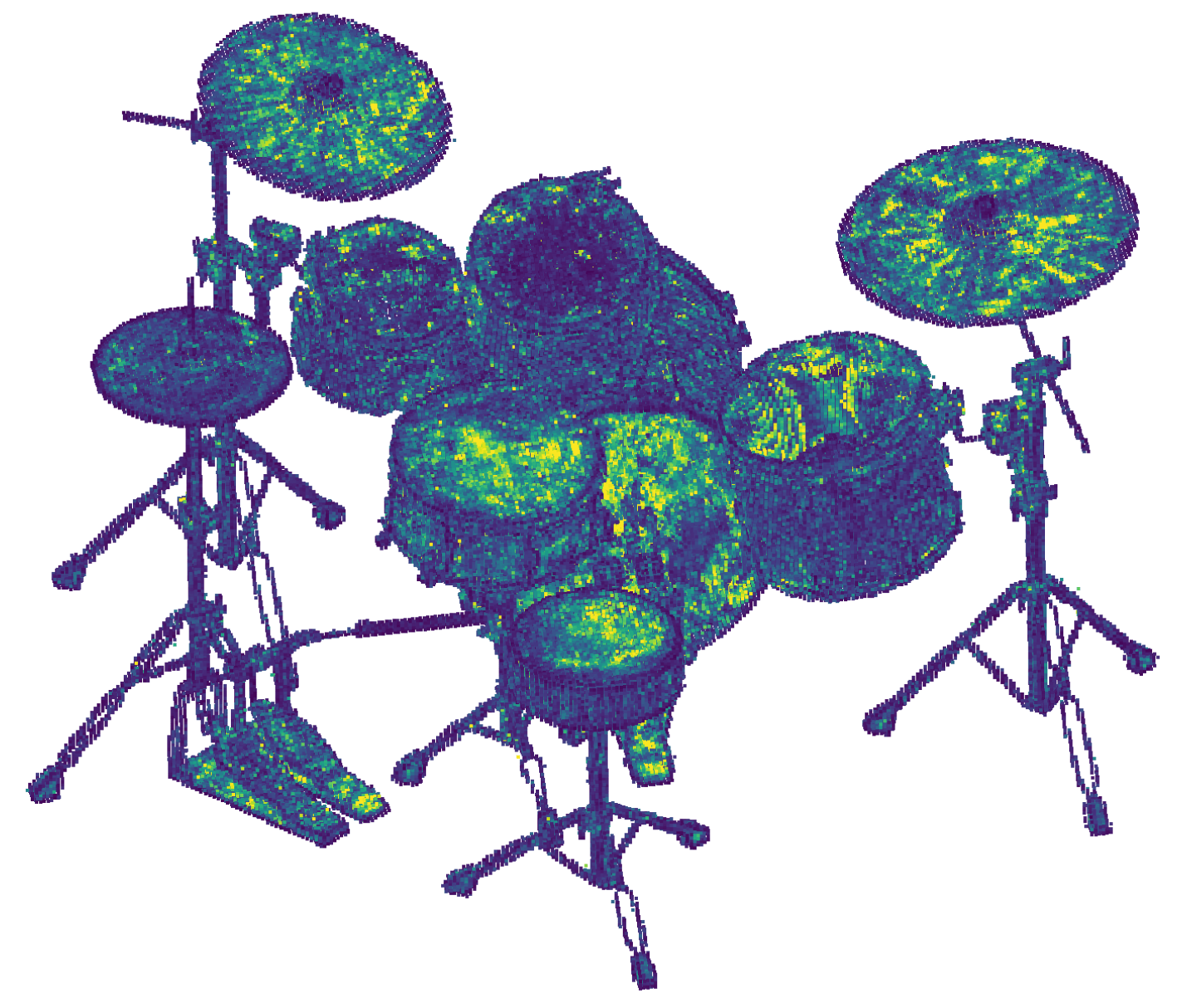}}
\subfigure{\label{fig:drums_noise_image_unc_}
	\includegraphics[width=0.15\linewidth]{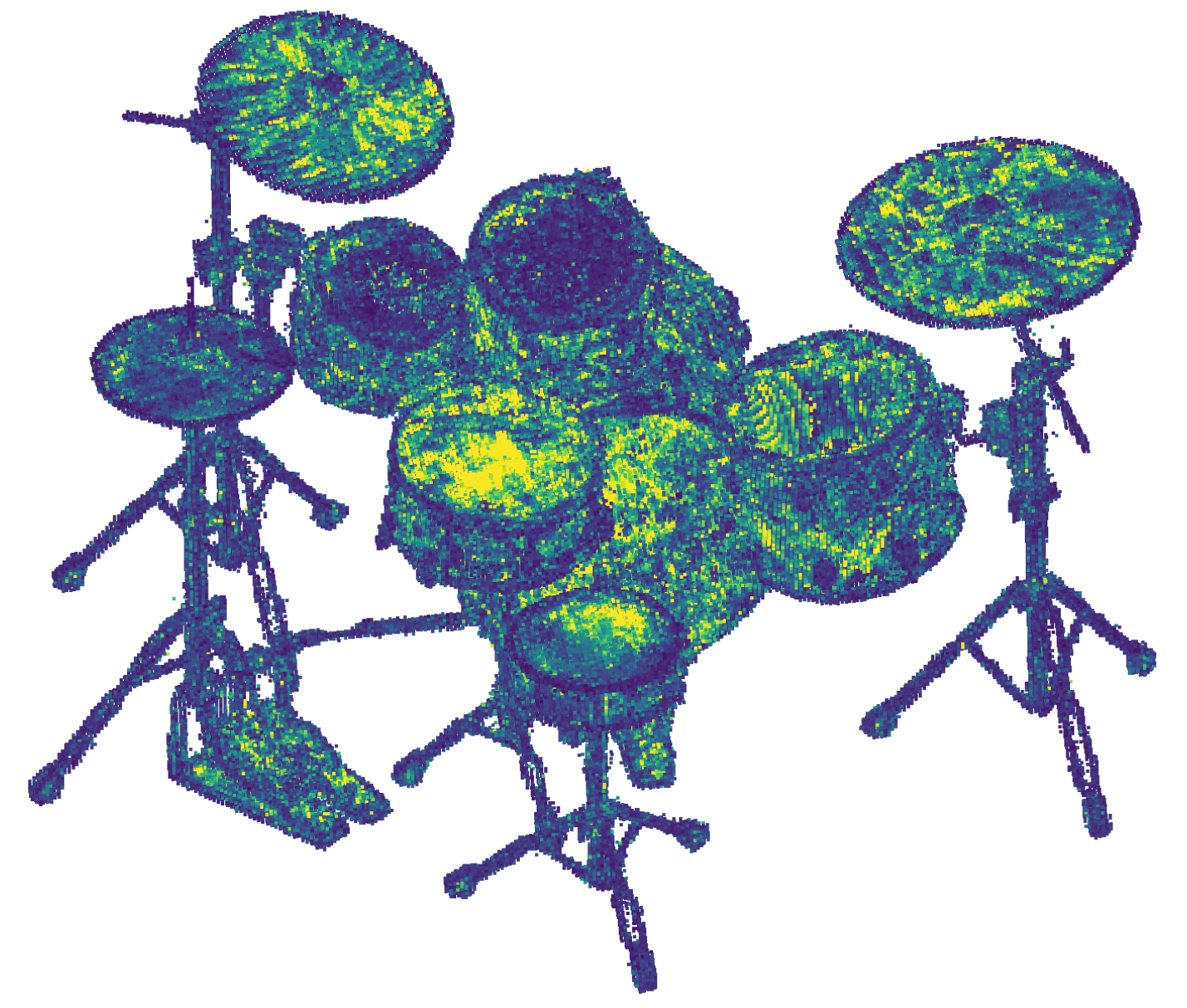}}
\subfigure{\label{fig:drums_noise_trans_unc_}
	\includegraphics[width=0.15\linewidth]{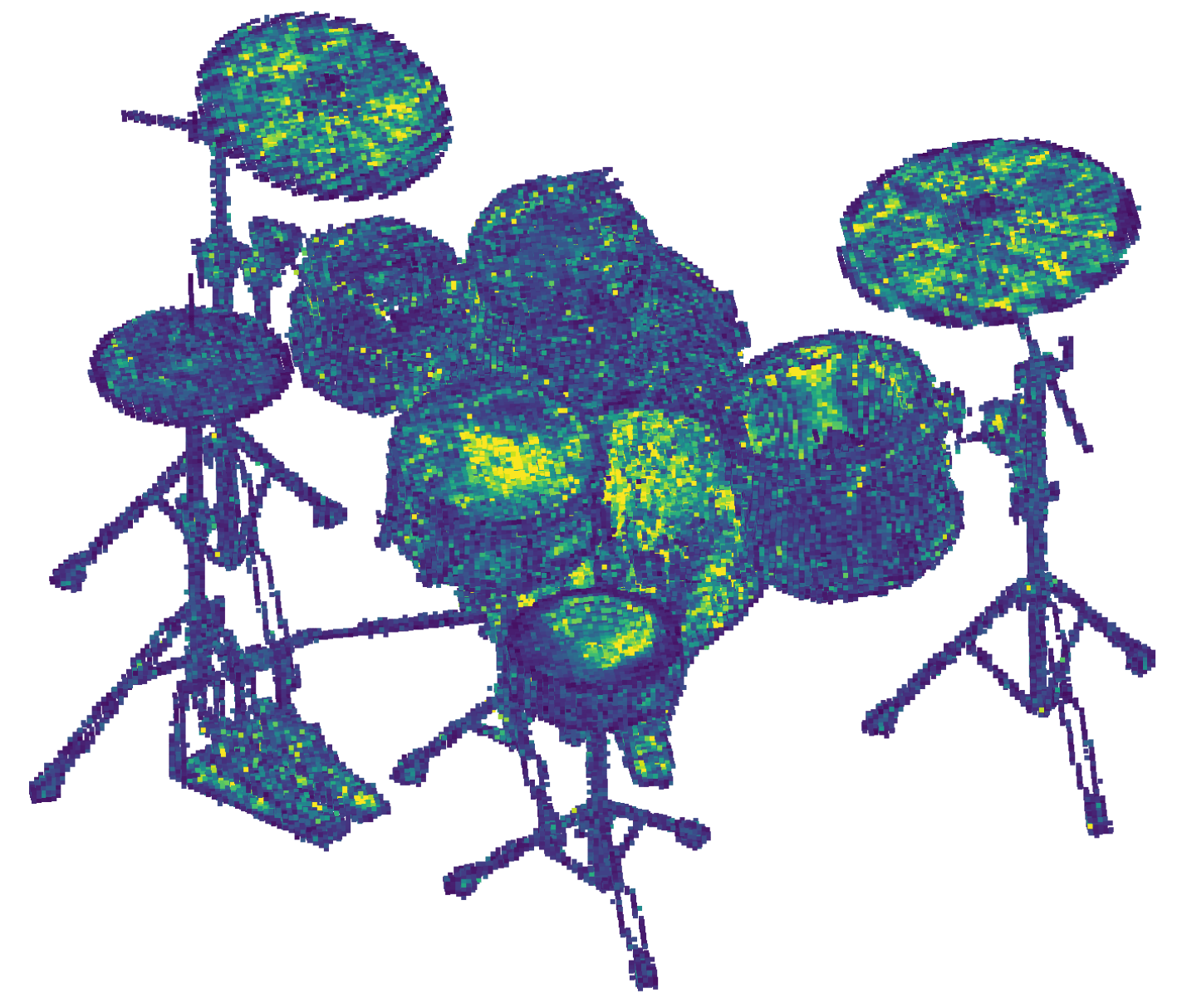}}
\subfigure{\label{fig:drums_noise_rot_unc_}
	\includegraphics[width=0.15\linewidth]{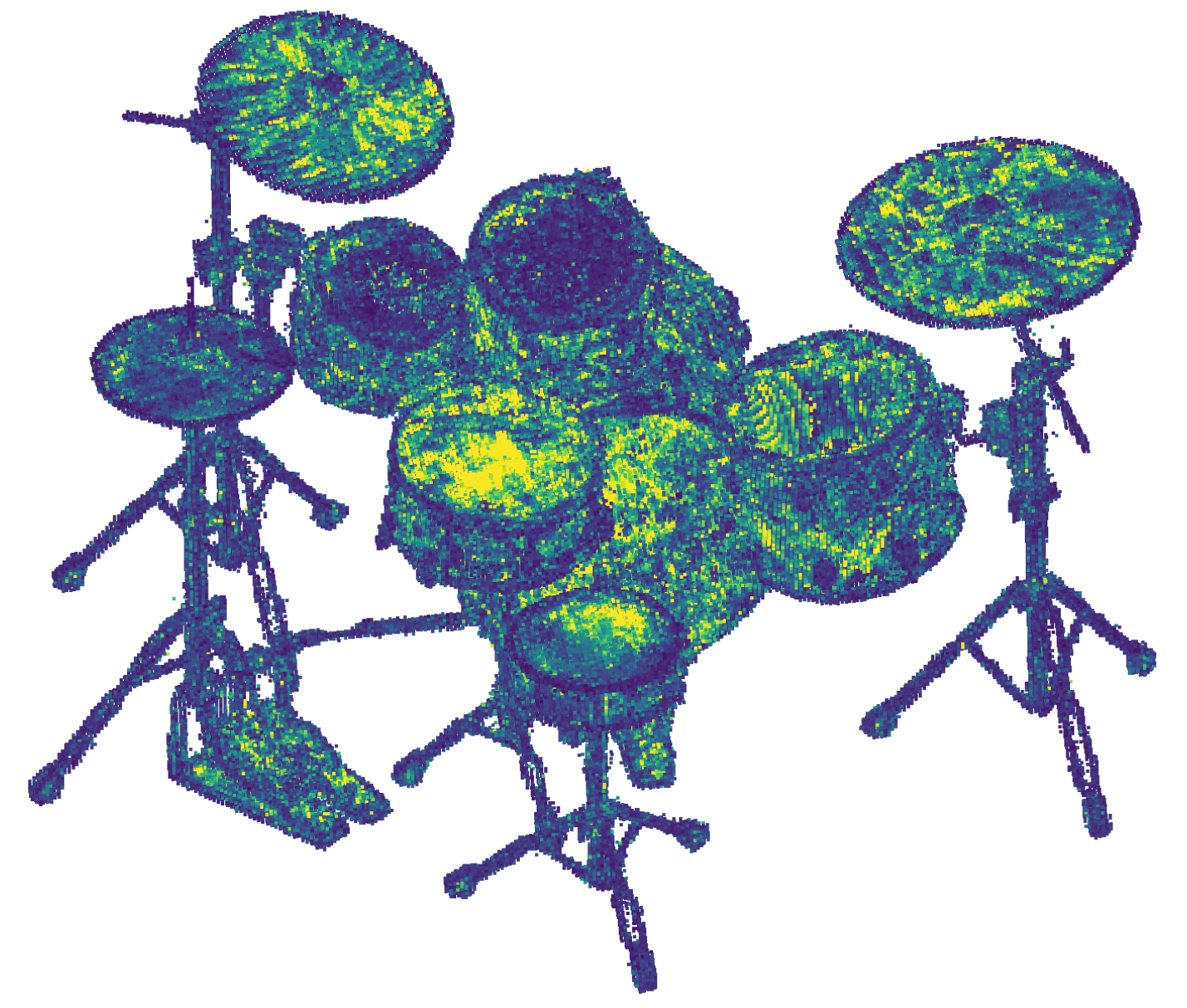}}\\
\rotatebox{90}{$\,\,\,\,\,\,\,\,\,\,\,\,\,\,\,\,$hotdog}
\subfigure{\label{fig:hotdog_member_}
	\includegraphics[width=0.15\linewidth]{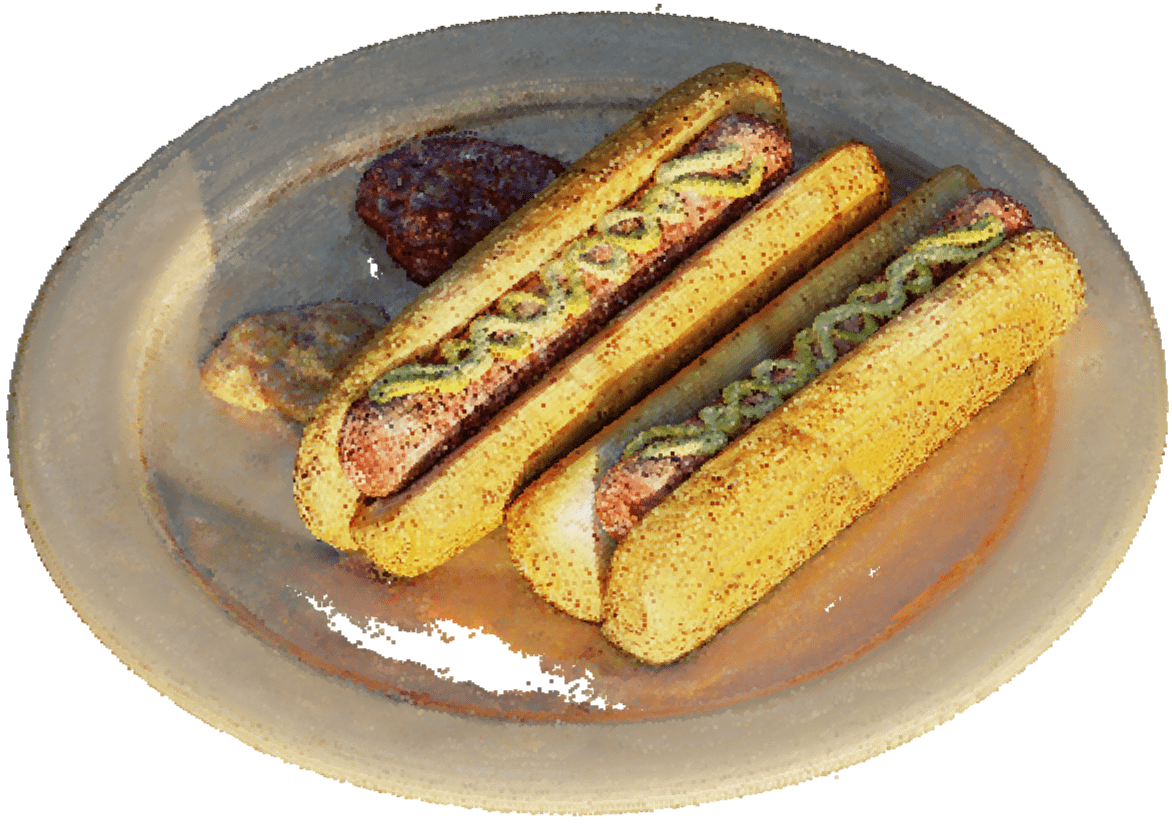}}
\subfigure{\label{fig:hotdog_ensemble_}  
     \includegraphics[width=0.15\linewidth]{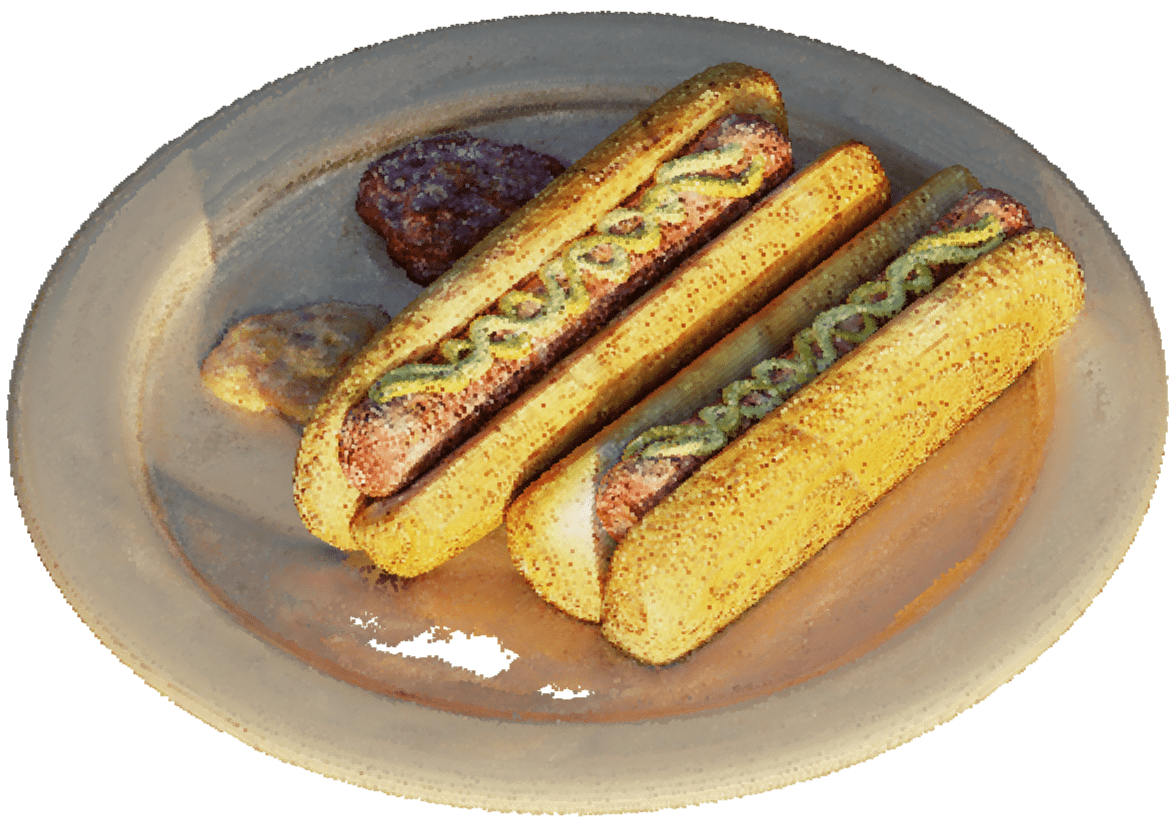}} 
\subfigure{\label{fig:hotdog_unc_}  
     \includegraphics[width=0.15\linewidth]{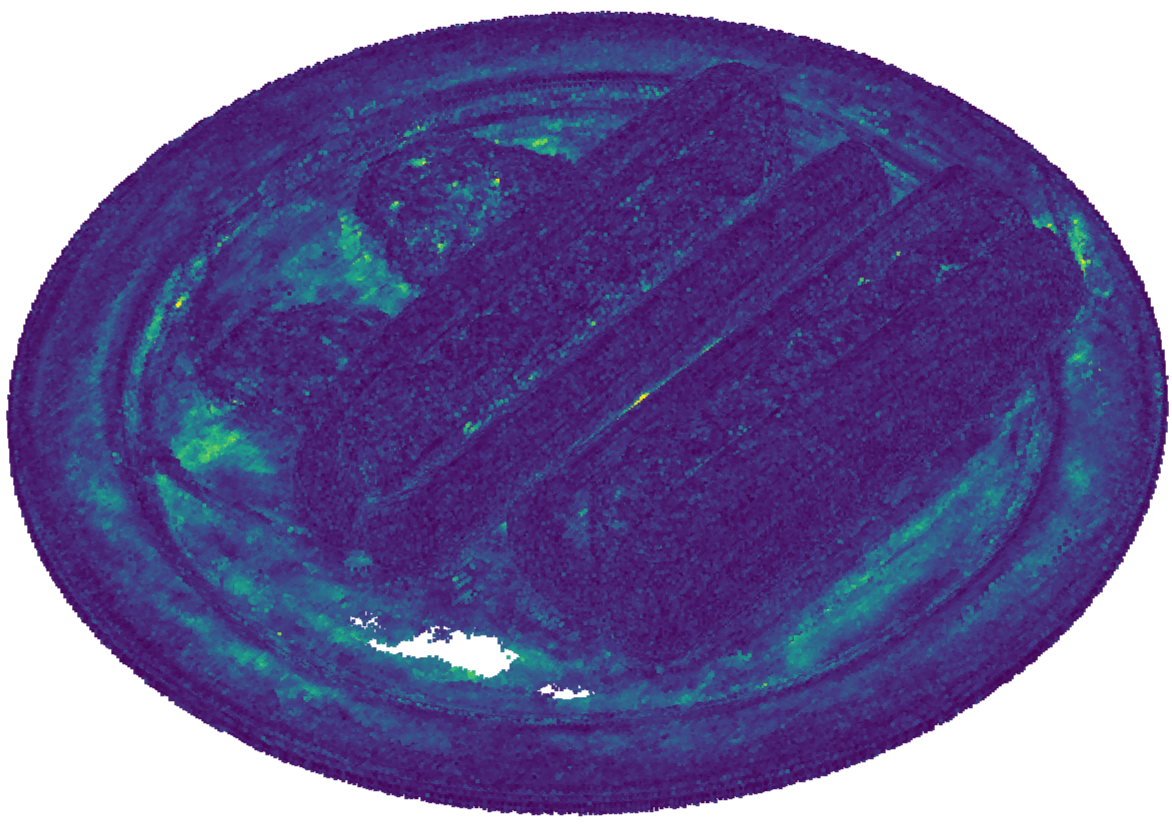}}
\subfigure{\label{fig:hotdog_noise_image_unc_}
	\includegraphics[width=0.15\linewidth]{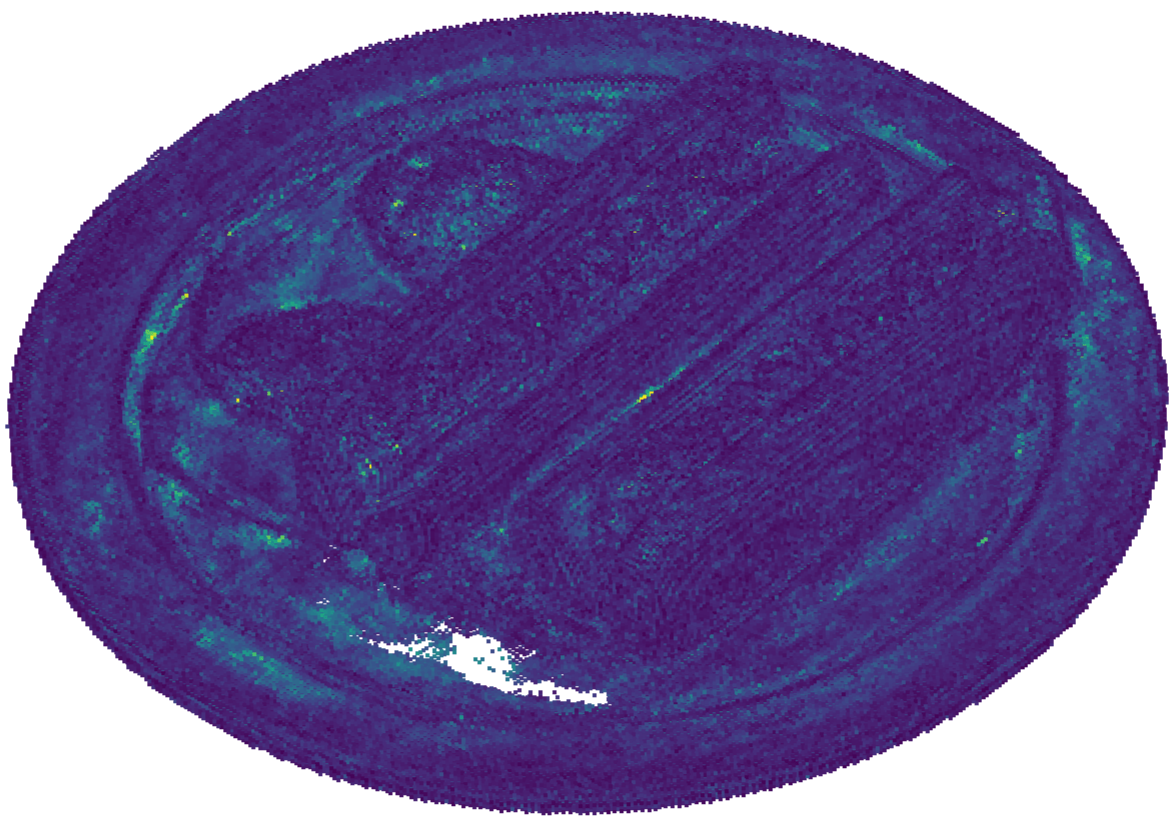}}
\subfigure{\label{fig:hotdog_noise_trans_unc_}
	\includegraphics[width=0.15\linewidth]{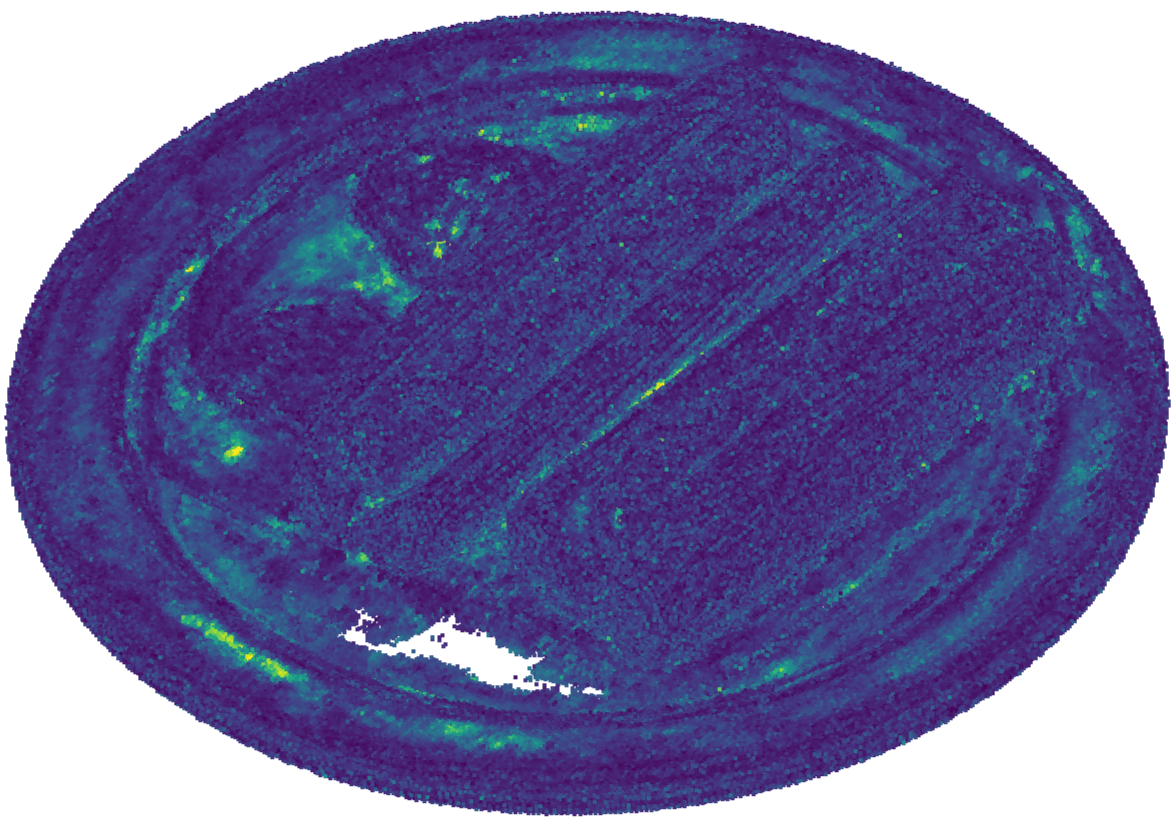}}
\subfigure{\label{fig:hotdog_noise_rot_unc_}
	\includegraphics[width=0.15\linewidth]{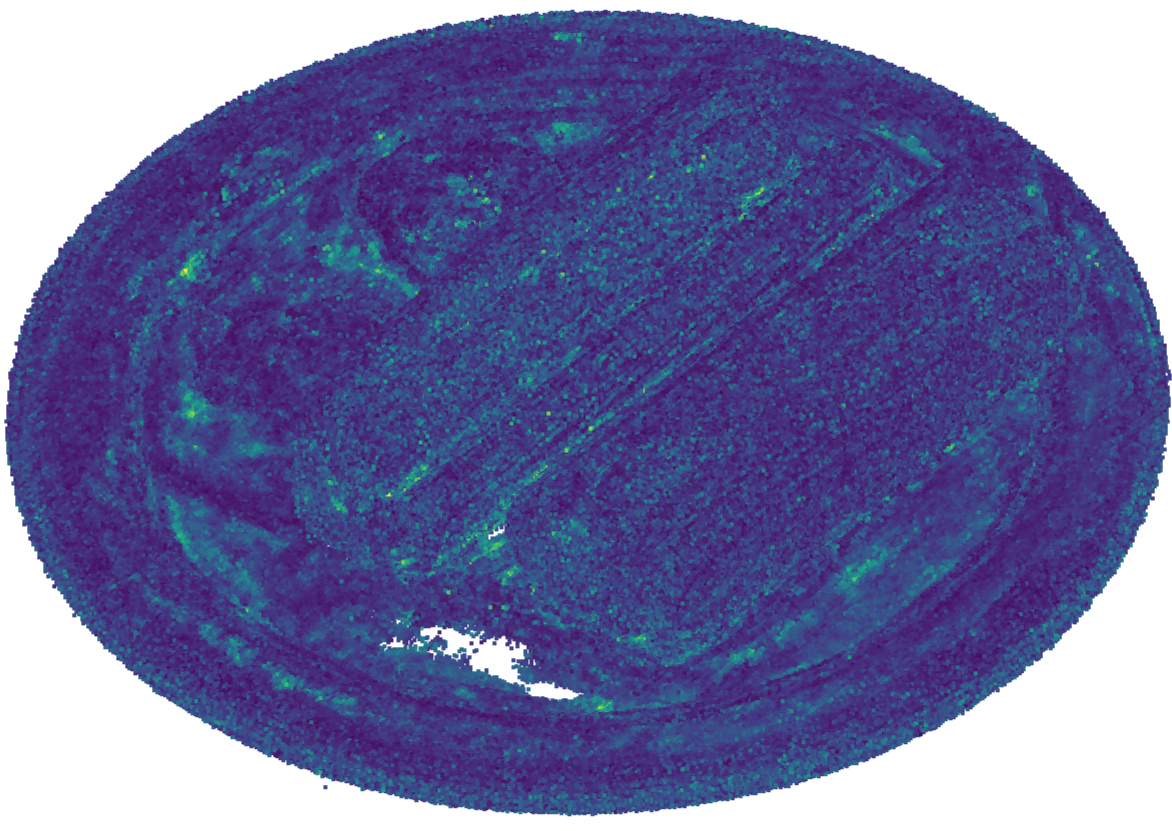}}\\
\rotatebox{90}{$\,\,\,\,\,\,\,\,\,\,\,\,\,\,\,\,$lego}
\subfigure{\label{fig:lego_member_}
	\includegraphics[width=0.16\linewidth]{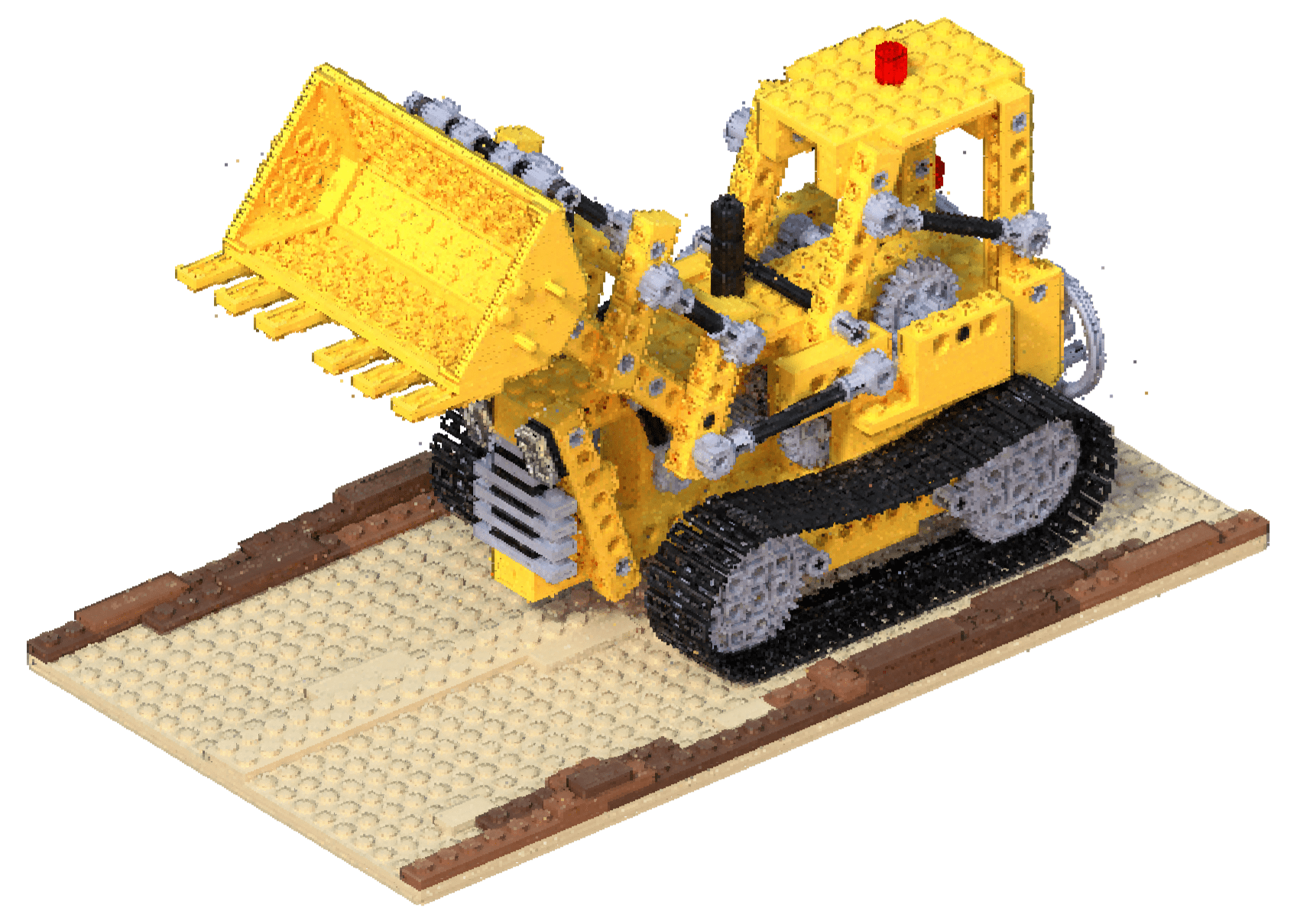}}
  	\hspace{-3mm}
\subfigure{\label{fig:lego_ensemble_}  
     \includegraphics[width=0.16\linewidth]{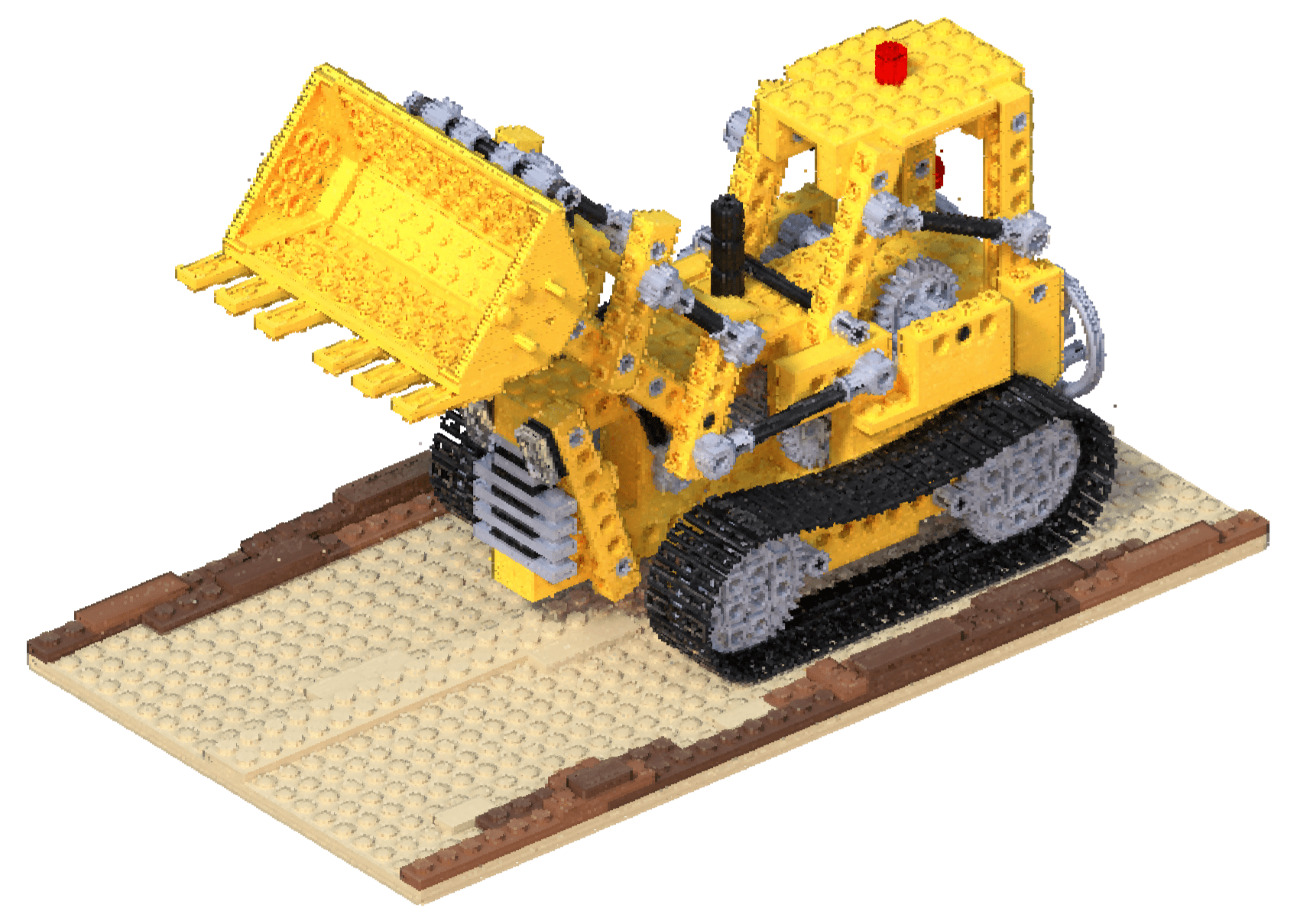}} 
      	\hspace{-3mm}
\subfigure{\label{fig:lego_unc_}  
     \includegraphics[width=0.16\linewidth]{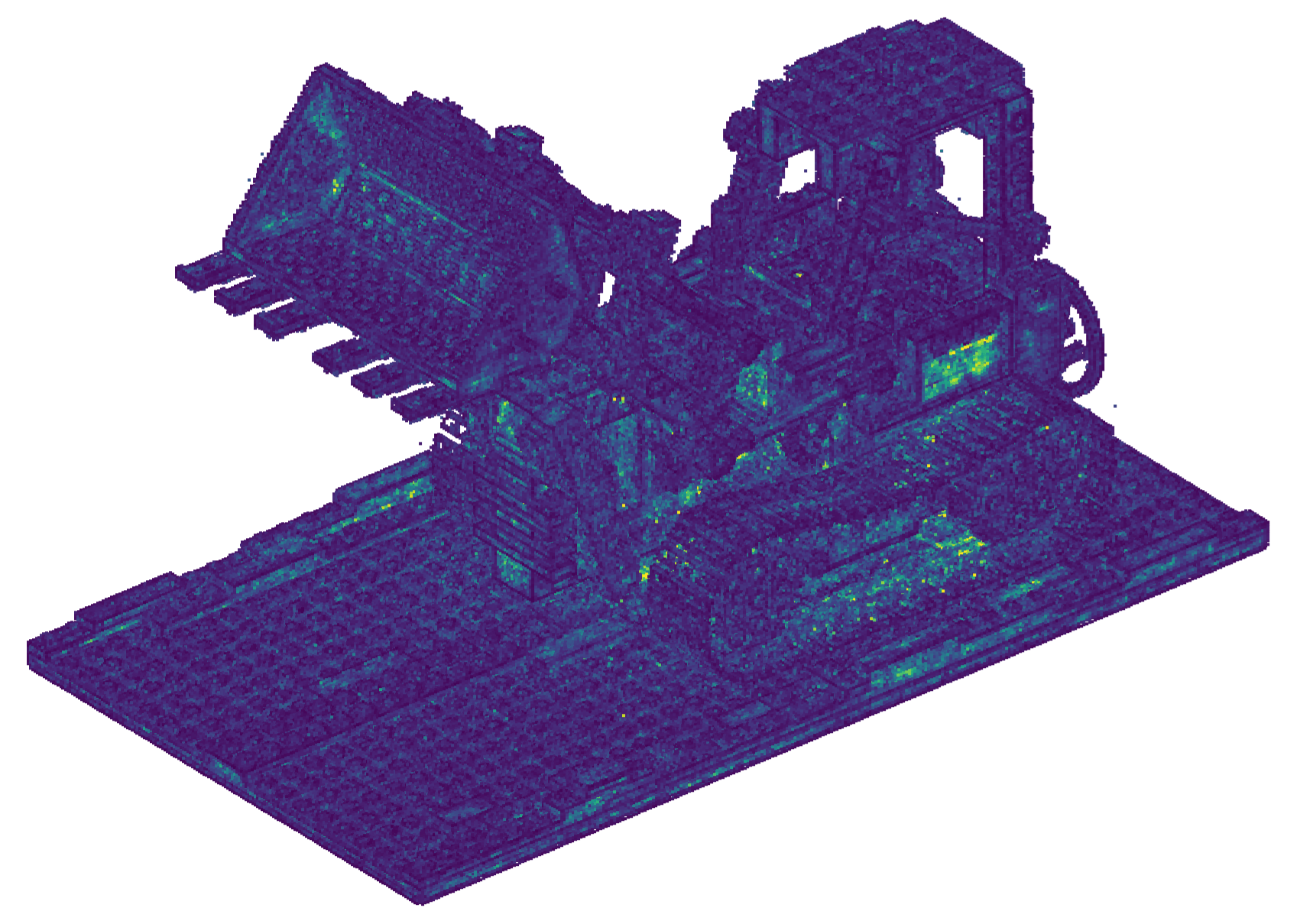}}
      	\hspace{-3mm}
\subfigure{\label{fig:lego_noise_image_unc_}
	\includegraphics[width=0.16\linewidth]{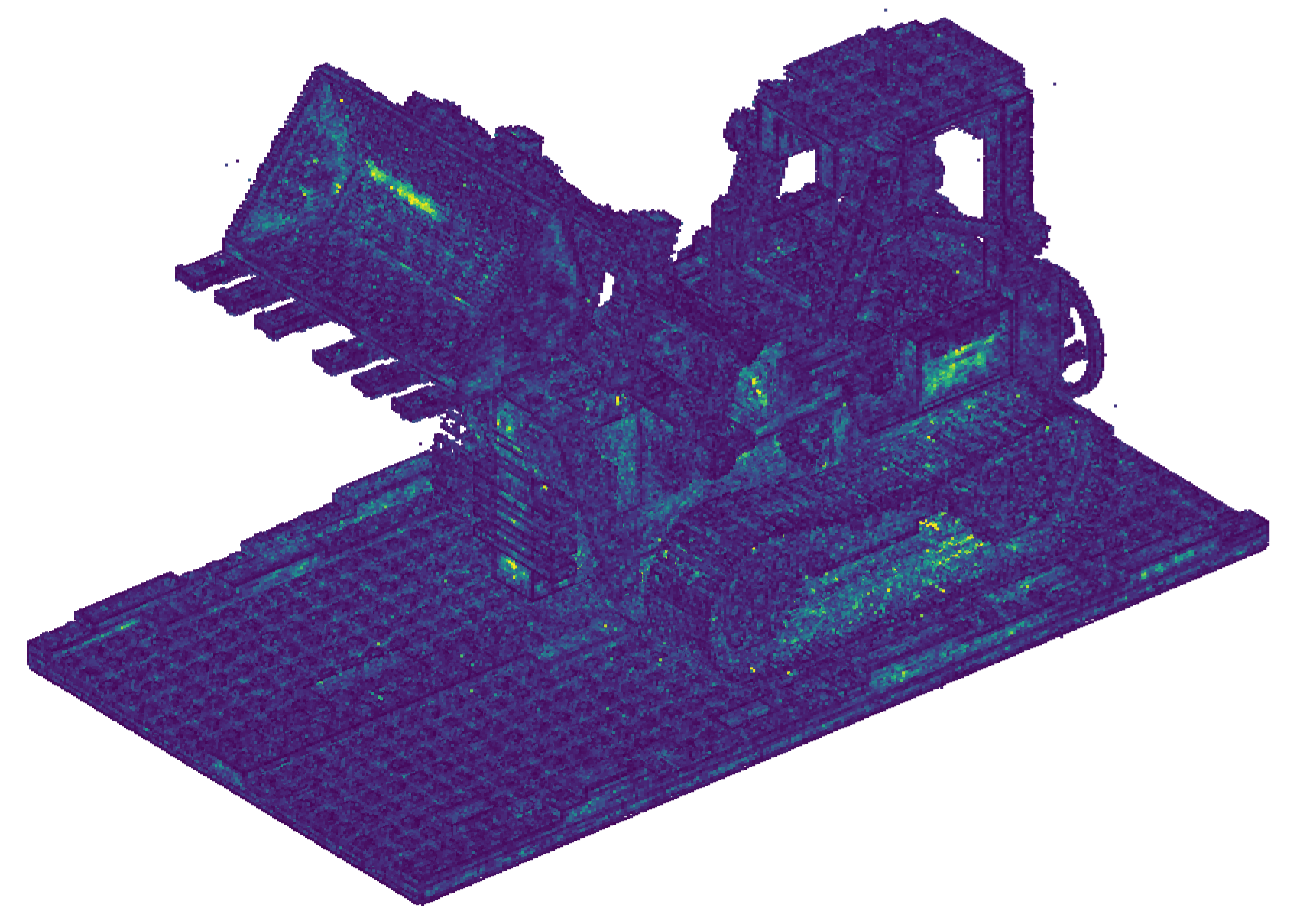}}
  	\hspace{-3mm}
\subfigure{\label{fig:lego_noise_trans_unc_}
	\includegraphics[width=0.16\linewidth]{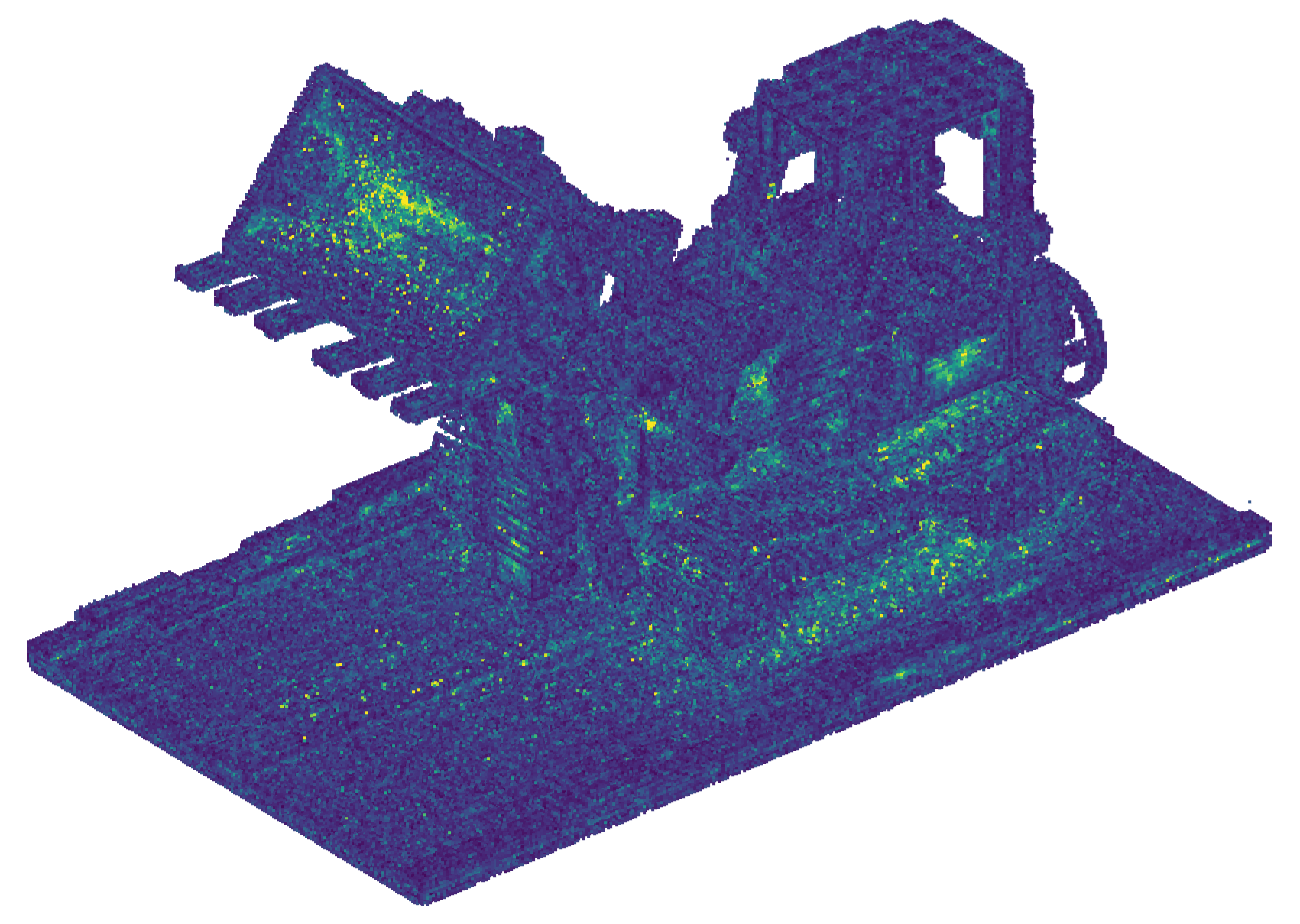}}
  	\hspace{-3mm}
\subfigure{\label{fig:lego_noise_rot_unc_}
	\includegraphics[width=0.16\linewidth]{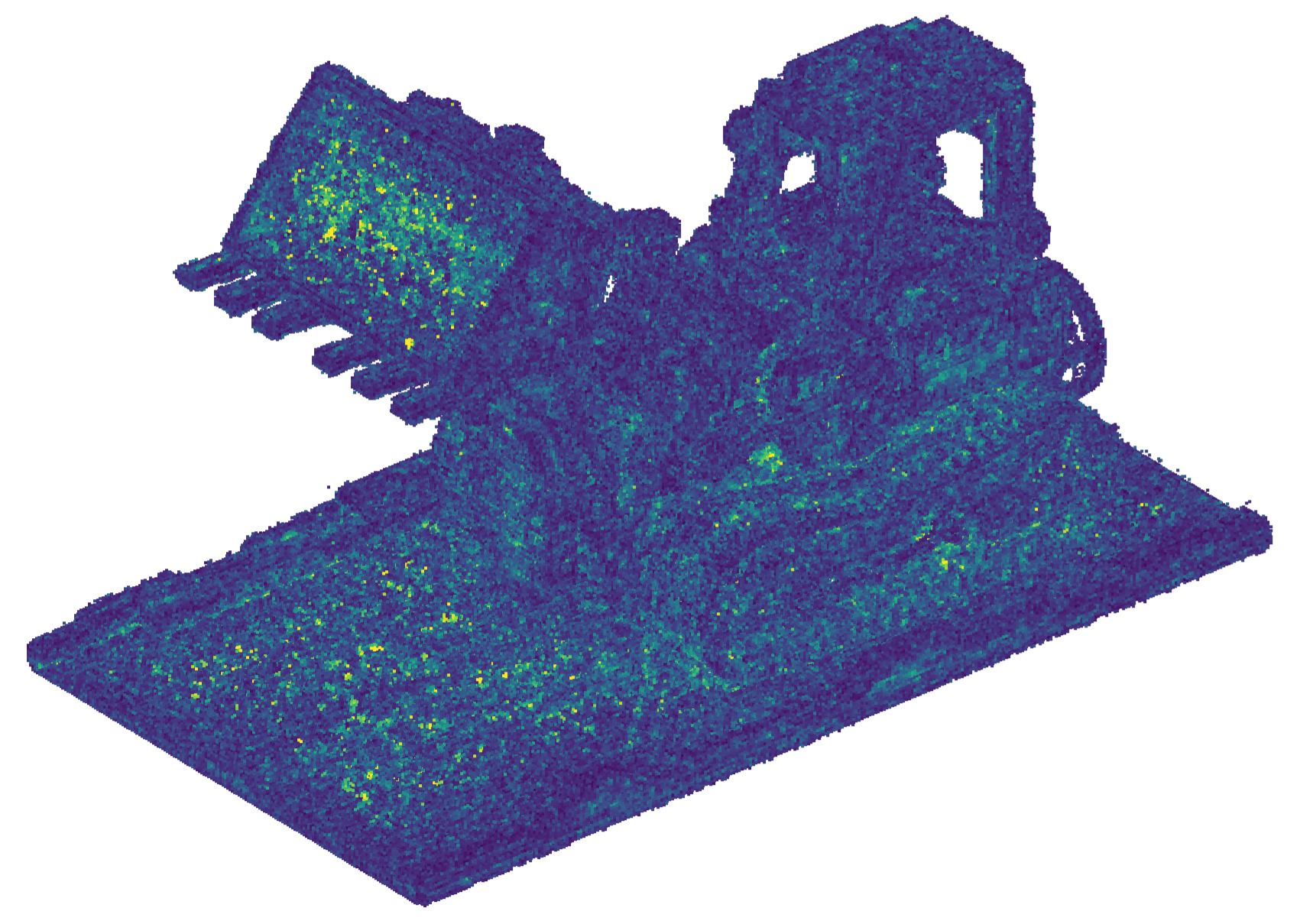}}\\
\rotatebox{90}{$\,\,\,\,\,\,$materials}
\subfigure{\label{fig:materials_member_}
	\includegraphics[width=0.15\linewidth]{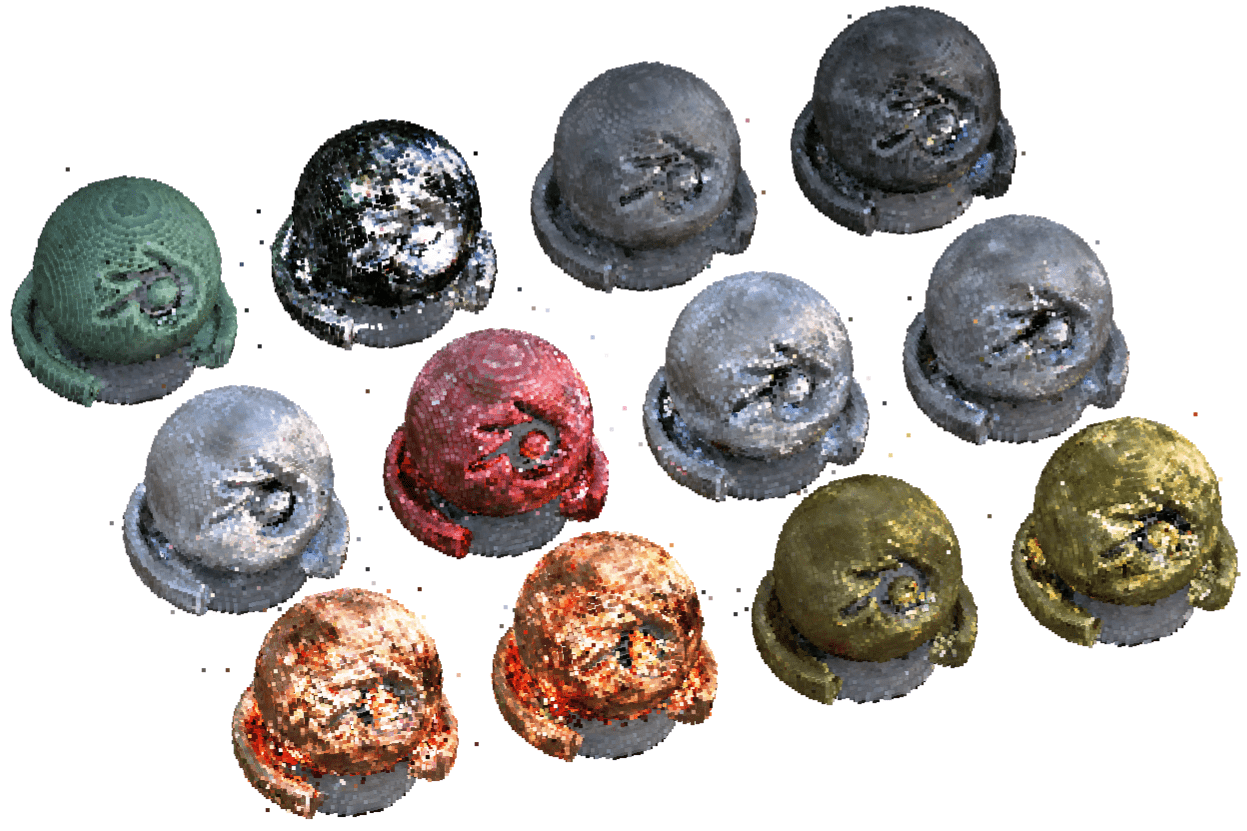}}
\subfigure{\label{fig:materials_ensemble_}  
     \includegraphics[width=0.15\linewidth]{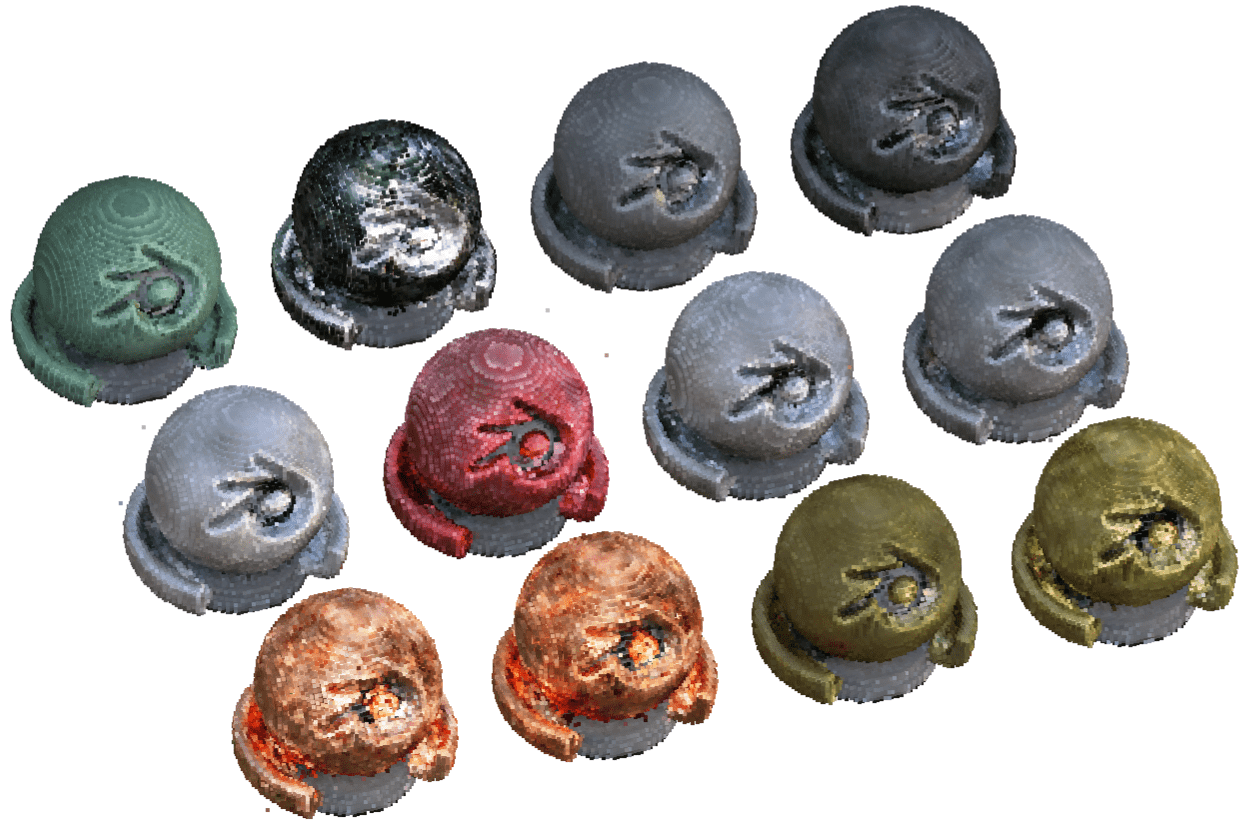}} 
\subfigure{\label{fig:materials_unc_}  
     \includegraphics[width=0.15\linewidth]{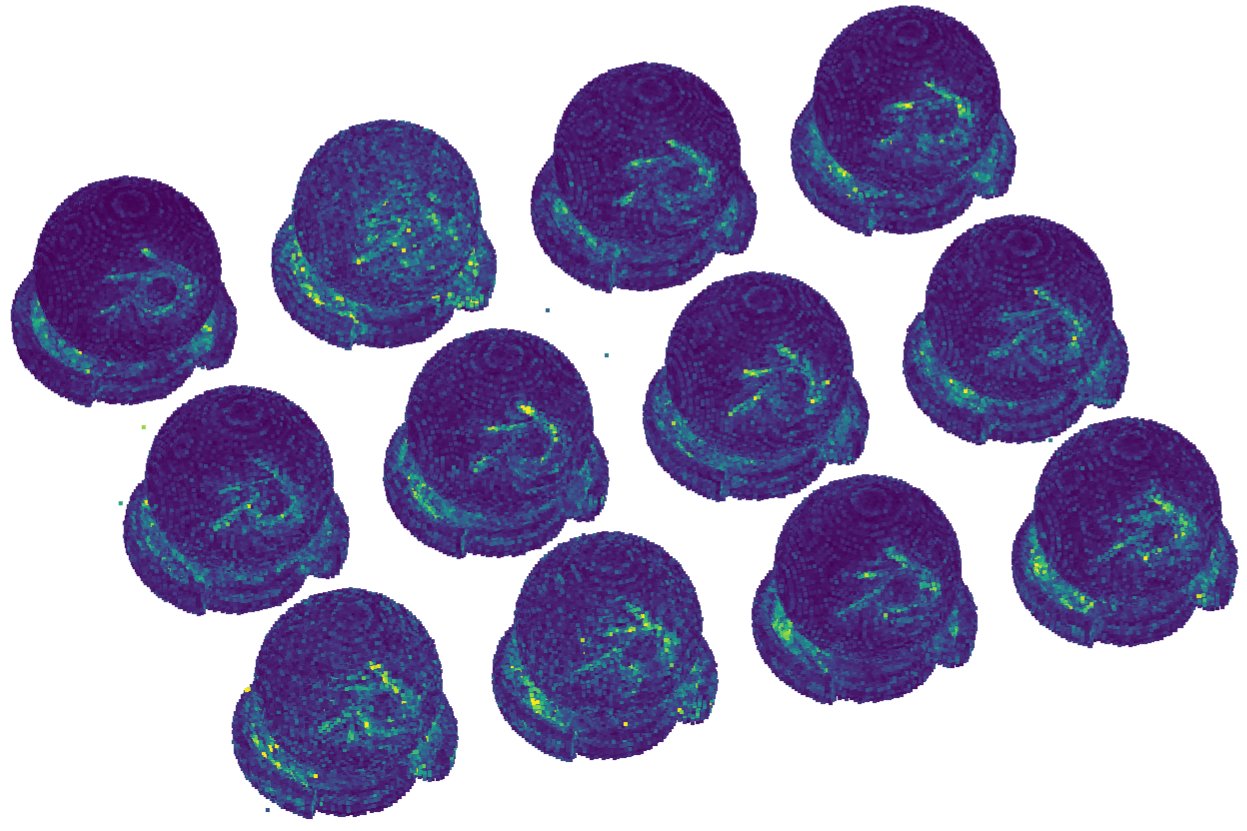}}
\subfigure{\label{fig:materials_noise_image_unc_}
	\includegraphics[width=0.15\linewidth]{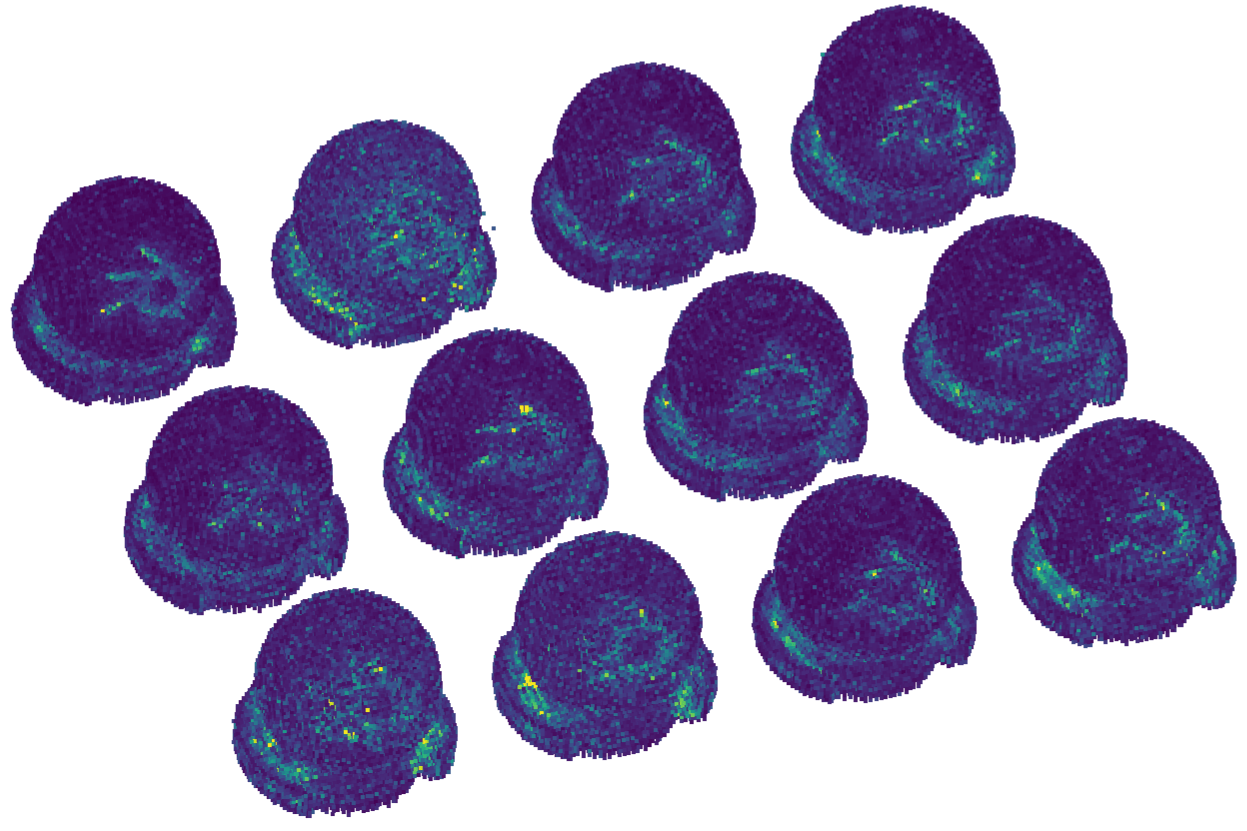}}
\subfigure{\label{fig:materials_noise_trans_unc_}
	\includegraphics[width=0.15\linewidth]{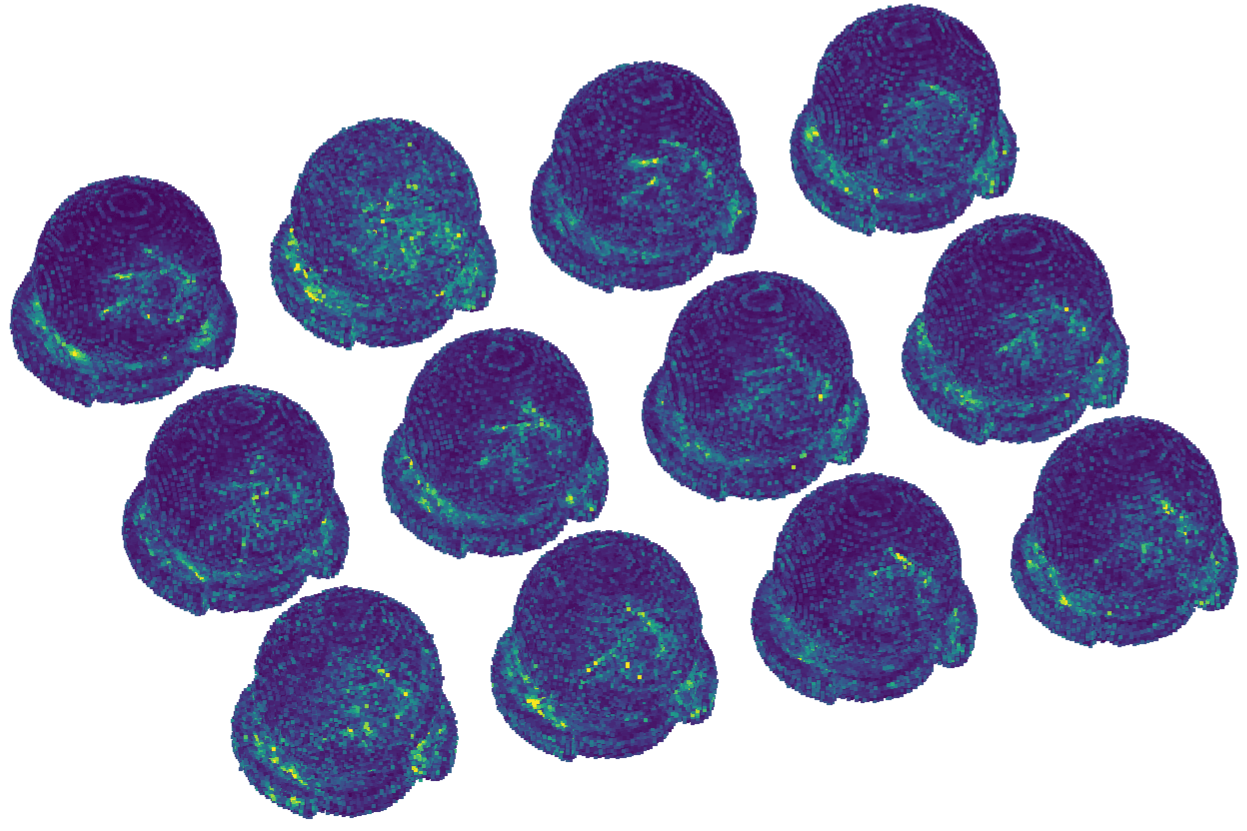}}
\subfigure{\label{fig:materials_noise_rot_unc_}
	\includegraphics[width=0.15\linewidth]{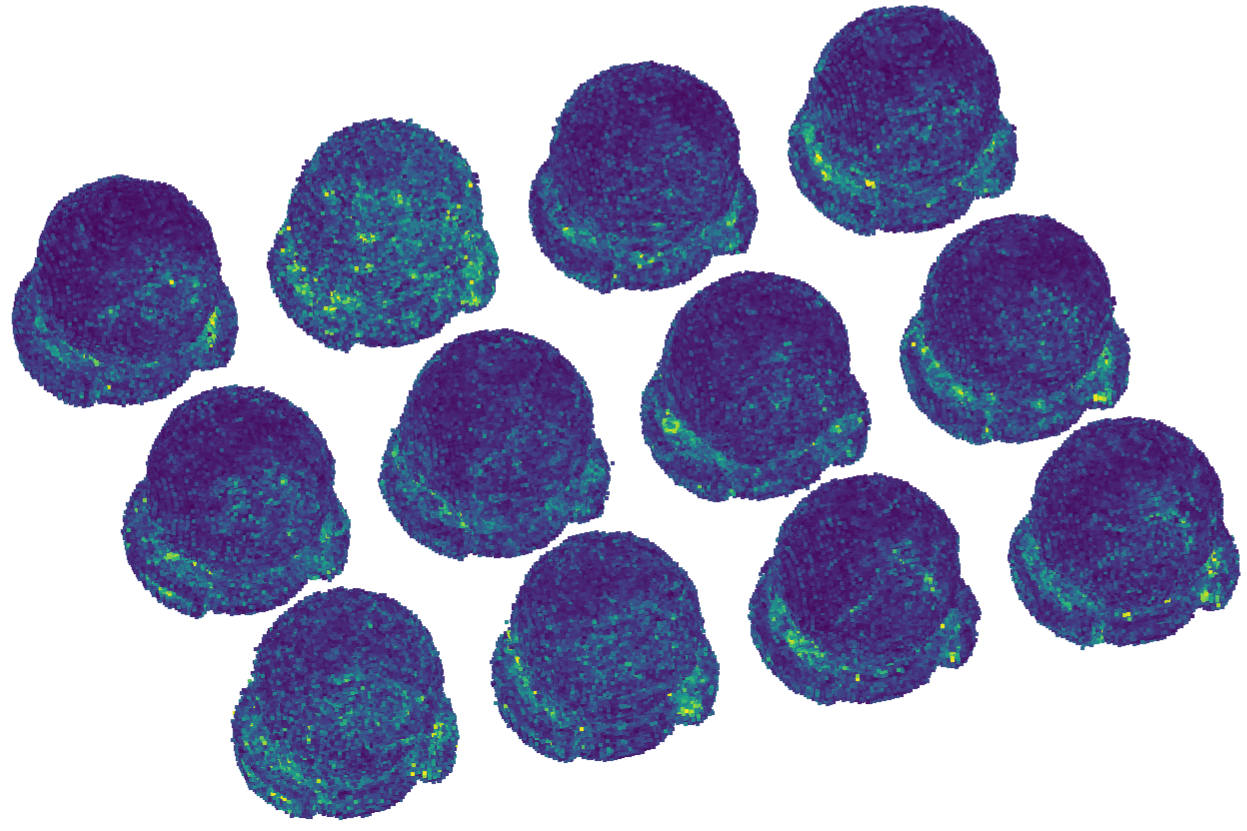}}\\
\rotatebox{90}{$\,\,\,\,\,\,\,\,\,\,\,\,\,\,\,\,$mic}
\subfigure{\label{fig:mic_member_}
	\includegraphics[width=0.15\linewidth]{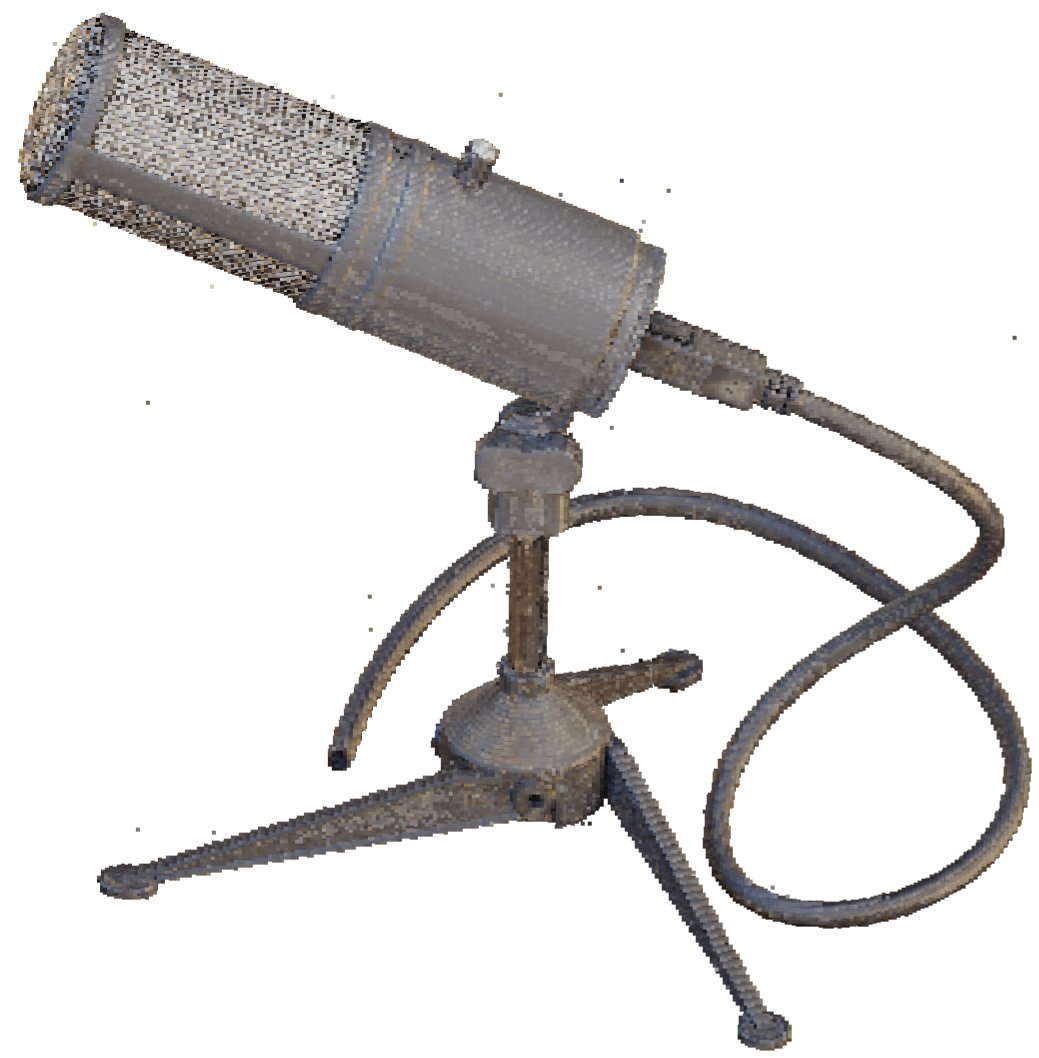}}
\subfigure{\label{fig:mic_ensemble_}  
     \includegraphics[width=0.15\linewidth]{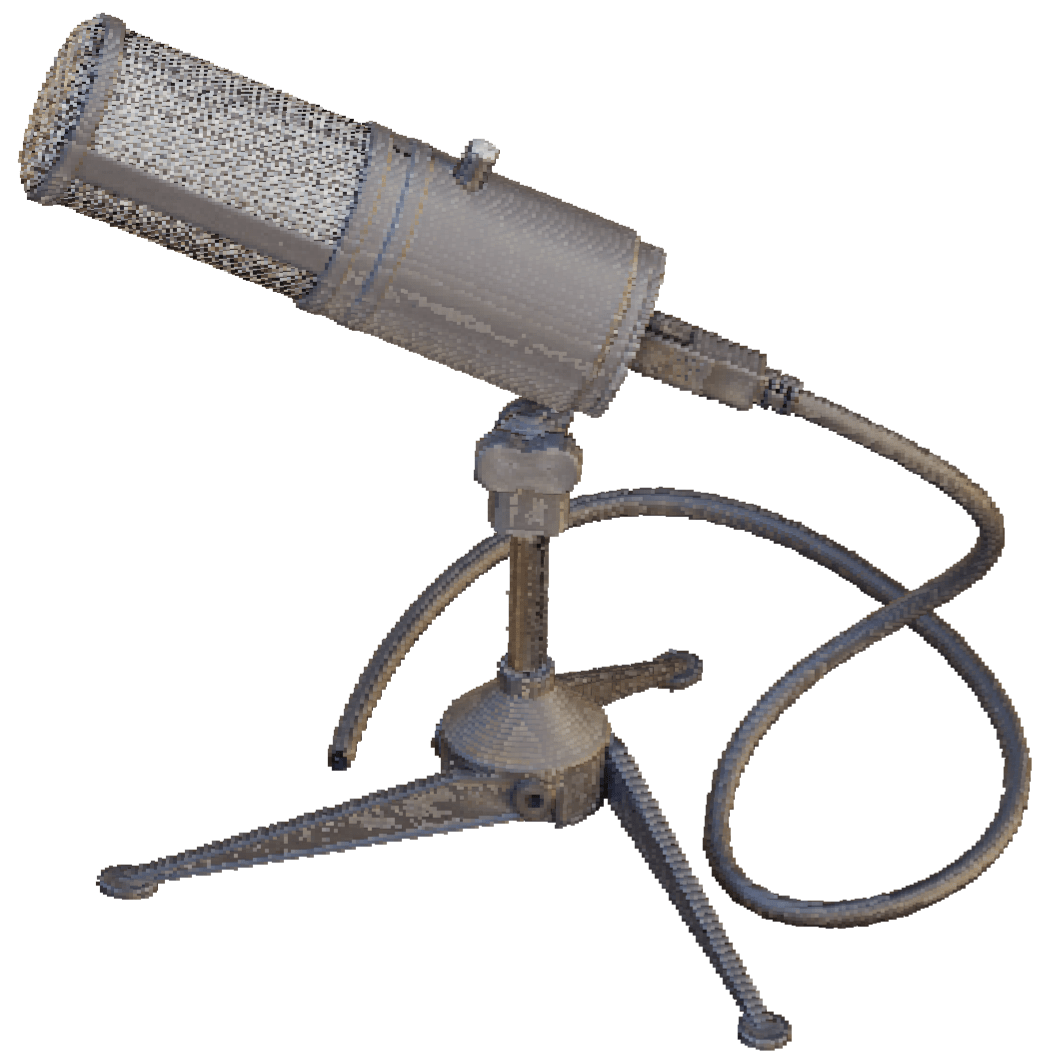}} 
\subfigure{\label{fig:mic_unc_}  
     \includegraphics[width=0.15\linewidth]{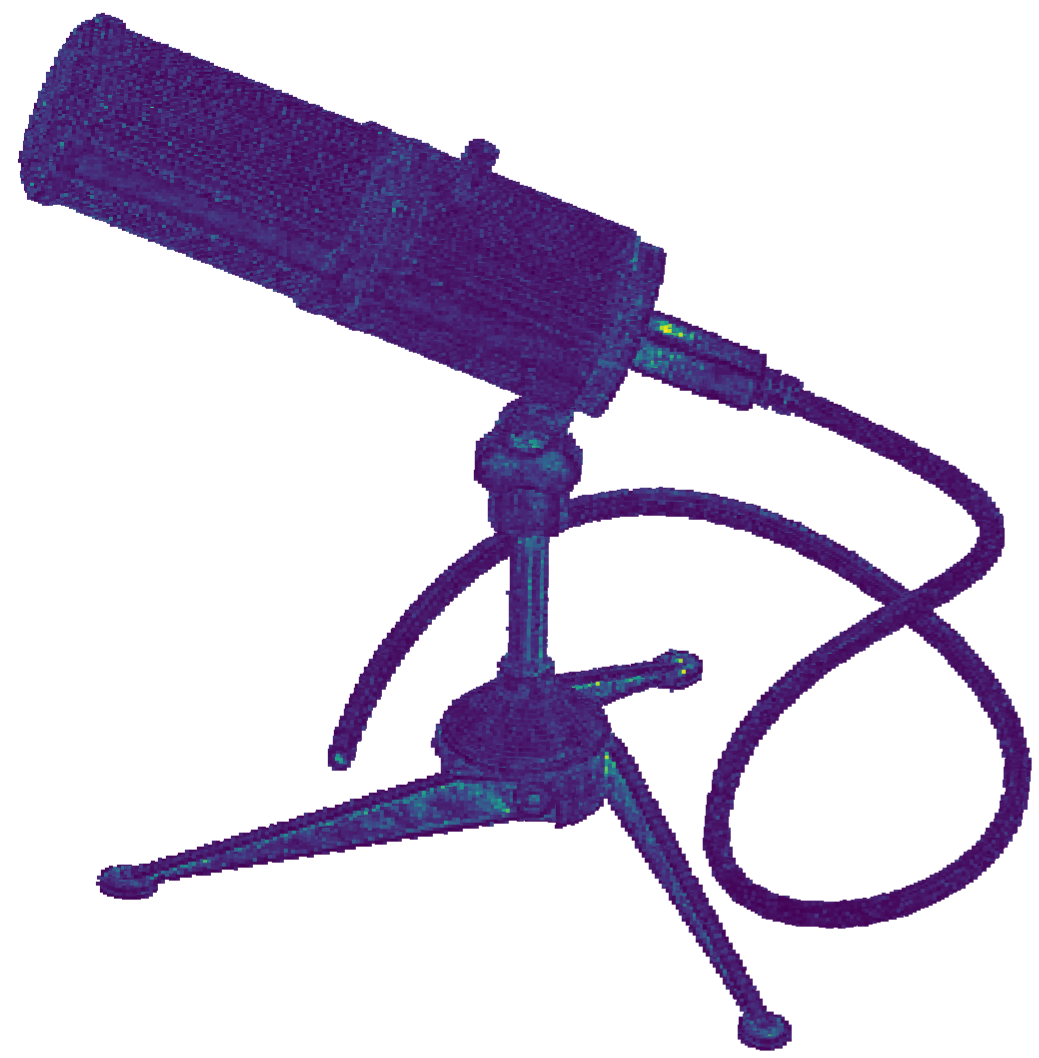}}
\subfigure{\label{fig:mic_noise_image_unc_}
	\includegraphics[width=0.15\linewidth]{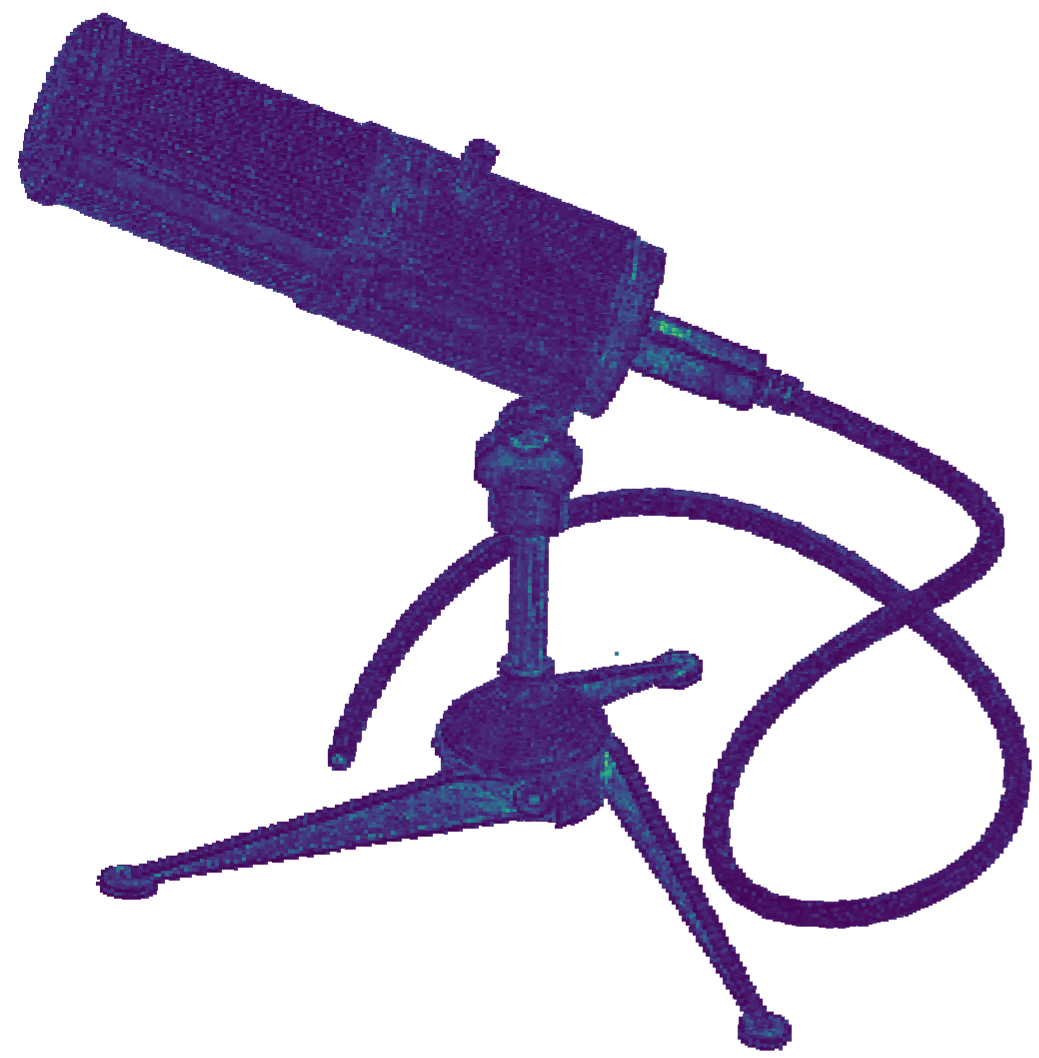}}
\subfigure{\label{fig:mic_noise_trans_unc_}
	\includegraphics[width=0.15\linewidth]{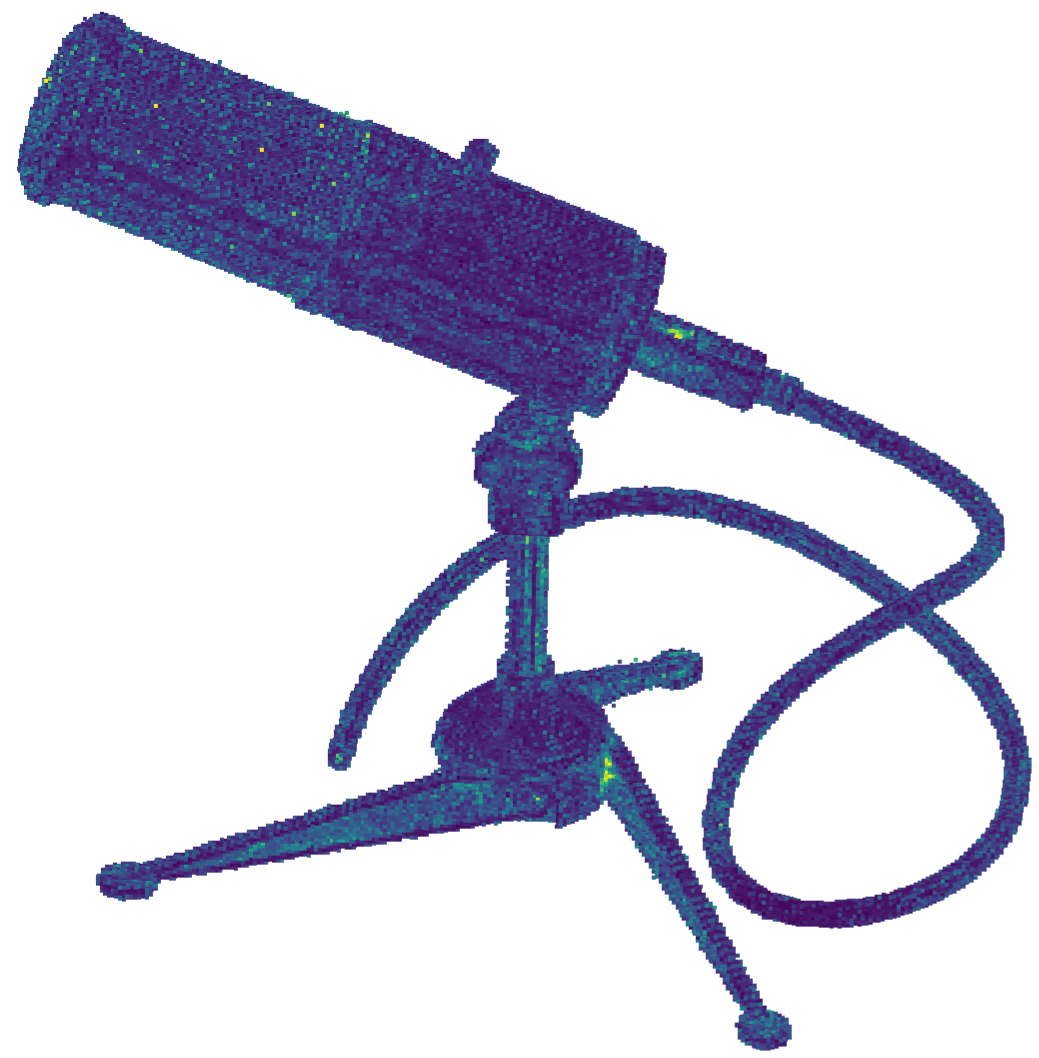}}
\subfigure{\label{fig:mic_noise_rot_unc_}
	\includegraphics[width=0.15\linewidth]{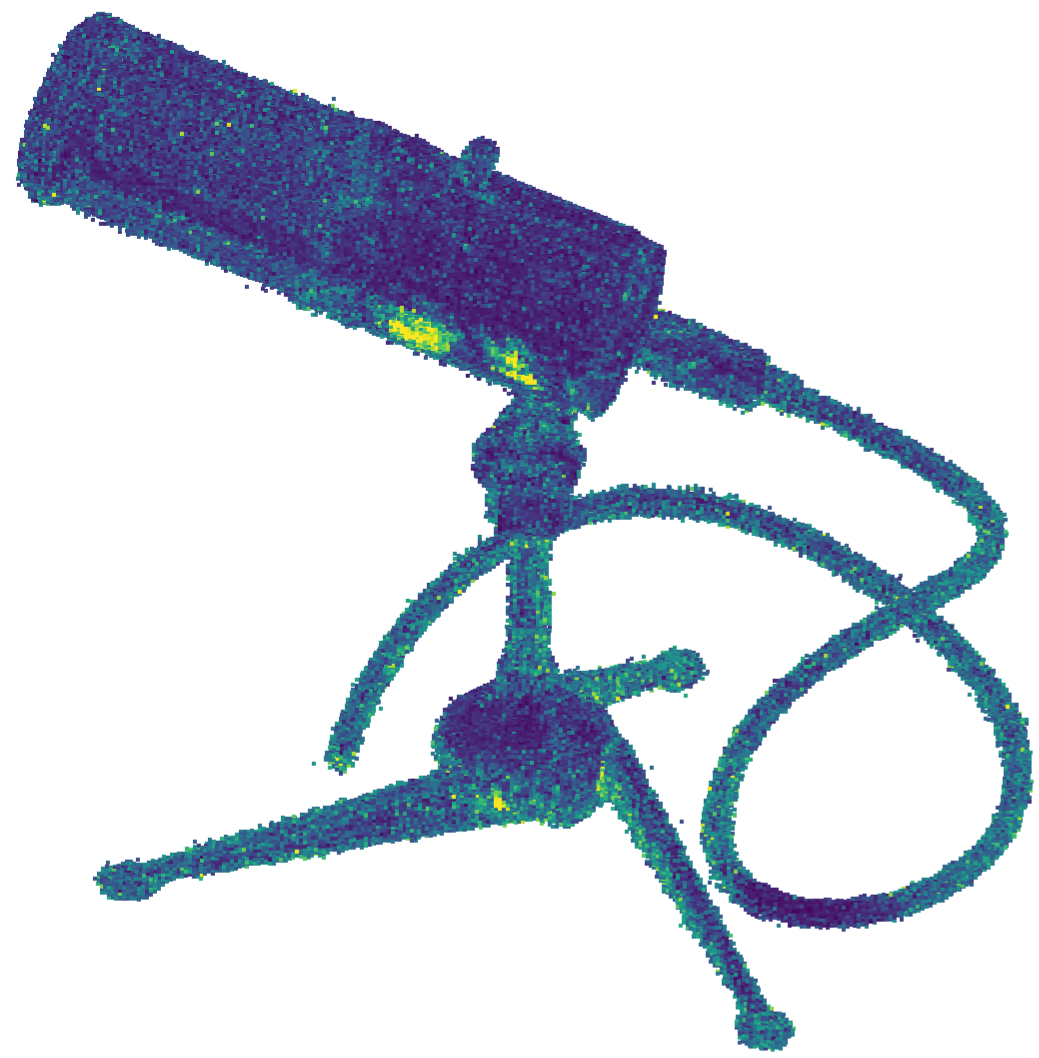}}\\
\rotatebox{90}{$\,\,\,\,\,\,\,\,\,\,\,\,\,\,\,\,$ship}
\subfigure{\label{fig:ship_member_}
	\includegraphics[width=0.15\linewidth]{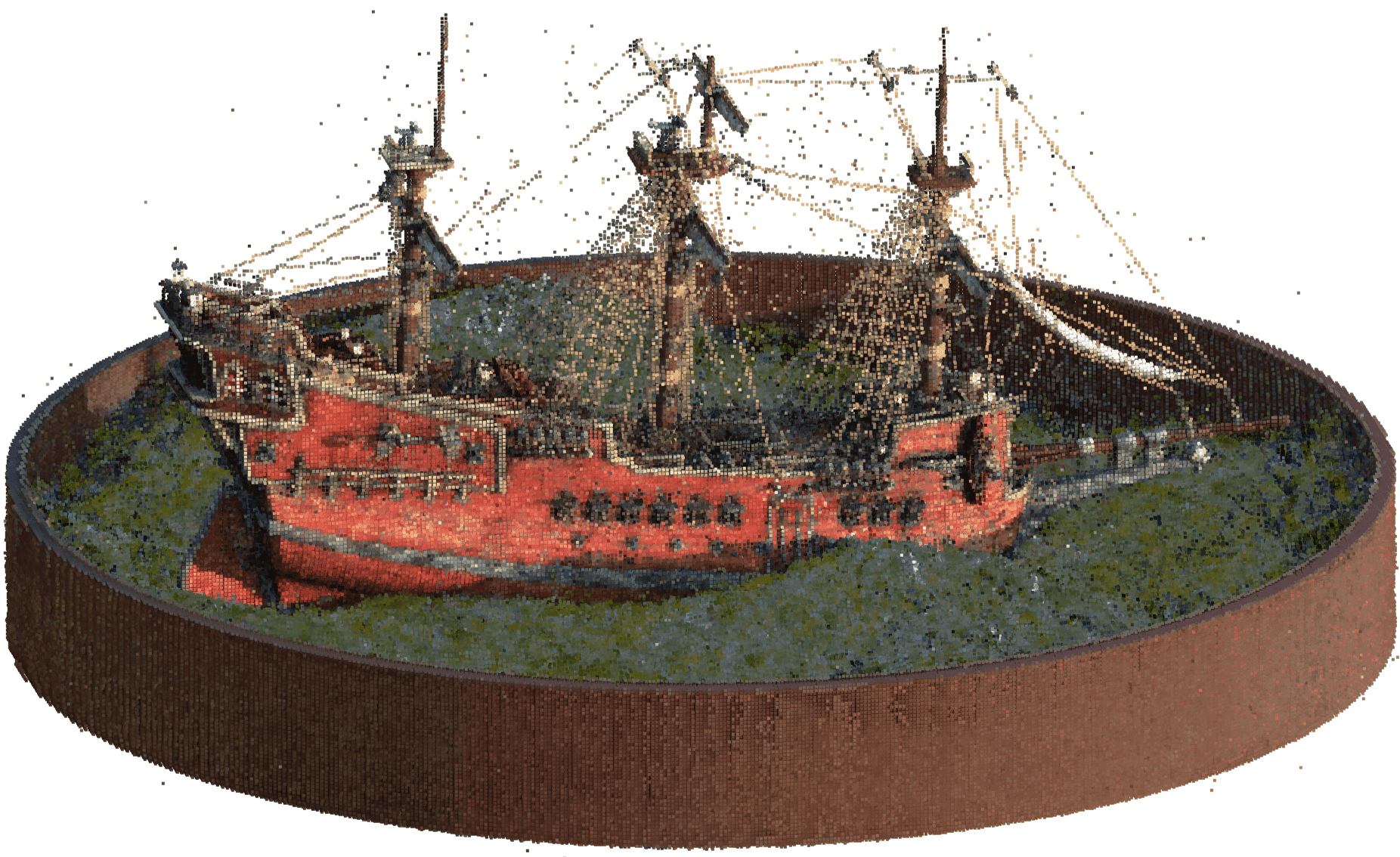}}
\subfigure{\label{fig:ship_ensemble_}  
     \includegraphics[width=0.15\linewidth]{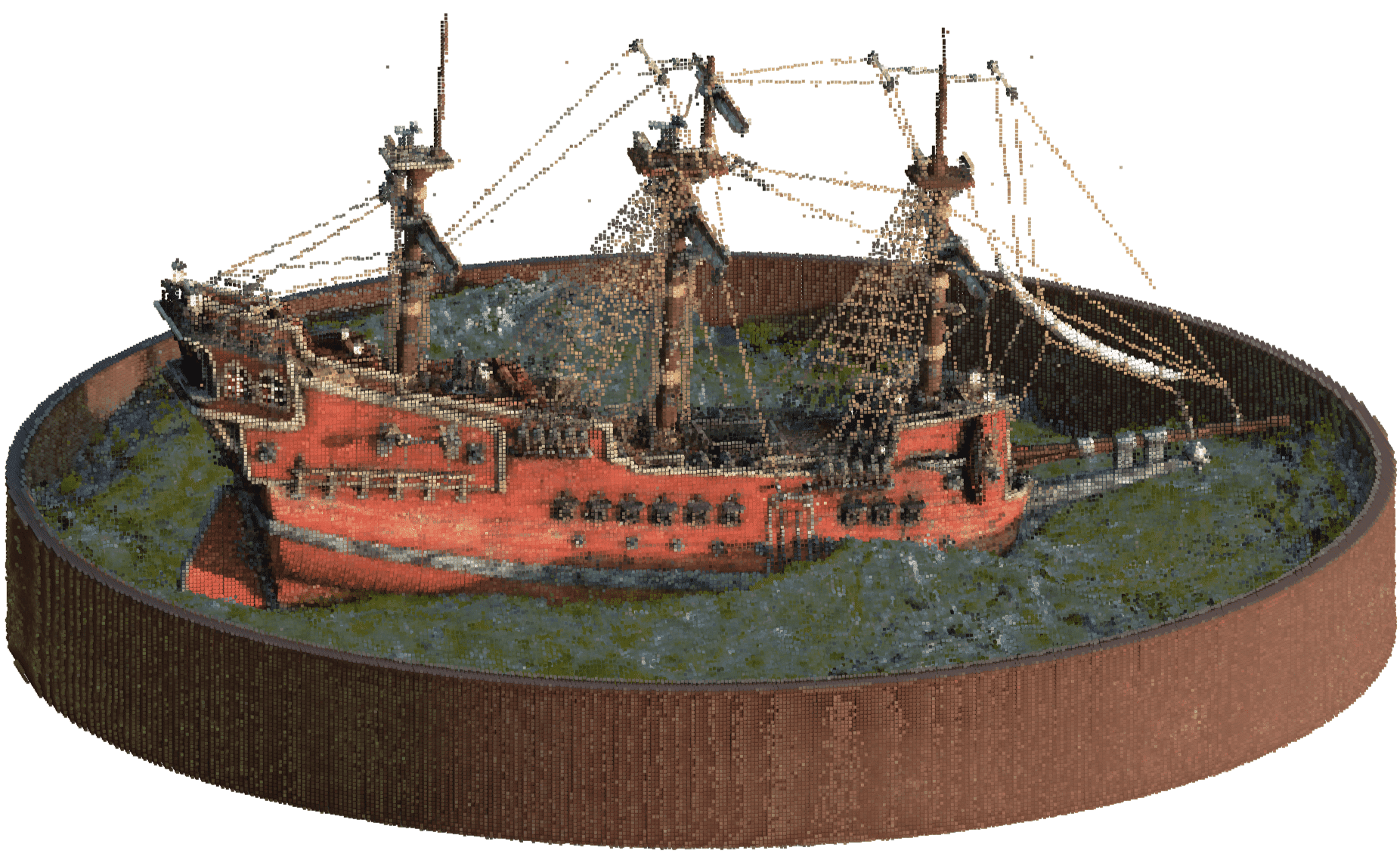}} 
\subfigure{\label{fig:ship_unc_}  
     \includegraphics[width=0.15\linewidth]{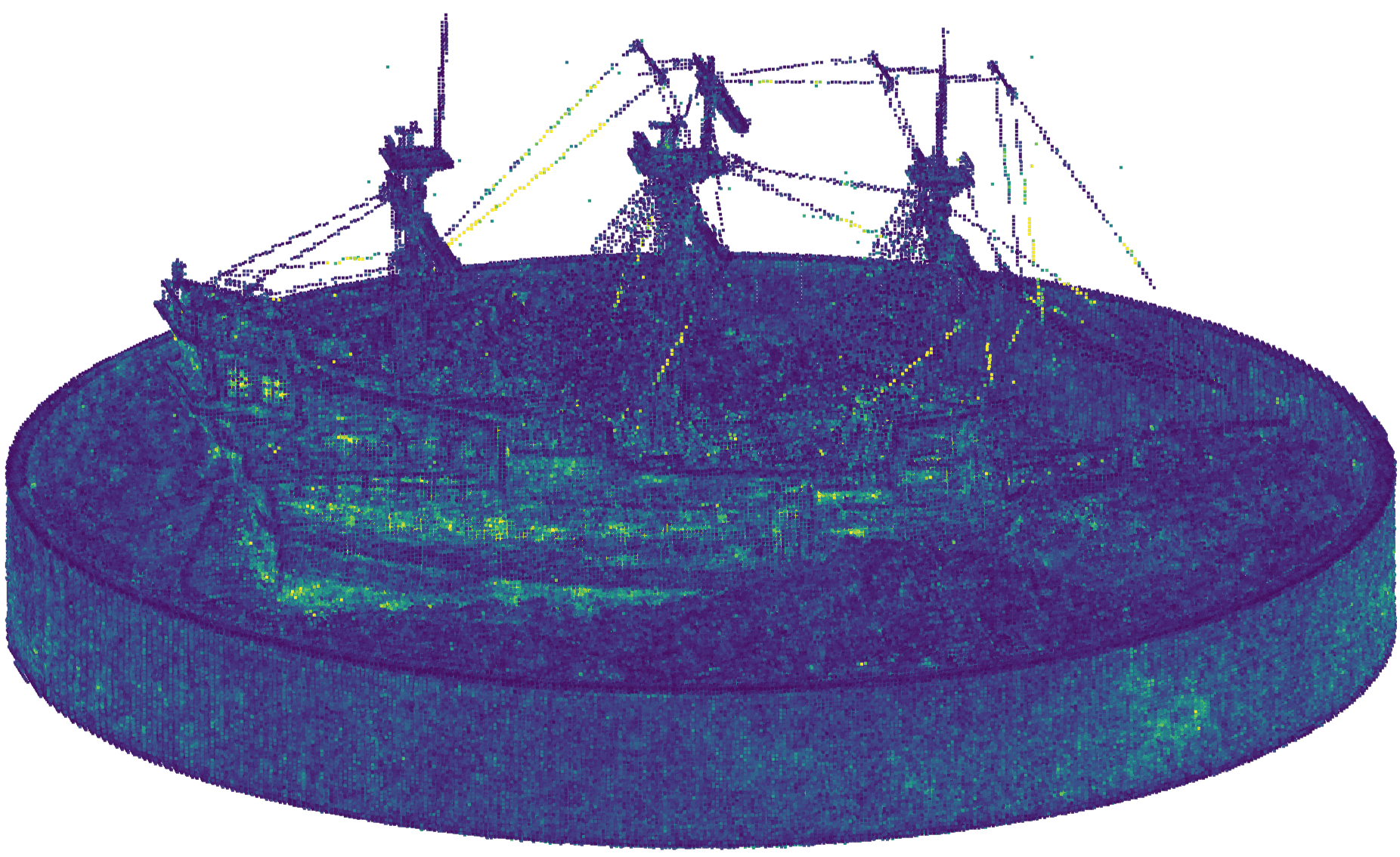}}
\subfigure{\label{fig:ship_noise_image_unc_}
	\includegraphics[width=0.15\linewidth]{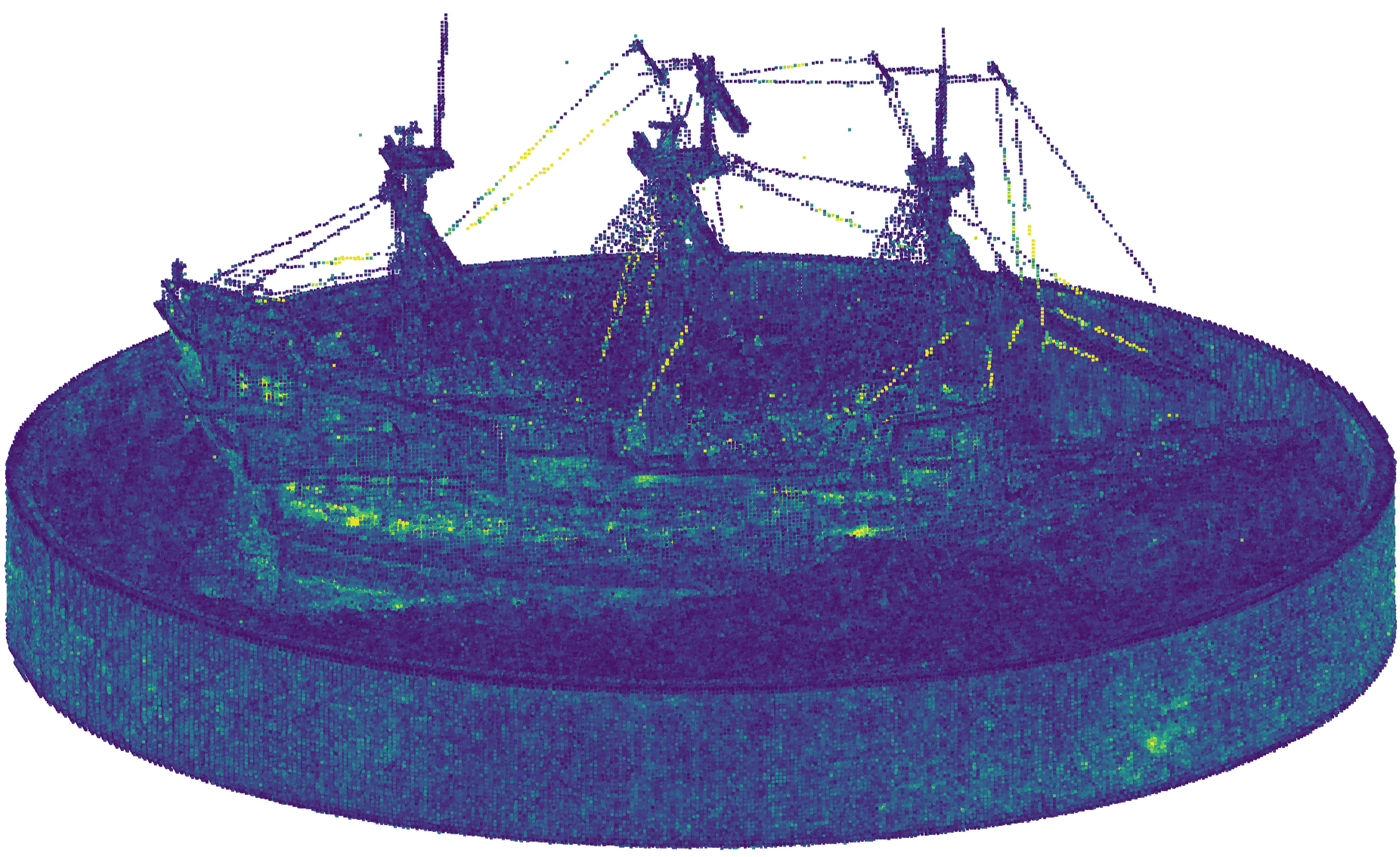}}
\subfigure{\label{fig:ship_noise_trans_unc_}
	\includegraphics[width=0.15\linewidth]{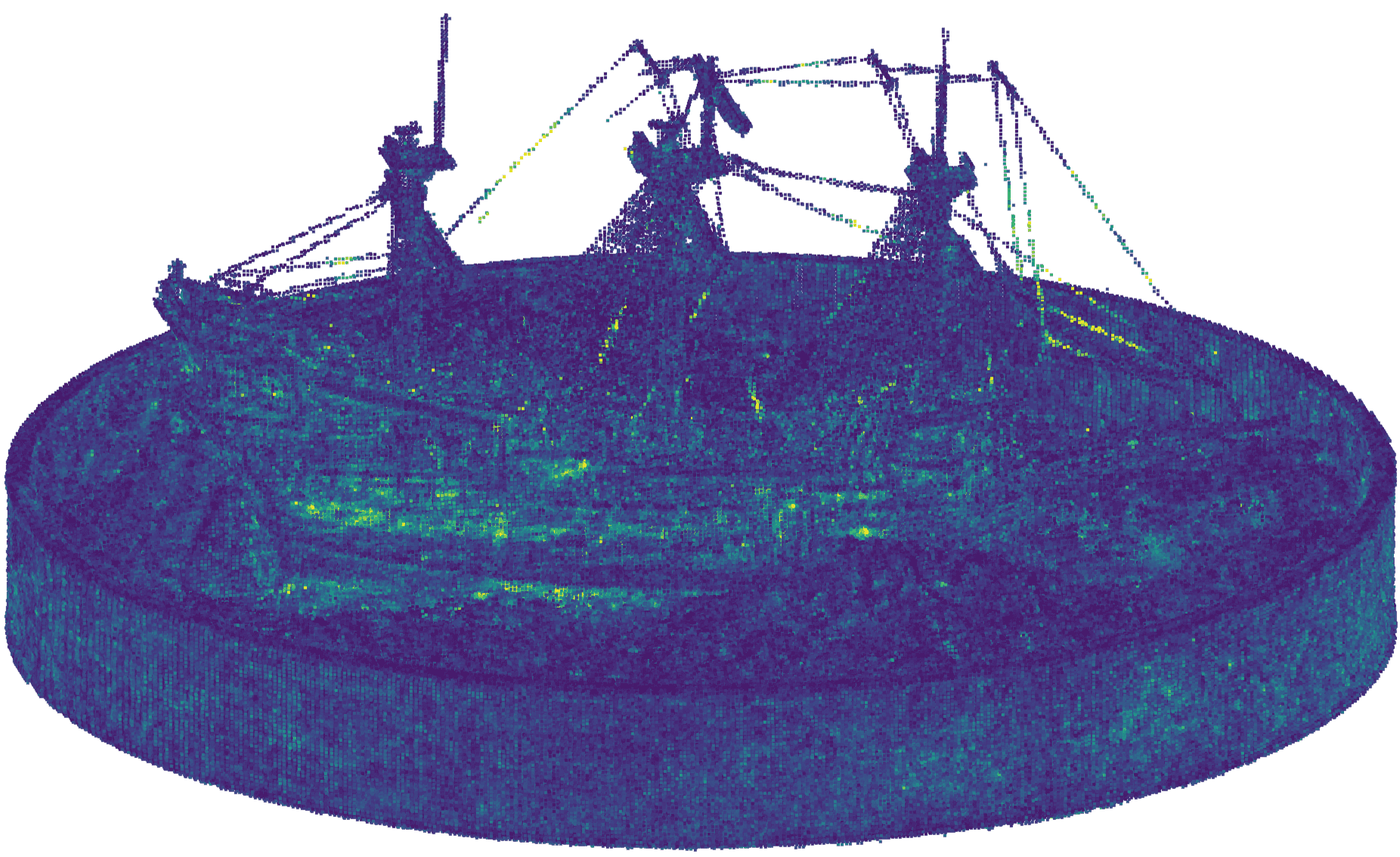}}
\subfigure{\label{fig:ship_noise_rot_unc_}
	\includegraphics[width=0.15\linewidth]{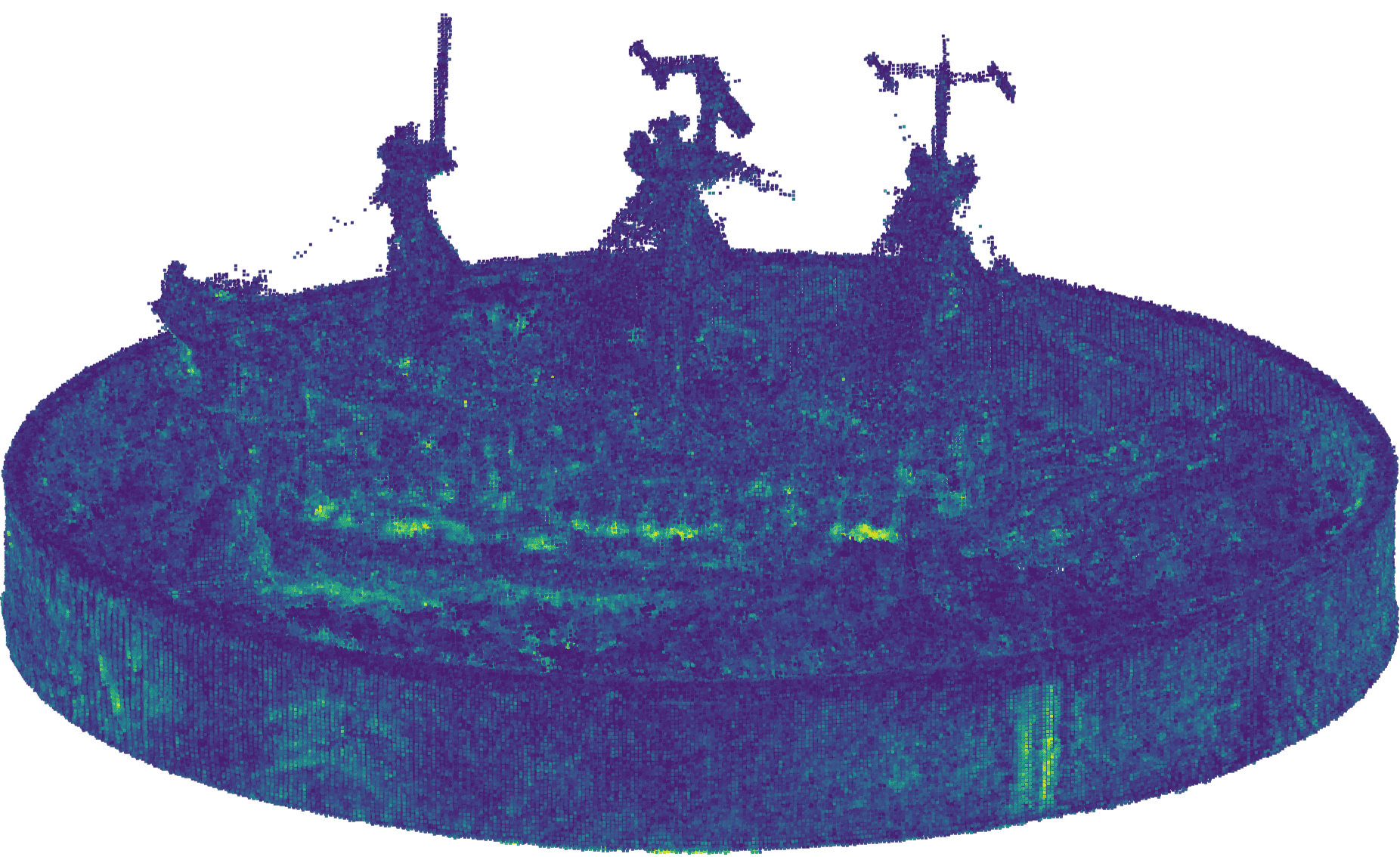}}\\
 	\vspace{-3mm}
\subfigure{
	\includegraphics[width=0.30\linewidth]{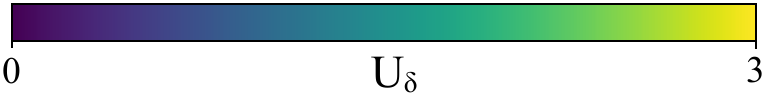}}
	\caption{Qualitative comparison on the NeRF synthetic dataset, showing points in the 3D grid with density above 15. Shown are the individual scenes for single NeRF (column 1), NeRF-Ensemble (column 2) as well as the density uncertainty \text{$\text{U}_{\delta}$} under different input data configurations: Original (column 3), image noise $\sigma_{\text{Im}}$ (column 4), pose noise $\sigma_{\text{t}}$ (column 5) and pose noise $\sigma_{\text{R}}$ (column 6). Density uncertainty values above 3 are set to 3 for clearer visualization.}
\label{fig:pointclouds_synthetic}
\end{figure}

\begin{figure}[H]
	\centering
\hspace{-0.25cm}
\raisebox{\dimexpr 0cm-\height}{NeRF}
\hspace{1cm}
\raisebox{\dimexpr 0cm-\height}{NeRF-Ensemble}
\hspace{0.8cm}
\raisebox{\dimexpr 0cm-\height}{Original}
\hspace{1.6cm}
\raisebox{\dimexpr 0cm-\height}{$\sigma_{\text{Im}}$}
\hspace{2.1cm}
\raisebox{\dimexpr 0cm-\height}{$\sigma_{\text{t}}$}
\hspace{2.1cm}
\raisebox{\dimexpr 0cm-\height}{$\sigma_{\text{R}}$}\\
	\vspace{3mm}
\rotatebox{90}{$\,\,\,\,\,\,\,\,\,\,\,\,\,\,\,\,$scan24}
\subfigure{\label{fig:Ficus_HoloLens_intern}
	\includegraphics[width=0.13\linewidth]{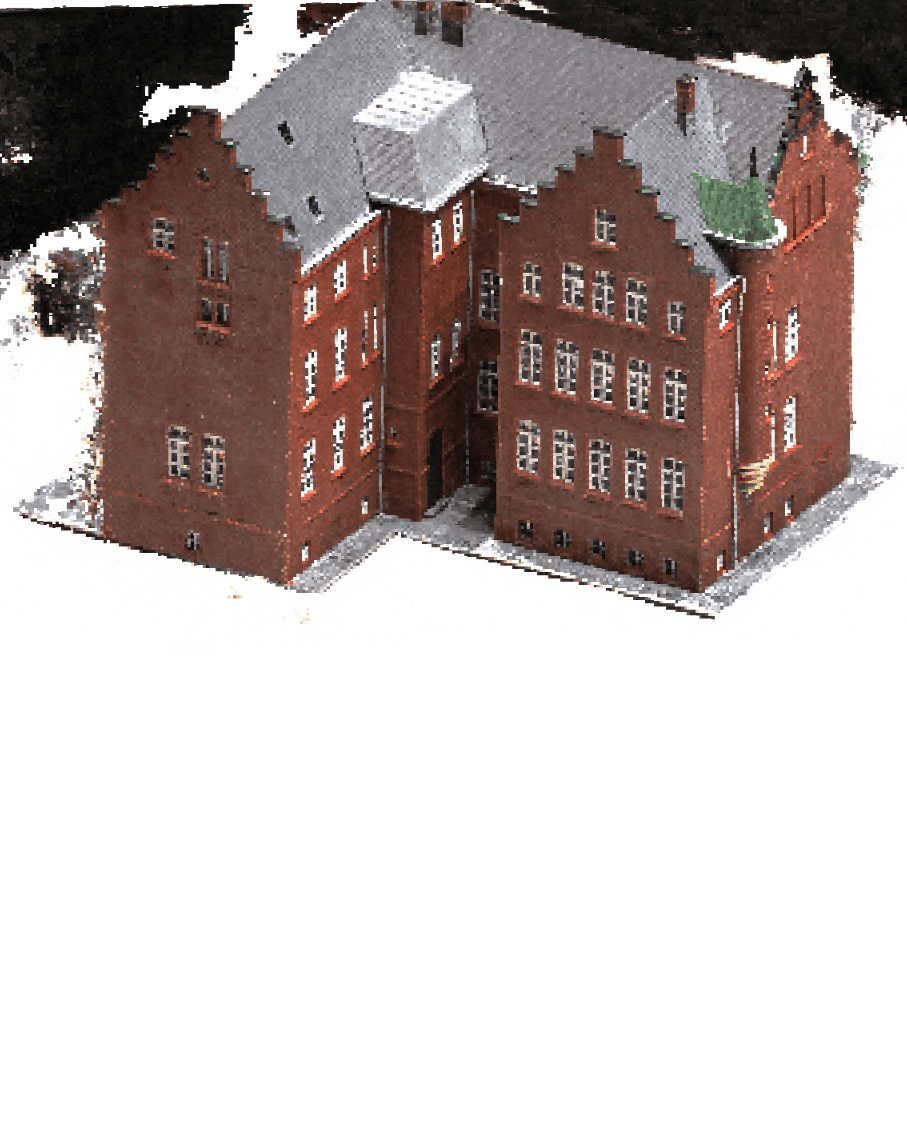}}
 	\hspace{0.31cm}
\subfigure{\label{fig:Ficus_HoloLens_intern_pose}  
     \includegraphics[width=0.13\linewidth]{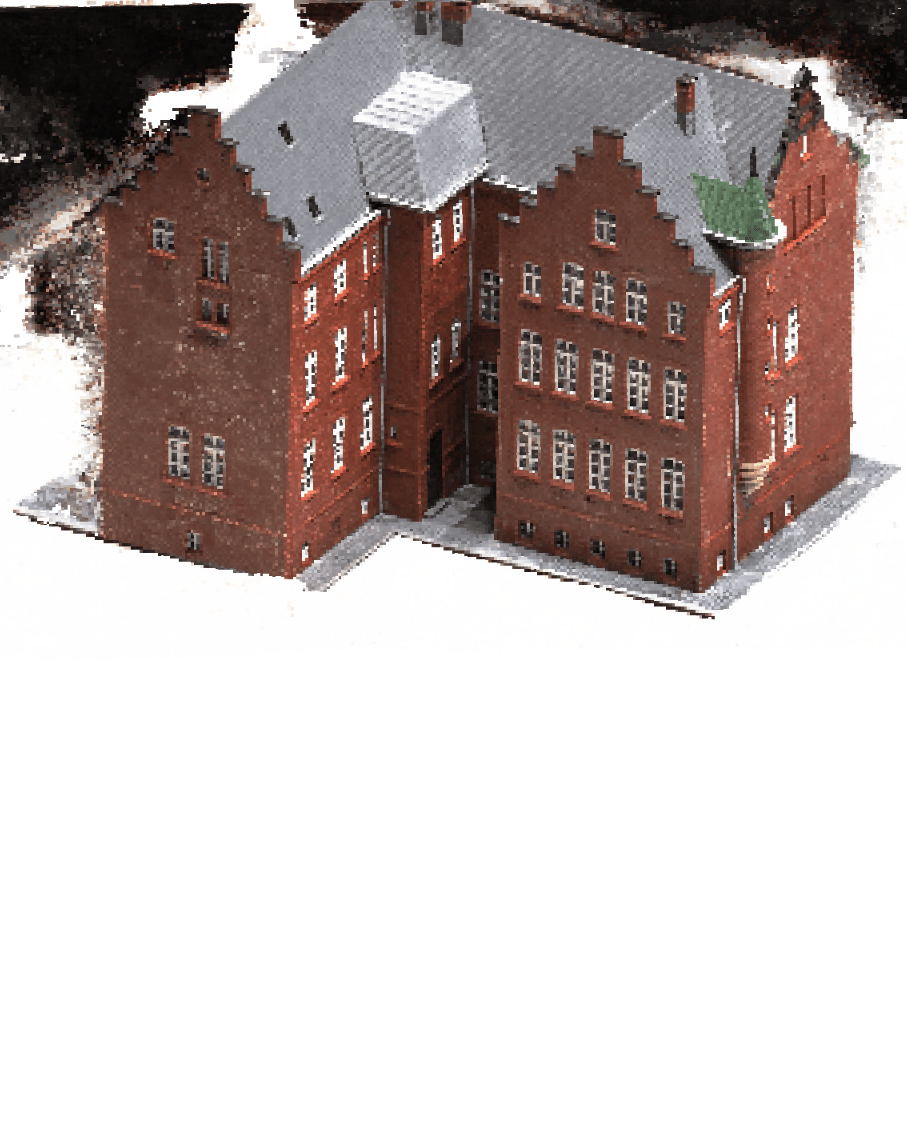}} 
      	\hspace{0.31cm}
\subfigure{\label{fig:Ficus_HoloLens_COLMAP_pose}  
     \includegraphics[width=0.13\linewidth]{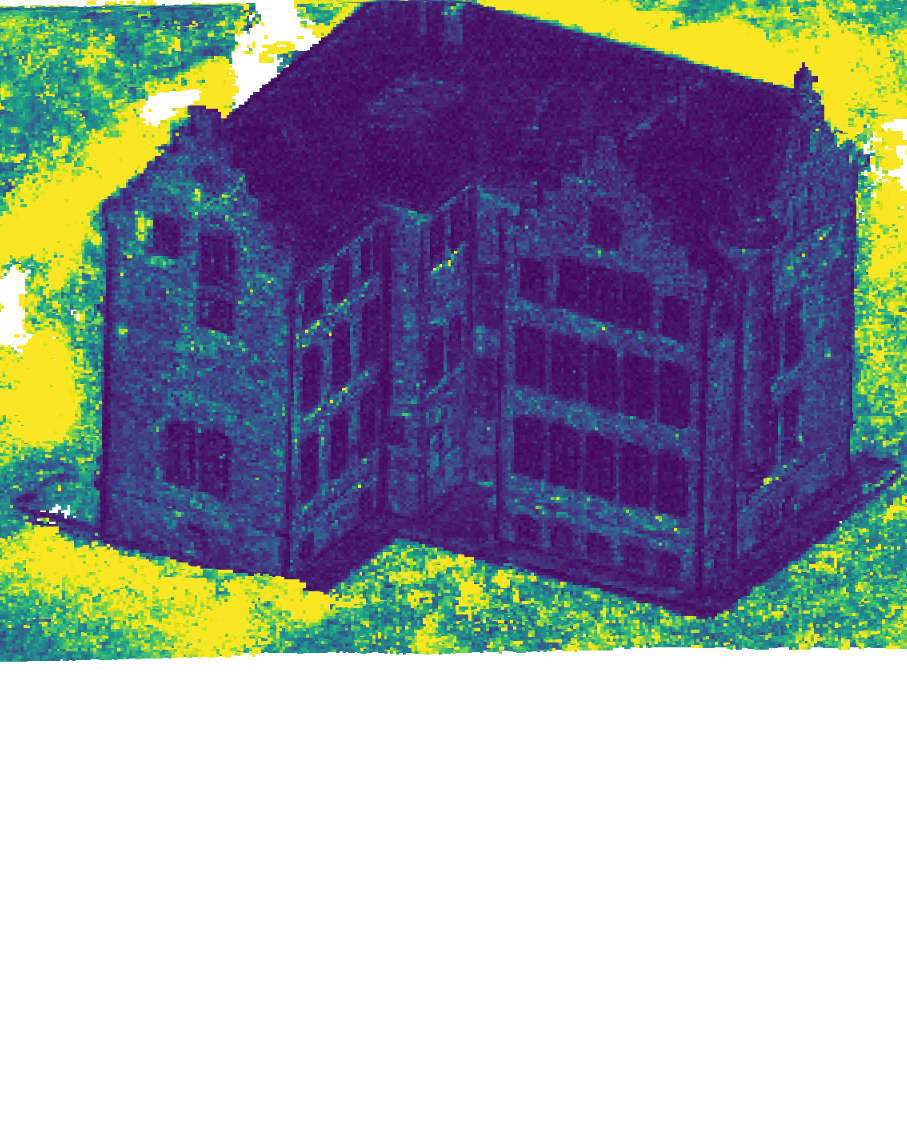}}
      	\hspace{0.31cm}
\subfigure{\label{fig:scan24_noise_image_unc_}
	\includegraphics[width=0.13\linewidth]{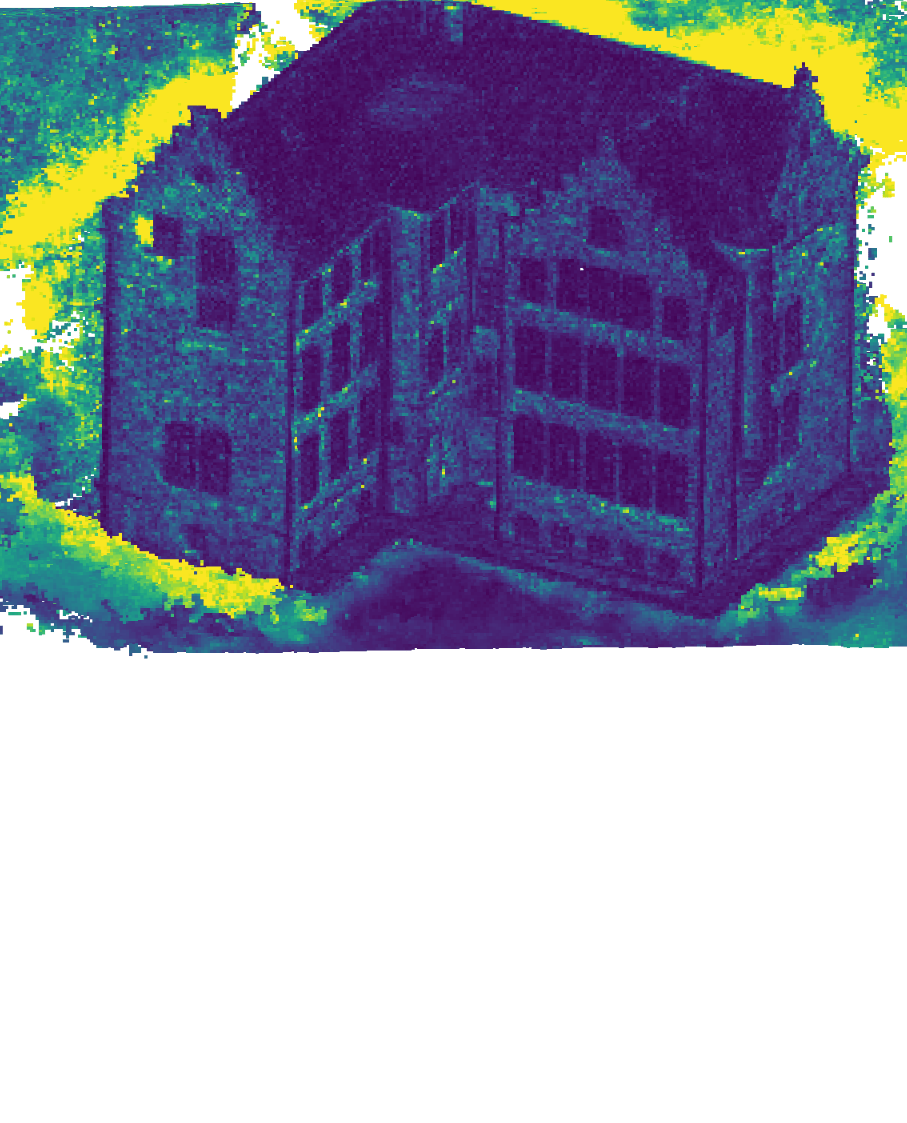}}
  	\hspace{0.31cm}
\subfigure{\label{fig:scan24_noise_trans_unc_}
	\includegraphics[width=0.13\linewidth]{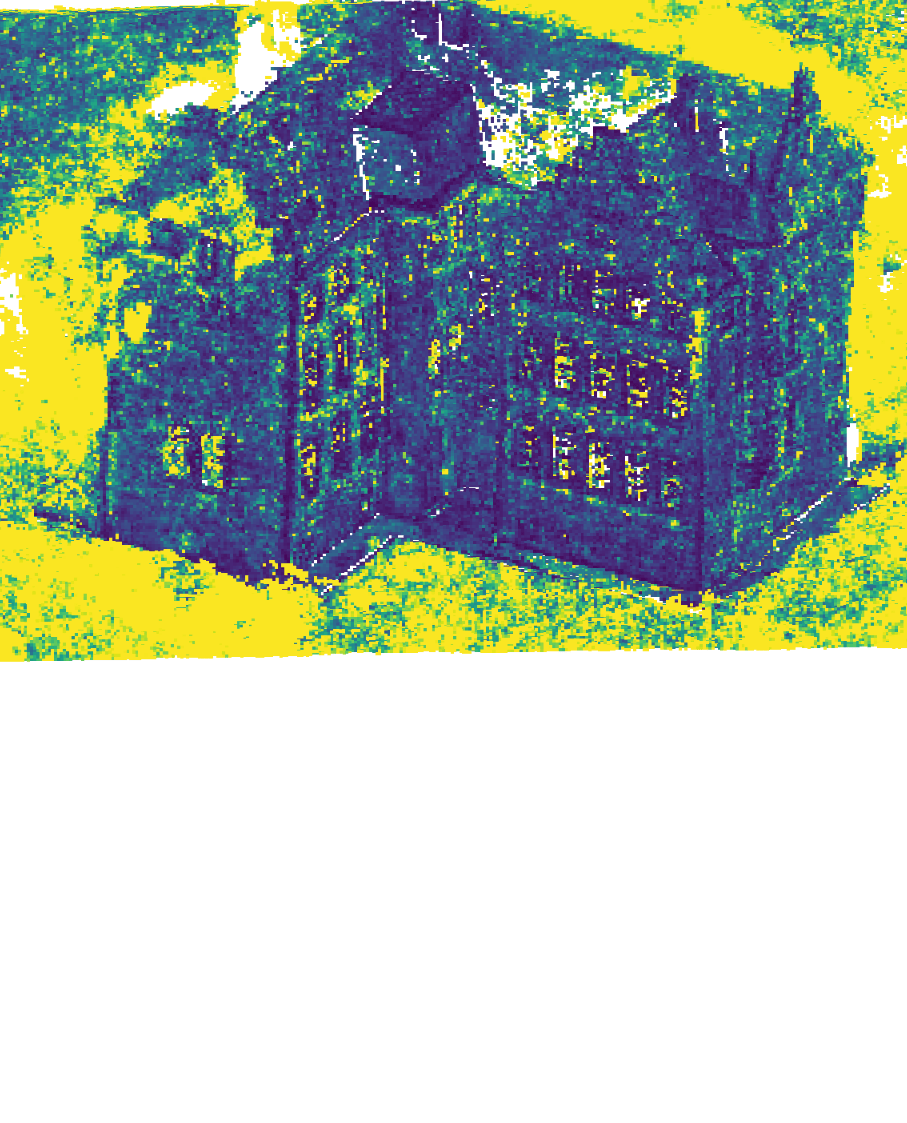}}
  	\hspace{0.31cm}
\subfigure{\label{fig:scan24_noise_rot_unc_}
	\includegraphics[width=0.13\linewidth]{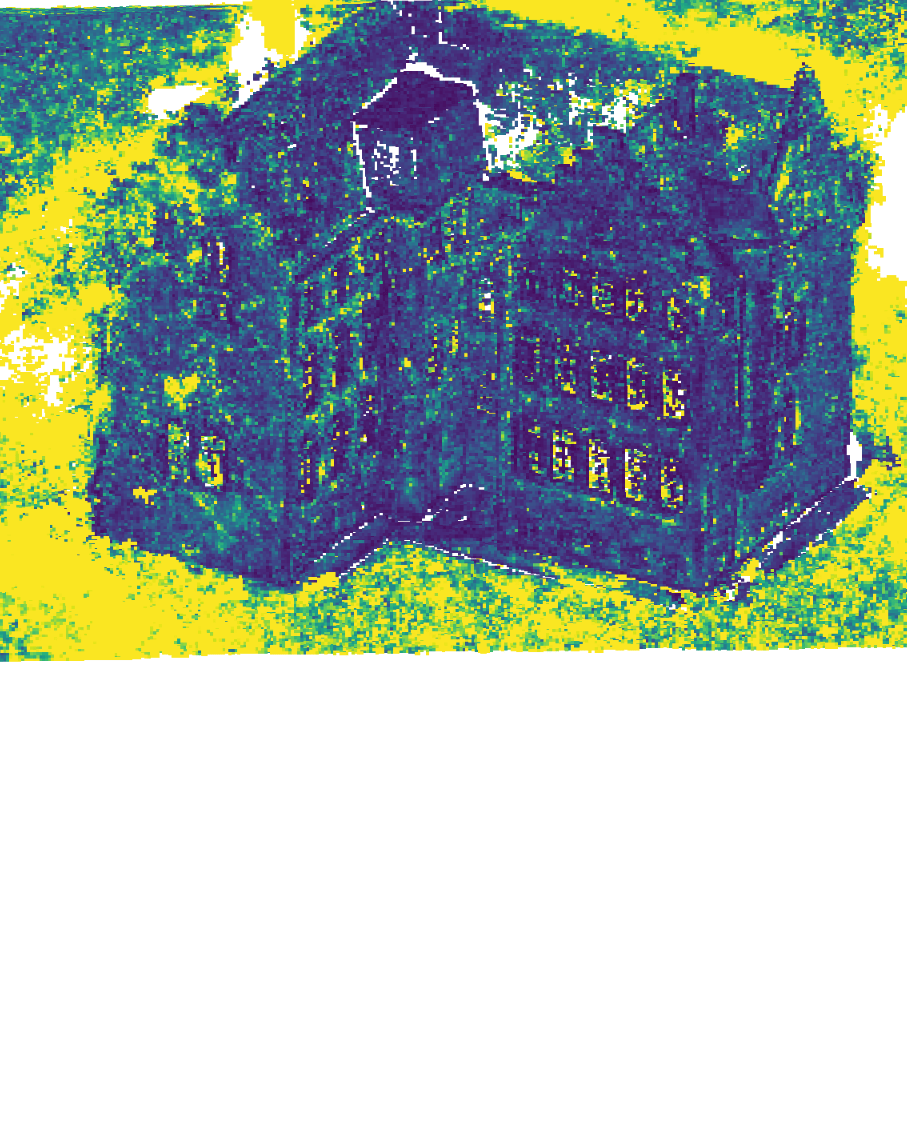}}\\
\vspace{-0.5cm}
\rotatebox{90}{$\,\,\,\,\,\,\,\,\,\,\,\,\,\,\,\,$scan37}
\subfigure{\label{fig:scan37_member_}
	\includegraphics[width=0.13\linewidth]{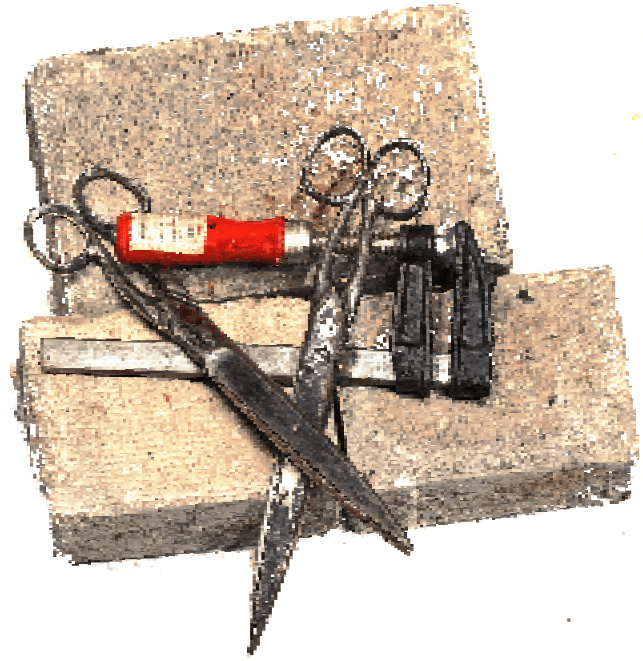}}
  	\hspace{0.31cm}
\subfigure{\label{fig:scan37_ensemble_}  
     \includegraphics[width=0.13\linewidth]{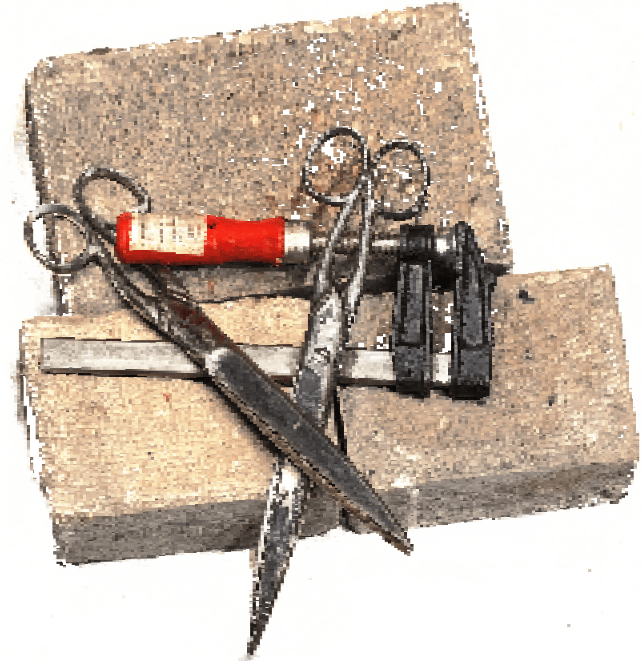}} 
      	\hspace{0.31cm}
\subfigure{\label{fig:scan37_unc_}  
     \includegraphics[width=0.13\linewidth]{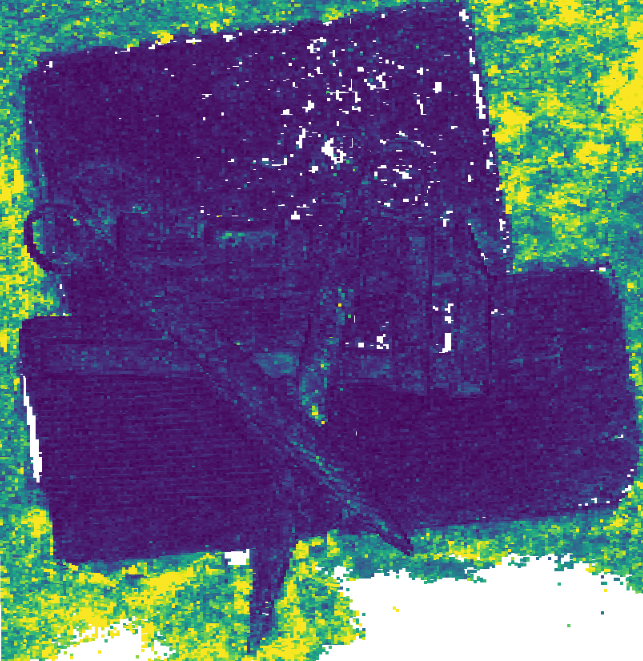}}
      	\hspace{0.31cm}
\subfigure{\label{fig:scan37_noise_image_unc_}
	\includegraphics[width=0.13\linewidth]{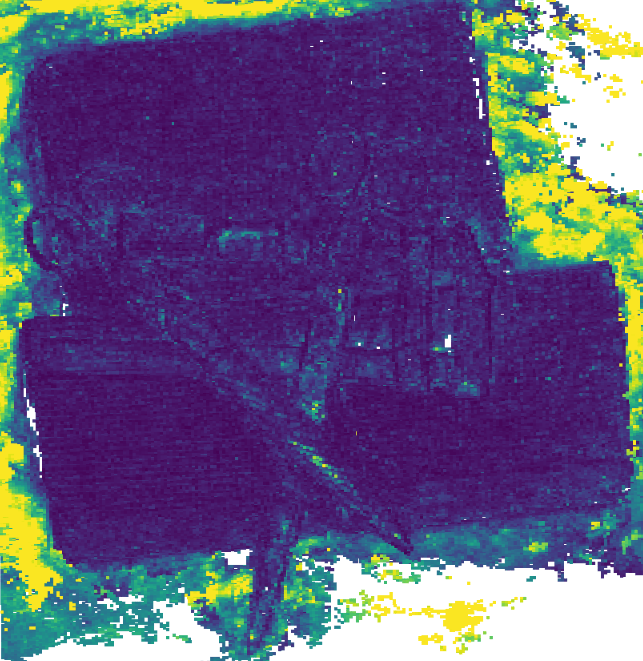}}
  	\hspace{0.31cm}
\subfigure{\label{fig:scan37_noise_trans_unc_}
	\includegraphics[width=0.13\linewidth]{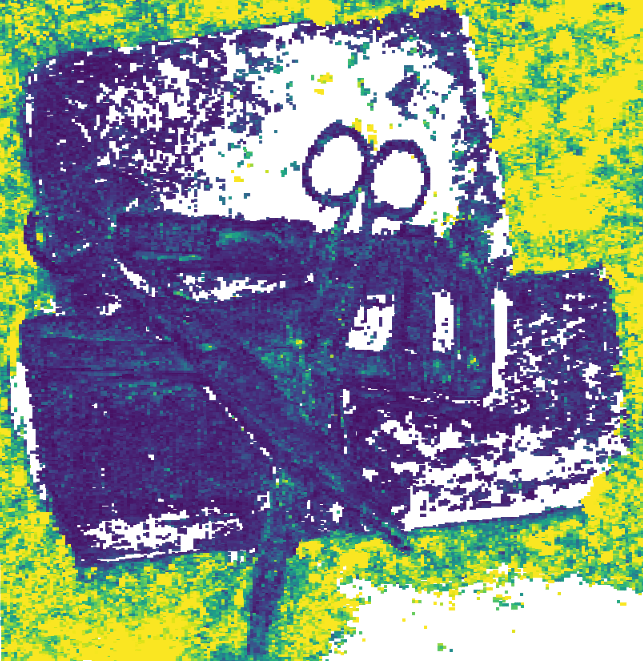}}
  	\hspace{0.31cm}
\subfigure{\label{fig:scan37_noise_rot_unc_}
	\includegraphics[width=0.13\linewidth]{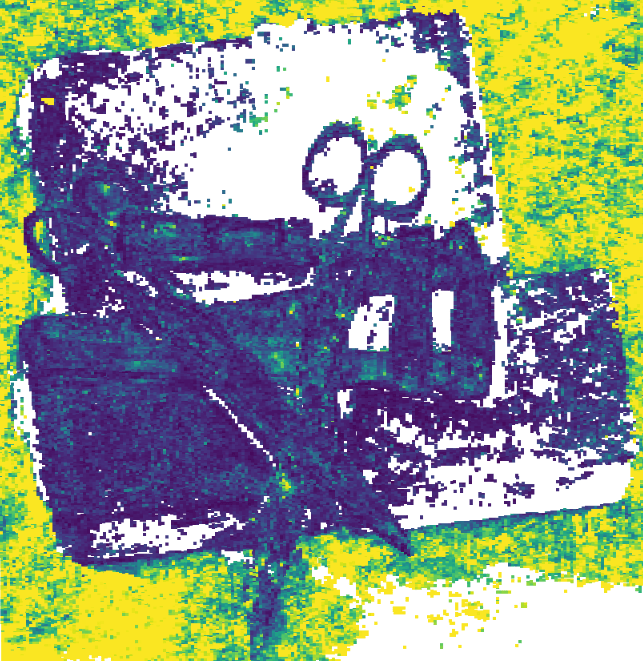}}\\
\rotatebox{90}{$\,\,\,\,\,\,\,\,\,\,\,\,\,\,\,\,$scan40}
\subfigure{\label{fig:scan40_member_}
	\includegraphics[width=0.13\linewidth]{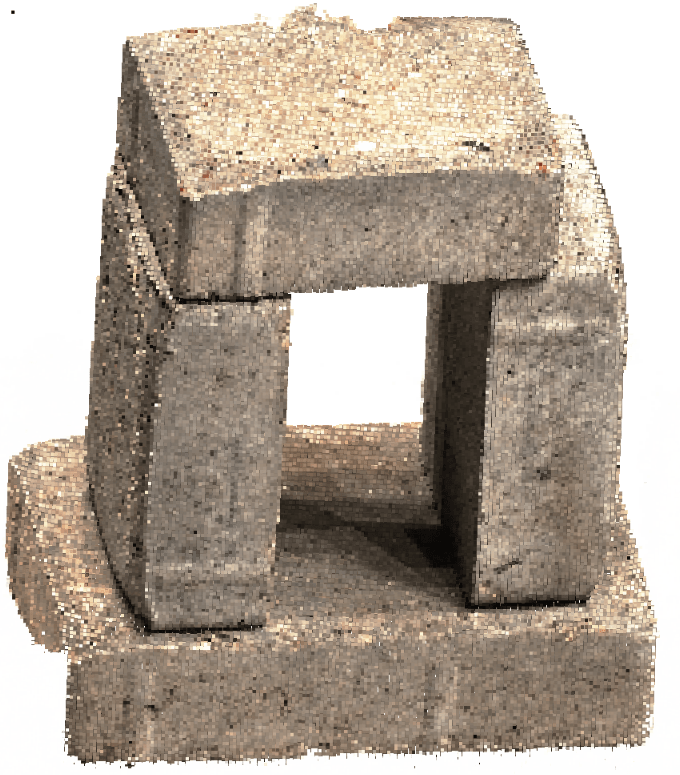}}
  	\hspace{0.3cm}
\subfigure{\label{fig:scan40_ensemble_}  
     \includegraphics[width=0.13\linewidth]{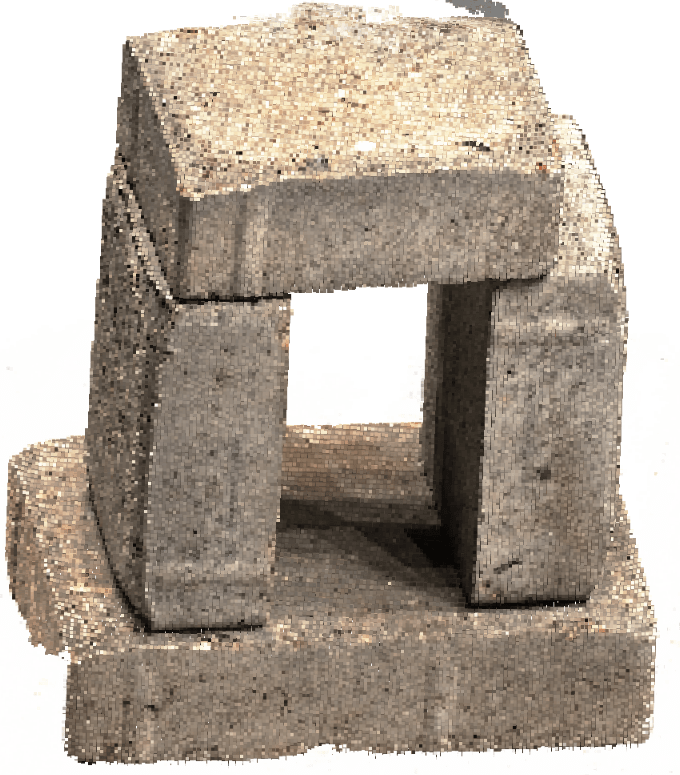}} 
      	\hspace{0.3cm}
\subfigure{\label{fig:scan40_unc_}  
     \includegraphics[width=0.13\linewidth]{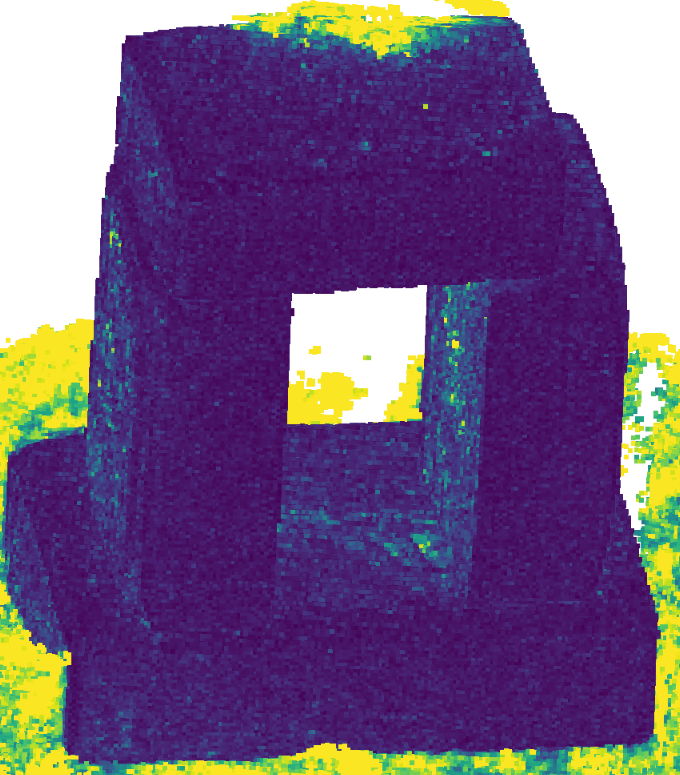}}
      	\hspace{0.3cm}
\subfigure{\label{fig:scan40_noise_image_unc_}
	\includegraphics[width=0.13\linewidth]{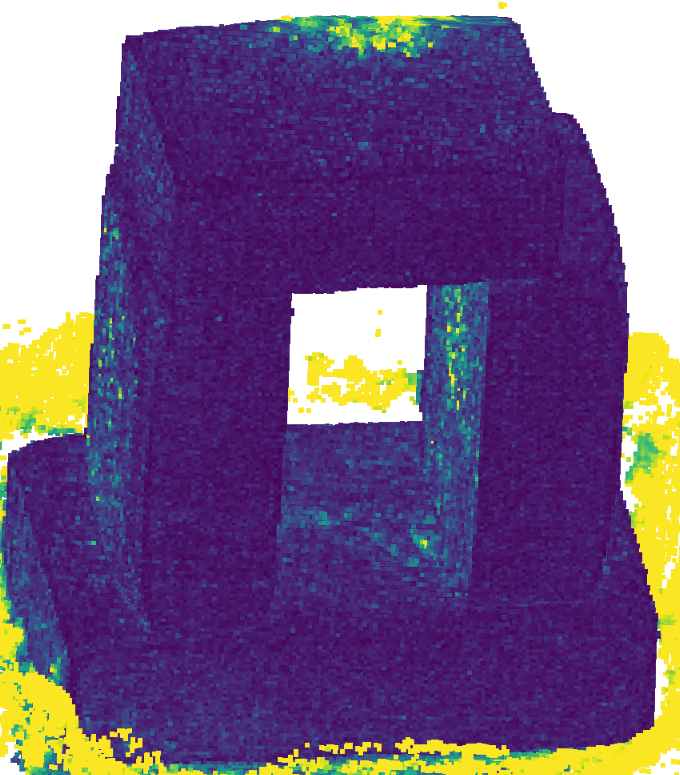}}
  	\hspace{0.3cm}
\subfigure{\label{fig:scan40_noise_trans_unc_}
	\includegraphics[width=0.13\linewidth]{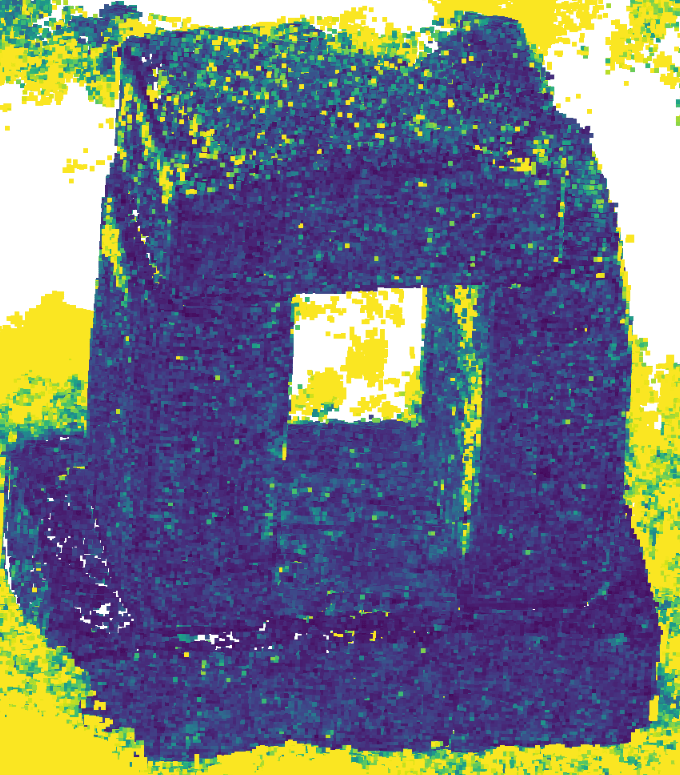}}
  	\hspace{0.3cm}
\subfigure{\label{fig:scan40_noise_rot_unc_}
	\includegraphics[width=0.13\linewidth]{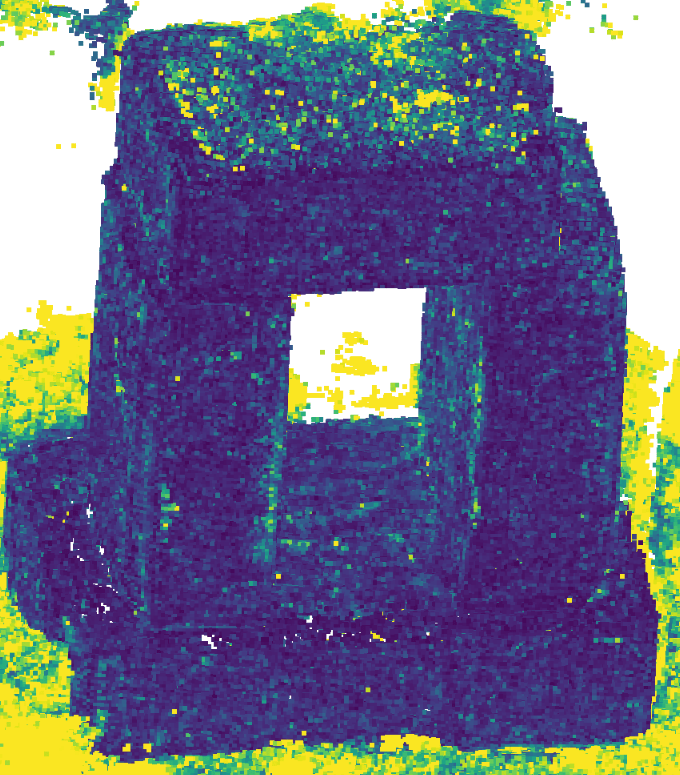}}\\
\rotatebox{90}{$\,\,\,\,\,\,\,\,\,\,\,\,\,\,\,\,$scan55}
\subfigure{\label{fig:scan55_member_}
	\includegraphics[width=0.13\linewidth]{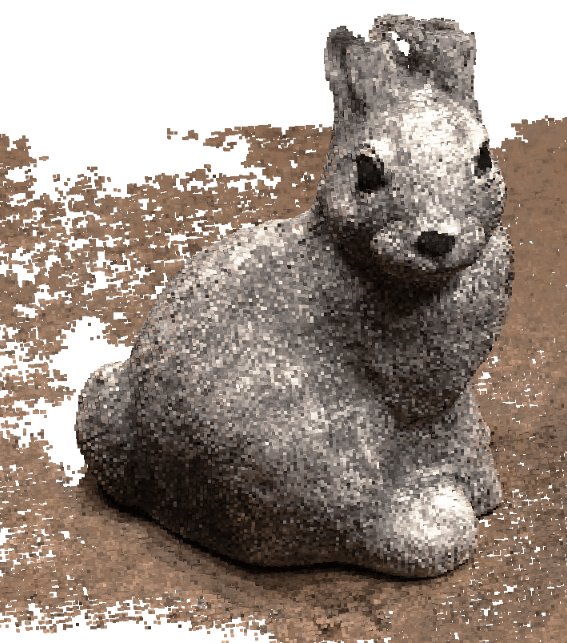}}
  	\hspace{0.3cm}
\subfigure{\label{fig:scan55_ensemble_}  
     \includegraphics[width=0.13\linewidth]{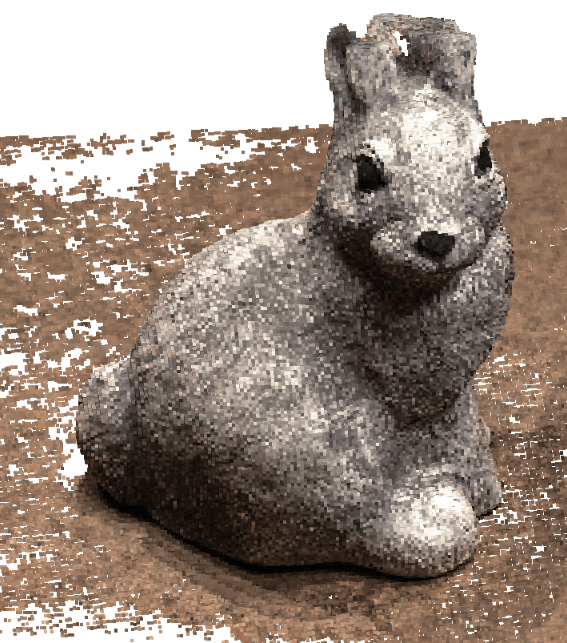}} 
      	\hspace{0.3cm}
\subfigure{\label{fig:scan55_unc_}  
     \includegraphics[width=0.13\linewidth]{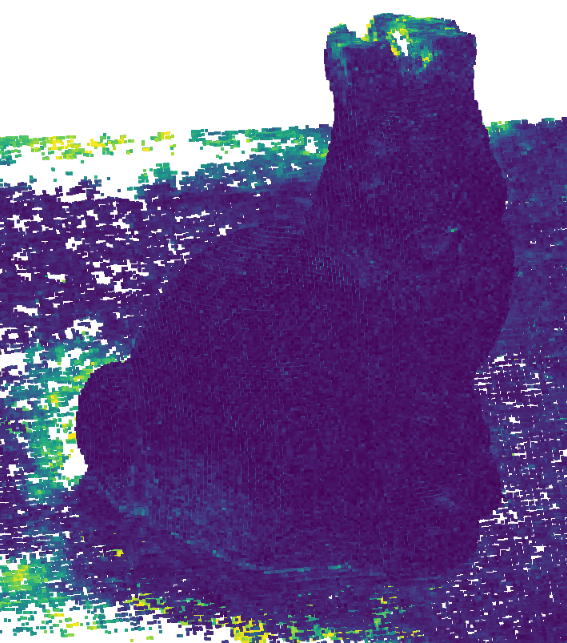}}
      	\hspace{0.3cm}
\subfigure{\label{fig:scan55_noise_image_unc_}
	\includegraphics[width=0.13\linewidth]{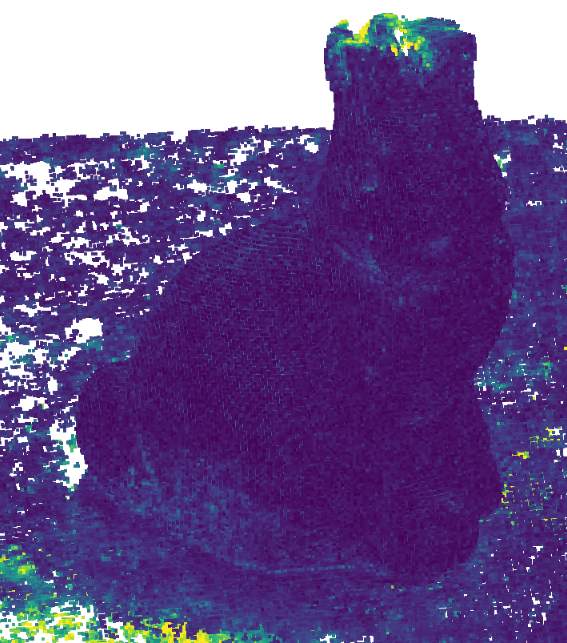}}
  	\hspace{0.3cm}
\subfigure{\label{fig:scan55_noise_trans_unc_}
	\includegraphics[width=0.13\linewidth]{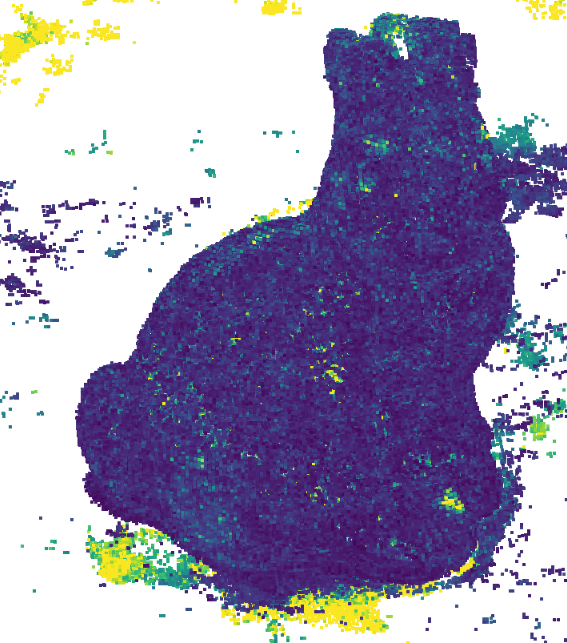}}
  	\hspace{0.3cm}
\subfigure{\label{fig:scan55_noise_rot_unc_}
	\includegraphics[width=0.13\linewidth]{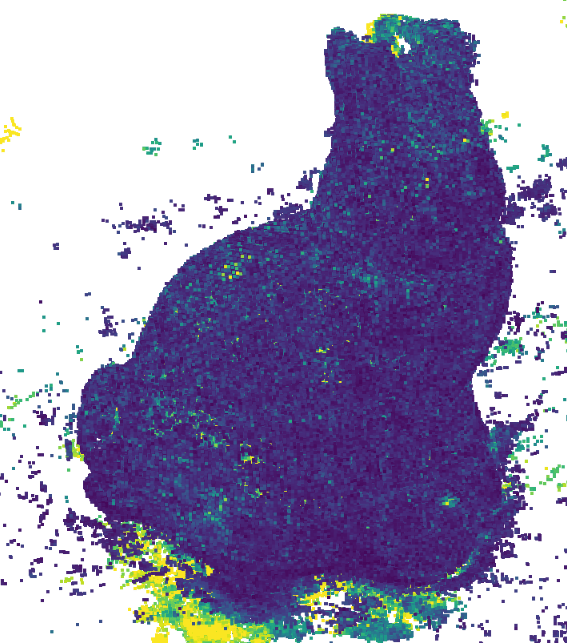}}\\
\rotatebox{90}{$\,\,\,\,\,\,\,\,\,\,\,\,\,\,\,\,$scan63}
\subfigure{\label{fig:scan63_member}
	\includegraphics[width=0.13\linewidth]{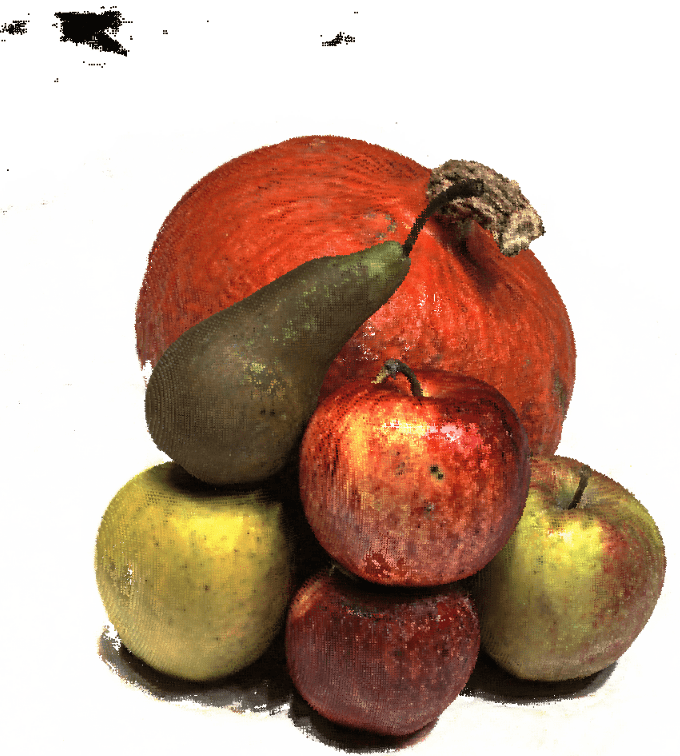}}
  	\hspace{0.3cm}
\subfigure{\label{fig:scan63_ensemble}  
     \includegraphics[width=0.13\linewidth]{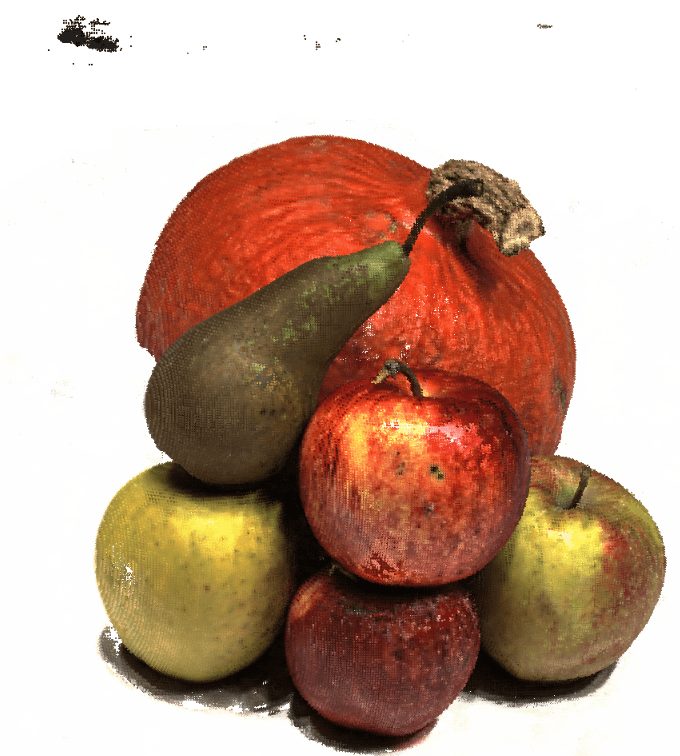}} 
      	\hspace{0.3cm}
\subfigure{\label{fig:scan63_ensemble_unc}  
     \includegraphics[width=0.13\linewidth]{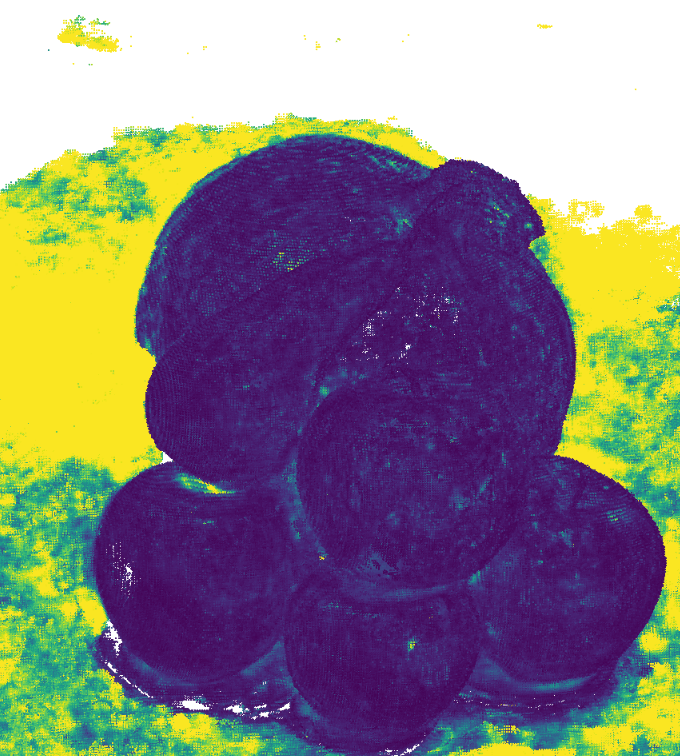}}
      	\hspace{0.3cm}
\subfigure{\label{fig:scan63_ensemble_images_unc}
	\includegraphics[width=0.13\linewidth]{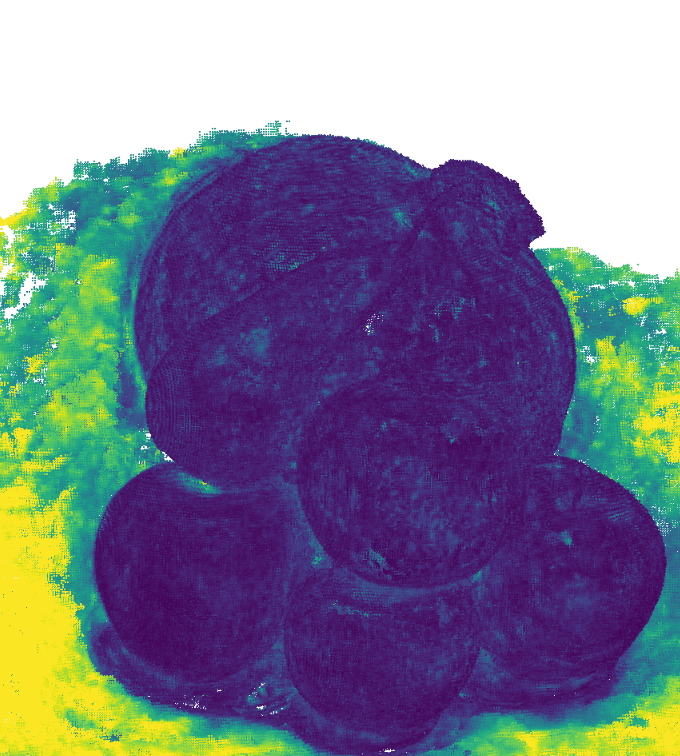}}
  	\hspace{0.3cm}
\subfigure{\label{fig:scan63_ensemble_trans_unc}
	\includegraphics[width=0.13\linewidth]{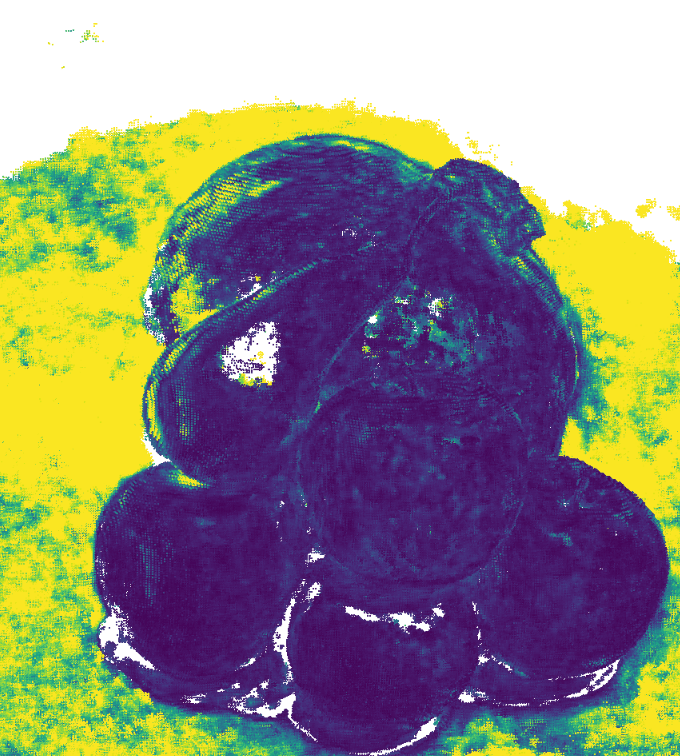}}
  	\hspace{0.3cm}
\subfigure{\label{fig:scan63_ensemble_rot_unc}
	\includegraphics[width=0.13\linewidth]{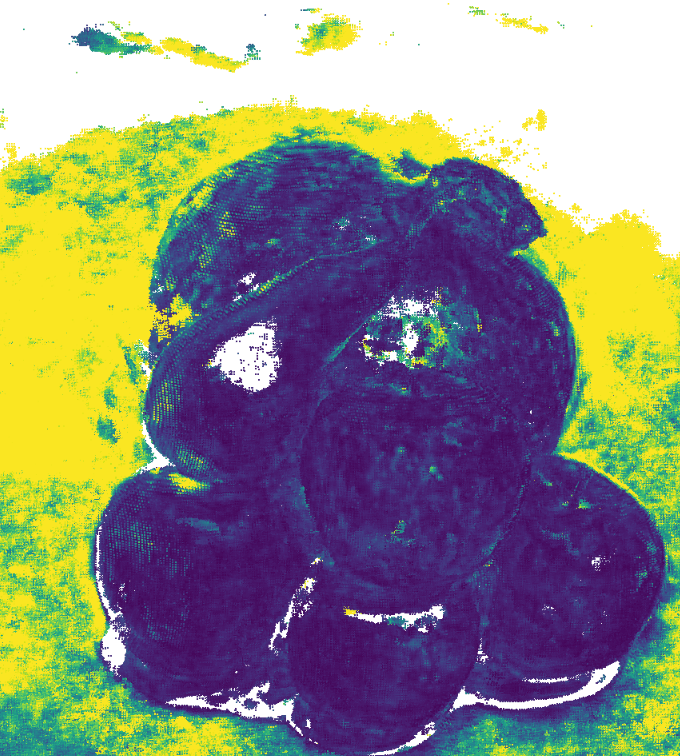}}\\
	\rotatebox{90}{$\,\,\,\,\,\,\,\,\,\,\,\,\,\,\,\,$scan114}
\subfigure{\label{fig:scan114_member_}
	\includegraphics[width=0.13\linewidth]{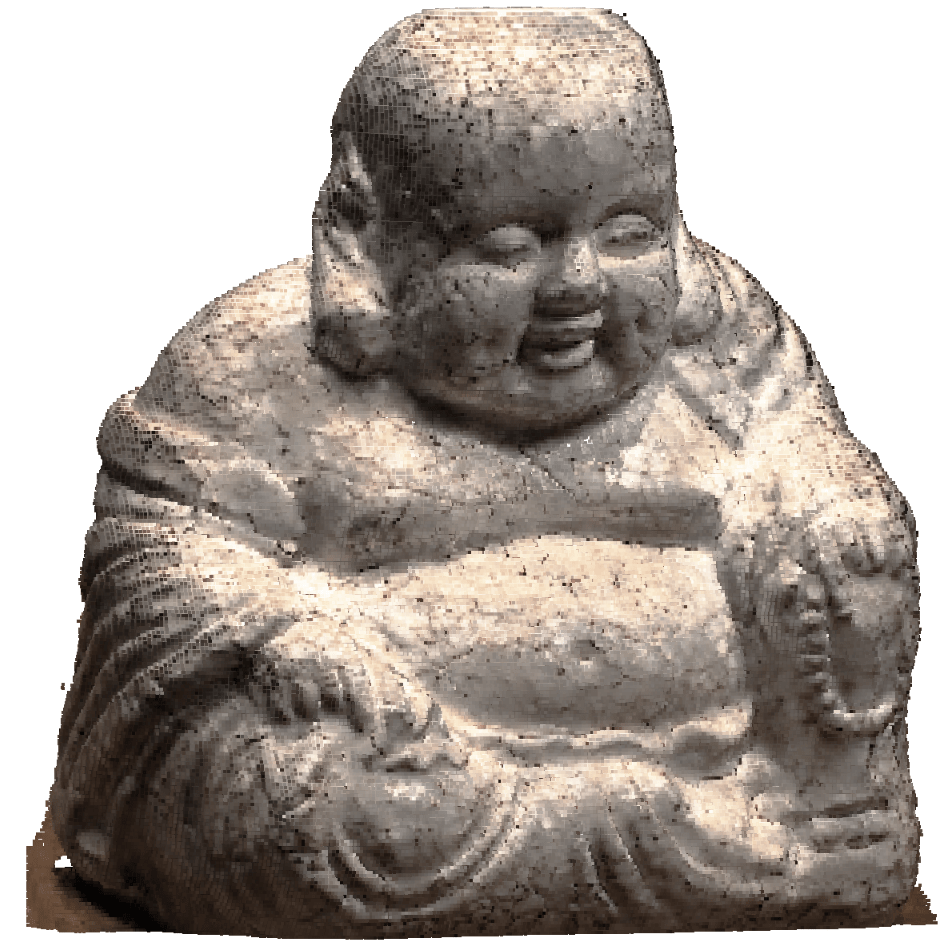}}
  	\hspace{0.3cm}
\subfigure{\label{fig:scan114_ensemble_}  
     \includegraphics[width=0.13\linewidth]{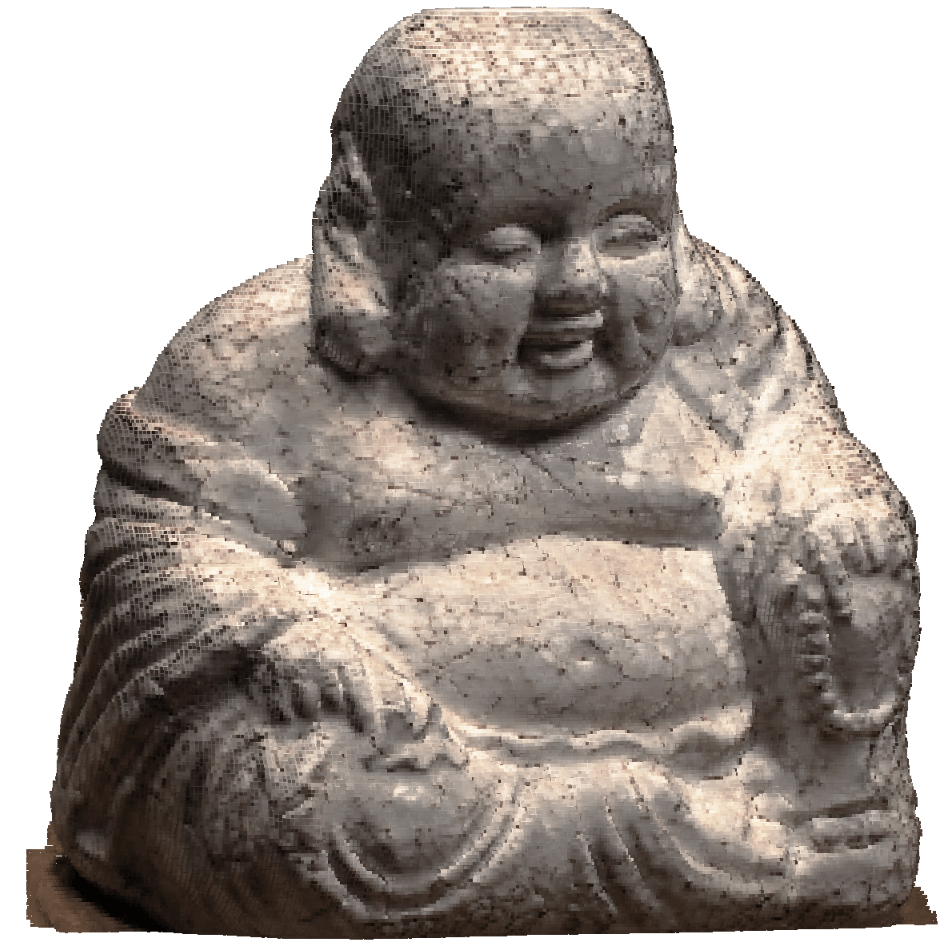}} 
      	\hspace{0.3cm}
\subfigure{\label{fig:scan114_unc_}  
     \includegraphics[width=0.13\linewidth]{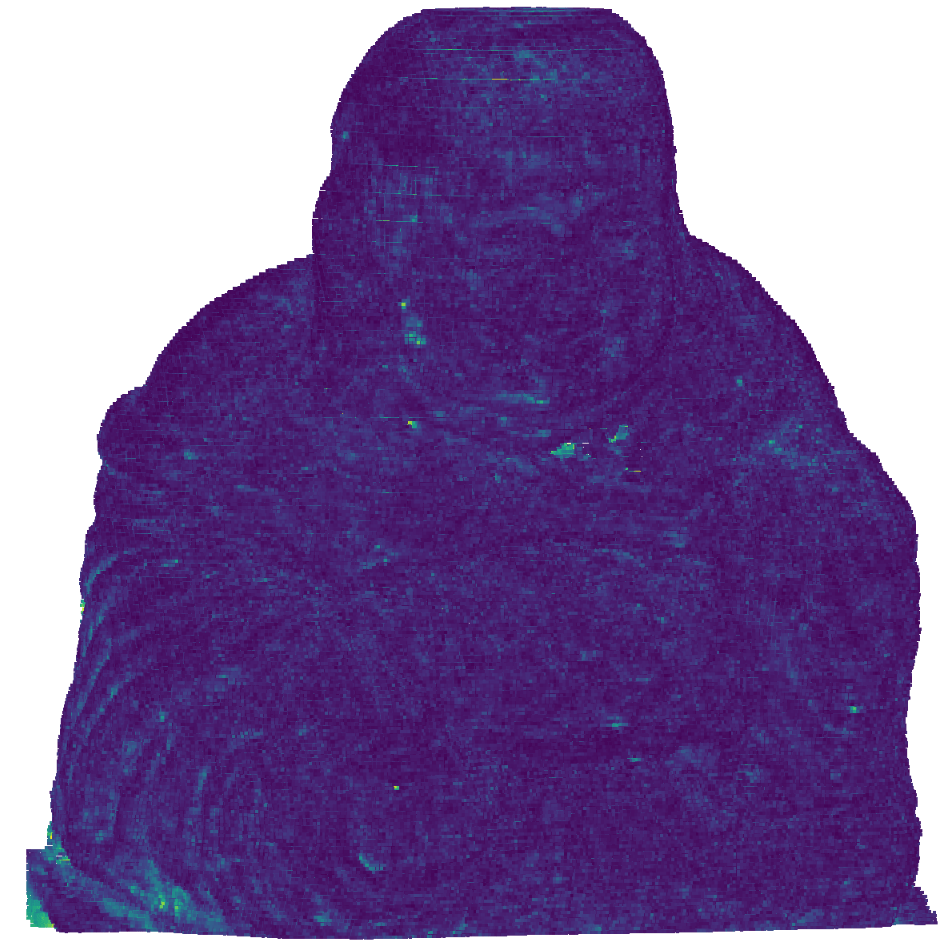}}
      	\hspace{0.3cm}
\subfigure{\label{fig:scan114_noise_image_unc_}
	\includegraphics[width=0.13\linewidth]{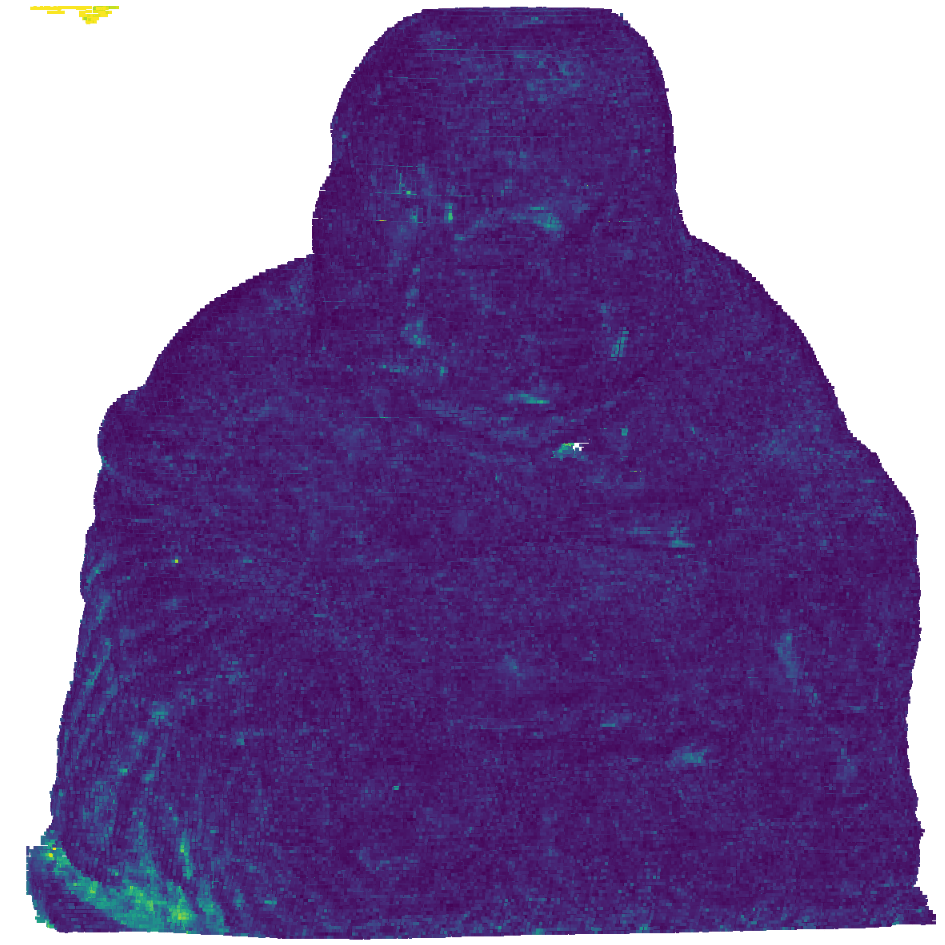}}
  	\hspace{0.3cm}
\subfigure{\label{fig:scan114_noise_trans_unc_}
	\includegraphics[width=0.13\linewidth]{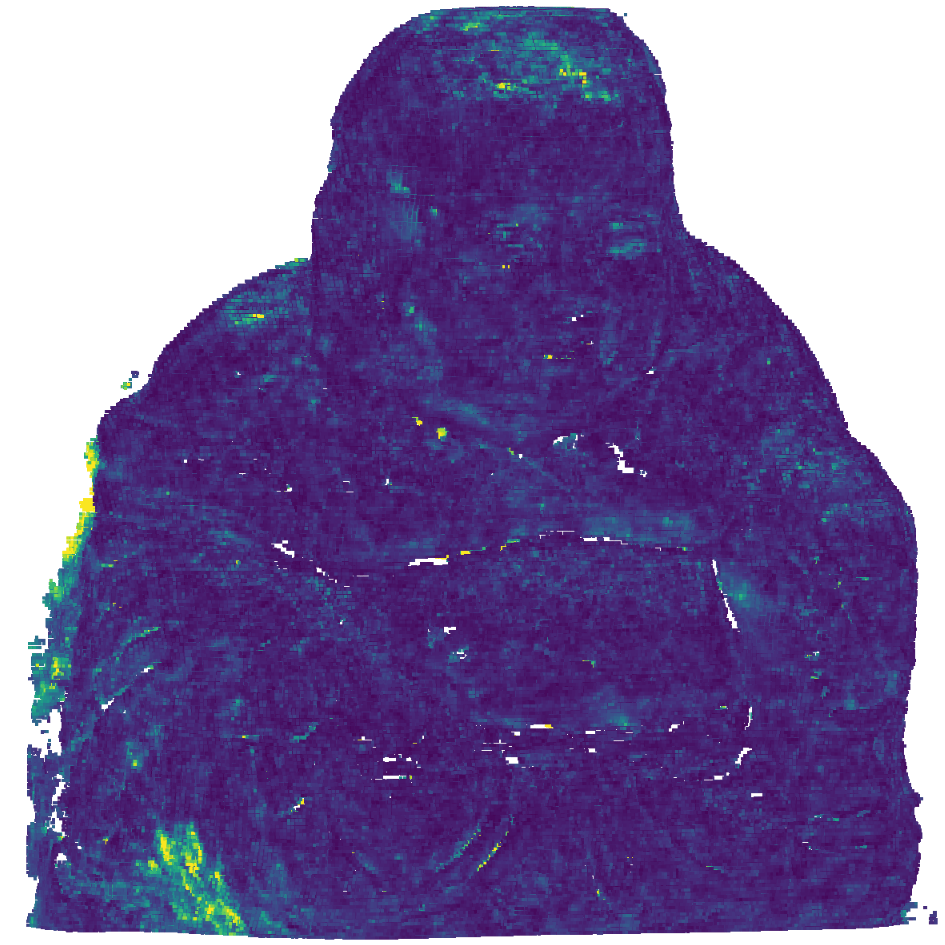}}
  	\hspace{0.3cm}
\subfigure{\label{fig:scan114_noise_rot_unc_}
	\includegraphics[width=0.13\linewidth]{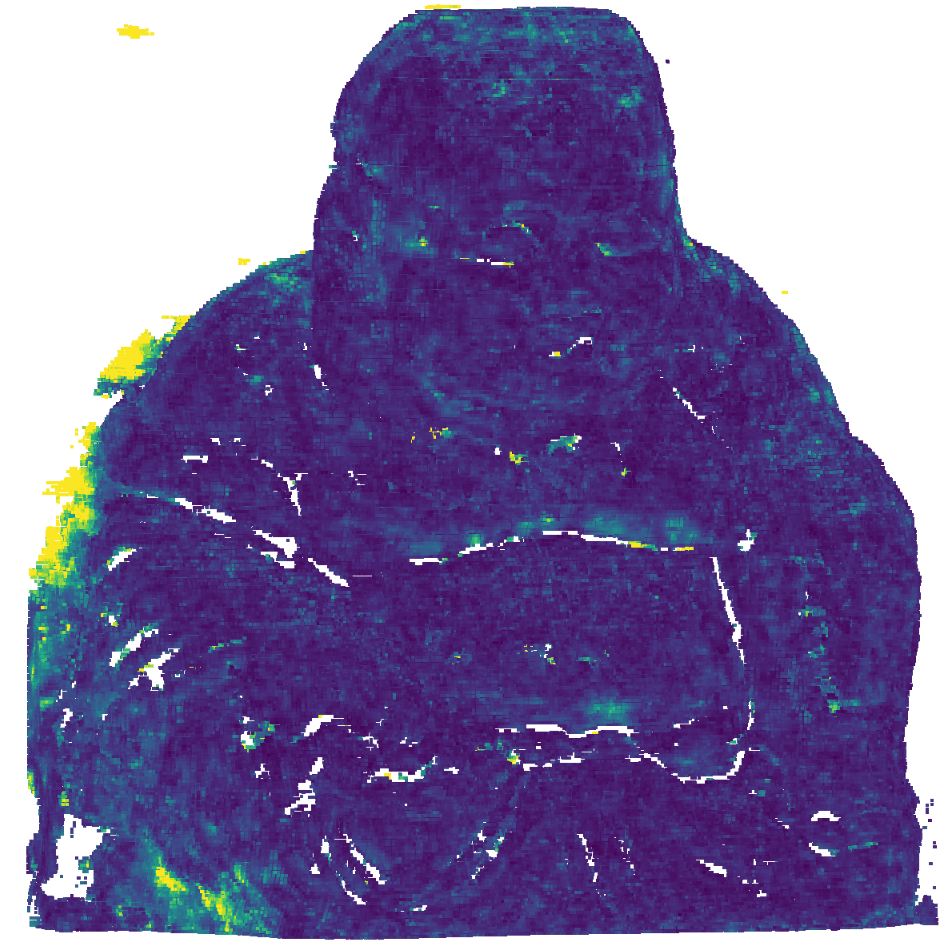}}\\
	\subfigure{
	\includegraphics[width=0.30\linewidth]{figures/colorbar_uncertainty_.pdf}}
	\caption{Qualitative comparison on the DTU dataset, showing points in the 3D grid with density above 15. Shown are the individual scenes for single NeRF (column 1), NeRF-Ensemble (column 2) as well as the density uncertainty \text{$\text{U}_{\delta}$} under different input data configurations: Original (column 3), image noise $\sigma_{\text{Im}}$ (column 4), pose noise $\sigma_{\text{t}}$ (column 5) and pose noise $\sigma_{\text{R}}$ (column 6). Density uncertainty values above 3 are set to 3 for clearer visualization.}
\label{fig:pointclouds_dtu}
\end{figure}

\begin{figure}[H]
	\centering
\rotatebox{90}{$\,\,\,\,\,$internal poses}
\subfigure[NeRF]{\label{fig:hololens_member_}
	\includegraphics[width=0.2\linewidth]{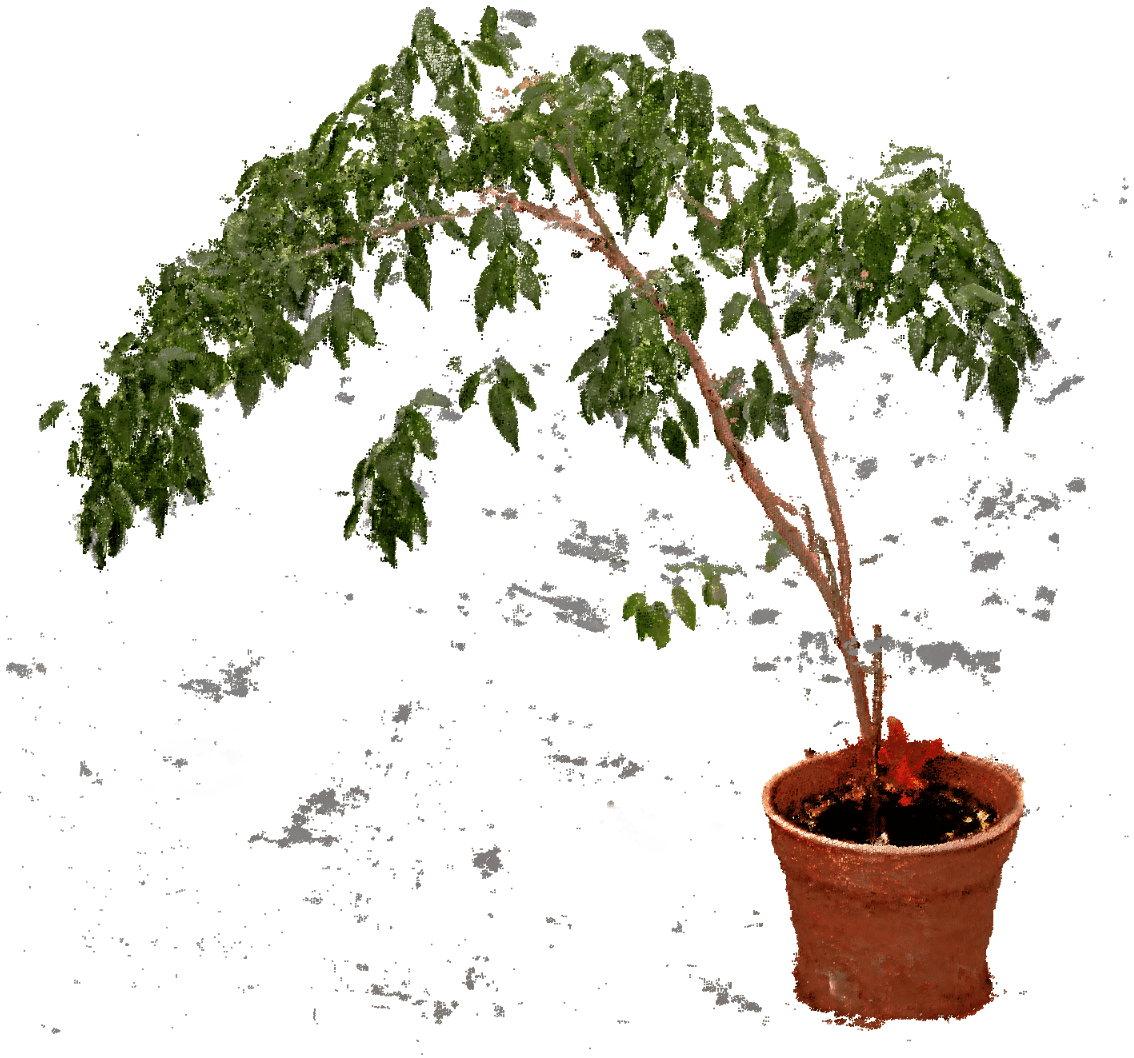}}
 \hspace{4mm}
\subfigure[NeRF-Ensemble]{\label{fig:hololens_ensemble_}  
     \includegraphics[width=0.2\linewidth]{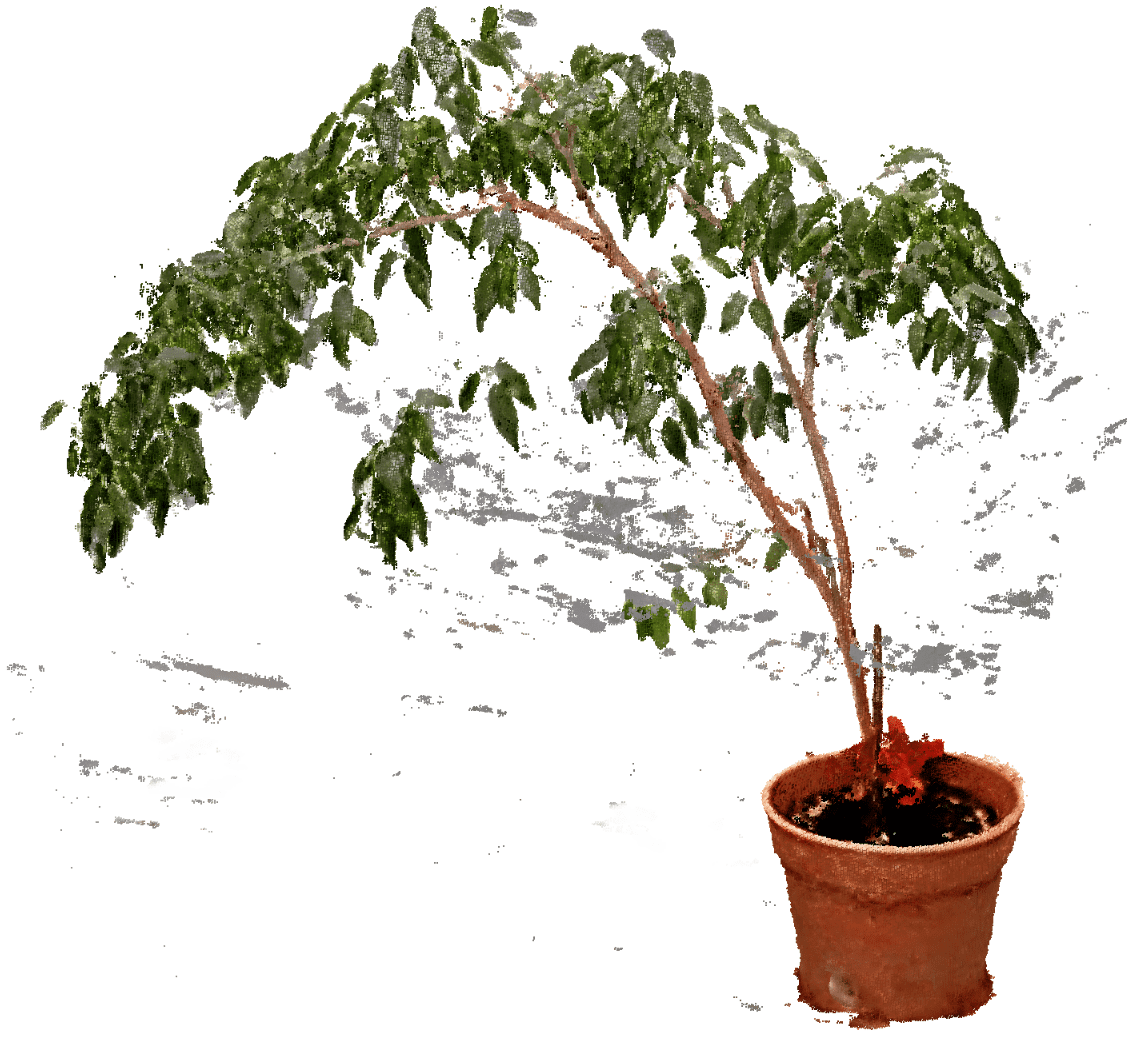}}
     \hspace{4mm}
\subfigure[density]{\label{fig:hololens_density_}
	\includegraphics[width=0.2\linewidth]{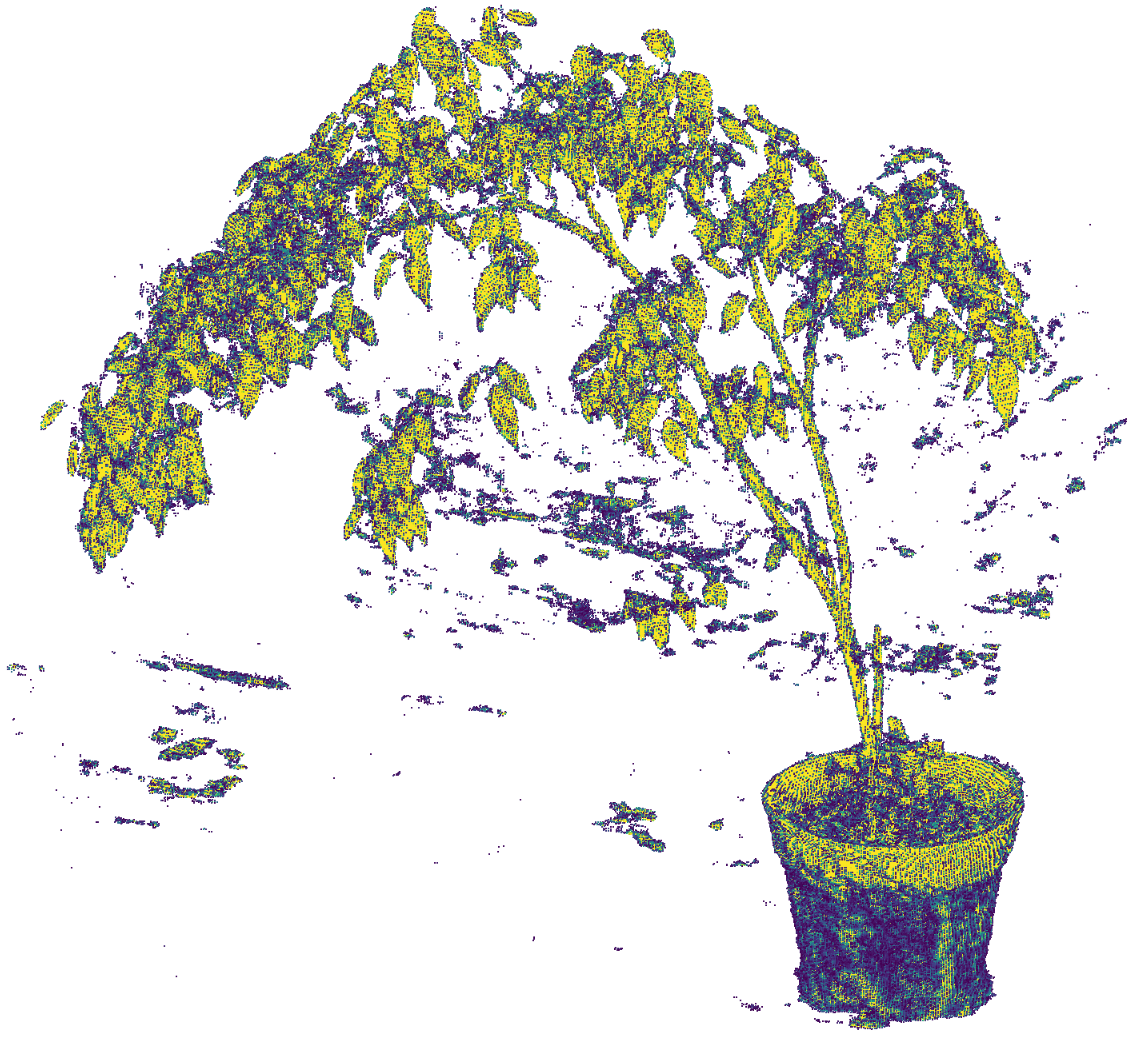}}
 \hspace{4mm}
\subfigure[uncertainty]{\label{fig:hololens_unc_}\includegraphics[width=0.2\columnwidth]{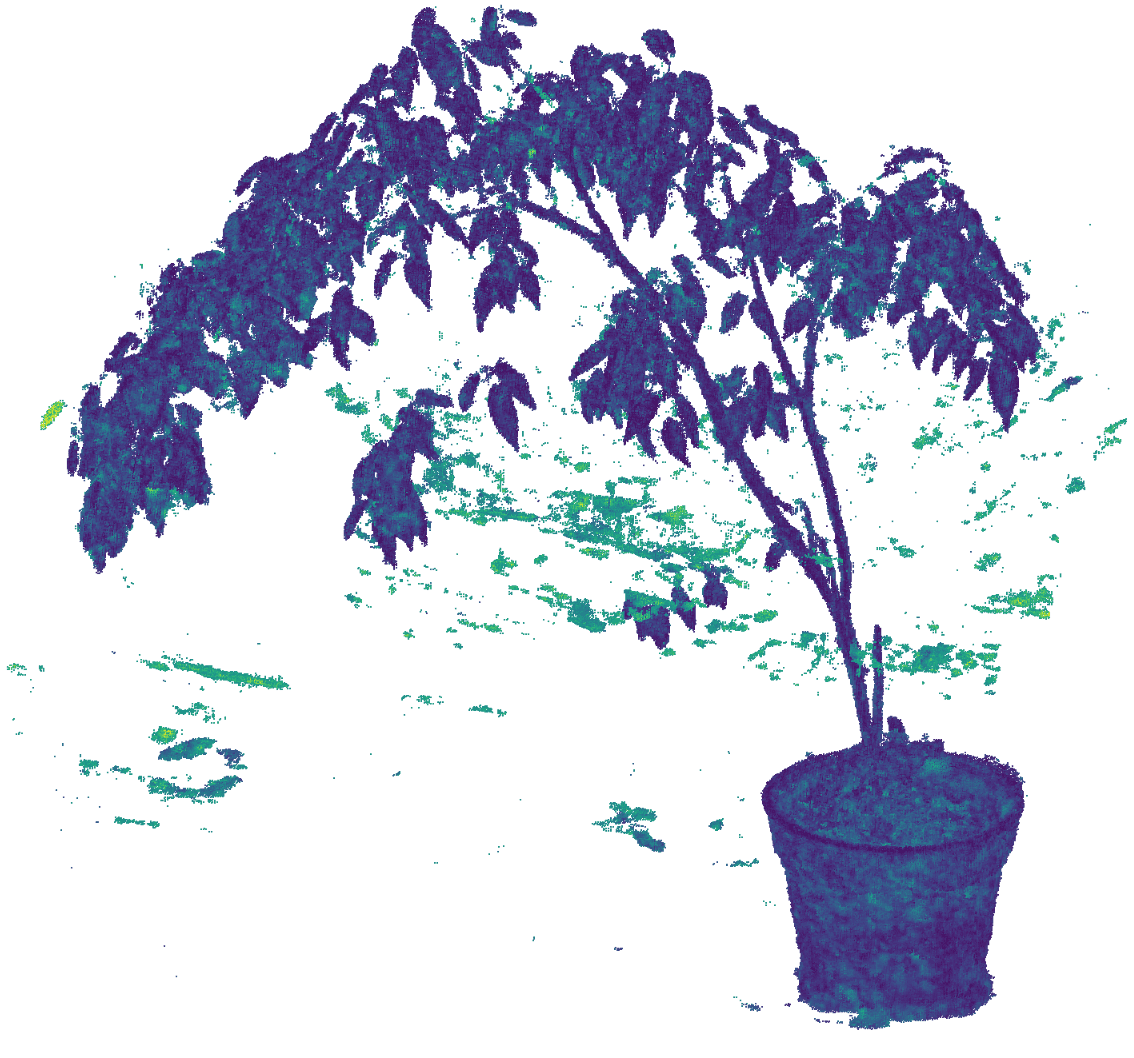}}\\
\rotatebox{90}{$\,\,\,\,\,$refined internal poses}
\subfigure[NeRF]{\label{fig:hololens_refined_member_}
	\includegraphics[width=0.2\linewidth]{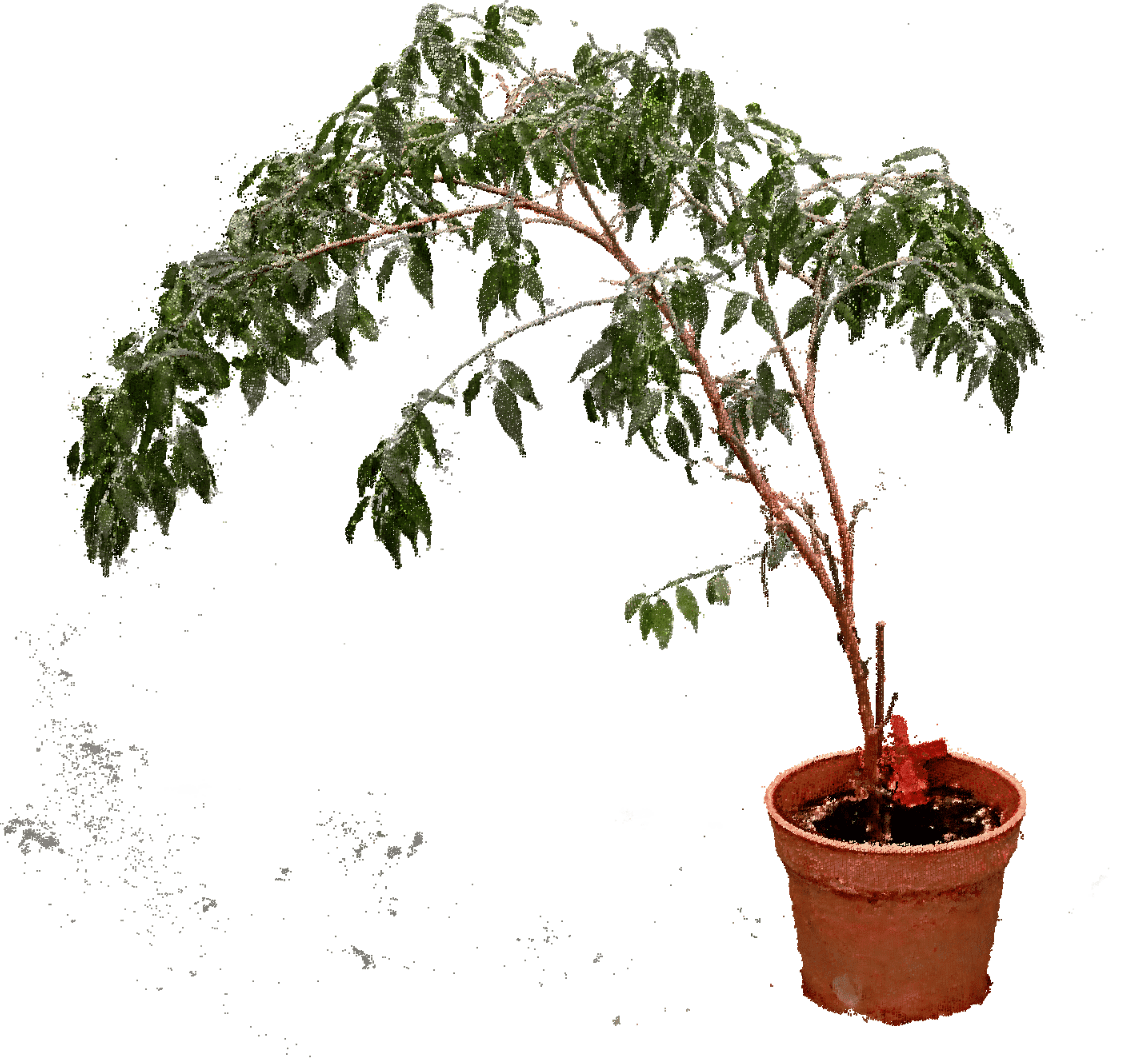}}
  \hspace{4mm}
\subfigure[NeRF-Ensemble]{\label{fig:hololens_refined_ensemble_}  
     \includegraphics[width=0.2\linewidth]{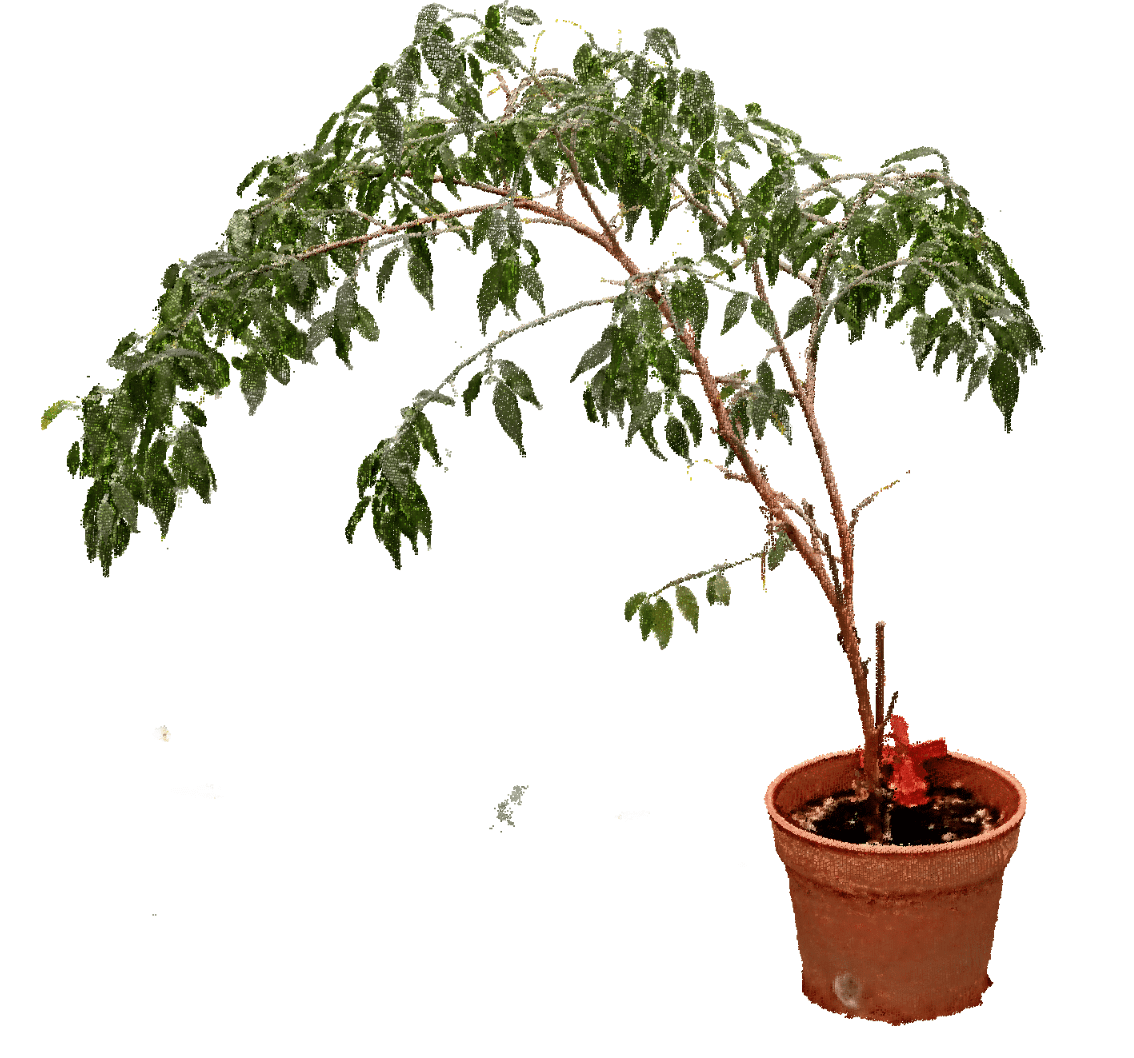}}
      \hspace{4mm}
\subfigure[density]{\label{fig:hololens_refined_density_}
	\includegraphics[width=0.2\linewidth]{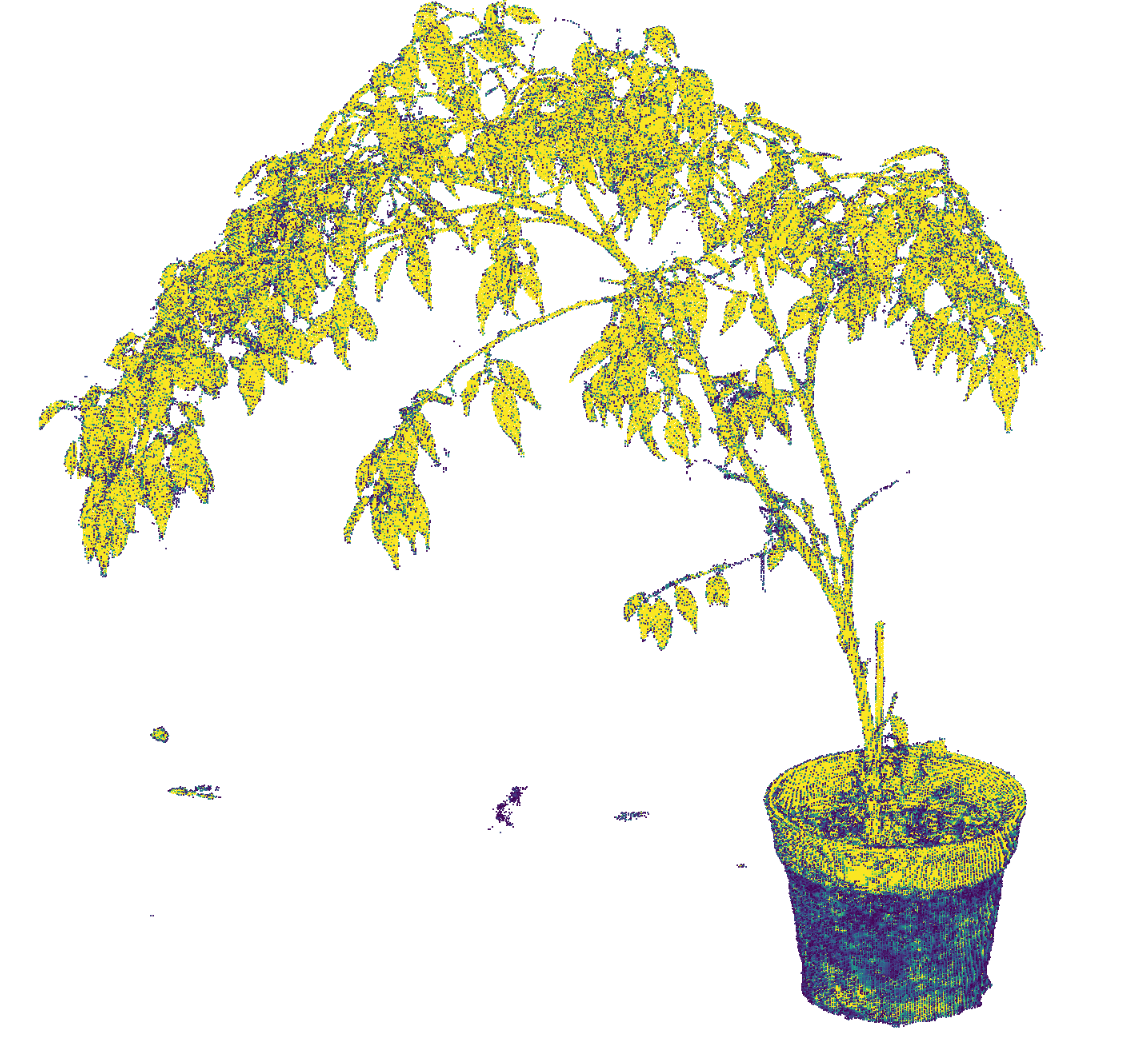}}
  \hspace{4mm}
\subfigure[uncertainty]{\label{fig:hololens_refined_unc_}\includegraphics[width=0.2\columnwidth]{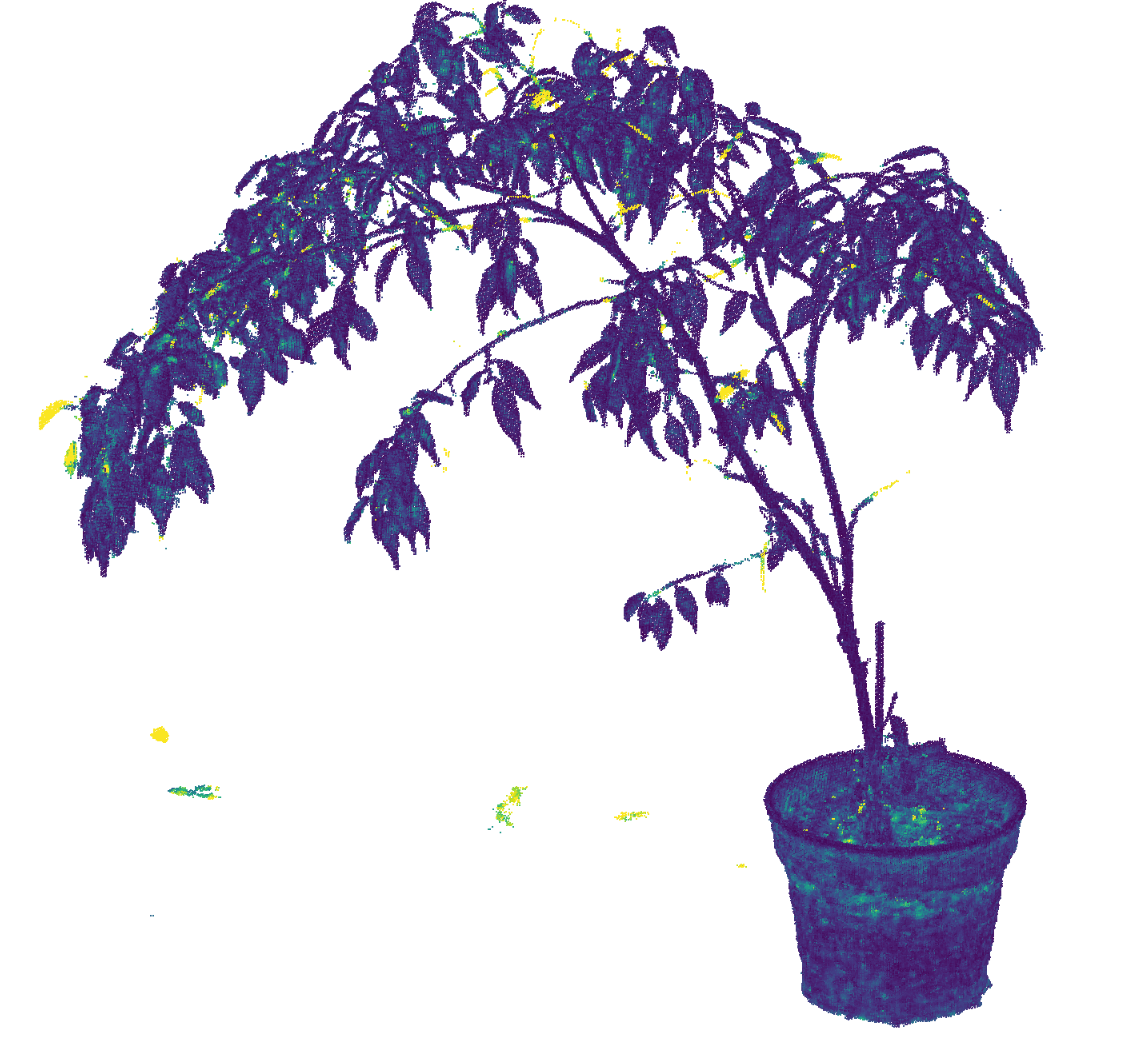}}
     \vspace{-1mm}\\
\subfigure{
	\includegraphics[width=0.2\linewidth]{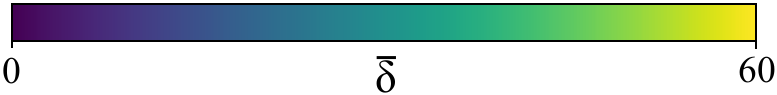}} \addtocounter{subfigure}{-1}
	\hspace{1cm}
\subfigure{
	\includegraphics[width=0.2\linewidth]{figures/colorbar_uncertainty_.pdf}}\addtocounter{subfigure}{-1}
	\vspace{-3mm}
	\caption{Qualitative comparison on the HoloLens data, showing points in the 3D grid with density above 15. For the reconstruction from internal HoloLens camera poses: NeRF \protect{\subref{fig:hololens_member_}}, NeRF-Ensemble \protect{\subref{fig:hololens_ensemble_}} and the respective mean density $\overline{\delta}$ \protect{\subref{fig:hololens_density_}} and density uncertainty $\text{U}_{\delta}$ \protect{\subref{fig:hololens_unc_}}. For the reconstruction from refined internal HoloLens camera poses: NeRF \protect{\subref{fig:hololens_refined_member_}}, NeRF-Ensemble \protect{\subref{fig:hololens_refined_ensemble_}} and the respective mean density $\overline{\delta}$ \protect{\subref{fig:hololens_refined_density_}} and density uncertainty $\text{U}_{\delta}$ \protect{\subref{fig:hololens_refined_unc_}}.}
\label{fig:pointclouds_hololens}
\end{figure}

The trends of the quantitative investigations are reflected in the visual results. Figure \ref{fig:pointclouds_synthetic} shows the reconstructed points in the 3D grid above density of 15 for the NeRF synthetic dataset. Visualized are results from a single NeRF, NeRF-Ensemble as well as the range of the density uncertainty for original input data and input data with additive Gaussian noise. 
It is evident that a low level of pose noise $\sigma_{\text{t}}$ = 0.01$\%$ (column 4) results in a scene with higher density uncertainty \text{$\text{U}_{\delta}$}. Pose noise $\sigma_{\text{R}}$ = 0.1° (column 5) also leads to a visually higher density uncertainty. In addition, the noise in the scene increases, resulting in more artifacts which were not present for data without noise (column 3) at the same density threshold.
A similar result is obtained for the DTU dataset in Figure \ref{fig:pointclouds_dtu}, where even a low level of pose noise $\sigma_{\text{t}}$ = 0.1\,mm and $\sigma_{\text{R}}$ = 0.01° leads to a significant increase in density uncertainty \text{$\text{U}_{\delta}$}. Again, the point clouds are noisier and contain more (foggy) artifacts. In addition, some of the points are missing, which corresponds to the quantitative results showing that the density for low-quality input data, especially poses, decreases significantly.
In all scenes, the image noise is barely noticeable in the density uncertainty \text{$\text{U}_{\delta}$} and visually only leads to slightly higher noise.
The HoloLens data, see Figure \ref{fig:pointclouds_hololens}, displays a higher density uncertainty \text{$\text{U}_{\delta}$} (see Figure \ref{fig:hololens_unc_}) for the internal camera poses, which is particularly evident in the artifacts of the scene. In addition, the density for this configuration is lower, see Figure \ref{fig:hololens_density_}. In contrast, the density for the refined internal camera poses is higher (Figure \ref{fig:hololens_refined_density_}), while the density uncertainty is lower (see Figure \ref{fig:hololens_refined_unc_}) and there are barely any (foggy) artifacts at all.
The analysis make it clear that apparent artifacts or fog have similar high density values as parts of the object and are therefore present in rendered images as well as reconstructed 3D scene. On the other hand, the results (see Figures \ref{fig:pointclouds_synthetic}, \ref{fig:pointclouds_dtu}, \ref{fig:pointclouds_hololens}) show that the density uncertainty of the network is high for these points.

\subsection{Scene Constraints: Acquisition constellation, occlusions, material properties} \label{sec:scene_constraints_results}
This Section presents the impact of the scene constraints and covers a qualitative evaluation of the density uncertainty. 

\subsubsection{\textbf{Density Uncertainty}}\label{sec:scene_constraints_results_unc}
The visual results show that the density uncertainty \text{$\text{U}_{\delta}$} of the network is varying for different areas of the object. This also applies to the case of non-noisy high-quality input data, which is visualized in column three of the Figures \ref{fig:pointclouds_synthetic}, \ref{fig:pointclouds_dtu} and \ref{fig:pointclouds_hololens}.  In these overview images, high density uncertainties \text{$\text{U}_{\delta}$} mainly relate to occluded areas of the scenes that are barely visible. Furthermore, it becomes apparent that the DTU dataset contains high density uncertainties \text{$\text{U}_{\delta}$} for many of the scenes, especially those with low-textured subsurfaces. For the NeRF synthetic dataset, material properties-dependent uncertainties occur as well.

\begin{figure}[H]
	\centering
\subfigure[]{\label{fig:aufnahmekonfiguration1}
	\includegraphics[width=0.26\linewidth]{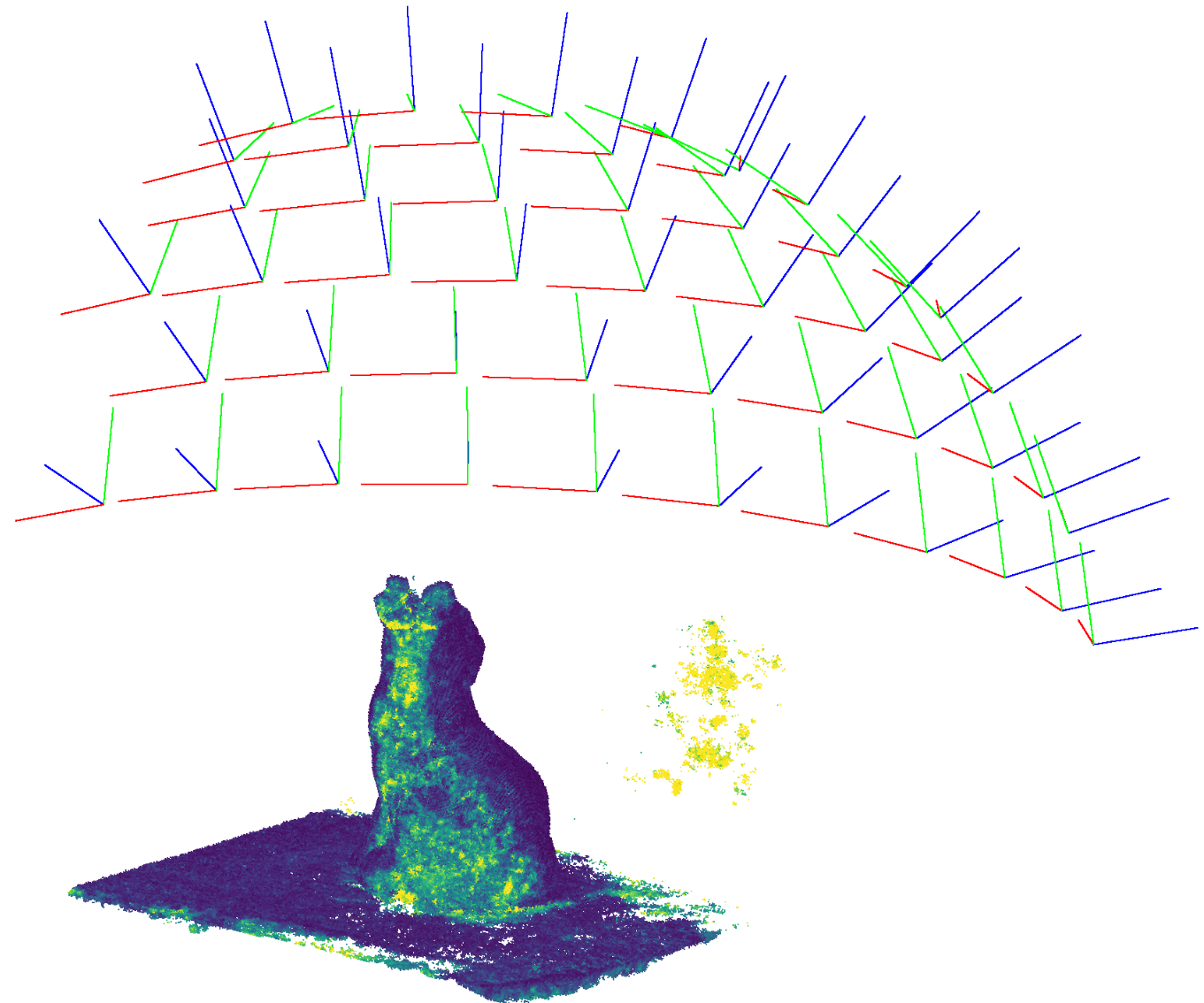}}
\subfigure[]{\label{fig:aufnahmekonfiguration3}
	\includegraphics[width=0.3\linewidth]{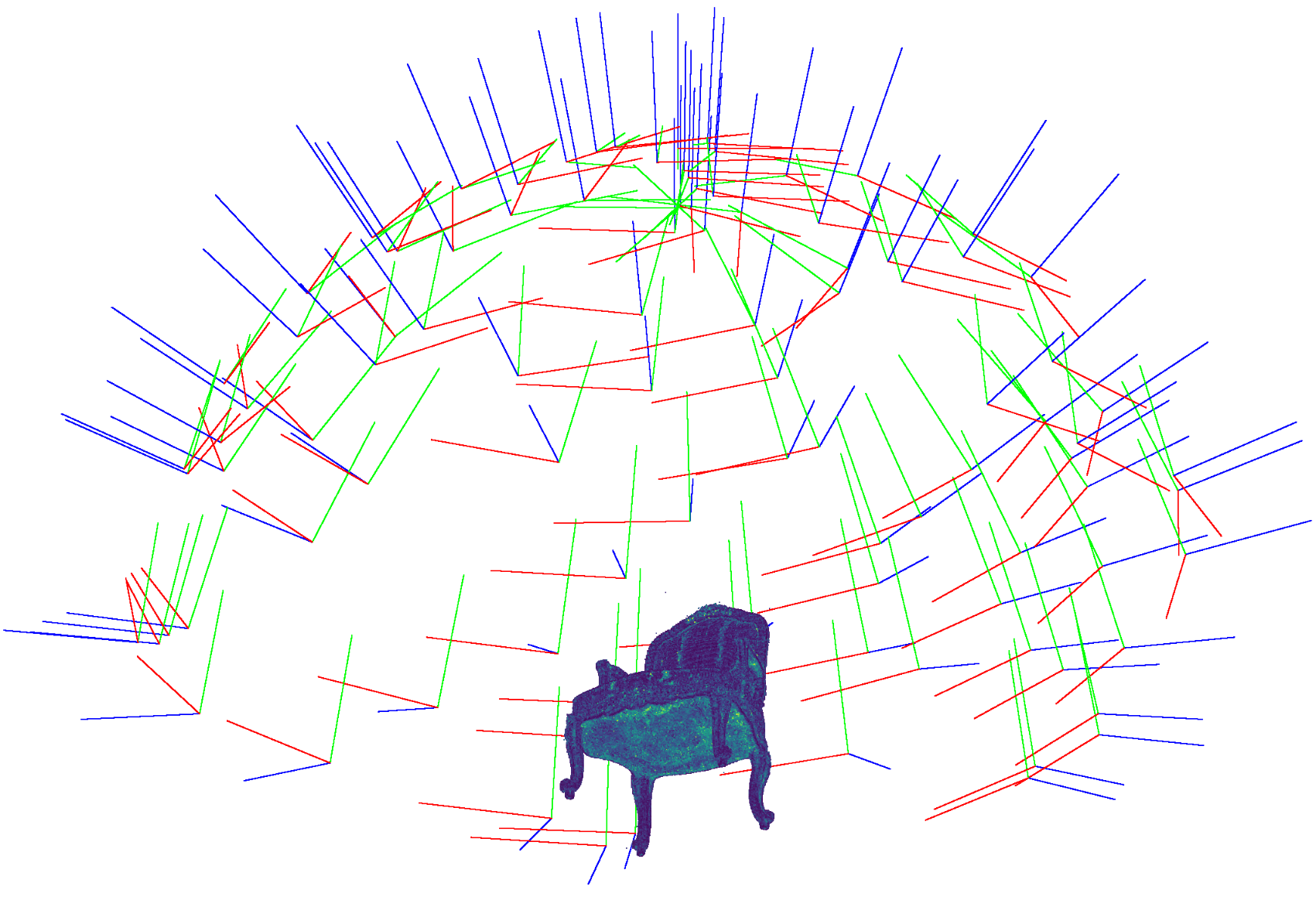}}
   \subfigure[]{\label{fig:aufnahmekonfiguration5}
    \vspace{0.2cm}
	\includegraphics[width=0.27\linewidth]{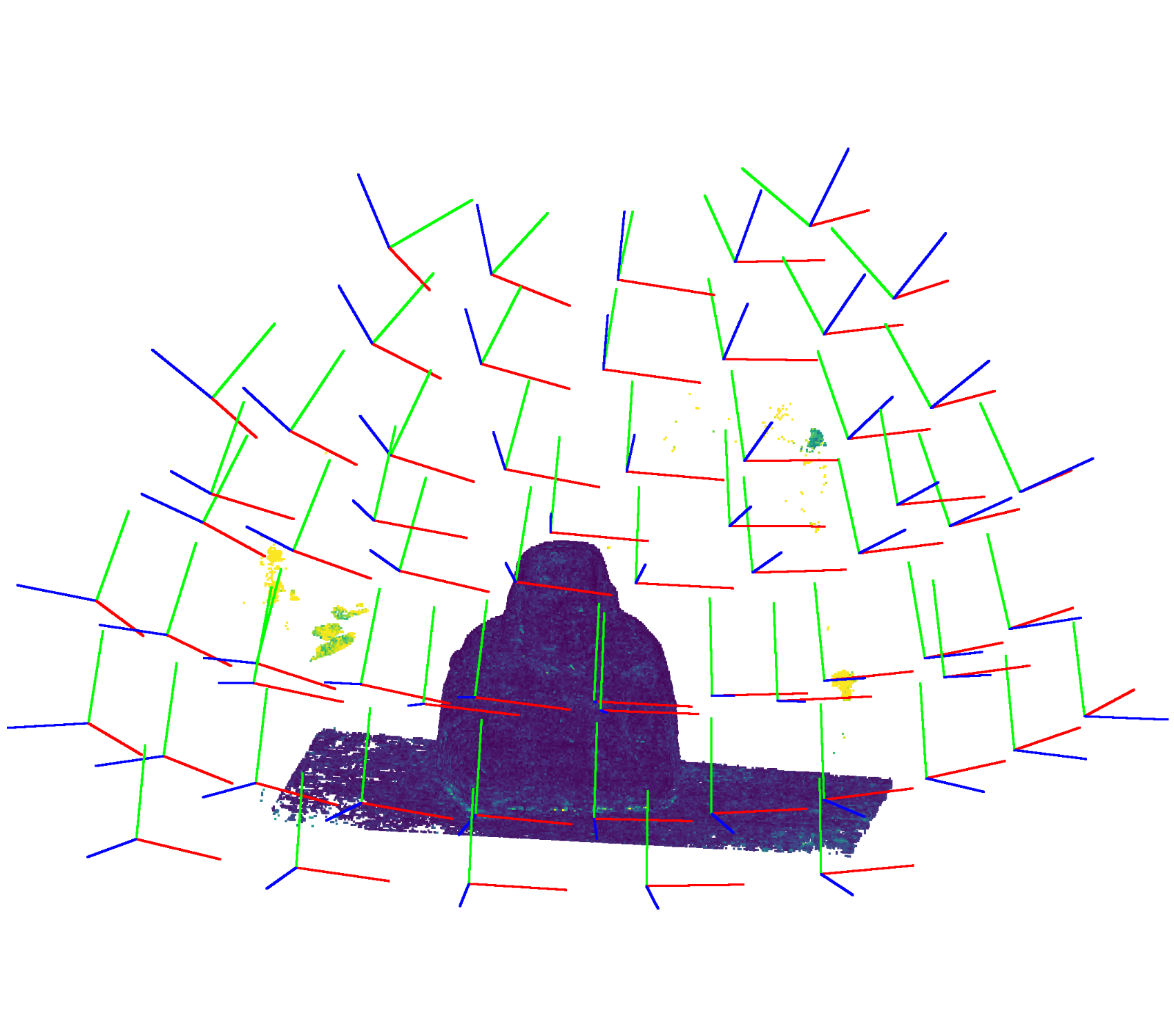}}\\
 \vspace{-0.8cm}
 \subfigure[]{\label{fig:aufnahmekonfiguration2}
	\includegraphics[width=0.26\linewidth]{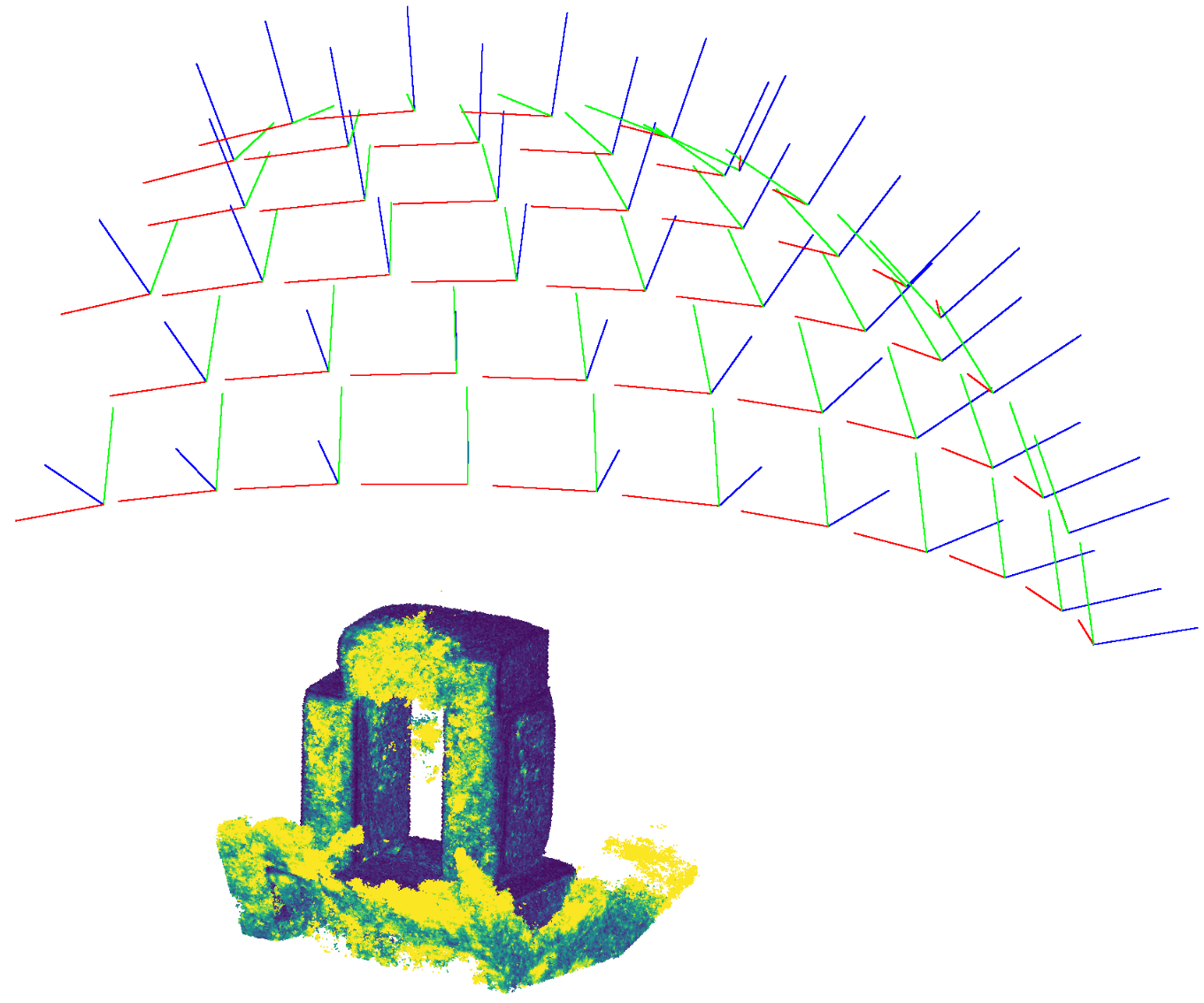}}
\subfigure[]{\label{fig:aufnahmekonfiguration4}
	\includegraphics[width=0.3\linewidth]{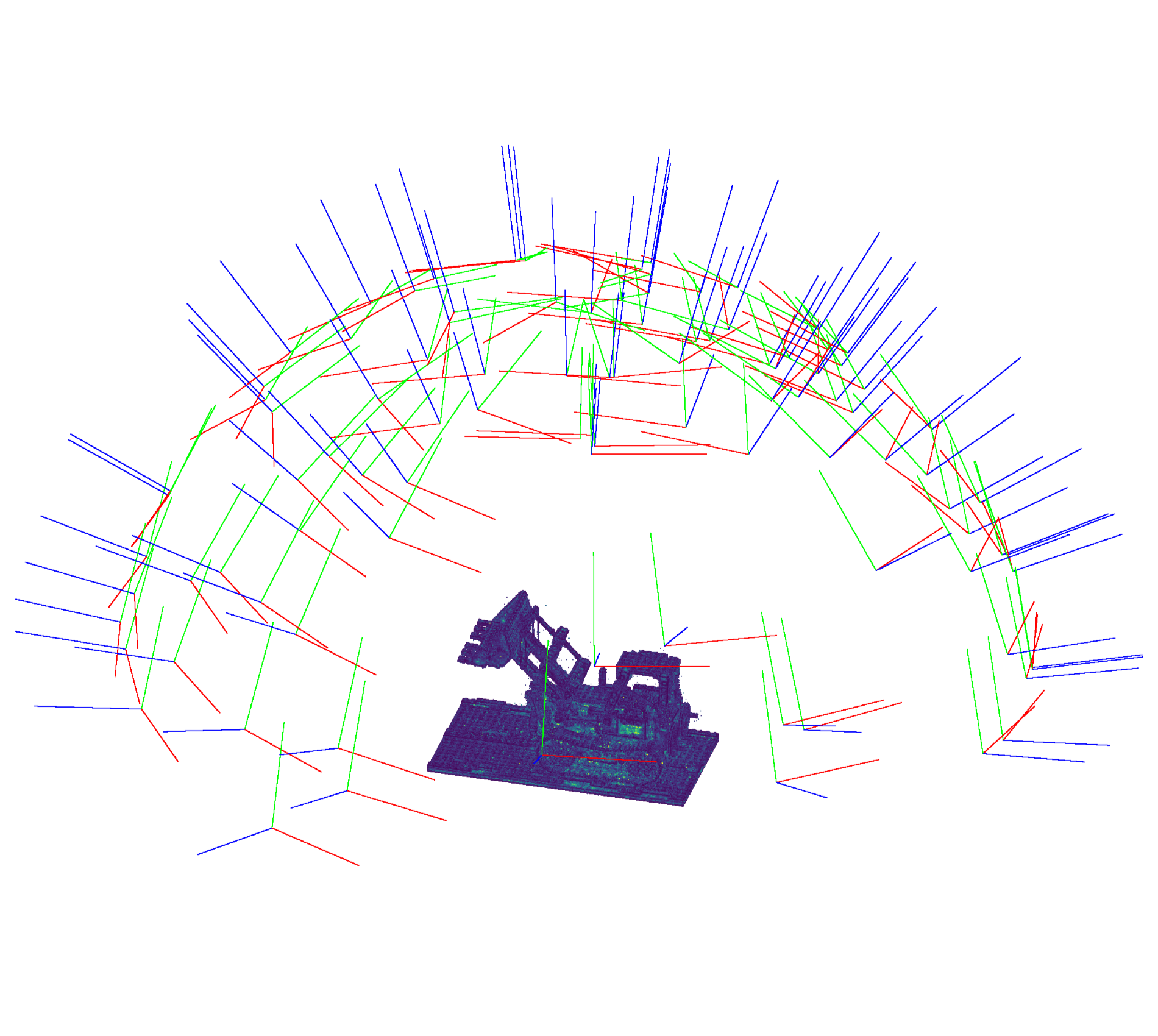}}
 \subfigure[]{\label{fig:aufnahmekonfiguration6}
	\includegraphics[width=0.27\linewidth]{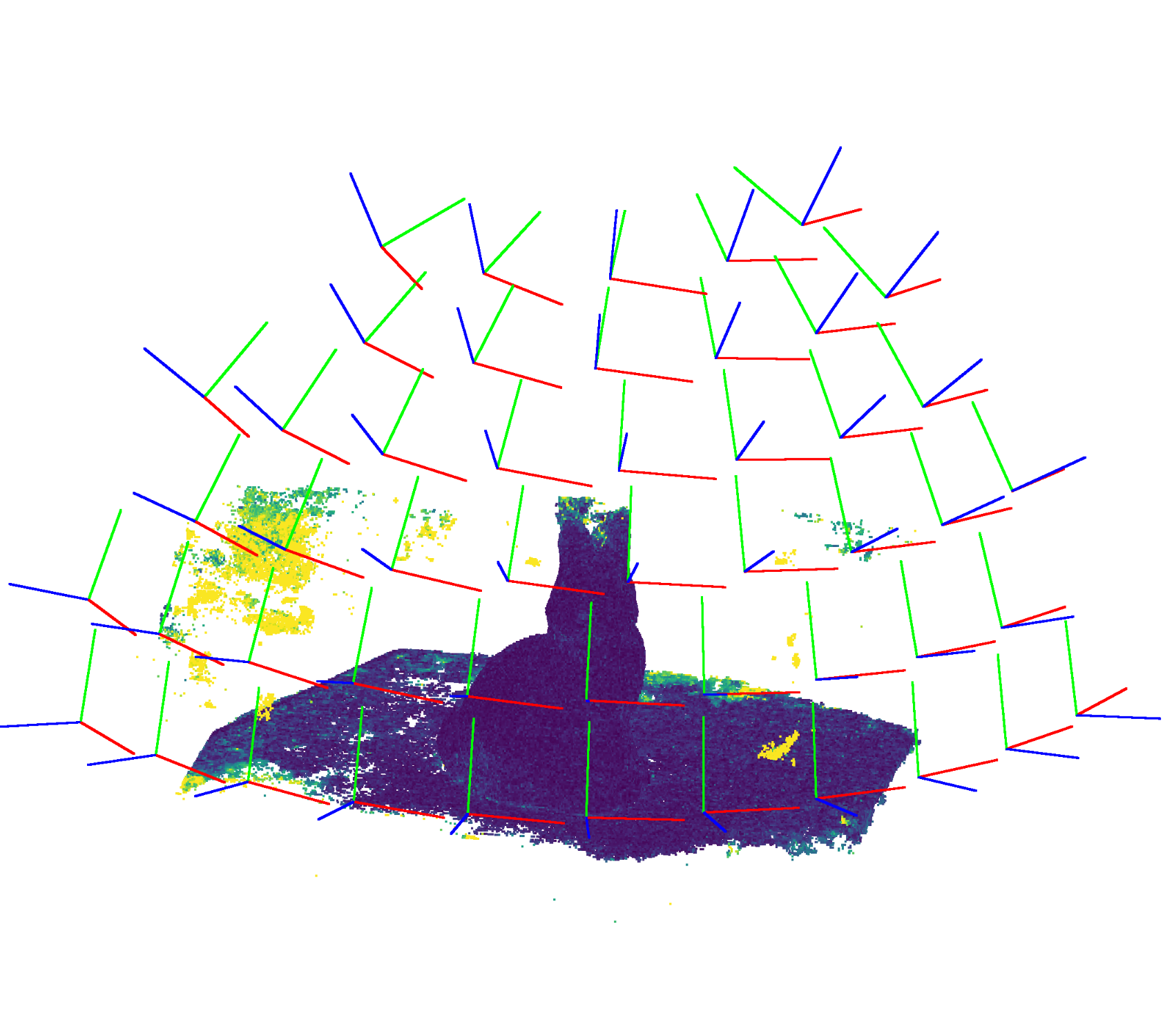}}\\
\subfigure{
	\includegraphics[width=0.2\linewidth]{figures/colorbar_uncertainty_.pdf}}\addtocounter{subfigure}{-1}
	\vspace{-2mm}
	\caption{Acquisition constellation and occlusions. Visualized is the density uncertainty \text{$\text{U}_{\delta}$}. The poses of the corresponding images are shown in their constellation with colored orientation of the 3 axes of the poses. Column 1 \protect{\subref{fig:aufnahmekonfiguration1}}, \protect{\subref{fig:aufnahmekonfiguration2}}: Back of the object which is occluded and not covered due to the acquisition constellation, which is an upper half hemisphere. Column 2 top \protect{\subref{fig:aufnahmekonfiguration3}}: Underside of the object that is not covered due to the acquisition constellation, which is an upper hemisphere. Column 2 bottom
 \protect{\subref{fig:aufnahmekonfiguration4}}: Object from the side, whereby there are gaps at the sides in the acquisition constellation and areas in the object that are barely visible due to occlusions from the scene geometry. Column 3
 \protect{\subref{fig:aufnahmekonfiguration5}}, \protect{\subref{fig:aufnahmekonfiguration6}}: (foggy) artifacts with high density uncertainty due to the acquisition constellation, which is an upper half hemisphere.}
\label{fig:aufnahmekonfiguration}
\end{figure}

Figure \ref{fig:aufnahmekonfiguration} provides a detailed representation of some scenes along with the associated acquisition constellations displayed as the three axes of the poses.
Hidden object parts due to the acquisition constellation and occlusions show higher density uncertainties \text{$\text{U}_{\delta}$}. For example, the scene lego, as well as the underside of the scene chair. Uncertainties due to poor coverage of the recordings, which result in occlusions, and the acquisition constellation in general should also be mentioned. This becomes particularly clear with the DTU dataset, whose camera poses are located on a one-sided hemisphere. The back of the object, which is not visible from the camera poses, is therefore not observed. This appears to be the case for subsurfaces without texture. Large white artifacts are reconstructed within the white surfaces in scenes scan24, scan37 and scan40. These are not visible in the rendered images due to the color, but the coordinates of the points contain estimated density values and are therefore present in reconstructed 3D point clouds, which have high uncertainties. This is not the case for scenes scan55 and scan114 with colored, slightly textured subsurfaces. 

\begin{figure}[H]
	\centering
	\vspace{-2mm}
\subfigure[density]{\label{fig:slice_chair_density}
	\includegraphics[width=0.12\linewidth]{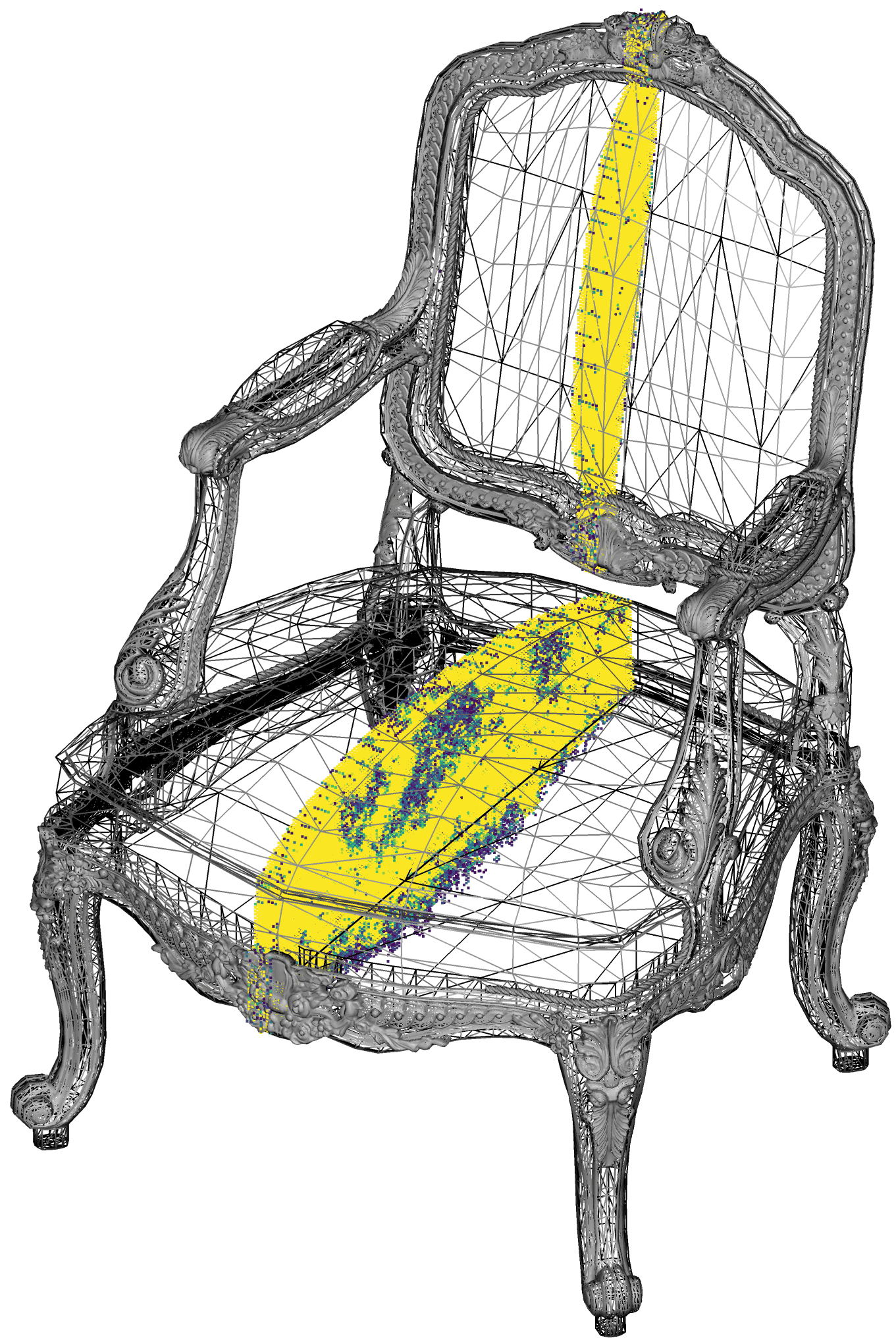}}
 	\hspace{0.5cm}
	\subfigure[uncertainty]{\label{fig:slice_chair_uncertainty}
	\includegraphics[width=0.12\linewidth]{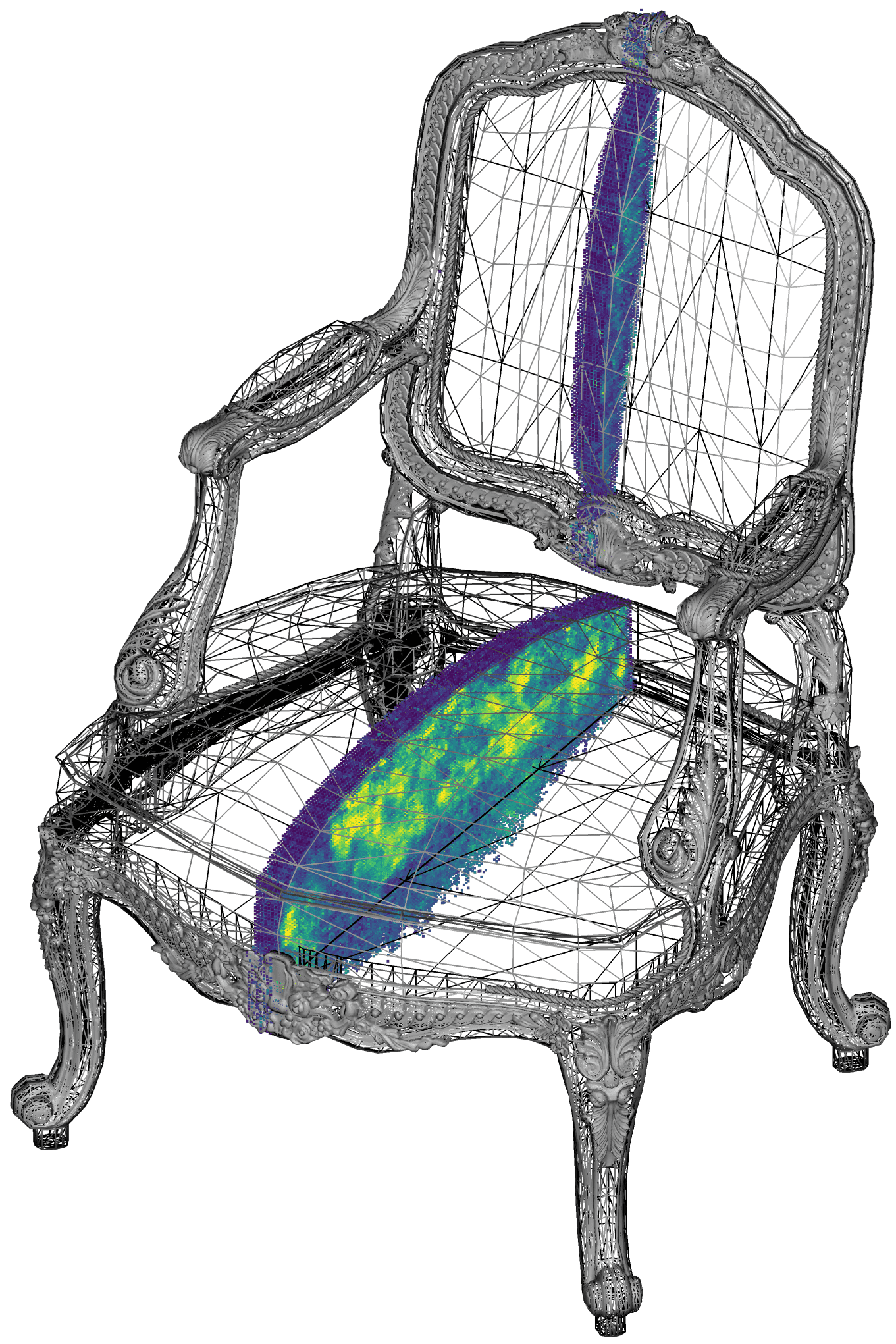}} 
	\hspace{1cm}
 \subfigure[density]{\label{fig:slice_scan40_density}
	\includegraphics[width=0.15\linewidth]{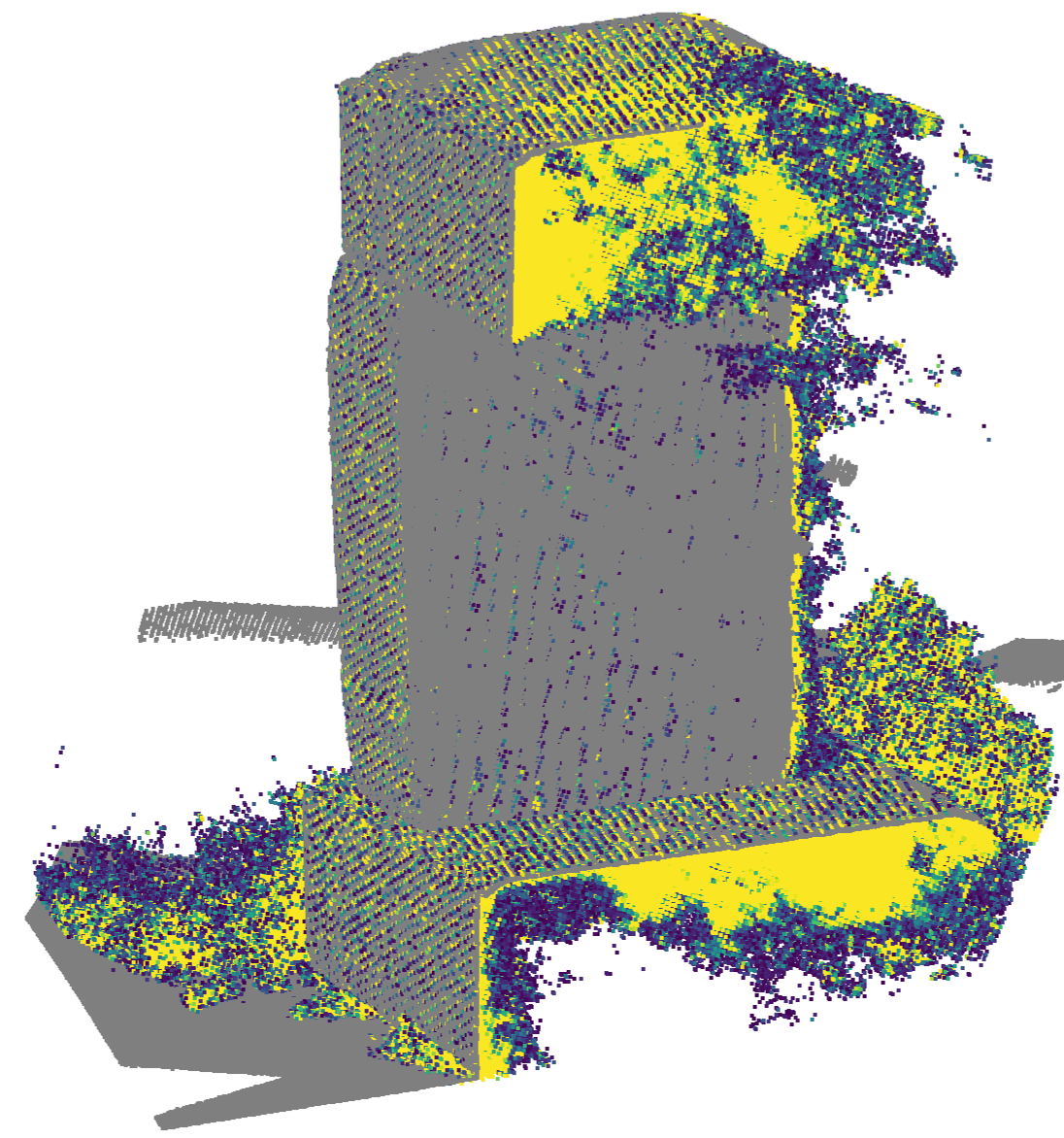}}
		\hspace{0.5cm}
	\subfigure[uncertainty]{\label{fig:slice_scan40_uncertainty}
	\includegraphics[width=0.15\linewidth]{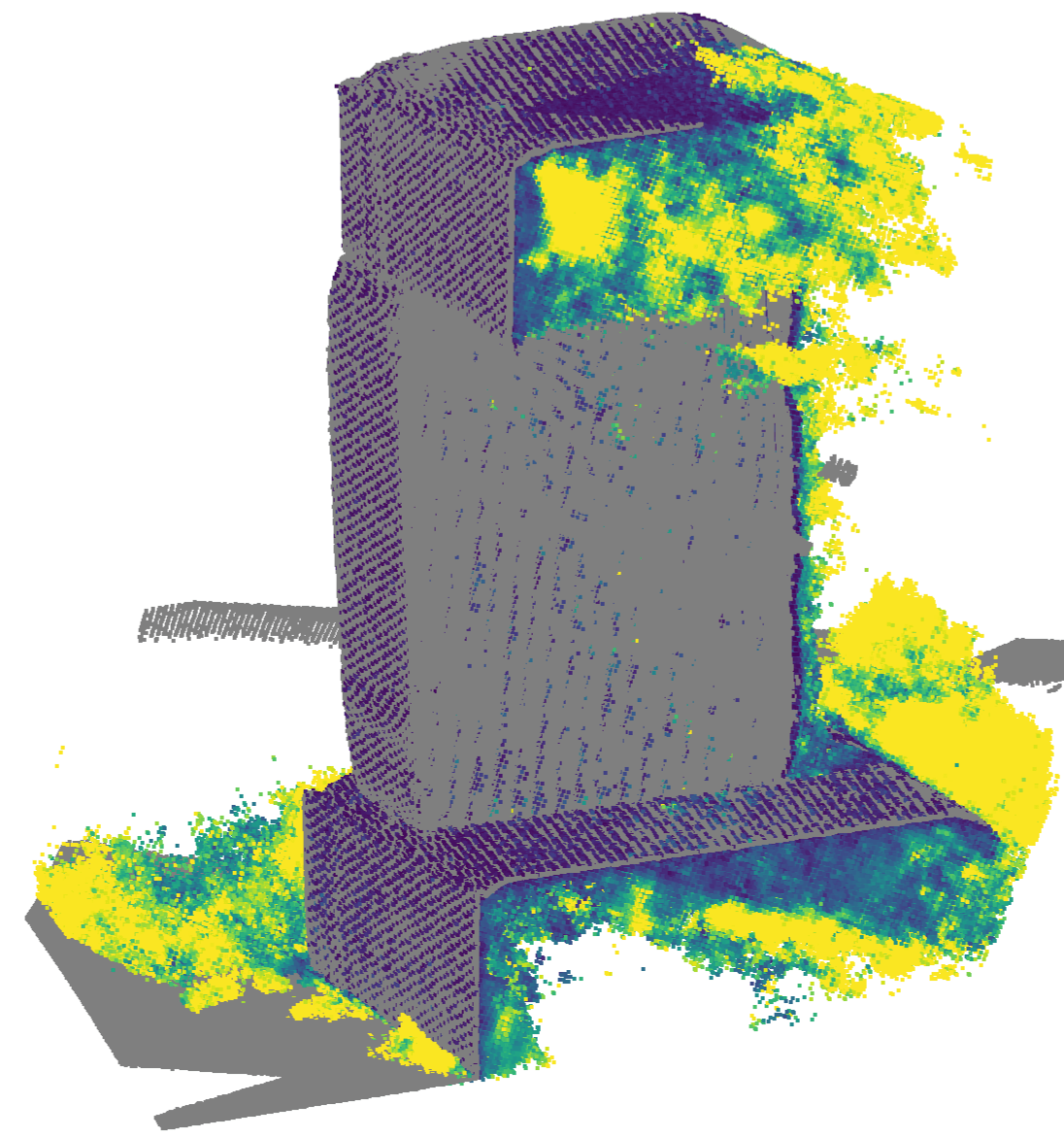}}
 \\
\subfigure{
	\includegraphics[width=0.2\linewidth]{figures/colorbar_uncertainty_.pdf}}
	\caption{Occlusions. Cross-sections through the objects. The mean density \text{$\text{U}_{\delta}$} and density uncertainty \text{$\text{U}_{\delta}$} are shown in comparison to the reference point clouds and meshes (in gray). The density in the object is usually high, while the density uncertainty \text{$\text{U}_{\delta}$} is higher in the interior compared to the surface of the object.}
\label{fig:slices}
\end{figure}

NeRFs predict the continuous density field in the entire 3D space and thus also inside the object, which are always occluded in the images. As the cross-sections in Figure \ref{fig:slices} show, a high density uncertainty \text{$\text{U}_{\delta}$} is also present within the objects compared to the object surface. 

\begin{figure}[H]
	\centering
	\subfigure[Training Image]{\label{fig:materials_chair_1}
	\includegraphics[width=0.13\columnwidth]{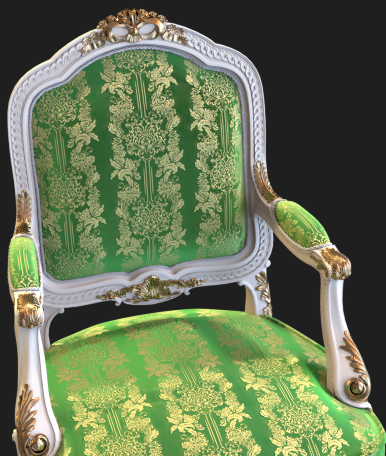}}
     \subfigure[Uncertainty]{\label{fig:materials_chair_2}  
     \includegraphics[width=0.13\columnwidth]{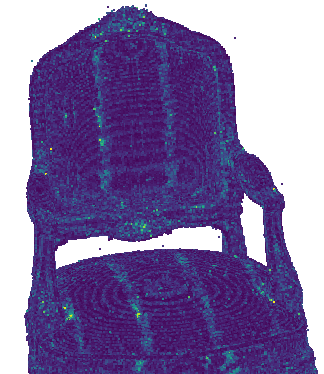}}	
     \subfigure[Training Image]{\label{fig:materials_drums_1}
	\includegraphics[width=0.15\columnwidth]{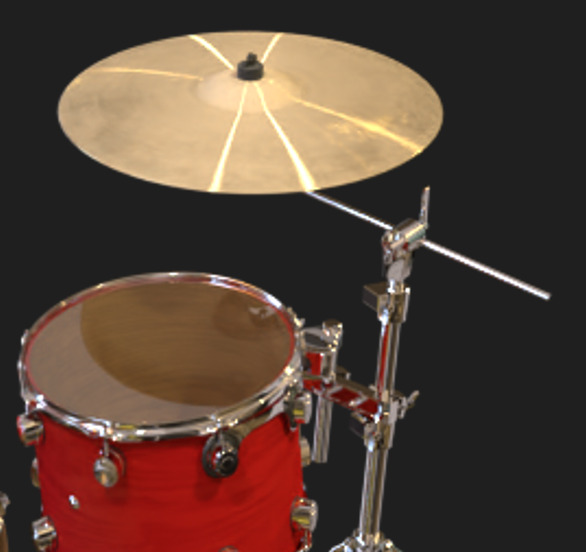}}
     \subfigure[Uncertainty]{\label{fig:materials_drums_2}  
     \includegraphics[width=0.15\columnwidth]{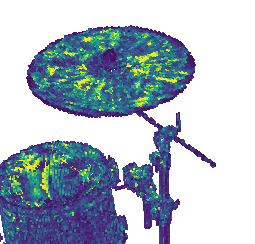}}
     \subfigure[Training Image]{\label{fig:materials_materials_1}
	\includegraphics[width=0.18\columnwidth]{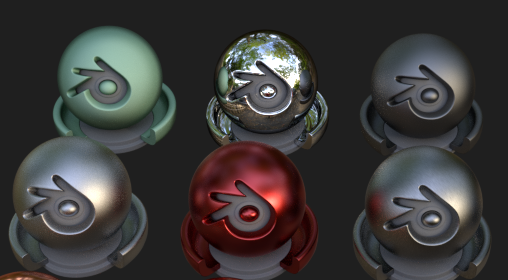}}
     \subfigure[Uncertainty]{\label{fig:materials_materials_2}  
     \includegraphics[width=0.18\columnwidth]{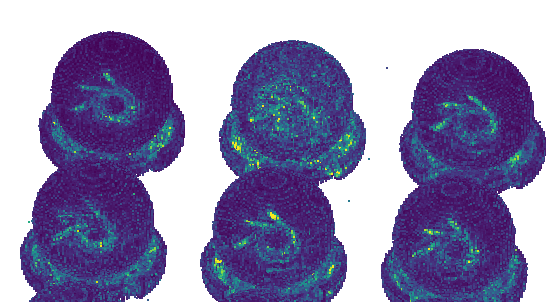}}	
     \\
\subfigure{
	\includegraphics[width=0.2\linewidth]{figures/colorbar_uncertainty_.pdf}}
	\vspace{-3mm}
	\caption{Material properties. Comparison of a detailed sections of the training images versus the density uncertainty \text{$\text{U}_{\delta}$}: Scene chair \protect{\subref{fig:materials_chair_1}} versus \protect{\subref{fig:materials_chair_2}} with different fabrics, scene drums \protect{\subref{fig:materials_drums_1}} versus \protect{\subref{fig:materials_drums_2}} with reflective or semi-transparent material, as well as scene materials \protect{\subref{fig:materials_materials_1}} versus \protect{\subref{fig:materials_materials_2}} with reflective material.}
\label{fig:materials}
\end{figure}

With regard to specific material properties, the results of the NeRF synthetic dataset in Figure \ref{fig:pointclouds_synthetic} are representative. It becomes evident that the density uncertainty \text{$\text{U}_{\delta}$} varies based on the material properties of the object within the scene.
For example, in scene chair, the stripes show higher uncertainties than the rest of the fabric. But also, materials like in scene drums and scene materials show higher uncertainties in areas that are reflective or semi-transparent as shown in detail in Figure \ref{fig:materials}.

\subsection{Outlook: Potential of NeRF-Ensembles}\label{sec:results_potential}
In addition to the density uncertainty quantification, we present further promising advantages that NeRF-Ensembles provide. Firstly, increased robustness by the usage of NeRF-Ensembles instead single NeRFs in Section \ref{sec:robustness}. Secondly, we provide a method for the removal of (foggy) artifacts through density uncertainties in Section \ref{sec:artifacts}, which inevitably results from our findings.

\subsubsection{\textbf{Robustness}}\label{sec:robustness}
\begin{figure}[H]
	\centering
	\subfigure[NeRF]{\label{fig:drums_member_cutout}
	\includegraphics[width=0.20\linewidth]{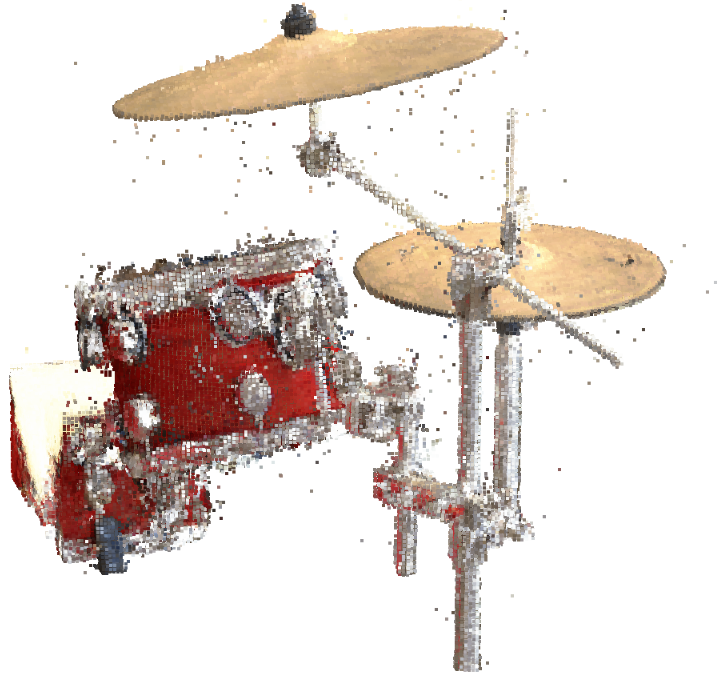}}
     \subfigure[NeRF-Ensemble]{\label{fig:drums_ensemble_cutout}  
     \includegraphics[width=0.20\linewidth]{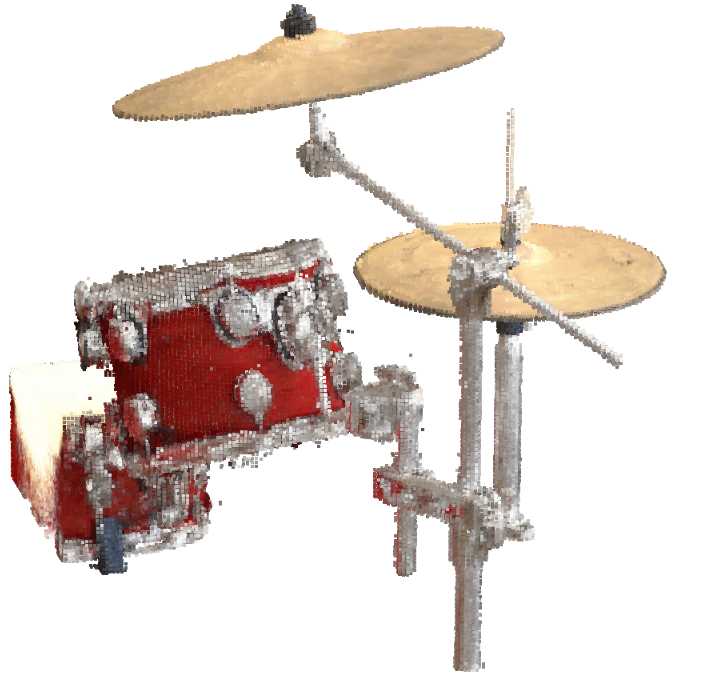}}	
     \subfigure[NeRF]{\label{fig:scan55_member_cutout}
	\includegraphics[width=0.20\linewidth]{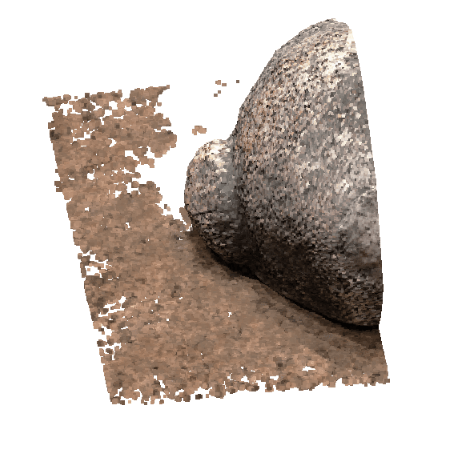}}
     \subfigure[NeRF-Ensemble]{\label{fig:scan55_ensemble_cutout}  
     \includegraphics[width=0.20\linewidth]{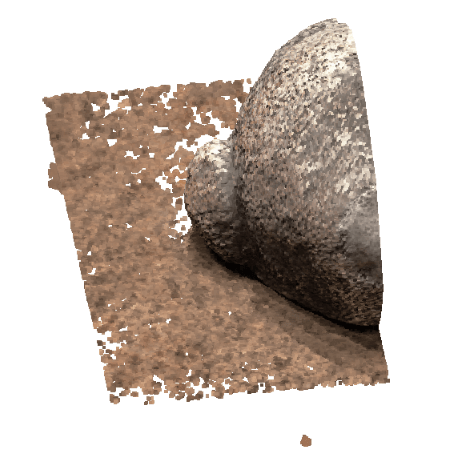}}\\
     	\subfigure[NeRF]{\label{fig:holo_refined_member_cutout_}
	\includegraphics[width=0.15\linewidth]{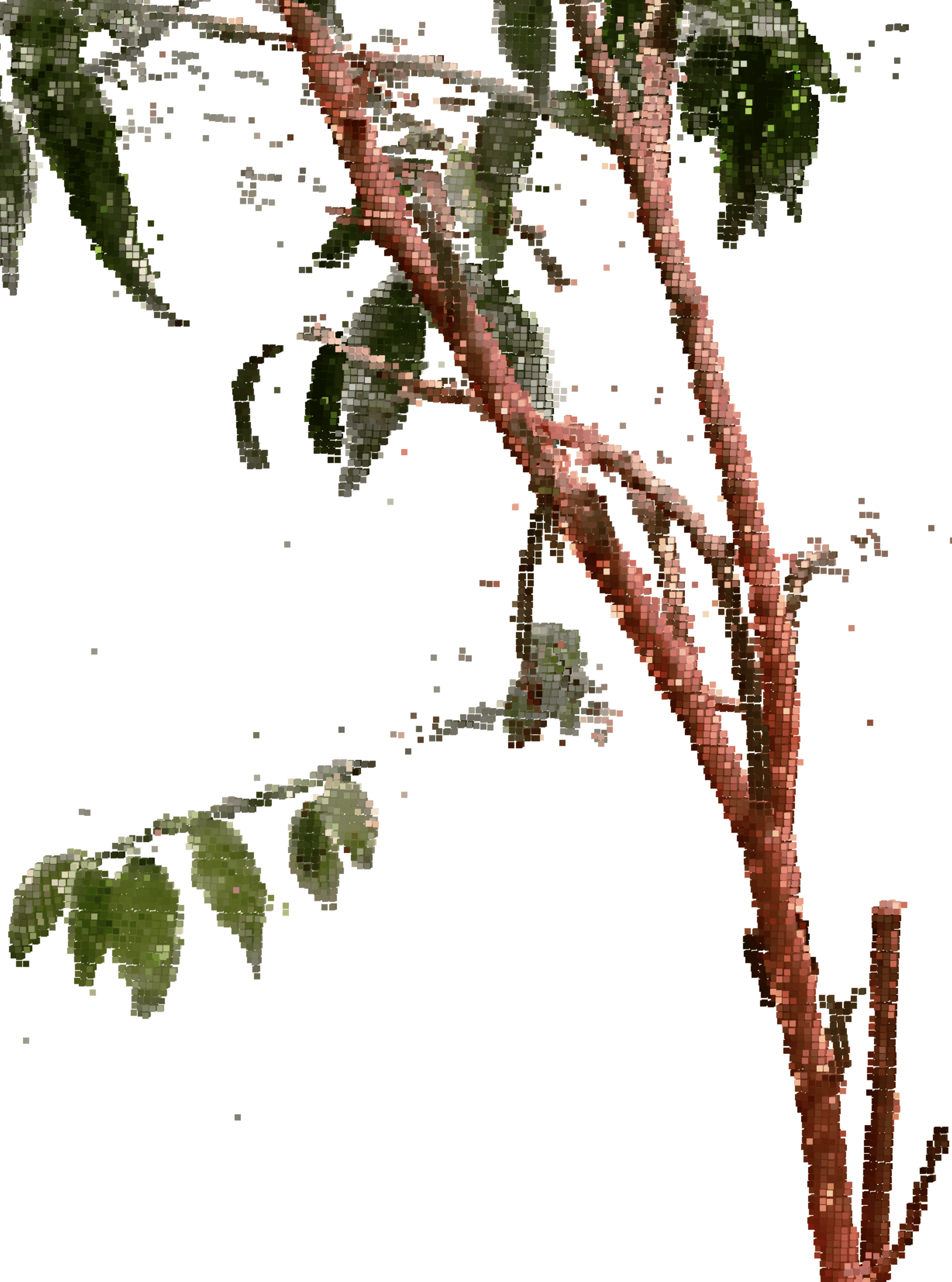}}
 \hspace{4mm}
     \subfigure[NeRF-Ensemble]{\label{fig:holo_refined_ensemble_cutout_}  
     \includegraphics[width=0.15\linewidth]{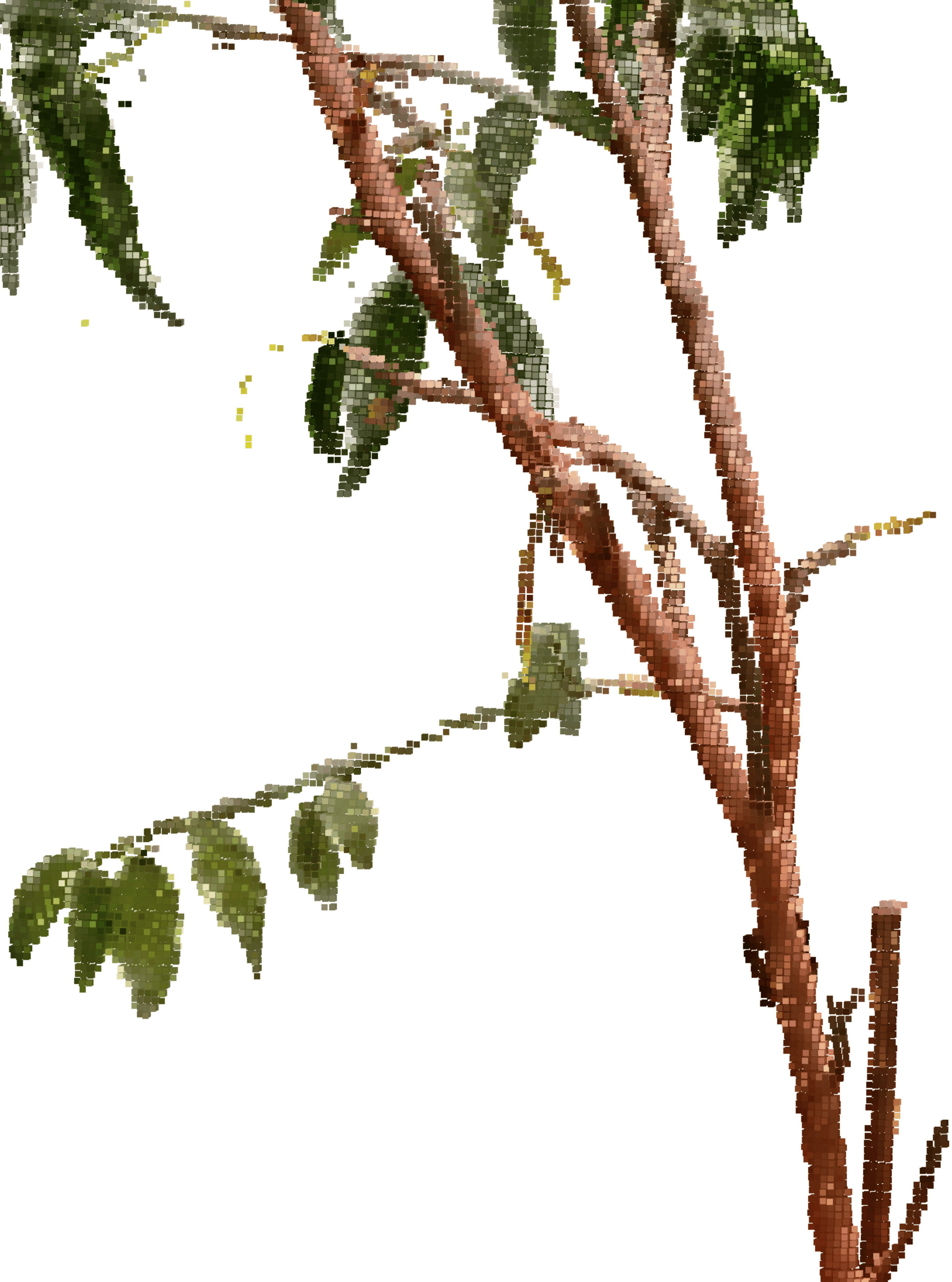}}
	\vspace{-3mm}
	\caption{Comparison of a detailed sections of the 3D point clouds each for a single NeRFs versus a NeRF-Ensemble: NeRF \protect{\subref{fig:drums_member_cutout}} versus NeRF-Ensemble \protect{\subref{fig:drums_ensemble_cutout}} for the nerf synthetic dataset, NeRF \protect{\subref{fig:scan55_member_cutout}} versus NeRF-Ensemble \protect{\subref{fig:scan55_ensemble_cutout}} for the DTU dataset, as well as NeRF \protect{\subref{fig:holo_refined_member_cutout_}} versus NeRF-Ensemble \protect{\subref{fig:holo_refined_ensemble_cutout_}} for the HoloLens data with refined poses.}
\label{fig:pointclouds_cutout}
\end{figure}

In addition to the quantification of the density uncertainty, NeRF-Ensembles show further advantages. This is shown in the previous qualitative results in Figures \ref{fig:pointclouds_synthetic}, \ref{fig:pointclouds_dtu}, \ref{fig:pointclouds_hololens} for each dataset by comparing the respective columns 1 for a NeRF and 2 for a NeRF-Ensemble.
NeRF-Ensembles can rely on the mean density per position in contrast to the density from a single NeRF. In doing so, NeRF-Ensemble lead to a reduction of small artifacts and outliers, as well as a higher object completeness. Figure \ref{fig:pointclouds_cutout} outlines this advantage of NeRF-Ensembles in detail for a scene from NeRF synthetic dataset, DTU dataset and HoloLens data. Small artifacts and outliers are removed (comparison of Figure \ref{fig:drums_member_cutout} to \ref{fig:drums_ensemble_cutout}), a higher completeness of low-textured surfaces is reached (comparison of Figure \ref{fig:scan55_member_cutout} to \ref{fig:scan55_ensemble_cutout}) as well as a higher completeness of fine structures (comparison of Figure \ref{fig:holo_refined_member_cutout_} to \ref{fig:holo_refined_ensemble_cutout_}).

\subsubsection{\textbf{Artifact Removal}}\label{sec:artifacts}

In general, the removal of so-called fog, ghostly, floater or foggy artifacts \cite{cleannerf, vipnerf, nerfbusters, artifacts_postprocessing} is of great interest when it comes to NeRFs. 
These foggy artifacts change their shape and appearance in the scene during training process. They also occur in the case of low-quality input data, such as poses \cite{jaeger2023hololens}. This is highlighted by our findings from Section \ref{sec:results_density_uncertainty}. In addition to low-quality poses, artifacts occur due to scene constraints from the acquisition constellation and occlusions, as Section \ref{sec:scene_constraints_results_unc} demonstrates.
Therefore, we expect that (foggy) artifacts in the 3D scene are subject to high density variations, resulting in a high density uncertainty of the NeRF output. In this context, we want to investigate a possible solution for their removal. \newline

Based on this, we propose a density uncertainty-dependent artifact removal based on NeRF-Ensembles. This is achieved by removing points based on a percentile of large density uncertainties in the implicit 3D space, which is discretized by the density grid.
The remaining 3D positions are those whose uncertainty is less than or equal to a maximum value of $p$-th percentile. Let $\sigma_{\delta}(\text{X})$ be the uncertainty of the density at position $\text{X}$, $\text{percentile}_{p}(\sigma_{\delta})$ represents the $p$-th percentile of the uncertainty of the discretized 3D grid. The set of remaining 3D positions $\text{X'}$ for which the uncertainty is less than or equal to the $p$-th percentile is given by the following Equation \ref{equ:4}:

\begin{equation}\label{equ:4}
\begin{aligned}
 \text{X'} = \left\{\text{X} \mid \sigma_{\delta}(\text{X}) \leq \text{percentile}_{p}(\sigma_{\delta}(\text{X})\right\}.
   \end{aligned}
\end{equation}

\begin{figure}[H]
	\centering
	\subfigure[rendered image]{\label{fig:artefacts_scan40_rendered}
	\includegraphics[width=0.225\linewidth]{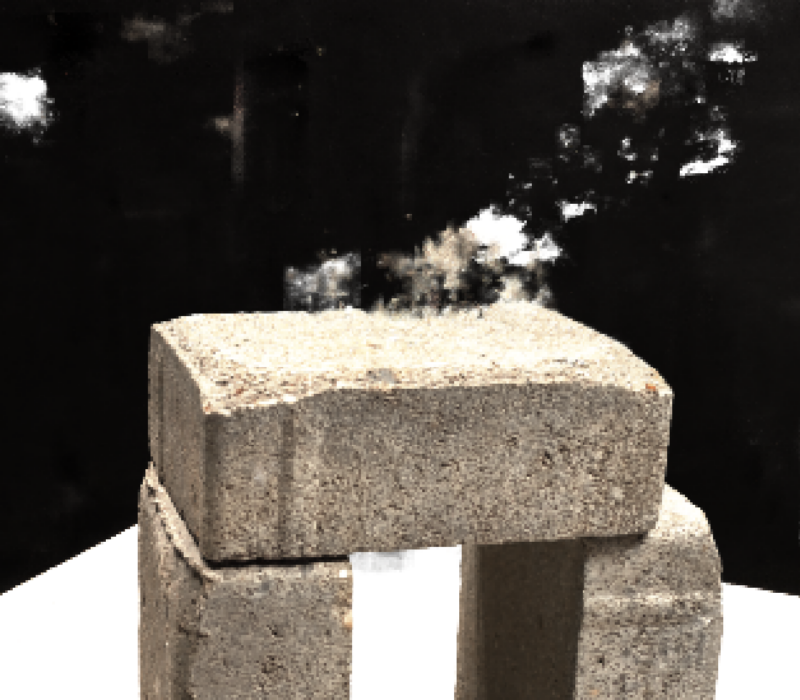}}
	\hspace{0.2cm}
     \subfigure[before artifact removal]{\label{fig:artefacts_scan40_before}  
     \includegraphics[width=0.225\linewidth]{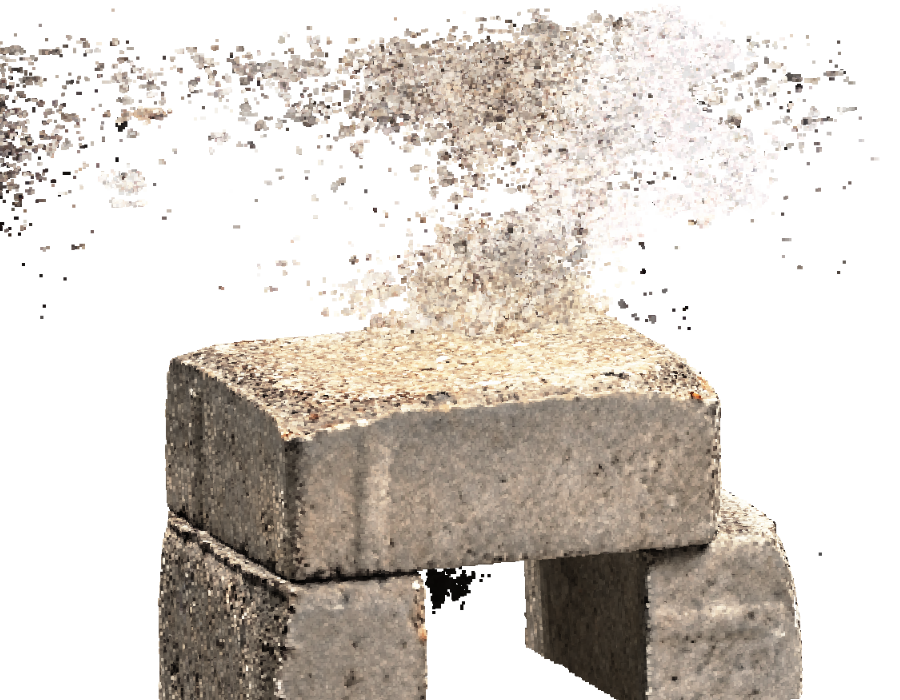}}	
     \subfigure[artifacts $>$90-th percentile]{\label{fig:artefacts_scan40_90th}
	\includegraphics[width=0.225\linewidth]{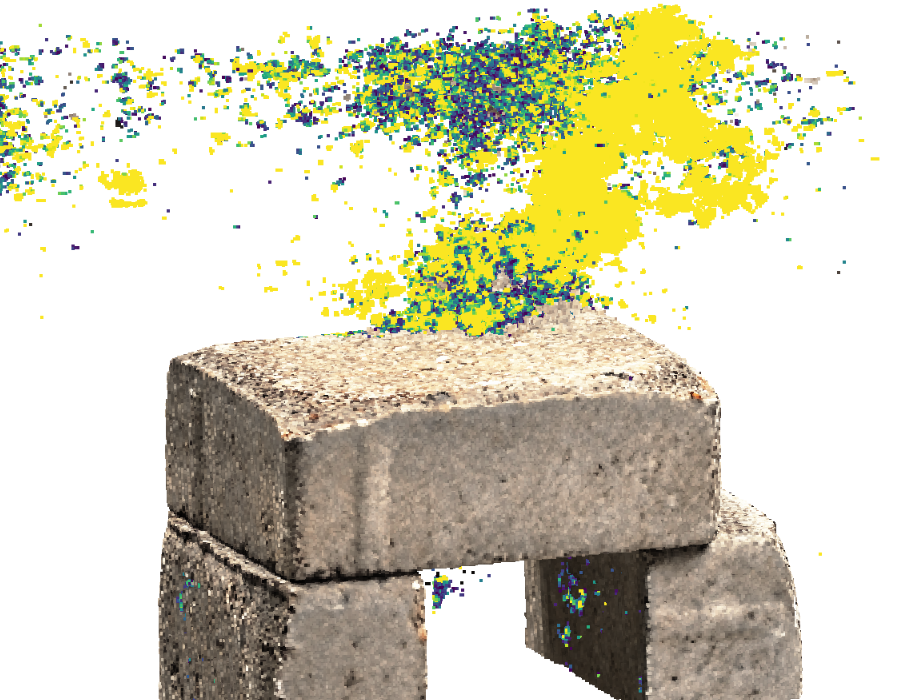}}
	\hspace{0.2cm}
     \subfigure[after artifact removal]{\label{fig:artefacts_scan40_after}  
     \includegraphics[width=0.225\linewidth]{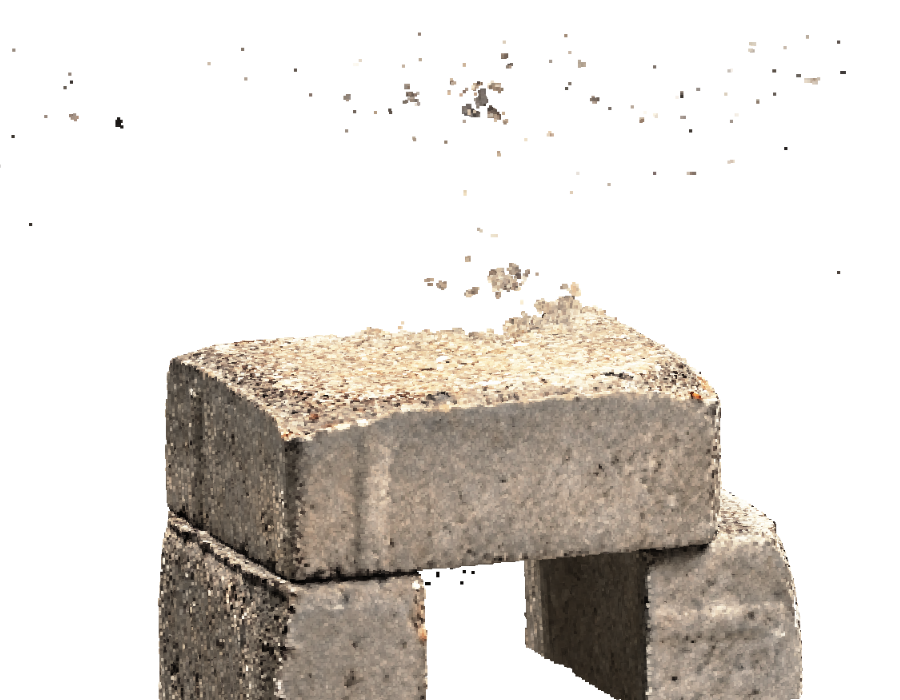}} \\
	\subfigure[rendered image]{\label{fig:artefacts_rendered}
	\includegraphics[width=0.225\linewidth]{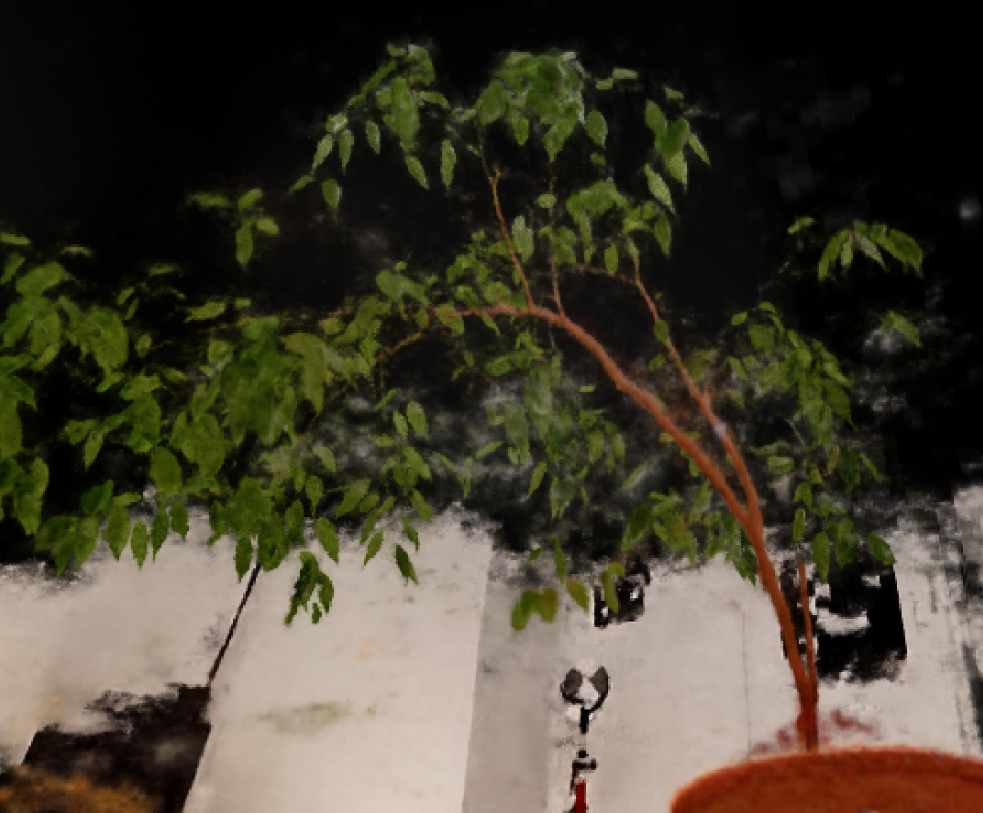}}
	\hspace{0.2cm}
     \subfigure[before artifact removal]{\label{fig:artefacts_before}  
     \includegraphics[width=0.225\linewidth]{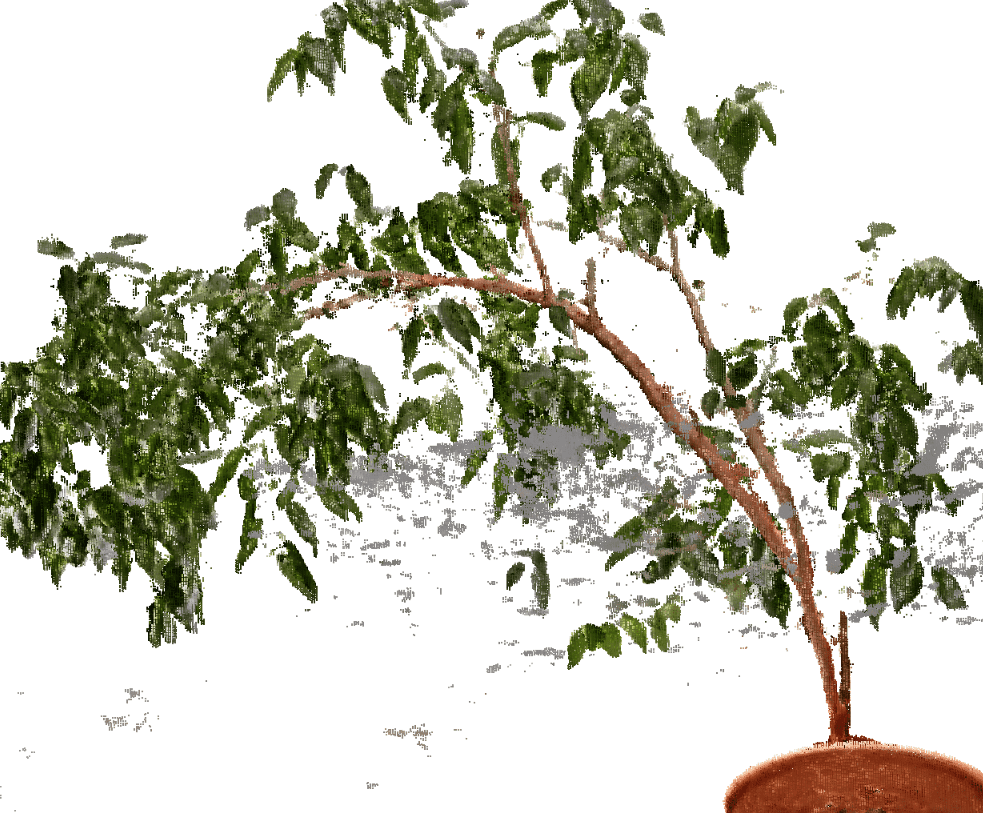}}	
     \subfigure[artifacts $>$90-th percentile]{\label{fig:artefacts_90th}
	\includegraphics[width=0.225\linewidth]{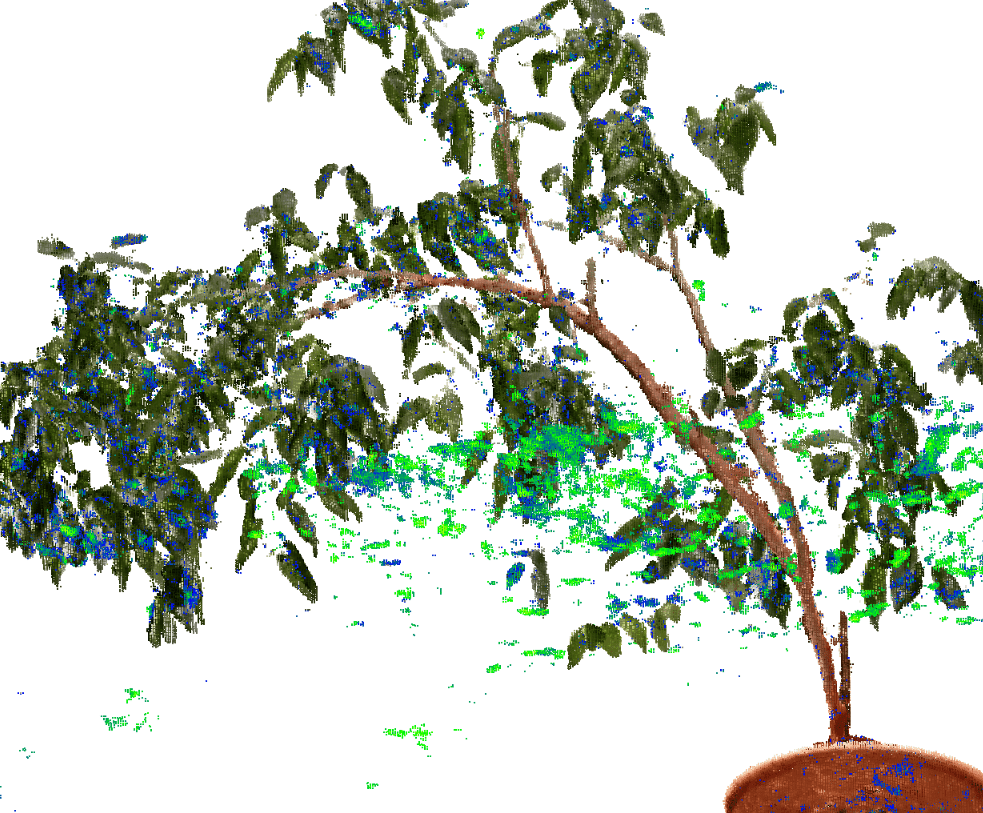}}
	\hspace{0.2cm}
     \subfigure[after artifact removal]{\label{fig:artefacts_after}  
     \includegraphics[width=0.225\linewidth]{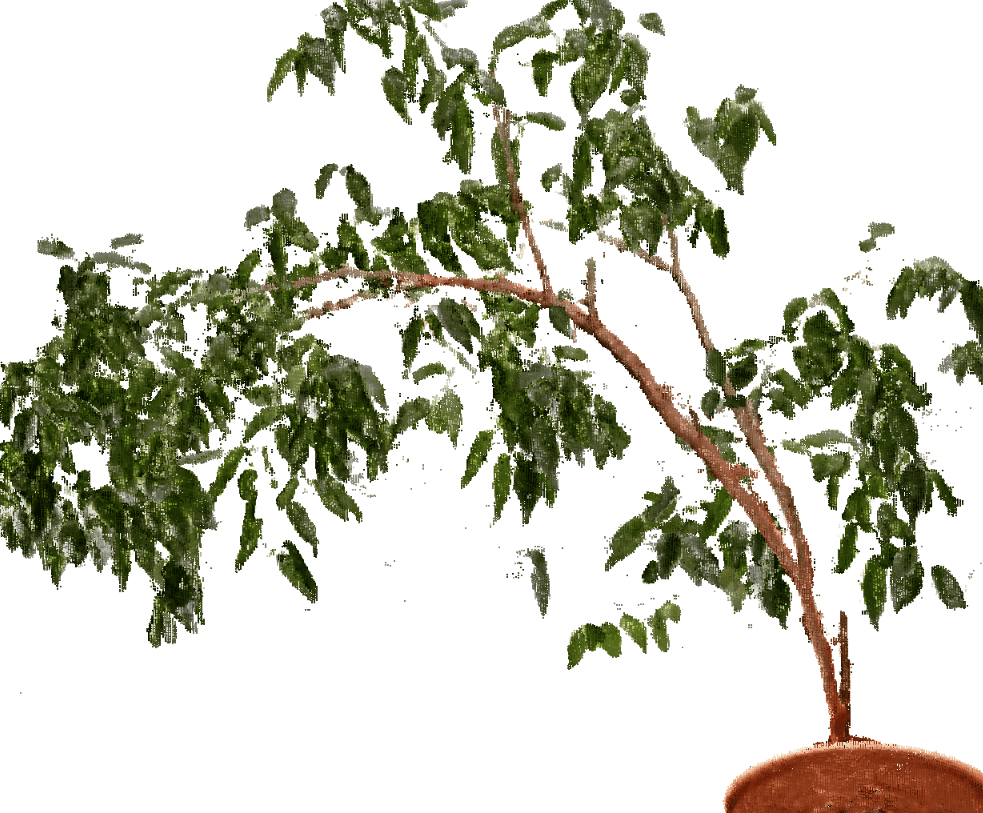}} \\
	\vspace{-3mm}
	\caption{Top: Artifacts from acquisition constellation. Bottom: Artifacts due to low-quality camera poses. From left to right: the rendered images \protect{\subref{fig:artefacts_scan40_rendered}}, \protect{\subref{fig:artefacts_rendered}}, the point clouds from the direct network output with the artifacts \protect{\subref{fig:artefacts_scan40_before}}, \protect{\subref{fig:artefacts_before}}, the artifacts over an density uncertainty percentile higher than 90 are highlighted in respective color \protect{\subref{fig:artefacts_scan40_90th}}, \protect{\subref{fig:artefacts_90th}}, the point clouds after artifact removal \protect{\subref{fig:artefacts_scan40_after}}, \protect{\subref{fig:artefacts_after}}.}
\label{fig:uncertainty_threshold}
\end{figure}

Exemplary, we perform a density uncertainty aware artifact removal on scenes in Figure \ref{fig:uncertainty_threshold}. We consider scene scan40 from the DTU dataset, where artifacts with high density uncertainties occur due to missing views in the acquisition constellation. And the HoloLens scene, where the artifacts occur in the case of low-quality poses.
We use the 90-th percentile of density uncertainty to remove the rough upper uncertainty outliers. The results are less noisy and free from artifacts that would have been present in corresponding 3D point cloud reconstructions and thus also in the rendered images.

\newpage

\section{Discussion}\label{sec:discussion}
So far, our research on NeRF-Ensembles takes a critical look at the complex interplay between the input data with data constraints, as well as the scene with scene constraints, and the resulting training performance as well as the density uncertainty. \newline

A key aspect concerns the data constraints given by the quality of the input data. Low-quality poses not only significantly degrade the training process, as indicated by a decrease in the Peak Signal-to-Noise Ratio. But also contribute to an increased density uncertainty of the NeRF output. Hereby, pose noise $\sigma_{\text{t}}$, $\sigma_{\text{R}}$ and $\sigma_{\text{t,R}}$ all affect the density uncertainty. Along with an increased density uncertainty, a significant decrease in density through low-quality input data was evident. Overall, data constraints lead to a noisier output with large (foggy) artifacts. In this respect, we see great potential for density uncertainty driven pose refinement methods for future work, given that only a few scene constraints are present.
Moreover, beyond the data constraints due to image noise and pose noise, our investigation extended to the impact of scene constraints on the density uncertainty. Thereby even with high-quality input data, variations and anomalies exist in the density uncertainty due to other factors. These factors are represented by acquisition constellations, such as a lack of views and occlusions, e.g. when images are captured from one side only. This leads to (foggy) artifacts and a high density uncertainty in these occluded areas. Moreover, material properties of the object, such as reflective surfaces, are a possible source of higher density uncertainty. \newline

Along with the density uncertainty quantification, we see a further potential in the application of NeRF-Ensembles. Firstly, there is an improved robustness from the mean density output of NeRF-Ensembles compared to the density from single NeRFs. The averaging of the density values across the ensemble members not only eliminates small artifacts and outliers, but also contributes to a more stabilized and higher completeness of the scene. Here are also possibilities for other metrics, such as median density values from NeRF-Ensembles. Secondly, a major improvement by NeRF-Ensembles is achieved in post-processing, where density uncertainty becomes crucial. The percentile-based density uncertainty outlier threshold effectively removes large (foggy) artifacts, improving the overall quality of NeRF output. In this context, we not only consider percentile-based removal as a possible solution, but also fixed density uncertainty thresholds. \newline Training multiple members within an ensemble for the uncertainty quantification requires additional training time. Nevertheless, we see the benefits such as enhanced robustness and artifact removal as a trade-off that justifies the training time of NeRF-Ensembles. \newline

Altogether, we can summarize the following main aspects to enable optimal 3D scene reconstructions with a high training performance regarding the novel view synthesis and a low density uncertainty of the network output: 1) High-quality input data, referring to images and poses. High-quality poses seem to be more important than non-noisy images in terms of the density uncertainty, i.e. the quality of the volumetric density. This is a crucial aspect, when generating 3D surface and point cloud reconstructions from NeRFs. Nonetheless, high-quality images are also important for proper rendering. 
2) Acquisition constellations and occlusions: Convergent acquisitions, spherically distributed around the object with as few gaps as possible between the views. 3) Well-textured subsurfaces and backgrounds. 4) The application of NeRF-Ensembles over a single NeRF to achieve a quantification of density uncertainty, to eliminate small outliers by mean density values as well as the density uncertainty-dependent removal of (foggy) artifacts in post-processing.

\section{Conclusion}\label{CONSLUSION}

In summary, we provided a survey investigating the density uncertainty of Neural Radiance Fields by using NeRF-Ensembles. Through our comprehensive study, we address the interplay of data and scene constraints, along with the training process and the resulting density uncertainty. Moreover, we demonstrate further potential of NeRF-Ensembles for enhanced robustness and artifact removal.\newline

The results emphasize the crucial impact of data constraints with the quality of the input data, which includes both images and poses. We observe a substantial impact on training performance, as measured by the Peak Signal-to-Noise Ratio. Furthermore, we reveal a correlation between data constraints and density uncertainty. Even with high-quality input data, variations in density uncertainty persist. These variations result from scene constraints, including acquisition constellations, occlusions, and material properties. This highlights the relevance of a detailed density uncertainty quantification in general. \newline

Furthermore, NeRF-Ensembles are suitable for enhancing the robustness due to their multiple votes from several NeRFs, comparable to well-established techniques like random forests \cite{randomforest}. The mean density eliminates small artifacts and outliers and contributes to the reconstruction with higher completeness. Moreover, NeRF-Ensembles offer practical applications in post-processing for the removal of large (foggy) artifacts through percentile-wise density uncertainty outlier elimination.\newline

Looking ahead, our findings not only contribute valuable insights into the application of NeRF-Ensembles by highlighting their potential to quantify the NeRF output, but also offer improved robustness and the possibility of density uncertainty-dependent artifact removal. This opens new opportunities for enhanced 3D reconstructions in the fields of computer graphics, computer vision, and photogrammetry.


\bibliographystyle{elsarticle-num}
\bibliography{authors}







\end{document}